\documentclass[10pt,twocolumn,letterpaper]{article}

\usepackage{iccv}
\usepackage{times}
\usepackage{epsfig}
\usepackage{graphicx}
\usepackage{amsmath}
\usepackage{amssymb}
\usepackage[accsupp]{axessibility}  

\usepackage[utf8]{inputenc} 
\usepackage[T1]{fontenc}    
\usepackage[dvipsnames]{xcolor}
\usepackage{url}            
\usepackage{booktabs}       
\usepackage{amsfonts}       
\usepackage{bbm}
\usepackage{nicefrac}       
\usepackage{microtype}      
\usepackage{lipsum}		
\usepackage{paralist}
\usepackage{amssymb}
\usepackage{caption}
\usepackage{array}
\usepackage{setspace}
\usepackage{subcaption}
\usepackage{bm}
\usepackage[acronym,smallcaps,nowarn,section,nogroupskip,nonumberlist]{glossaries}
\usepackage{soul}
\usepackage{arydshln} 
\usepackage{pifont}
\usepackage{multirow}
\usepackage{enumitem}

\usepackage[pagebackref=true,breaklinks=true,colorlinks,bookmarks=false]{hyperref}
\usepackage[nameinlink,capitalise,sort&compress]{cleveref}

\iccvfinalcopy 

\def\iccvPaperID{2101} 

\ificcvfinal\fi

\makeglossaries
\newcolumntype{P}[1]{>{\centering\arraybackslash}m{#1}}
\makeatletter
\newcommand\footnoteref[1]{\protected@xdef\@thefnmark{\ref{#1}}\@footnotemark}
\makeatother

\definecolor{red}{HTML}{E41A1C}
\definecolor{orange}{HTML}{FF7F00}
\definecolor{yellow}{HTML}{FFC020}
\definecolor{green}{HTML}{4DAF4A}
\definecolor{blue}{HTML}{377EB8}
\definecolor{purple}{HTML}{984EA3}

\hypersetup{%
  colorlinks,
  linkcolor={violet!70!black},
  citecolor={YellowOrange!70!black},
  urlcolor={Aquamarine!70!black}
}

\crefname{figure}{Figure}{Figures}
\crefname{table}{Table}{Tables}
\crefname{subtable}{Table}{Tables}
\crefname{section}{Section}{Sections}
\Crefname{section}{Section}{Sections}
\crefformat{equation}{Eq. (#2#1#3)}
\crefformat{enumi}{(#2#1#3)}

\newcommand{\predictsymbol}{^*}
\newcommand{\cleanvideo}{\bar{\bm{v}}}

\newcommand{\cluttervideo}{\bm{v}}
\newcommand{\cluttervideoframe}{v}
\newcommand{\cleanvideolabel}{p}
\newcommand{\cluttervideolabel}{p}
\newcommand{\cluttervideoprediction}{y\predictsymbol_{\text{mode}}}

\newcommand{\clutterframelabel}{y}
\newcommand{\clutterframepredictions}{\bm{y}\predictsymbol}
\newcommand{\clutterframeprediction}{y\predictsymbol}
\newcommand{\contextset}{\mathcal{C}}
\newcommand{\targetset}{\mathcal{T}}
\newcommand{\objectset}{\mathcal{P}}

\newcommand{\allcleanvideos}{\overline{\mathcal{V}}}
\newcommand{\allcluttervideos}{\mathcal{V}}

\newcommand{\trainusers}{\mathcal{K}^\text{train}}
\newcommand{\testusers}{\mathcal{K}^\text{test}}

\DeclareMathOperator*{\argmax}{arg\,max}
\DeclareMathOperator*{\argmin}{arg\,min}
\DeclareMathOperator*{\given}{\,|\,}

\newcommand{\user}{\kappa}

\newcommand{\cmark}{\ding{51}}%
\newcommand{\xmark}{\ding{55}}%

\glsdisablehyper{}
\newacronym{ML}{ml}{machine learning}
\newacronym{SGD}{sgd}{stochastic gradient descent}
\newacronym{BVI}{bvi}{blind or visually-impaired}
\newacronym{MLP}{mlp}{multi-layer perceptron}
\newacronym{TOR}{tor}{teachable object recogniser}
\newacronym{FPS}{fps}{frames per second}
\newacronym{SOTA}{sota}{state-of-the-art}
\newacronym{PII}{pii}{personally identifiable information}

\begin{document}

\title{ORBIT: A Real-World Few-Shot Dataset for Teachable Object Recognition}

\author{Daniela Massiceti\(^1\)
        \quad
        Luisa Zintgraf\(^2\)
        \quad
        John Bronskill\(^3\)
        \quad
        Lida Theodorou\(^4\)\\
        \and
        Matthew Tobias Harris\(^4\)
        \quad
        Edward Cutrell\(^1\)
        \quad
        Cecily Morrison\(^1\)
        \quad
        Katja Hofmann\(^1\)
        \quad
        Simone Stumpf\(^4\)\\\\
        {\normalsize \(^1\)\textit{Microsoft Research} \quad \(^2\)\textit{University of Oxford} \quad \(^3\)\textit{University of Cambridge} \quad \(^4\)\textit{City, University of London}}
}

\maketitle
\ificcvfinal\thispagestyle{empty}\fi

\begin{abstract}
    \vspace{-0.65em}
    \noindent Object recognition has made great advances in the last decade, but predominately still relies on many high-quality training examples per object category. In contrast, learning new objects from only a few examples could enable many impactful applications from robotics to user personalization. Most few-shot learning research, however, has been driven by benchmark datasets that lack the high variation that these applications will face when deployed in the real-world. To close this gap, we present the ORBIT dataset and benchmark, grounded in the real-world application of teachable object recognizers for people who are blind/low-vision. The dataset contains 3,822 videos of 486 objects recorded by people who are blind/low-vision on their mobile phones. The benchmark reflects a realistic, highly challenging recognition problem, providing a rich playground to drive research in robustness to few-shot, high-variation conditions. We set the benchmark's first state-of-the-art and show there is massive scope for further innovation, holding the potential to impact a broad range of real-world vision applications including tools for the blind/low-vision community. We release the dataset at~\href{https://doi.org/10.25383/city.14294597}{https://doi.org/10.25383/city.14294597} and benchmark code at~\href{https://github.com/microsoft/ORBIT-Dataset}{https://github.com/microsoft/ORBIT-Dataset}.
\end{abstract}

\glsresetall

\vspace{-1.5em}
\section{Introduction}
\vspace{-0.5em}

Object recognition systems have made spectacular advances in recent years~\cite{szegedy2017inception,touvron2020fixing,tan2019efficientnet,ren2015faster,he2017mask,mahajan2018exploring,redmon2017yolo9000} however, most systems still rely on training datasets with 100s to 1,000s of high-quality, labeled examples per object category.
These demands make training datasets expensive to collect, and limit their use to all but a few application areas.
\begin{figure}
    \centering
    \begin{subfigure}[t]{0.5\textwidth}
        \centering
        \includegraphics[width=0.22\textwidth]{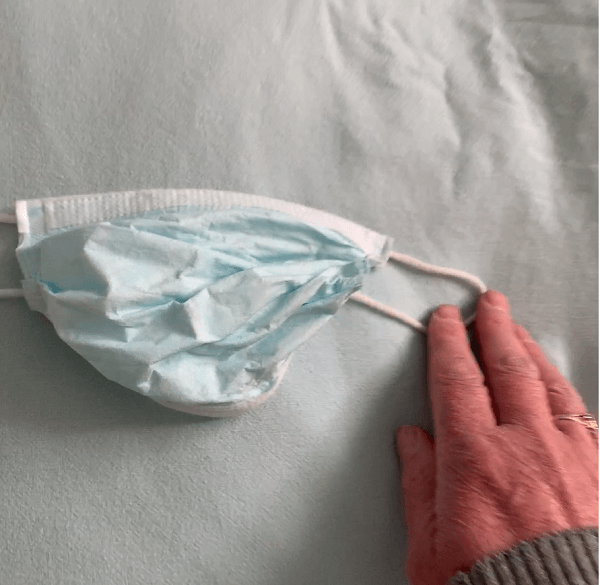}
        \includegraphics[width=0.22\textwidth]{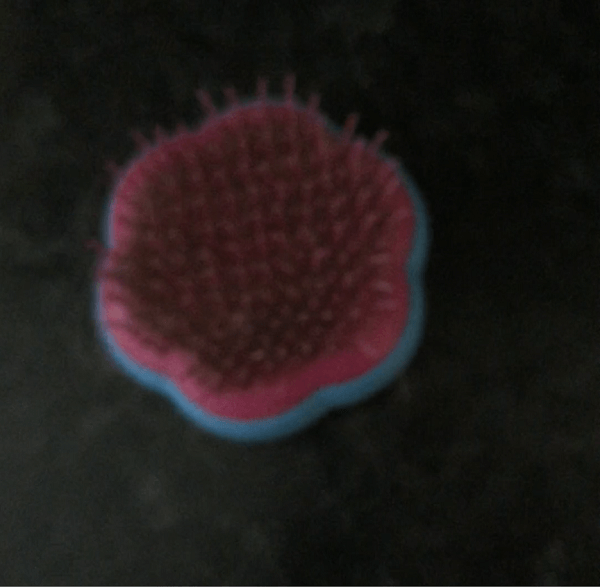}
        \includegraphics[width=0.22\textwidth]{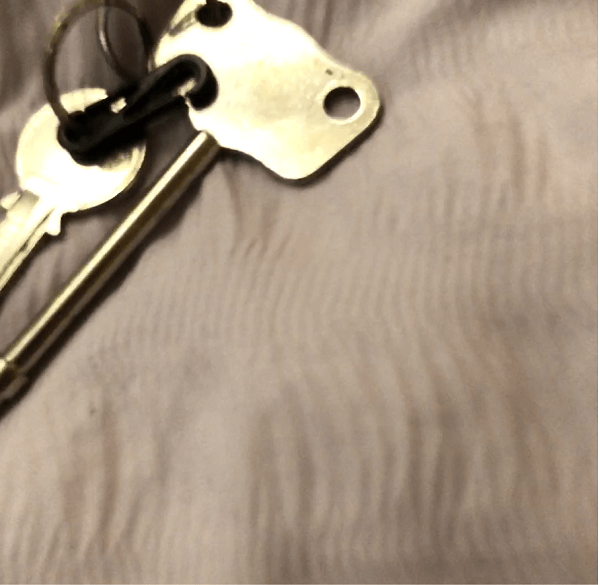}
        \includegraphics[width=0.22\textwidth]{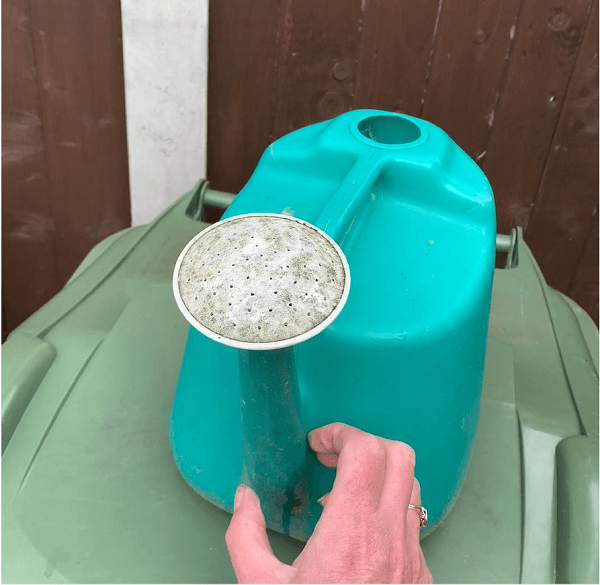}
        \caption{Frames from clean videos}
    \end{subfigure}
    \vspace{0.5em}
    \begin{subfigure}[t]{0.5\textwidth}
        \centering
        \includegraphics[width=0.22\textwidth]{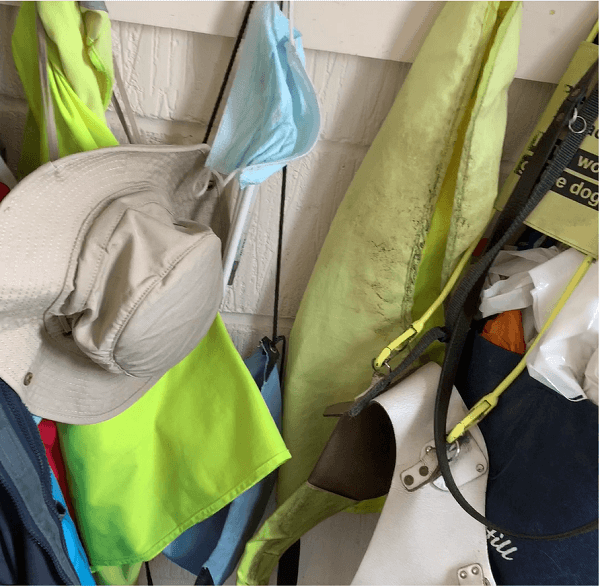}
        \includegraphics[width=0.22\textwidth]{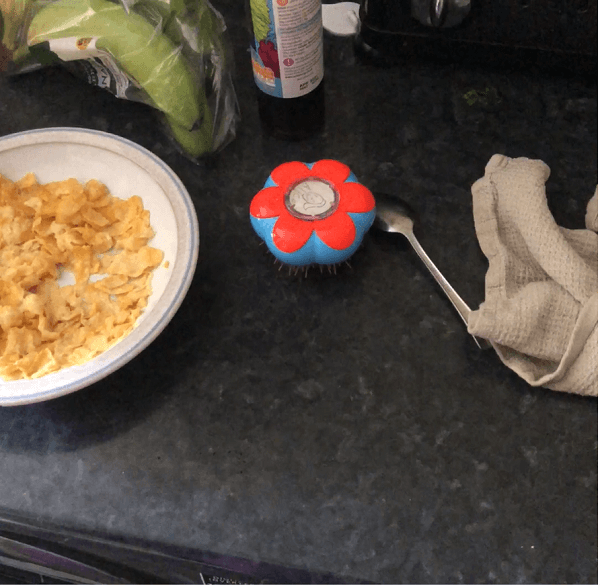}
        \includegraphics[width=0.22\textwidth]{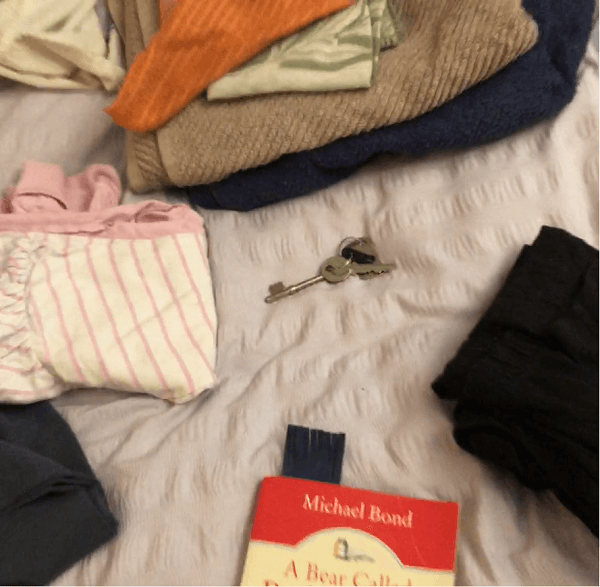}
        \includegraphics[width=0.22\textwidth]{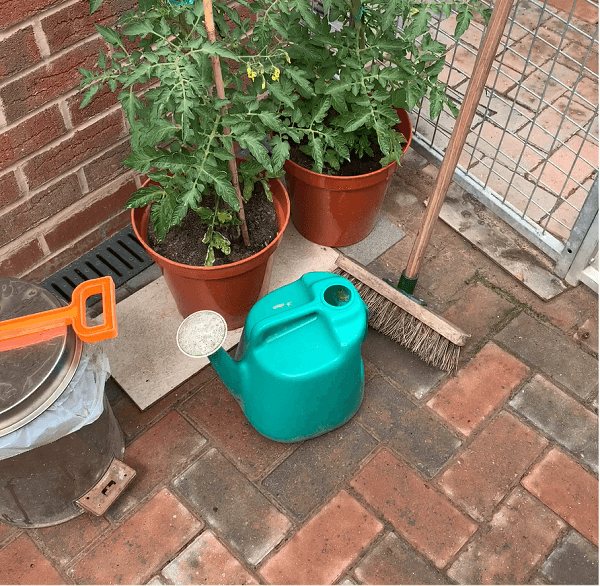}
        \caption{Frames from clutter videos}
    \end{subfigure}
    \vspace{-1.5em}
    \caption{High-variation examples in the ORBIT dataset -- a facemask, hairbrush, keys, and watering can. Full videos in the supplementary material. Further examples in~\cref{fig:more-qual-frame-examples}.}
    \label{fig:qual-frame-examples}
    \vspace{-1.8em}
\end{figure}

Few-shot learning aims to reduce these demands by training models to recognize completely novel objects from only a few examples~\cite{finn2017model,vinyals2016matching,snell2017prototypical,bertinettometa,requeima2019fast,gordon2018metalearning, tian2020rethinking}.
This will enable recognition systems that can adapt in real-world, dynamic scenarios, from self-driving cars to applications where users provide the training examples themselves.
Meta-learning algorithms which ``learn to learn''~\cite{thrun2012learning,finn2017model,vinyals2016matching,gordon2018metalearning} hold particular promise toward this goal with recent advances opening exciting possibilities for light-weight, adaptable recognition.  

Most few-shot learning research, however, has been driven by datasets that lack the high variation --- in number of examples per object and quality of those examples (framing, blur, etc.; see~\cref{tab:meta-dataset-comparison}) --- that recognition systems will likely face when deployed in the real-world.
%
%
Key datasets such as Omniglot~\cite{lake2011one,vinyals2016matching} and \emph{mini}ImageNet~\cite{vinyals2016matching}, for example, present highly structured benchmark tasks which assume a fixed number of objects and training examples per object.
Meta-Dataset~\cite{triantafillou2019meta}, another key dataset, poses a more challenging benchmark task of adapting to novel \emph{datasets} given a small (random) number of training examples.
Its constituent datasets~\cite{lake2011one,quickdraw,ILSVRC15,Lin14COCO,nilsback2008flowers,wah2011birds,cimpoi14describing}, however, mirror the high-quality images of Omniglot and \emph{mini}ImageNet, leaving robustness to the noisy frames that would be streamed from a real-world system unaddressed.
While these datasets have catalyzed research in few-shot learning, state-of-the-art performance is now relatively saturated and leaves reduced scope for algorithmic innovation~\cite{hu2020ptmap,chen2019self,park2019meta}.

To drive further innovation in few-shot learning for real-world impact, there is a strong need for datasets that capture the high variation inherent in real-world applications.
We motivate that both the dataset and benchmark task should be grounded in a potential real-world application to bring real-world recognition challenges to life in their entirety.
An application area that neatly encapsulates a few-shot, high-variation scenario are \glspl{TOR} for people who are blind/low-vision~\cite{lee2019hands,kacorri2017teachable}. 
Here, a user can customize an object recognizer by capturing a small number of (high-variation) training examples of essential objects on their mobile phone.
The recognizer is then trained (in deployment) on these examples such that it can recognize the user's objects in novel scenarios.
As a result, \glspl{TOR} capture a microcosm of highly challenging and realistic conditions that can be used to drive research in real-world recognition tasks, with the potential to impact a broad range of applications beyond just tools for the blind/low-vision community.

We introduce the ORBIT dataset~\cite{orbitdataset2021}, a collection of videos recorded by people who are blind/low-vision on their mobile phones, and an associated few-shot benchmark grounded in \glspl{TOR}.
Both were designed in collaboration with a team of \gls{ML}, human-computer interaction, and accessibility researchers, and will enable the \gls{ML} community to
\begin{inparaenum}[1)]
\item accelerate research in few-shot, high-variation object recognition, and
\item explore new research directions in few-shot \emph{video} recognition.
\end{inparaenum}
We intend both as a rich playground to drive research in robustness to challenging, real-world conditions, a step beyond what curated few-shot datasets and structured benchmark tasks can offer, and to ultimately impact a broad range of real-world vision applications.
In summary, our contributions are:

\vspace{-1.5em}
\noindent\paragraph{1. ORBIT benchmark dataset.}
The ORBIT benchmark dataset~\cite{orbitdataset2021} (\cref{sec:orbit-benchmark-dataset}) is a collection of 3822 videos of 486 objects recorded by 77 blind/low-vision people on their mobile phones and can be downloaded at \href{https://doi.org/10.25383/city.14294597}{https://doi.org/10.25383/city.14294597}.
Examples are shown in~\cref{fig:qual-frame-examples,fig:more-qual-frame-examples}.
Unlike existing datasets~\cite{ILSVRC15,deng2009imagenet,Lin14COCO,vinyals2016matching, triantafillou2019meta}, ORBIT show objects in a wide range of realistic conditions, including when objects are poorly framed, occluded by hands and other objects, blurred, and in a wide variation of backgrounds, lighting, and object orientations.
%
\vspace{-1.5em}
\noindent\paragraph{2. ORBIT teachable object recognition benchmark.}
We formulate a few-shot benchmark on the ORBIT dataset (\cref{sec:orbit-benchmark}) that is grounded in \glspl{TOR} for people who are blind/low-vision.
Contrasting existing few-shot (and other) works, the benchmark proposes a novel user-centric formulation which measures personalization to individual users.
It also incorporates metrics that reflect the potential computational cost of real-world deployment on a mobile device.
These and the benchmark's other metrics are specifically designed to drive innovation for realistic settings.

\vspace{-1.5em}
\noindent\paragraph{3. \Gls{SOTA} on the ORBIT benchmark.}
We implement 4 few-shot learning models that cover the main classes of approach in the field, extend them to videos, and establish the first \gls{SOTA} on the ORBIT benchmark (\cref{sec:experiments}).
We also perform empirical studies showing that training on existing few-shot learning datasets is \emph{not} sufficient for good performance on the ORBIT benchmark (\cref{tab:orbit-baselines}) leaving significant scope for algorithmic innovation in few-shot techniques that can handle high-variation data.

Code for loading the dataset, computing benchmark metrics, and running the baselines is available at \href{https://github.com/microsoft/ORBIT-Dataset}{https://github.com/microsoft/ORBIT-Dataset}.

\begin{table*}[t]
    \centering
    \hspace{-2em}
    \scalebox{0.72}{
    \begin{tabular}{ll|P{3cm}P{3cm}P{3cm}P{3.1cm}P{3.4cm}}
        & & Omniglot~\cite{lake2011one} & \emph{mini}ImageNet~\cite{vinyals2016matching} & Meta-Dataset~\cite{triantafillou2019meta} & TEgO~\cite{lee2019hands} & \textbf{ORBIT Benchmark} \\
        \cmidrule{2-7}
        &\textbf{Data type} & Image & Image & Image & Image & Video frames  \\
        &\textbf{\# classes} & 1623 & 100 & 4934 & 19 & 486 \\
        &\textbf{\# samples/class} & 20 & 600 & 6-340,029 & 180-487 & 33-3,600 \\
        &\textbf{\# total samples} & 32,460 & 60,000 & 52,764,077 & 11,930 & 2,687,934 \\
        &\textbf{Goal} & Image classification & Image classification & Image classification & Image classification & Frame classification \\
        &\textbf{Task} & Fixed shot/way & Fixed shot/way & Random shot/way & Fixed shot/way & Random shot/way \\
        &\textbf{Source} & Turk & Web & Web & Mobile phone & Mobile phone \\
        &\textbf{Data collectors} & Sighted (20) & Sighted & Sighted & Sighted (1) Blind (1) & Blind (67) \\
        \cmidrule{2-7}
        \multirow{6}{*}{\rotatebox[origin=c]{90}{\textbf{High-variation}}} 
        \multirow{6}{*}{\rotatebox[origin=c]{90}{\textbf{features}}}
        & Unbalanced classes & \xmark & \xmark & \cmark & \xmark & \cmark \\
        & Lighting variation & \xmark & \cmark & \cmark & \xmark & \cmark \\
        & Background variation & \xmark & \cmark & \cmark & \cmark* & \cmark \\
        & Viewpoint variation & \xmark & \xmark & \xmark & \cmark & \cmark \\
        & Ill-framed objects & \xmark & \xmark & \xmark & \cmark & \cmark \\
        & Blur & \xmark & \xmark & \xmark & \xmark & \cmark \\
    \end{tabular}}
    \vspace{0.em}
    \caption{Comparison of few-shot learning datasets. Note, the ORBIT benchmark dataset is a subset of all videos contributed by collectors (see~\cref{app:sec:dataset-prep}). *Collected in 2 controlled environments -- 1 uniform background, 1 cluttered space.}
    \label{tab:meta-dataset-comparison}
    \vspace{-1em}
\end{table*}

\section{Related Work}

\noindent\textbf{Few-shot learning datasets.} Omniglot~\cite{lake2011one,vinyals2016matching}, \emph{mini}ImageNet~\cite{vinyals2016matching}, and Meta-Dataset~\cite{triantafillou2019meta} have driven recent progress in few-shot learning.
Impressive gains have been achieved on Omniglot and \emph{mini}ImageNet~\cite{vinyals2016matching,hu2020ptmap,chen2019self,park2019meta}, however results are now largely saturated and highly depend on the selected feature embedding.
Meta-Dataset, a dataset of 10 datasets, formulates a more challenging task where whole \emph{datasets} are held-out, but these datasets contain simple and clean images, such as clipart drawings of characters/symbols~\cite{lake2011one,vinyals2016matching,quickdraw}, and ImageNet-like images~\cite{Lin14COCO,ILSVRC15,nilsback2008flowers,wah2011birds,cimpoi14describing} showing objects in uniform lighting, orientations, and camera viewpoints.
The ORBIT dataset and benchmark presents a more challenging few-shot task with high-variation examples captured in real-world scenarios.

\noindent\textbf{High-variation datasets.}
Datasets captured by users in real-world settings are naturally high-variation~\cite{barbu2019objectnet,SomethingSomething,Damen2018EPICKITCHENS,koh2020wilds,core50dataset,kacorri2017teachable,sosa2017glassene, gurari2018vizwiz}, but none collected thus far explicitly target few-shot object recognition.
\emph{ObjectNet}~\cite{barbu2019objectnet} is a test-only dataset of challenging images (e.g. unusual orientations/backgrounds) for ``many-shot" classification.  
\emph{Something-Something}~\cite{SomethingSomething} and \emph{EPIC-Kitchens}~\cite{Damen2018EPICKITCHENS} are video datasets collected by users with mobile and head-mounted cameras, respectively, but are focused on action recognition based on many examples and ``action captions''.
\emph{Core50}~\cite{core50dataset} is a video dataset captured on mobile phones for a continual learning recognition task. In contrast to ORBIT, the videos are high quality (captured by sighted people, with well-lit centered objects).
%
Other high-variation datasets include those collected by people who are blind/low-vision~\cite{kacorri2017teachable,sosa2017glassene,gurari2018vizwiz} (see \href{https://incluset.com}{\emph{IncluSet}} for a repository of accessibility datasets~\cite{Kacorri2020Incluset}) however, most are not appropriate for few-shot learning.
\emph{TeGO}~\cite{kacorri2017teachable} contains mobile phone images of 19 objects taken by only 2 users (1 sighted, 1 blind) in 2 environments (1 uniform background, 1 cluttered scene). It validates the \gls{TOR} use-case, but is too small to deliver a robust, deployable system.
\emph{VizWiz}~\cite{gurari2018vizwiz}, although larger scale (31,173 mobile phone images contributed by 11,045 blind/low-vision users) targets image captioning and question-answering tasks, and is not annotated with object labels.
The ORBIT dataset and benchmark is motivated by the lack of datasets that have the scale and structure required for few-shot, high-variation real-world applications, and adds to the growing repository of datasets for accessibility.

\section{ORBIT Benchmark Dataset}
\label{sec:orbit-benchmark-dataset}

Our goal is to drive research in recognition tasks under few-shot, high-variation conditions so that deployed few-shot systems are robust to such conditions.
Toward this goal, we focus on a real-world application that serves as a microcosm of a few-shot, high-variation setting --- \glspl{TOR} for people who are blind/low-vision -- and engage the blind/low-vision community in collecting a large-scale dataset.

The collection took place in two phases, and collectors recorded and submitted all videos (completely anonymously) via an accessible iOS app (see~\cref{app:sec:orbit-camera-app}).
The collection protocol was designed and validated through extensive user studies~\cite{theodorou2021disability} and led to the key decision to capture \textit{videos} rather than images of objects.
This was based on the hypothesis that a video increases a blind collector's chances of capturing frames that contained the object while reducing the time/effort cost to the collector, compared to multiple attempts at a single image.
The study was approved by the City, University of London Research Ethics Committee.
The full data collection protocol is described in~\cref{app:sec:data-collection-protocol} and a datasheet~\cite{gebru2018datasheets} for the dataset is included in~\cref{app:sec:datasheet}.

We summarize the benchmark dataset in~\cref{tab:benchmark-dataset-summary} and describe it in detail below (see~\cref{app:sec:dataset-prep} for dataset preparation, and~\cref{app:sec:extended-benchmark-dataset-summary} for example clips).
The benchmark dataset is used to run the benchmark described in~\cref{sec:orbit-benchmark}.

\noindent\textbf{Number of collectors.}
Globally, 77 collectors contributed to the ORBIT benchmark dataset.
Collectors who contributed only 1 object were merged to enforce a minimum of 3 objects per user such that the per-user classification task was a minimum of 3-way, resulting in an effective 67 users.

\begin{table*}[t]
    \centering
    \scalebox{0.8}{
    \begin{tabular}{c|ccccccccc}
    & \textbf{Collectors} &  \textbf{Objects} & \textbf{Videos} & \multicolumn{3}{c}{\textbf{Videos per object}} & \multicolumn{3}{c}{\textbf{Frames per video}}\\
    & & &  & mean/std & 25/75\(^\text{th}\) perc. & min/max & mean/std & 25/75\(^\text{th}\) perc. & min/max \\
    \cmidrule{1-10}
    \textbf{Total} & 67 & 486 & 3822 & 7.9/4.8 & 7.0/7.0 & 3.0/46.0 & 703.3/414.1 & 396.2/899.0 & 33.0/3600.0 \\
    \cmidrule{1-10}
    \textbf{Clean} & & & 2996 & 6.2/4.6 & 5.0/6.0 & 2.0/44.0 & 771.3/420.6 & 525.8/900.0 & 33.0/3600.0 \\
    \textbf{Clutter} & & & 826 & 1.7/1.5 & 1.0/2.0 & 1.0/13.0 & 456.7/272.9 & 248.5/599.0 & 40.0/3596.0\\
    \cmidrule{1-10}
    \textbf{Per-collector} & 1 & 7.3/2.8 & 57.0/47.4 & 7.5/4.0 & 6.6/7.4 & 3.4/38.4 & 728.8/208.8 & 609.4/808.2 & 213.1/1614.3 \\
    \cmidrule{1-10}
    \textbf{Clean} & & & 44.7/44.0 & 5.8/3.9 & 4.8/6.0 & 2.4/36.5 & 809.9/244.7 & 664.7/898.5 & 219.3/1872.6 \\ 
    \textbf{Clutter} & & & 12.3/10.8 & 1.8/1.1 & 1.0/2.0 & 1.0/9.9 & 728.8/208.8 & 609.4/808.2 & 213.1/1614.3\\ 
    \end{tabular}}
    \vspace{-0.5em}
    \caption{ORBIT benchmark dataset.}
    \label{tab:benchmark-dataset-summary}
    \vspace{-1em}
\end{table*}

\begin{figure}[ht!]
    \centering
    \begin{subfigure}[t]{\columnwidth}
        \centering
        \includegraphics[width=\textwidth, trim=0.2cm 0.4cm 1cm 3cm, clip]{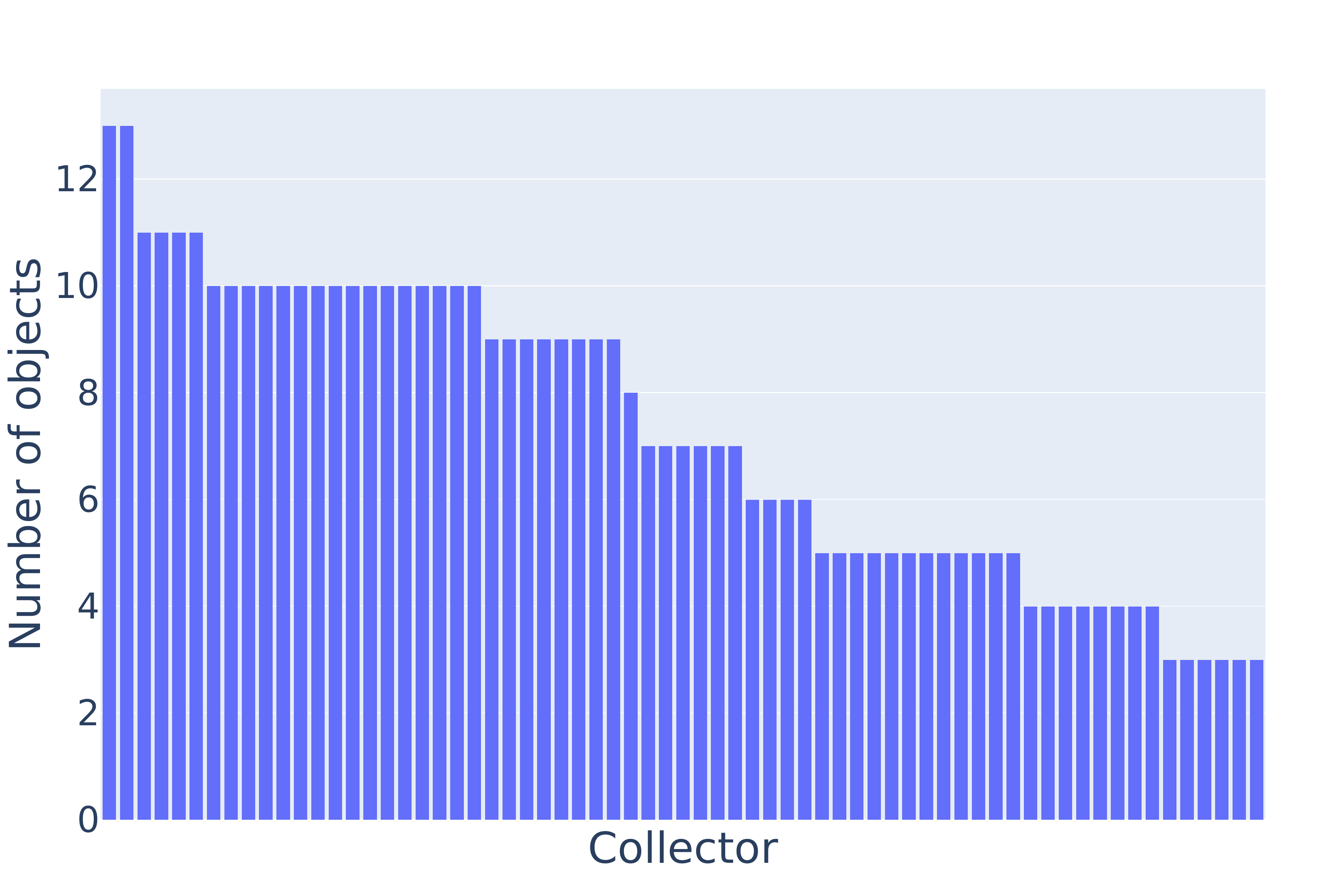}
        \vspace{-1.6em}
        \caption{Number of objects per collector.}
        \label{fig:num-objects-all-main}
    \end{subfigure}\\
    \begin{subfigure}[t]{\columnwidth} 
        \centering
        \includegraphics[width=\textwidth, trim=0.2cm 0.4cm 1cm 1cm, clip]{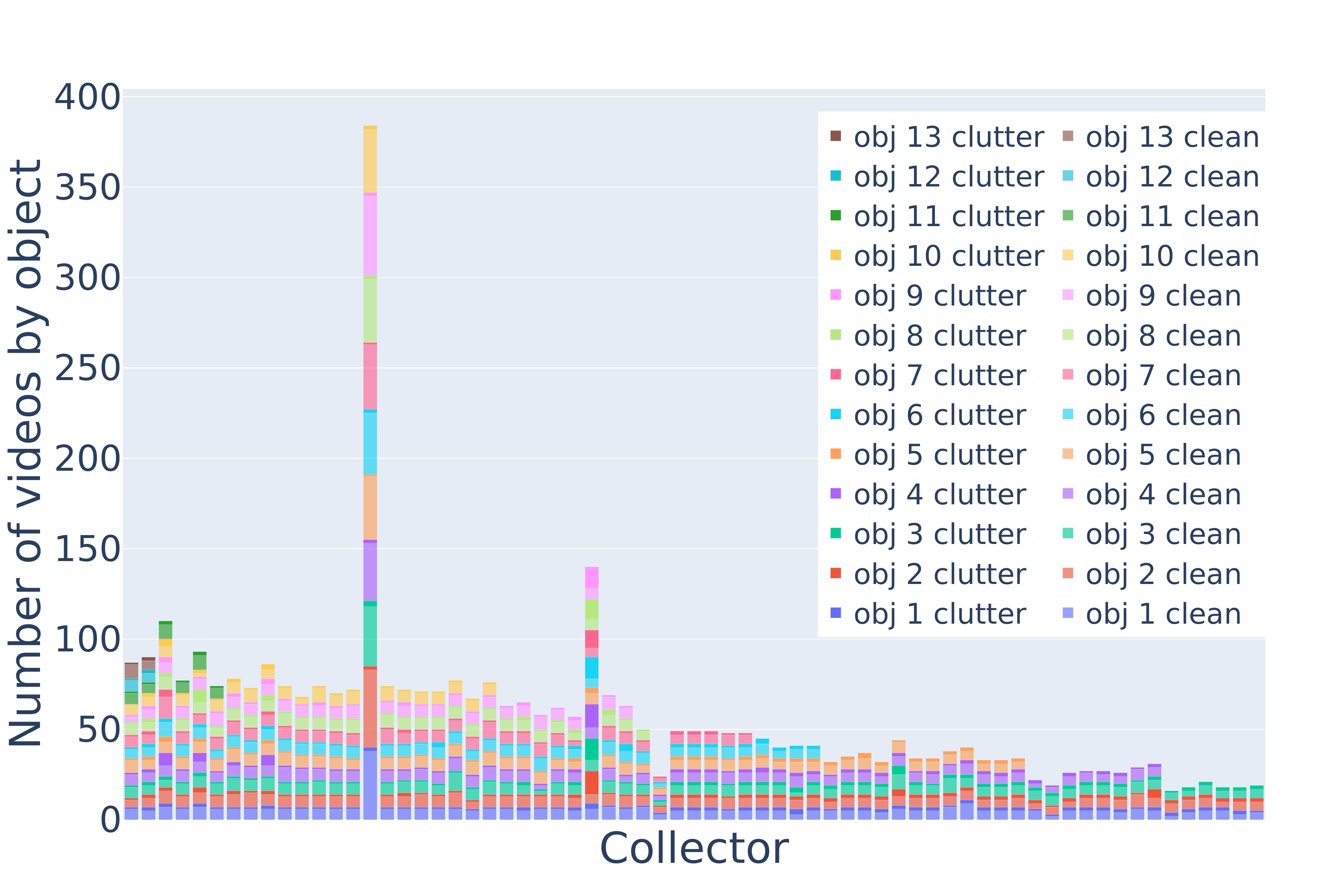}
        \vspace{-1.6em}
        \caption{Number of videos (stacked by object) per collector.}
        \label{fig:num-vids-by-object-all-main}
    \end{subfigure}
    \caption{Number of objects and videos across 67 collectors.}
    \label{fig:benchmark-dataset-summary}
    \vspace{-1em}
\end{figure}

\noindent\textbf{Numbers of videos and objects.}
Collectors contributed a total of 486 objects and 3,822 videos (2,687,934 frames, 83GB).
2,996 videos showed the object in isolation, referred to as \emph{clean} videos, while 826 showed the object in a realistic, multi-object scene, referred to as \emph{clutter} videos. 
We collected both types to match what a \gls{TOR} will encounter in the real-world (see~\cref{sec:evaluation-modes}).
Each collector contributed on average 7.3 (\(\pm\)2.8) objects, with 5.8 (\(\pm\)3.9) clean videos and 1.8 (\(\pm\)1.1) clutter videos per object.
\cref{fig:benchmark-dataset-summary} shows the number of objects (\ref{fig:num-objects-all-main}) and number of videos per collector (\ref{fig:num-vids-by-object-all-main}).
We discuss the impact of the 2 collectors who contributed more videos than the average collector in~\cref{app:sec:dataset-prep-set-creation}.

\noindent\textbf{Types of objects.}
Collectors provided object labels for each video contributed.
Objects covered course-grained categories (e.g. remote, keys, wallet) as well as fine-grained categories (e.g. Apple TV remote, Virgin remote, Samsung TV remote control).
For summarization purposes, we clustered the objects based on object similarity and observe a long-tailed distribution (see~\cref{fig:benchmark-cluster-histo-all}).
The largest clusters contained different types of remotes/controls, keys, wallets/purses, guidecanes, doors, airpods, headphones, mobile phones, watches, sunglasses and Braille readers.
More than half of the clusters contained just 1 object.
The clustering algorithm and cluster contents are included in~\cref{app:sec:object-clustering}.

\noindent\textbf{Bounding box annotations.}
Since the clutter videos could contain multiple objects, we provide bounding box annotations around the target object in all clutter videos (available in the code repository).
We use these to compute the proportion of time the target object spends in- versus out-of-frame per video, and show this in~\cref{app:fig:bounding-box-summary-allsets} averaged over all clutter videos per collector.
On average, the target object is in-frame for \(\sim\)95\% of any given clutter video.

\noindent\textbf{Video lengths.}
Video lengths depended on the recording technique required for each video type (see~\cref{app:sec:data-collection-protocol}).
On average, clean videos were 25.7s (\(\sim\)771 frames at 30 \acrshort{FPS}), and clutter videos were 15.2s (\(\sim\)457 frames at 30 \acrshort{FPS}).

\noindent\textbf{Unfiltered ORBIT dataset.}
Some collectors did not meet the minimum requirements to be included in the benchmark dataset (e.g. an object did not have both clean and clutter videos).
The benchmark dataset was therefore extracted from a larger set of 4733 videos (3,161,718 frames, 97GB) of 588 objects contributed by 97 collectors.
We summarize the unfiltered dataset in~\cref{app:orbit-full-summary}.

\section{Teachable Object Recognition Benchmark}
\label{sec:orbit-benchmark}

The ORBIT dataset can be used to explore a wide set of real-world recognition tasks from continual learning~\cite{core50dataset, lomonaco2020rehearsal} to video segmentation~\cite{lee2011key, perazzi2017learning, lu2020learning}.
In this paper, we focus on few-shot object recognition from high-variation examples and present a realistic and challenging few-shot benchmark grounded in \glspl{TOR} for people who are blind/low-vision.

In~\cref{sec:pors}, we describe how a \gls{TOR} works, mapping it to a few-shot learning problem, before presenting the benchmark's evaluation protocol and metrics in~\cref{sec:eval-protocol}.

\subsection{Teachable Object Recognition}
\label{sec:pors}

We define a \gls{TOR} as a generic recognizer that can be customized to a user's personal objects using a small number of training examples -- in our case, videos -- which the user has captured themselves.
The 3 steps to realizing a \gls{TOR} are:

\begin{compactenum}[(1)]
\item \textbf{Train.} 
    A recognition model is trained on a large dataset of objects where each object has only a few examples.
    The model can be optimized to either
    \begin{inparaenum}[i)]
    \item directly recognize a set of objects~\cite{tian2020rethinking, chen2020new} or
    \item learn \textit{how} to recognize a set of objects (i.e. meta-learn)~\cite{finn2017model, snell2017prototypical,vinyals2016matching,requeima2019fast}.
    \end{inparaenum} 
    This happens before deploying the model in the real world.
\item \textbf{Personalize.} 
    A real-world user captures a few examples of a set of their personal objects. The deployed model is trained on this user's objects using just these examples.
\item \textbf{Recognize.} 
    The user employs their now-personalized recognizer to identify their personal objects in novel (test) scenarios. 
    As the user points their recognizer at a scene, it delivers frame-by-frame predictions.
\end{compactenum}

\subsubsection{\glspl{TOR} as a few-shot learning problem}
\label{sec:few-shot-for-pors}

\begin{figure}[ht!]
    \centering
    \vspace{-1em}
    \includegraphics[width=0.5\textwidth, trim=0 0 10cm 0,clip]{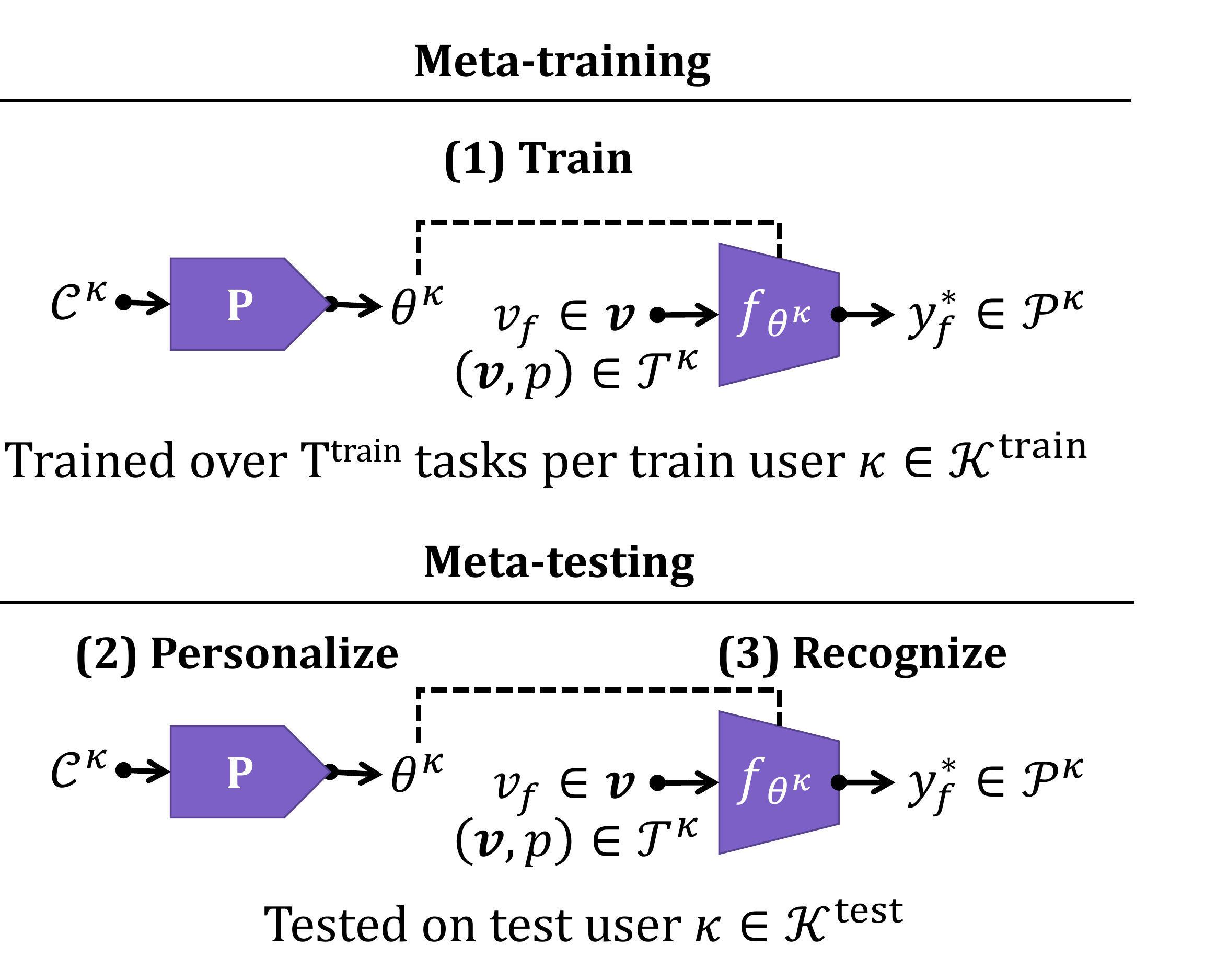}
    \vspace{-2em}
    \caption{Teachable object recognizers cast as a few-shot learning problem. \(P\) is the personalization method, for example, several gradient steps using a optimization-based approach, or parameter generation using a model-based approach (see~\cref{sec:baselines}).}
    \label{fig:few-shot-diagram}
    \vspace{-1em}
\end{figure}

The \textbf{(1) train} step of a \gls{TOR} can be mapped to the `meta-training' phase typically used in few-shot learning set-ups.
The \textbf{(2) personalize} and \textbf{(3) recognize} steps can be mapped to `meta-testing'  (see~\cref{fig:few-shot-diagram}).
With this view, we now formalize the teachable object recognition task, drawing on nomenclature from the few-shot literature~\cite{finn2017model,snell2017prototypical,requeima2019fast,gordon2018metalearning}.

We construct a set of train users \(\trainusers\) and test users \(\testusers\) (\(\trainusers\, \cap\,\testusers = \varnothing\)) akin to the train and test object classes used in few-shot learning.
A user \(\user\) has a set of personal objects \(\objectset^\user\) that they want a recognizer to identify, setting up a \(|\objectset^\user|\)-way classification problem.
To this end, the user captures a few videos of each object, together called the user's ``context'' set \(\contextset^\user = \{(\cleanvideo, p)_i\}_{i=1}^N\), where \(\cleanvideo\) is a context video, \(p\in \objectset^\user\) is its object label, and \(N\) is the total number of the user's context videos.
The goal is to use \(\contextset^\user\) to learn a recognition model \(f_{\theta^\user}\) that can identify the user's objects, where \(\theta^\user\) are the model parameters specific to user \(\user\).

Once personalized, the user can point their recognizer at novel ``target'' scenarios to receive per-frame predictions:
\begin{align}\label{eq:frame-predict}
    \clutterframeprediction_f = 
    \argmax_{y_f \in \objectset^\user}\, 
    f_{\theta^\user} (\cluttervideoframe_{f})
    \quad\quad
    \cluttervideoframe_f \in \cluttervideo
    \quad\quad
    (\cluttervideo, p) \in \targetset^\user
\end{align}
where \(\cluttervideoframe_f\) is a target frame, \(\cluttervideo\) is a target video, \(\targetset^\user\) is all the user's target videos, and \(\clutterframelabel_f \in \objectset^\user\) is the frame-level label.\footnote{Note, \(y_f = p\) where \(p\in\objectset^\user\) is the video-level object label}

Following the typical paradigm, during meta-training (i.e. the \textbf{train} step), multiple tasks are sampled per user \(\user \in \trainusers\) where a task is a random sub-sample of the user's \(\contextset^\user\) and \(\targetset^\user\) (see \cref{app:sec:task-sampling}).
The recognition model can be trained on these tasks using an episodic~\cite{finn2017model,snell2017prototypical,vinyals2016matching,requeima2019fast} or non-episodic approach~\cite{chen2020new,tian2020rethinking,laenen2020episodes}.
We formalize both in the context of \glspl{TOR} in~\cref{app:sec:episodic-vs-non}.
Then, at meta-testing, one task is sampled per test user \(\user \in \testusers\)
containing \emph{all} the user's context and target videos.
For each test user, the recognizer is personalized using all their context videos \(\contextset^\user\) (i.e. the \textbf{personalize} step), and then evaluated on each of the user's target videos in \(\targetset^\user\) (i.e. the \textbf{recognize} step).
In the following section, we discuss this evaluation protocol.

\subsection{Evaluation protocol}
\label{sec:eval-protocol}

ORBIT's evaluation protocol is designed to reflect how well a \gls{TOR} will work in the hands of a real-world user --- both in terms of performance and computational cost to personalize.
To achieve this, we test (and train) in a user-centric way where tasks are sampled \emph{per-user} (that is, only from a given user's objects and its associated context/target videos).
This contrasts existing few-shot (and other) benchmarks, and offers powerful insights into how well a meta-trained \gls{TOR} can personalize to a single user.
\vspace{-1em}

\subsubsection{Train/validation/test users}

The user-centric formulation in~\cref{sec:few-shot-for-pors} calls for a disjoint set of train users \(\trainusers\) and test users \(\testusers\). 
We therefore separate the 67 ORBIT collectors into 44 train users and 17 test users, with the remaining 6 marked as validation users \(\mathcal{K}^\text{val}\).
To ensure the test case is sufficiently challenging, we enforce that test (and validation) users have a minimum of 5 objects (see further details in~\cref{app:sec:dataset-prep-set-creation}).
The total number of objects in the splits are 278/50/158, respectively.
We report statistics for each set of train/validation/test users in~\cref{app:sec:extended-benchmark-dataset-summary}, mirroring those over all users in~\cref{sec:orbit-benchmark-dataset}.
\vspace{-1em}

\subsubsection{Evaluation modes}
\label{sec:evaluation-modes}

We establish 2 evaluation modes:\\
\noindent\textbf{Clean video evaluation (\textbf{\textsc{cle-ve}}).}
We construct a test user's context set \(\contextset^\user\) from their clean videos, and target set \(\targetset^\user\) from a \textit{held-out} set of their clean videos.
This mode serves as a simple check that the user's clean videos can be used to recognize the user's objects in novel `simple' scenarios when the object is in isolation.

\noindent\textbf{Clutter video evaluation (\textbf{\textsc{clu-ve}}).}
We construct a test user's context set \(\contextset^\user\) from their clean videos, and target set \(\targetset^\user\) from their clutter videos.
This mode matches the real-world usage of a \gls{TOR} where a user captures clean videos to register objects, and needs to identify those objects in complex, cluttered environments.
We consider {\sc clu-ve} to be ORBIT's primary evaluation mode since it most closely matches how a \gls{TOR} will be used in the real-world.
\vspace{-1em}

\subsubsection{Evaluation metrics}
\label{sec:eval-metrics}

\begin{table*}[t]
    \vspace{-1em}
    \centering
    \setlength{\tabcolsep}{4pt}
    \hspace{-1em}
    \scalebox{0.9}{
    \begin{tabular}{P{0.28\textwidth}P{0.28\textwidth}P{0.23\textwidth}P{0.25\textwidth}}
       \cmidrule{1-4}
        \textbf{\textsc{frame accuracy} (\(\uparrow\))} & \textbf{\textsc{frames-to-recognition} (\(\downarrow\))} & \multicolumn{2}{c}{\textbf{\textsc{video accuracy} (\(\uparrow\))}} \\
        $\begin{array}{l} 
           \cfrac{1}{|\targetset^\text{all}|}\sum\limits_{(\cluttervideo, \cluttervideolabel) \in \targetset^\text{all}} \frac{ \sum\limits_{f=1}^{|\cluttervideo|} \mathbbm{1} [\clutterframeprediction_f\!=\!\cluttervideolabel]}{|\cluttervideo|}
        \end{array}$
        &
        $\begin{array}{c}
            \cfrac{1}{|\targetset^\text{all}|}\sum\limits_{(\cluttervideo, \cluttervideolabel) \in \targetset^\text{all}} 
            \frac{\underset{\cluttervideoframe_f \in \cluttervideo}{\argmin}\, \clutterframeprediction_f = \cluttervideolabel }{|\cluttervideo|} 
        \end{array}$
        & 
        $\begin{array}{c} 
            \cfrac{1}{|\targetset^\text{all}|}\sum\limits_{(\cluttervideo, \cluttervideolabel) \in \targetset^\text{all}} \mathbbm{1} \Big[ \cluttervideoprediction = \cluttervideolabel \Big]
        \end{array}$
        &
        {\small \(\cluttervideoprediction\!= \!\underset{p \in \objectset^\user}{\argmax}\! \sum\limits_{f=1}^{|\cluttervideo|} \mathbbm{1} [\clutterframeprediction_f = p]\)} \\
        \cmidrule{1-4}
    \end{tabular}}
    \vspace{-1em}
    \caption{ORBIT evaluation metrics. Symbols \(\uparrow\) / \(\downarrow\) indicate up / down is better, respectively. \(\targetset^\text{all}\) is the set of all target videos pooled across all tasks for all test users in \(\testusers\).}
    \label{tab:eval-metric-eqs}
    \vspace{0em}
\end{table*}

For a test user \(\user \in \testusers\), we evaluate their personalized recognizer \(f_{\theta^\user}\) on each of their target videos.
We denote a target video of object \(p \in \objectset^\user\) as \(\cluttervideo = [\cluttervideoframe_1, \dots, \cluttervideoframe_F]\), and its frame predictions as \(\clutterframepredictions = [\clutterframeprediction_1, \dots, \clutterframeprediction_F]\), where \(F\) is the number of frames and \(\clutterframeprediction_f \in \objectset^\user\).
We further denote \(\cluttervideoprediction\) as the video's most frequent frame prediction.
For a given target video, we compute its:
\begin{description}[itemsep=-0.4em]
    \item [Frame accuracy:] the number of correct frame predictions, by the total number of frames in the video.
    \item [Frames-to-recognition (\textbf{\textsc{ftr}}):] the number of frames (w.r.t. the first frame \(\cluttervideoframe_1\)) before a correct prediction is made, by the total number of frames in the video.
    \item [Video accuracy:] 1, if the video-level prediction equals the video-level object label, \(\cluttervideoprediction = \cluttervideolabel\), otherwise 0. 
\end{description}
\vspace{-0.5em}
\noindent We compute these metrics for each target video in all tasks for all users in \(\testusers\).
We report the average and 95\% confidence interval of each metric over this flattened set of videos, denoted \(\targetset^\text{all}\) (see equations in~\cref{tab:eval-metric-eqs}).
We also compute a further 2 computational cost metrics:
\vspace{-0.5em}
\begin{description}[itemsep=-0.4em]
    \item [\textsc{macs} to personalize:] number of Multiply-Accumulate operations (\textsc{macs})  to compute a test user's personalized parameters \(\theta^\user\) using their context videos \(\contextset^\user\), reported as the average over all tasks pooled across test users.
    \item [Number of parameters:] total parameters in recognizer.  
\end{description}
\vspace{-0.5em}
We flag frame accuracy as ORBIT's primary metric because it most closely matches how a \gls{TOR} will ultimately be used.
The remaining metrics are complementary: \textsc{ftr} captures how long a user would have to point their recognizer at a scene before it identified the target object (with fewer frames being better) while video accuracy summarizes the predictions over a whole video.
\textsc{macs} to personalize provides an indication whether personalization could happen directly on a user's device or a cloud-based service is required, each impacting how quickly a recognizer could be personalized.
The number of parameters indicates the storage and memory requirements of the model on a device, and if cloud-based, the bandwidth required to download the personalized model. 
It is also useful to normalize performance by model capacity.

\section{Experimental analyses and results}
\label{sec:experiments}

\subsection{Baselines \& training set-up}

\paragraph{Baselines.}
\label{sec:baselines}

There are 3 main classes of few-shot learning approaches.
In \emph{metric-based} approaches, a per-class embedding is computed using the (labeled) examples in the context set, and a target example is classified based on its distance to each~\cite{snell2017prototypical,vinyals2016matching}.
In \emph{optimization-based} approaches, the model takes many~\cite{yosinski2014transferable,tian2020rethinking,chen2020new} or few \cite{finn2017model,zintgraf2018cavia,bertinettometa} gradient steps on the context examples, and the updated model then classifies the target examples.
%
%
Finally, in \emph{amortization-based} approaches, the model uses the context examples to directly generate the parameters of the classifier 
which is then used to classify a target example~\cite{requeima2019fast,gordon2018metalearning}.

We establish baselines on the ORBIT dataset across these 3 classes.
Within the episodic approaches, we choose Prototypical Nets~\cite{snell2017prototypical} for the metric family, MAML~\cite{finn2017model} for the optimization family, and CNAPs~\cite{requeima2019fast} for the amortization family.
We also implement a non-episodic fine-tuning baseline following~\cite{tian2020rethinking,chen2020new} who show that it can rival more complex methods.
This selection of models offers good coverage over those that are competitive on current few-shot learning image classification benchmarks.
For all implementation details of these baselines see~\cref{app:sec:baselines}.

\vspace{-1em}
\paragraph{Video representation.}
\label{sec:sampling-video-frames}

In~\cref{sec:few-shot-for-pors}, tasks are constructed from the context and target videos of a given user's objects.
We sample clips from each video and represent each clip as an average over its (learned) frame-level features.
For memory reasons, we do not sample all clips from a video. 
Instead, during meta-training, we randomly sample \(S^\text{train}\) non-overlapping clips, each of \(L\) contiguous frames, from both context and target videos.
Each clip is averaged and treated as an `element' in the context/target set, akin to an image in typical few-shot image classification.
During meta-testing, however, following~\cref{sec:eval-protocol} and~\cref{eq:frame-predict}, we must evaluate a test user's personalized recognizer on \emph{every} frame in \emph{all} of their target videos.
We, therefore, sample \emph{all} overlapping clips in a target video, where a clip is an \(L\)-sized buffer of each frame plus its short history.
Ideally, this should also be done for context videos, however, due to memory reasons, we sample \(S^\text{test}\) non-overlapping \(L\)-sized clips from each context video, similar to meta-training.
In our baseline implementations, \(S^\text{train}=4\), \(S^\text{test}=8\), and \(L=8\) (for further details see~\cref{app:sec:task-sampling,app:sec:task-hyperparameters}).

How frames are sampled during training/testing, and how videos are represented is flexible.
The evaluation protocol's only strict requirement is that a model outputs a prediction for every frame from every target video for every test user.

\vspace{-1em}
\paragraph{Number of tasks per test user.}

Because context videos are sub-sampled during meta-testing, a test user's task contains a random set, rather than \emph{all}, context clips.
To account for potential variation, therefore, we sample 5 tasks per test user, and pool all their target videos into \(\targetset^\text{all}\) for evaluation.
If memory was not a constraint, following~\cref{sec:few-shot-for-pors}, we would sample one task per test user which contained \emph{all} context and \emph{all} target clips.

\begin{table*}[t]
    \setlength\tabcolsep{3pt}
    \centering
    \scalebox{0.7}{
    \begin{tabular}{l|P{1.9cm}P{1.8cm}P{1.8cm}P{2.4cm}|P{1.9cm}P{1.8cm}P{1.9cm}P{2.4cm}|P{2.4cm}P{1.6cm}}
    & \multicolumn{4}{c}{\large \textbf{Clean Video Evaluation (\textsc{cle-ve})}} & \multicolumn{4}{c}{\large \textbf{Clutter Video Evaluation (\textsc{clu-ve})}} \\
    \cmidrule{2-9}
    \textsc{\textbf{model}} & \textsc{\textbf{frame acc}} & \textsc{\textbf{ftr}} & \textsc{\textbf{video acc}} & \textsc{\textbf{macs to personalize}} & \textsc{\textbf{frame acc}} & \textsc{\textbf{ftr}} & \textsc{\textbf{video acc}} & \textsc{\textbf{macs to personalize}} & \textsc{\textbf{method to personalize}} & \textsc{\textbf{\# params}}\\
    \cmidrule{1-11}
    \textbf{Best possible} & - & - & - & - & \textbf{95.31 (1.37)} & \textbf{0.00 (0.00)} & \textbf{100.00 (0.00)} & - & - & - \\
    \cmidrule{1-11}
    ProtoNets~\cite{snell2017prototypical} & 65.16 (1.96) & \textbf{7.55 (1.35)} & \textbf{81.88 (2.51)} & \textbf{2.82 \(\times\) 10\textsuperscript{12}} & \textbf{50.34 (1.74)} & \textbf{14.93 (1.52)} & \textbf{59.93 (2.48)} & \textbf{3.53 \(\times\) 10\textsuperscript{12}} & \textbf{1 forward pass} & \textbf{11.17M} \\
    CNAPs~\cite{requeima2019fast} & 66.15 (2.08) & \textbf{8.40 (1.40)} & \textbf{79.56 (2.63)} &  3.09 \(\times\) 10\textsuperscript{12} & \textbf{51.47 (1.81)} & 17.87 (1.69) & \textbf{59.53 (2.48)} & 3.87 \(\times\) 10\textsuperscript{12} & \textbf{1 forward pass} & 12.75M \\
    MAML~\cite{finn2017model} & \textbf{70.58 (2.10)} & \textbf{8.62 (1.56)} & \textbf{80.88 (2.56)} & 84.63 \(\times\) 10\textsuperscript{12} & \textbf{51.67 (1.88)} & 20.95 (1.84) & 57.87 (2.50) & 105.99 \(\times\) 10\textsuperscript{12} & 15 gradient steps & \textbf{11.17M} \\
    FineTuner~\cite{tian2020rethinking} & \textbf{69.47 (2.16)} & \textbf{7.82 (1.54)} & \textbf{79.67 (2.62)} & 282.09 \(\times\) 10\textsuperscript{12} & \textbf{53.73 (1.80)} & \textbf{14.44 (1.50)} & \textbf{63.07 (2.44)} & 353.30 \(\times\) 10\textsuperscript{12} & 50 gradient steps & \textbf{11.17M} \\
    \end{tabular}}
    \vspace{-0.5em}
    \caption{Baselines on the ORBIT Dataset. Results are reported as the average (95\% confidence interval) over all target videos pooled from 85 test tasks (5 tasks per test user, 17 test users). Best possible scores are computed using bounding box annotations which are available for the clutter videos (see~\cref{app:sec:extended-benchmark-dataset-summary,app:fig:bounding-box-summary-allsets}).
    }
    \label{tab:orbit-baselines}
    \vspace{-1em}
\end{table*}

\subsection{Analyses} \label{sec:evaluation:analysis}

\noindent\textbf{Baseline comparison.}
Performance is largely consistent across the baseline models in both \textsc{cle-ve} and \textsc{clu-ve} modes (see~\cref{tab:orbit-baselines}).
In \textsc{cle-ve}, all methods are equivalent in frame accuracy, \textsc{ftr} and video accuracy, except for ProtoNets and CNAPs which trail slightly in frame accuracy.
Comparing this to \textsc{clu-ve}, we see overall performance drops of 10-15 percentage points.
Here, models are overall equivalent on frame and video accuracy, however ProtoNets and FineTuner lead in \textsc{ftr}.
Further, absolute \textsc{clu-ve} scores are in the low 50s.
Looking at the best possible bounds (computed using the bounding box annotations, see~\cref{app:fig:bounding-box-summary-test}) suggests that there is ample scope for improvement and motivates the need for approaches that can handle distribution shifts from clean (context) to real-world, cluttered scenes (target), and are robust to high-variation data more generally.

In computational cost, ProtoNets has the lowest cost to personalize requiring only a single forward pass of a user's context videos, while FineTuner has the highest, requiring 50 gradient steps.
This, along with the total number of parameters (which are similar across models), suggests that ProtoNets and CNAPs would be better suited to deployment on a mobile device. 

\noindent\textbf{Meta-training on other few-shot learning datasets.}
A meta-trained model should, in principle, have the ability to learn \emph{any} new object (from any dataset) with only a few examples.
We investigate this by meta-training the baseline models on Meta-Dataset~\cite{triantafillou2019meta} using its standard task sampling protocol and then testing them on the ORBIT dataset (i.e. personalizing to test users with no training).
We adapt the meta-trained models to videos by taking the average over frame features in clips sampled from context and target videos (see~\cref{sec:sampling-video-frames}).
In~\cref{tab:metadataset-test}, we see that even on the easier, clean videos (\textsc{cle-ve}), performance is notably lower than the corresponding baselines in~\cref{tab:orbit-baselines} (for \textsc{clu-ve} see~\cref{app:tab:metadataset-test}).
MAML and CNAPs perform particularly poorly while ProtoNets and FineTuner fare slightly better, however, are still 6-8 percentage points below their above counterparts in frame accuracy.
This suggests that even though much progress has been made on existing few-shot benchmarks, they are not representative of real-world conditions and models trained on them may struggle to learn new objects when only high-variation examples are available.

\begin{table}[t]
    \centering
        \scalebox{0.85}{
        \begin{tabular}{l|ccc}
        \textsc{\textbf{model}} & \textsc{\textbf{frame acc}} & \textsc{\textbf{ftr}} & \textsc{\textbf{video acc}} \\
         \cmidrule{1-4}
         ProtoNets~\cite{snell2017prototypical} & \textbf{58.98 (2.23)} & \textbf{11.55 (1.79)} & \textbf{69.17 (3.01)} \\ 
         CNAPs~\cite{requeima2019fast} & 51.86 (2.49) & 20.81 (2.33) & 60.77 (3.18)\\
         MAML~\cite{finn2017model} & 42.55 (2.67) & 37.28 (2.99) & 46.96 (3.25) \\ 
         FineTuner~\cite{tian2020rethinking} & \textbf{61.01 (2.24)} & \textbf{11.53 (1.82)} & \textbf{72.60 (2.91)} \\ 
        \end{tabular}
        }
    \caption{
    \textsc{cle-ve} performance when meta-training on Meta-Dataset and meta-testing on ORBIT (for \textsc{clu-ve} see~\cref{app:tab:metadataset-test}). Even on clean videos, models perform poorly compared to when meta-training on ORBIT (\cref{tab:orbit-baselines}) suggesting that existing few-shot datasets may be insufficient for real-world adaptation. 
    }
    \label{tab:metadataset-test}
    \vspace{-1.5em}
\end{table}

\begin{figure}[t]
    \includegraphics[width=0.47\textwidth]{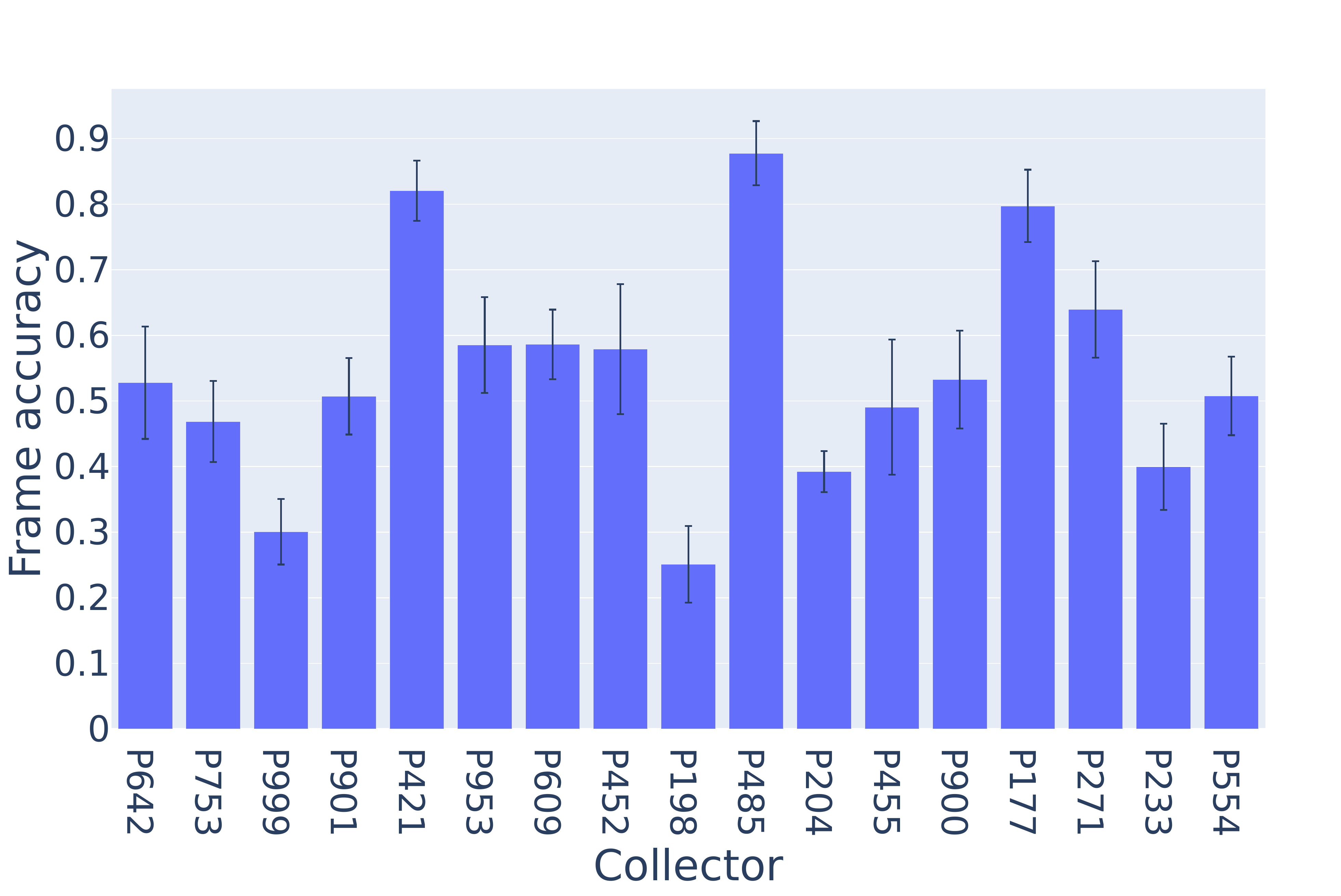}
    \caption{\textsc{clu-ve} frame accuracy varies widely across test users (error bars are 95\% confidence intervals) with ProtoNets~\cite{snell2017prototypical}. For other metrics and models see~\cref{fig:orbit-baselines-per-user-all}.}
    \label{fig:orbit-baselines-per-user}
    \vspace{-1.5em}
\end{figure}

\noindent\textbf{Per-user performance.}
In addition to averaging over \(\targetset^\text{all}\), the benchmark's user-centric paradigm allows us to average \emph{per-user} (i.e.\ over just their target videos).
This is useful because it provides a measure of how well a meta-trained \gls{TOR} would personalize to an individual real-world user.
In~\cref{fig:orbit-baselines-per-user} however, we show that ProtoNets' personalization is not consistent across users, for some going as low as 25\% in frame accuracy (for other metrics/models see~\cref{fig:orbit-baselines-per-user-all}).
A \gls{TOR} should be able adapt to \emph{any} real-world user, thus future work should not only aim to boost performance on the metrics but also reduce variance across test users.

\begin{figure}
    \centering
    \includegraphics[width=0.47\textwidth]{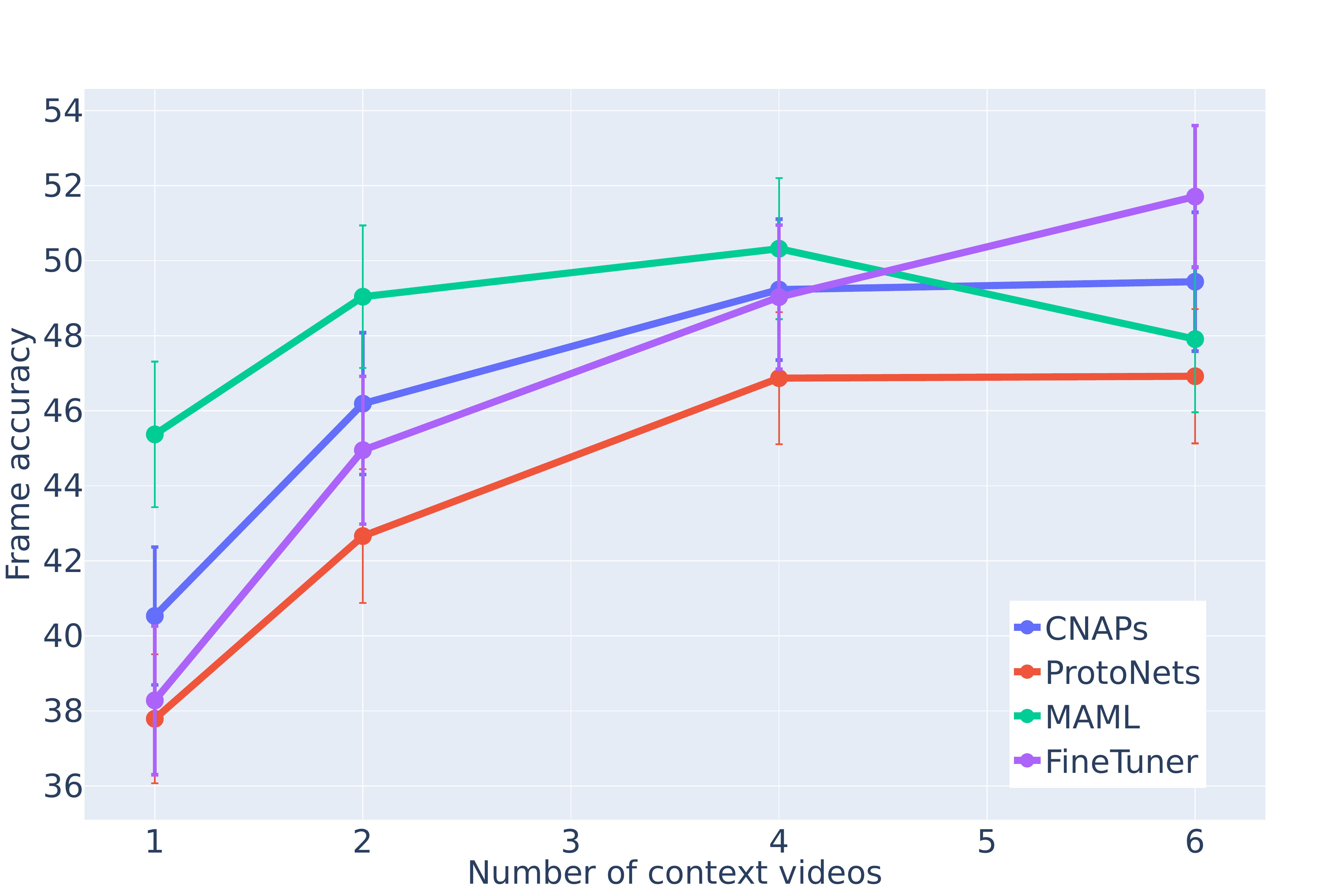}
    \caption{Meta-training with more context videos per object leads to better \textsc{clu-ve} performance. Frames are sampled from an increasing number of clean videos per object using the number of clips per video (\(S^\text{train}\)) to keep the total number of context frames fixed per train task. }
    \label{fig:num-context-vids-analysis}
    \vspace{-1.5em}
\end{figure}

\begin{figure}
    \centering
    \includegraphics[width=0.47\textwidth]{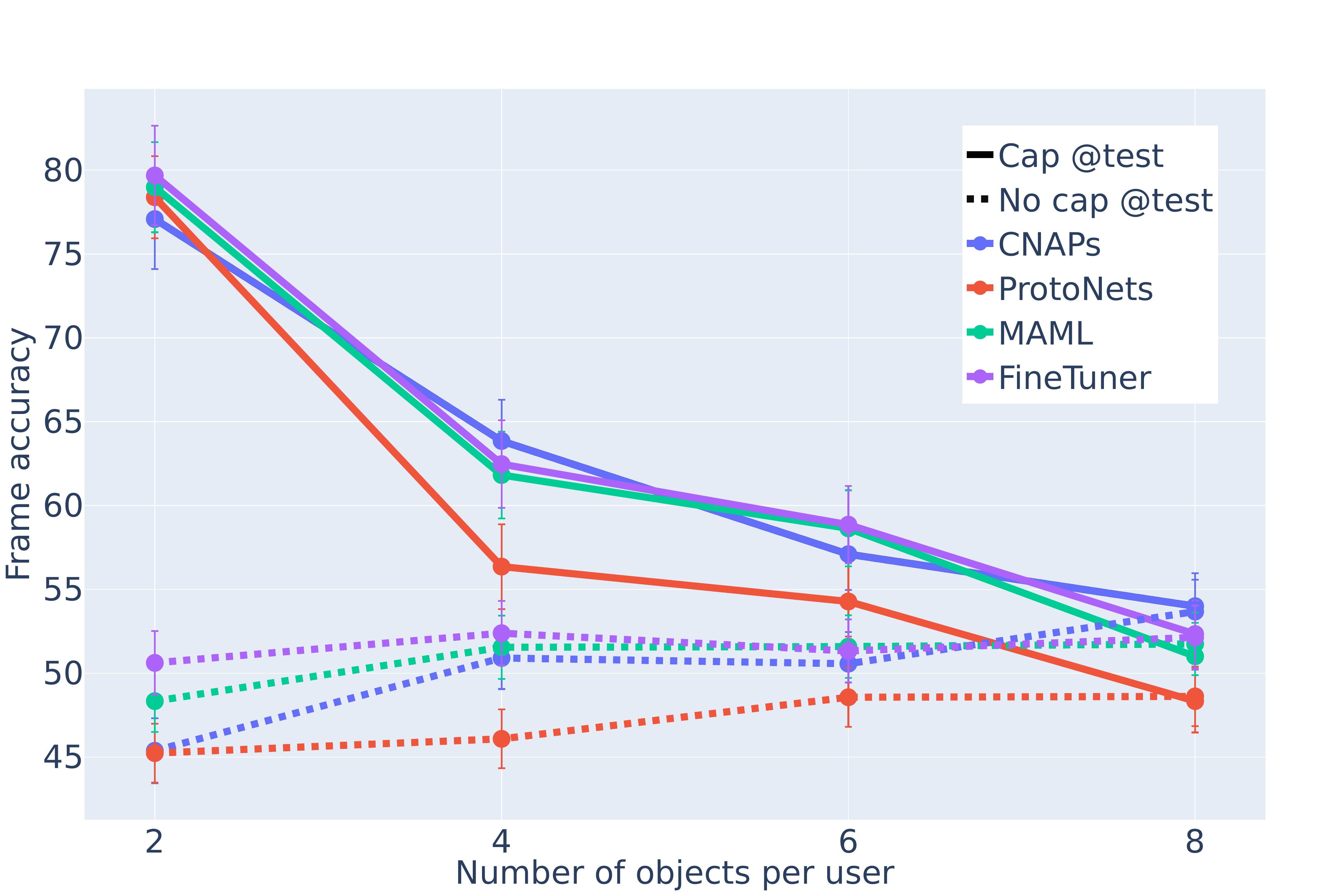}
    \caption{Meta-training and -testing with more objects per user poses a harder recognition problem (solid line), however, meta-training with fewer objects than encountered at meta-testing (dashed line) shows only a small \textsc{clu-ve} performance drop compared to~\cref{tab:orbit-baselines}, suggesting that models may be able to adapt to more objects in the real-world.}
    \label{fig:num-objects-analysis}
    \vspace{-1.5em}
\end{figure}

\noindent\textbf{Train task composition.}
Finally, we investigate the impact of the number of context videos per object (\cref{fig:num-context-vids-analysis}), and the number of objects per user (\cref{fig:num-objects-analysis}) sampled in train tasks on \textsc{clu-ve} frame accuracy.
In the first case, we expect that with more context videos per object, the more diversity the model will see during meta-training, and hence generalize better at meta-testing to novel (target) videos.
To test this hypothesis, we fix a quota of \(96\) frames per object in each train task and sample these frames from increasing numbers of context videos.
Frame accuracy increases with more context videos, but overall plateaus between 4-6 context videos per object.
Looking at the number of objects sampled per user next, we cap all train user's objects at \(\{2, 4, 6, 8\}\), respectively, when meta-training.
We then meta-test in two ways:
\begin{inparaenum}[1)]
\item we keep the caps in place on the test users, and
\item we remove the caps.
\end{inparaenum}
For 1), we see reducing accuracy for increasing numbers of objects, as is expected -- classifying between 8 objects is harder than classifying between 2.
For 2), we see a significant drop in accuracy relative to 1) suggesting that meta-training with fewer objects than would be encountered at meta-testing is detrimental.
This is an important real-world consideration since it is likely that over months/years, a user will accumulate many more objects than is currently present per user in the ORBIT dataset.
Overall, however, training with a cap of 6 or more objects yields roughly equivalent performance to that reported in~\cref{tab:orbit-baselines} where no caps are imposed during training.
Since ORBIT test users have up to 12 objects (see~\cref{fig:benchmark-num-objects-test}), our results suggest that a minimum of half the number of ultimate objects for a test user may be sufficient for meta-training.
We repeat these analyses for the other metrics in~\cref{app:fig:num-context-vids-analysis,app:fig:num-objects-analysis}, and include the corresponding tables in~\cref{tab:num-context-vids-analysis,tab:num-objects-analysis}.
We also investigate the impact of the number of tasks sampled per train user, included in \cref{app:sec:extended-analysis}.

\section{Discussion} \label{sec:discussion}

We present the ORBIT dataset and benchmark, both grounded in the few-shot application of \glspl{TOR} for people who are blind/low-vision.
Our baseline performance and further analyses demonstrate, however, that current few-shot approaches struggle on realistic, high-variation data.
This gap offers opportunities for new and exciting research, from making models robust to high-variation video data to quantifying the uncertainty in model predictions.
More than just pushing the state-of-the-art in existing lines of thought, the ORBIT dataset opens up new types of challenges that derive from systems that will support human-AI partnership.
We close by discussing three of these unique characteristics.

ORBIT's user-centric formulation provides an opportunity to measure how well the ultimate system will work in the hands of real-world users.
This contrasts most few-shot (and other) benchmarks which retain no notion of the end-user.
Our results show that the baselines do not perform consistently across users.
%
%
In the real-world, the heterogeneity of users, their objects, videoing techniques and devices will make this even more challenging.
It will therefore be important for models to quantify, explain and ultimately minimize variation across users, particularly as models are deployed in a wider variety of scenarios outside the high-income countries in which the dataset was collected.

Directly involving users in collecting a dataset intended to drive \gls{ML} research comes with challenges: user-based datasets are harder to scale than web-scraped datasets~\cite{deng2009imagenet,Lin14COCO,triantafillou2019meta} and users need an understanding of the potential system in order to contribute useful data.
Building the system first would address these challenges, but it cannot be done without algorithmic innovation (which itself requires the dataset).
The ORBIT dataset is a starting point and can be used to build the first generation of \glspl{TOR}, which can be deployed and themselves be used to collect more real-world data to drive a cycle of innovation between dataset and application. 

Finally, grounding in a real-world application encourages innovation in new directions to meet the real-world conditions of deployment.
This could range from new models that are lightweight enough to be personalized directly on a user’s phone to new research problems like handling the scenario when none of a user’s objects are in the frame.

In conclusion, the ORBIT dataset and benchmark aims to shape the next generation of recognition tools for the blind/low-vision community starting with \glspl{TOR}, and to improve the robustness of vision systems across a broad range of other applications.

\subsection*{Acknowledgments}
\noindent The ORBIT Dataset is funded by Microsoft AI for Accessibility.
LZ is supported by the 2017 MSR PhD Scholarship Program and 2020 MSR EMEA PhD Award. JB is supported by the EPSRC Prosperity Partnership EP/T005386/1.
We thank VICTA, RNC, RNIB, CNIB, Humanware, Tekvision School for the Blind, BlindSA, NFB, and AbilityNet.
Finally, we thank Emily Madsen for help with the video validation, and all the ORBIT collectors for their time and contributions.

{\small
\bibliographystyle{ieee_fullname}
\bibliography{references}
}

\clearpage
\appendix
\setcounter{figure}{0}
\setcounter{table}{0}
\setcounter{equation}{0}
\renewcommand{\thefigure}{A.\arabic{figure}}
\renewcommand{\thetable}{A.\arabic{table}}

\section{Unfiltered ORBIT dataset}

\subsection{Dataset collection protocol}
\label{app:sec:data-collection-protocol}

The unfiltered ORBIT dataset was collected in two phases, the first with blind/low-vision collectors based only in the UK, and the second with blind/low-vision collectors globally~\cite{theodorou2021disability}. 
Collectors were recruited via blind charities and networks, and were screened for blindness/low-vision before beginning the collection task.
Following this, they were sent instructions for collecting the videos using an accessible iOS app (see~\cref{app:sec:orbit-camera-app}).

\subsubsection{Phase 1 protocol}

Collectors were asked to record videos for at least 10 objects, including at least 2 large or immovable objects.
For each of these objects, they were asked to record two types of videos, train and test which corresponded to clean and clutter, respectively.
Collectors were asked to adopt specific video recording techniques to help capture their objects in-frame. 

For clean videos, the collector was asked to place the object on a clear surface with no other objects present and record a video of the object, repeating this on at least 6 different surfaces.
Collectors were given step-by-step instructions to record these using a zoom-out technique: 
\begin{compactenum}
\item Keep one hand on the surface next to the object as an anchor point to help aim the camera at the object
\item Hold the phone in your other hand and bring it as close as possible to the object
\item Start recording, then slowly draw the phone away from the object until it reaches your body at shoulder height
\item Rotate the object so that a different side is facing you, while returning the phone close to your anchor hand
\item Repeat this at least 3 times for different sides of the object, then stop recording
\end{compactenum}
We helped collectors time each rotation by playing a tick sound every 5 seconds, and asking them to record a new side of the object between ticks.
To prevent inadvertently long videos, we automatically ended recordings after 2 minutes.

For large or immovable objects, collectors were asked to position themselves relative to the object so that they were recording different aspects or angles of the object.
Again, they were asked to place an anchor hand on three different parts or positions of the object, and then draw their phone slowly towards their body.
We asked them to record the object from 3 different angles, repeated this twice, to provide a total of 6 clean videos. 

For clutter videos, collectors were asked to construct a cluttered scene based on a real-life situation in which the recognizer might be used --- for example, a surface on which someone else may have placed an object.
The scene needed to include at least 5 other distractor objects that did not include any of the collector's selected objects.
For immovable objects like a front door, we asked them to position some likely objects around or in front of it, such as packages, umbrellas, or shoes. 
Collectors were asked to record their clutter videos using 2 different techniques: a zoom-out and a panning technique, recording at least one video with each technique for each of their objects.
For the zoom-out technique, similar to the clean videos, they were asked to place an anchor hand in the scene, and then draw their phone slowly towards their body, before ending the recording.
For the panning technique, collectors were asked to place an anchor hand in the scene, making sure the selected object was not directly in front of them.
They were then asked to remove their anchor hand, and pan over the scene at shoulder height from right to left by turning their upper body.

\subsubsection{Phase 2 protocol}
In Phase 2, we reduced the number of objects to at least 5 per collector, with no specification for large or immovable objects.
We also reduced the number of videos to 5 clean (taken on 5 different surfaces) and 2 clutter videos.
For clean videos, collectors were asked to use the zoom-out technique, showing at least four sides of the object. 
For clutter videos, instead of creating a real-world scene as in Phase 1, collectors were asked to record the object, including its surroundings, in a scene where it would normally be found.
For simplicity, the panning technique was dropped, and collectors were asked to record both clutter videos using only the zoom-out technique.
For the second clutter video, they were simply asked to record it from a different position.
Finally, we reduced the cut-off video time to 30 and 20 seconds for clean and clutter videos, respectively.

\subsection{ORBIT Camera iOS app}
\label{app:sec:orbit-camera-app}

\begin{figure}[ht!]
  \centering
  \includegraphics[width=0.48\textwidth]{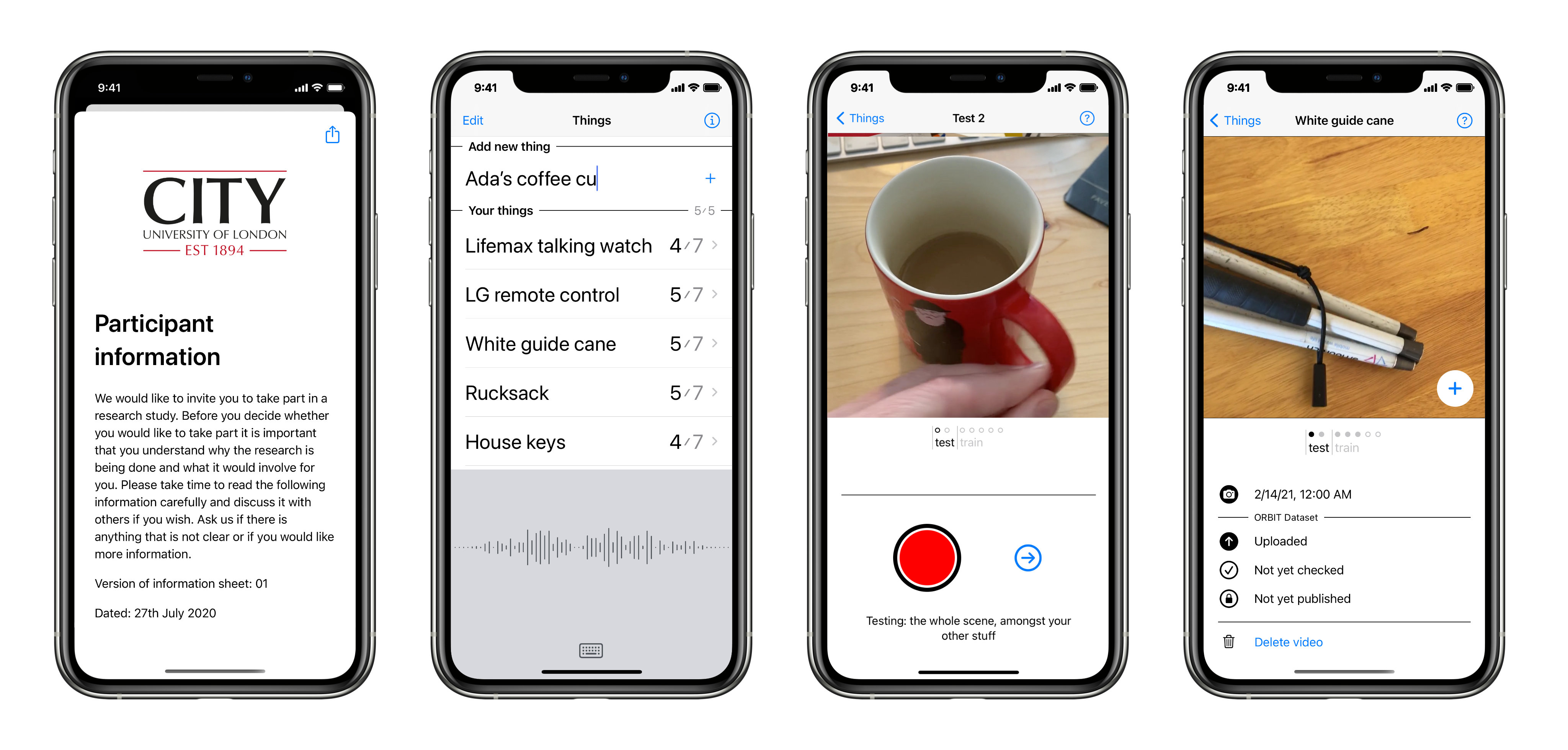}
  \caption{ORBIT Camera iOS app: Informed consent screen (left). `Things' screen showing a collector's personal objects (center-left). `Thing' screen for a specific object (center-right). `Thing' screen after recording a video (right).}
  \label{fig:orbit-camera-app-screeshots}
  \vspace{-1em}
\end{figure}

The dataset was collected using a custom-built iOS app (see~\cref{fig:orbit-camera-app-screeshots}) that followed Apple's accessibility guidelines and was tested with blind/low-vision users.
On first use, the collector was asked to read an accessible participant information and consent form, and then provide their explicit consent.
Once provided, collectors could access the app's home screen and start to add objects to their library.
This was done by entering an object’s name into a text box at the top of the home screen.
This name then became the label for all that object's videos. 
The collector could then navigate to a second screen to 
\begin{inparaenum}[i)]
\item record new videos, and 
\item review videos already recorded 
\end{inparaenum}
for that object.
Here, the collector could also delete and re-record videos already stored.
Status indicators were available for each recorded video, including whether it was uploaded to the data server, validated, and visible in the public dataset.

All contributed videos were uploaded to a data server which provided administrative functionality.
Administrators were able to set-up data collection periods, review videos uploaded by collectors, and mark them as reviewed and safe for export.
All videos were manually checked to ensure
\begin{inparaenum}[i)]
\item the labeled object was present in the video,
\item the video did not contain any \gls{PII}, and 
\item the object as well as object name were appropriate.
\end{inparaenum}
If a video did not meet all of these criteria, it was removed from the server and the collector received a notification via the app to re-record the video.
All \gls{PII} obtained via the consent screen was encrypted, and decryption keys were held by administrators outside the server.

Code for the app and back-end data server is open-sourced at \href{https://github.com/orbit-a11y}{https://github.com/orbit-a11y}.

\subsection{Unfiltered dataset summary}
\label{app:orbit-full-summary}

The unfiltered ORBIT dataset contains 4733 videos (3,161,718 frames, 97GB) of 588 objects, collected by 97 people who are blind/low-vision~\cite{orbitdataset2021}.
We summarize it in~\cref{tab:full-dataset-summary} and describe it in more detail below.

\vspace{-1em}
\paragraph{Number of collectors}
97 blind/low-vision collectors contributed to the unfiltered ORBIT dataset.
\begin{table*}[ht!]
    \centering
    \scalebox{0.8}{
    \begin{tabular}{l|ccccccccc}
    & \textbf{Collectors} &  \textbf{Objects} & \textbf{Videos} & \multicolumn{3}{c}{\textbf{Videos per object}} & \multicolumn{3}{c}{\textbf{Frames per video}}\\
    & & &  & mean/std & 25/75\(^\text{th}\) perc. & min/max & mean/std & 25/75\(^\text{th}\) perc. & min/max \\
    \textbf{Total} & 97 & 588 & 4733 & 8.0/5.5 & 7.0/8.0 & 1.0/49.0 & 668.0/411.1 & 355.0/898.0 & 2.0/3600.0 \\
    \cmidrule{1-10}
    \textbf{Clean} & & 576 & 3356 & 5.8/4.4 & 5.0/6.0 & 1.0/44.0 & 763.0/416.7 & 504.0/900.0 & 2.0/3600.0 \\
    \textbf{Clutter-zoom-out} & & 519 & 873 & 1.7/1.4 & 1.0/2.0 & 1.0/14.0 & 452.3/270.7 & 241.0/598.0 & 3.0/3596.0\\
    \textbf{Clutter-pan} & & 330 & 504 & 1.5/1.6 & 1.0/1.0 & 1.0/12.0 & 408.9/310.4 & 186.0/521.0 & 8.0/2684.0\\
    \cmidrule{1-10}
    \textbf{Per-collector} & 1 & 6.1/3.8 & 48.8/53.7 & 6.7/4.5 & 5.0/7.7 & 1.0/41.5 & 667.2/222.2 & 550.6/808.4 & 73.0/1467.2\\
    \cmidrule{1-10}
    \textbf{Clean} & & 6.1/3.8 & 35.1/41.2 & 5.0/3.6 & 4.0/6.0 & 1.0/36.6 & 774.9/287.1 & 648.9/898.7 & 134.5/1872.6\\
    \textbf{Clutter-zoom-out} & & 6.0/3.8 & 10.3/10.6 & 1.8/1.4 & 1.0/2.0 & 1.0/12.0 & 465.1/191.8 & 329.2/597.1 & 73.0/1258.8\\ 
    \textbf{Clutter-pan} & & 6.1/3.8 & 9.8/11.4 & 1.6/1.6 & 1.0/1.4 & 1.0/8.6 & 419.5/231.4 & 239.0/574.3 & 164.5/1208.4\\
    \end{tabular}}
    \caption{Unfiltered ORBIT dataset.}
    \label{tab:full-dataset-summary}
\end{table*}

\begin{figure*}[ht!]
     \centering
     \begin{subfigure}[t]{\textwidth}
         \centering
         \includegraphics[width=\textwidth]{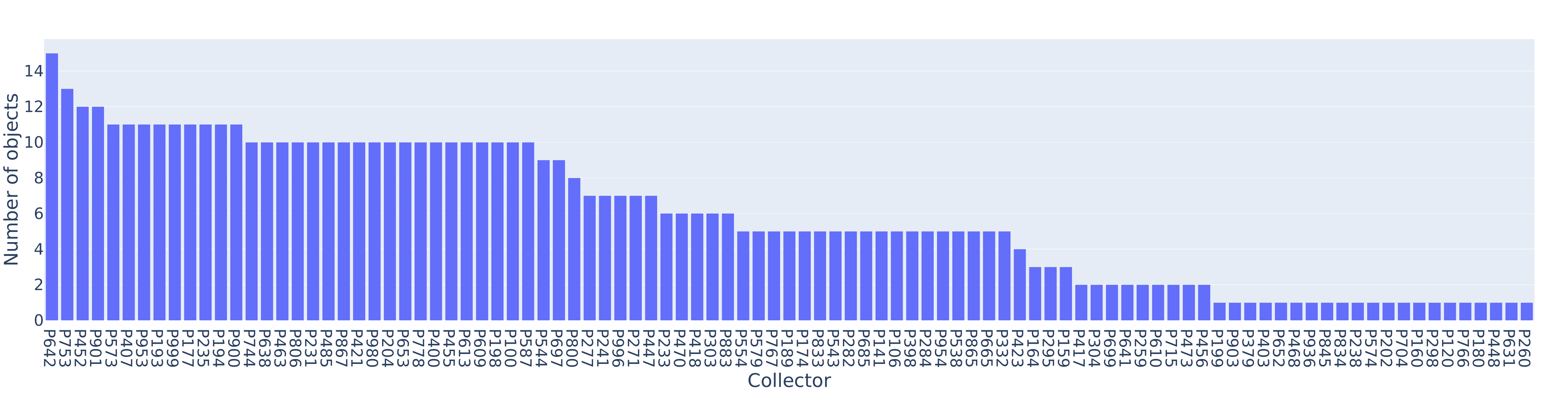}
         \caption{Number of objects per collector.}
         \label{fig:num-objects-all-full}
         \vspace{1em}
     \end{subfigure}
     \begin{subfigure}[t]{\textwidth} 
         \centering
         \includegraphics[width=\textwidth]{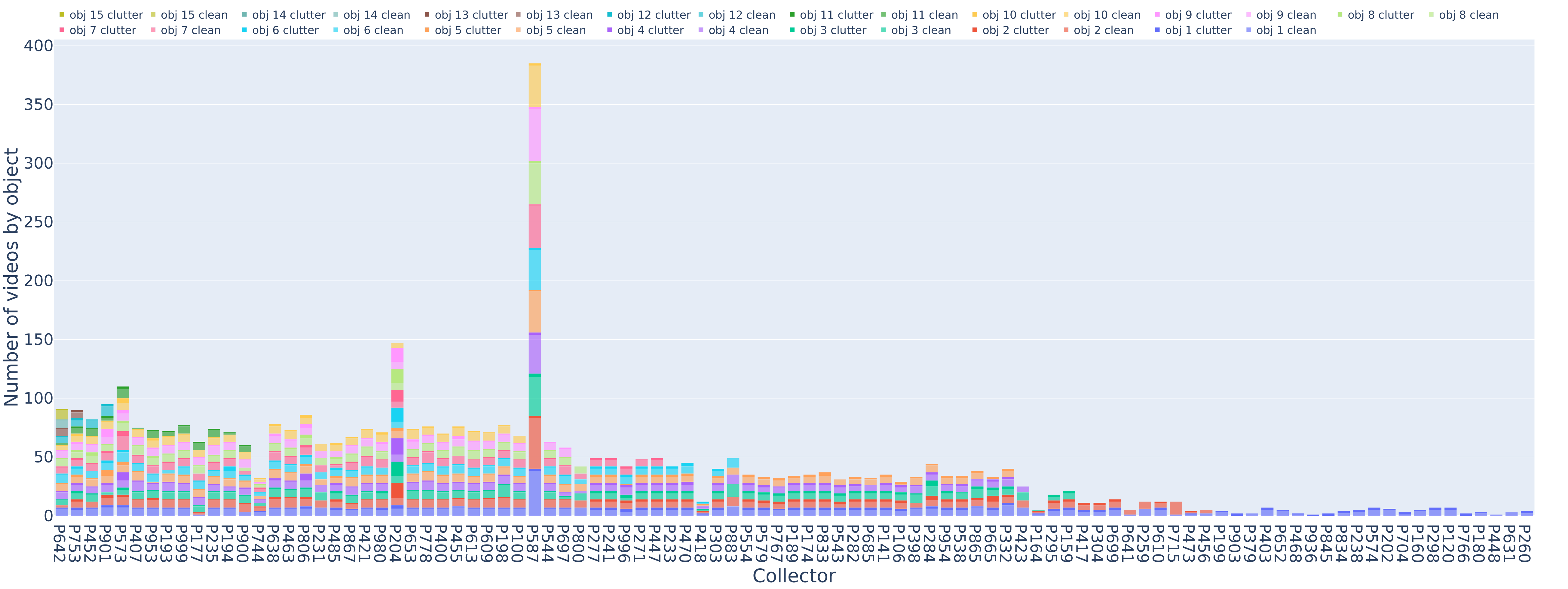}
         \caption{Number of videos (stacked by object) per collector.}
         \label{fig:num-vids-by-object-all-full}
     \end{subfigure}
     \caption{Number of objects and videos across 97 collectors in the unfiltered ORBIT dataset.}
     \label{fig:full-dataset-summary}
     \vspace{1em}
\end{figure*}

\vspace{-1em}
\paragraph{Numbers of videos and objects}
Collectors contributed a total of 588 objects and 4,733 videos.
%
3,356 of these videos showed the object in isolation, referred to as \emph{clean} videos, while 1,377 showed the object in realistic, multi-object scenes, referred to as \emph{clutter} videos.
Of the clutter videos, 873 were recorded with the zoom-out technique, and 504 with the panning technique (see video techniques in~\cref{app:sec:data-collection-protocol}).
Each collector contributed on average 6.1 (\(\pm 3.8\)) objects, with 34.6 (\(\pm 41.5\)) clean videos, 9.0 (\(\pm 10.5\)) clutter-zoom-out videos, and 5.2 (\(\pm 10.5\)) clutter-pan videos per object.
\cref{fig:full-dataset-summary} shows the number of objects (\ref{fig:num-objects-all-full}) and number of videos per collector (\ref{fig:num-vids-by-object-all-full}).

\vspace{-1em}
\paragraph{Types of objects}
For summarization purposes, we clustered objects based on object similarity and show their long-tailed distribution in~\cref{fig:full-cluster-histo-all}.
The clustering algorithm and full contents of each cluster are included in~\cref{app:sec:object-clustering}.

\section{ORBIT benchmark dataset preparation}
\label{app:sec:dataset-prep}

The ORBIT benchmark dataset is a subset of the unfiltered ORBIT dataset and was selected to meet the requirements of the ORBIT benchmark (\cref{sec:orbit-benchmark}).
The following sections detail the steps to prepare the benchmark dataset.

\subsection{File structure}
\label{app:sec:file-structure}

The following summarizes the file hierarchy of the ORBIT benchmark dataset:
\begin{compactitem}[>]
    \item \texttt{mode (train/validation/test)}
    \begin{compactitem}[>]
        \item \texttt{user} \(\user\)
        \begin{compactenum}[>]
            \item \texttt{object} \(p \in \objectset^\user\)
            \begin{compactitem}[>]
                \item \texttt{clean} 
                \begin{compactenum}[>]
                    \item \texttt{video} \(\cleanvideo_i \in \allcleanvideos^\user_p\)
                    \begin{compactitem}
                        \item \texttt{frame} \(v_1\)
                        \item \texttt{...}
                        \item \texttt{frame} \(v_{F_i}\)
                    \end{compactitem}
                \end{compactenum}
                \item \texttt{clutter}
                \begin{compactenum}[>]
                    \item \texttt{video} \(\cluttervideo_i \in \allcluttervideos^\user_p\)
                \end{compactenum}
            \end{compactitem}
        \end{compactenum}
    \end{compactitem}
\end{compactitem}

\subsection{Video filtering}
\label{app:sec:dataset-prep-pre-proc}

From the 97 collectors who submitted 4,733 videos (3,356 clean, 873 clutter, 504 clutter-pan) in the unfiltered dataset, the videos of 77 collectors were selected for the benchmark dataset --- a total of 3,822 videos (2,996 clean, 826 clutter, 0 clutter-pan) of 486 objects (see \cref{tab:benchmark-dataset-summary,fig:benchmark-dataset-summary}).
The following steps detail the video selection procedure:

\begin{compactenum}
\item All clutter-pan videos were removed (504 videos, 4 objects, 0 collectors). This was done to maintain consistency because the panning technique was explored in the first, but not the second, phase of the dataset collection. Clutter-zoom-out videos are, therefore, referred to as simply clutter videos in the benchmark dataset.
\item Videos shorter than 1 second were removed (20 videos, 0 objects, 0 collectors). Extremely short videos were assumed to be a mistaken recording.
\item Objects with \(<2\) clean videos and \(<1\) clutter video were remove (387 videos, 101 objects, 20 collectors). These limits were the minimum numbers required for the benchmark evaluation. 
\end{compactenum}

\subsection{Train/validation/test sets}
\label{app:sec:dataset-prep-set-creation}

Of the 77 collectors remaining, 12 contributed only 1 object.
To avoid discarding these videos, we merged these collectors together (or with others who had only contributed 2 objects) to enforce a minimum of 3 objects per collector\footnote{See merged collectors in \texttt{data/orbit\_benchmark\_users \_to\_split.json} in the code repository}.
This yielded an effective total of 67 benchmark users.
These 67 users were then split into train/validation/test users (44/6/17, respectively\footnote{See \texttt{data/orbit\_benchmark\_mode\_splits.json} in the code repository}) according to the following criteria:

\paragraph{Validation/test user criteria}
We enforced that validation and test users should have at least 5 objects to ensure the test case was sufficiently challenging.
In total, 53 out of 67 collectors met this criteria, from which we randomly sampled 17 test users and 6 validation users.
Note, one test user (P204) contributed \(\sim\)1 order of magnitude more clutter videos per object than the average test user.
Since the \textsc{clu-ve} evaluation mode averages performance over all clutter videos pooled from all tasks across all test users (i.e. \(\targetset^\text{all}\)), results are likely slightly skewed toward P204.
This can be addressed by also reporting the per-user scores as in~\cref{fig:orbit-baselines-per-user}.
Note, P204's clean videos were inline with the average, thus \textsc{cle-ve} performance remains unaffected.

\paragraph{Train user criteria}
The remaining 44 collectors were marked as train users (14 of which did not meet the above criteria).
Note, one train user (P587) captured \(\sim\)1 order of magnitude more clean (though not more clutter) videos than the average train user.
Since models are trained on a fixed number of tasks per train user and each task is capped in the number of clean videos sampled per object (see~\cref{app:sec:task-sampling}), the results are not skewed by P587.

\subsection{Video pre-processing}
Videos were split into frames using \texttt{ffmpeg} at a rate of \(30\) \acrshort{FPS}.
Frames were also re-sized from \(1080 \times 1080\) pixels to \(84 \times 84\).
This was done for \textsc{gpu} memory purposes and future work will look toward scaling to larger images.

\section{Extended benchmark dataset summary}
\label{app:sec:extended-benchmark-dataset-summary}

\begin{figure*}[ht!]
    \centering
    \begin{subfigure}[t]{\textwidth}
        \centering
        \hspace{-1.2em}
        \includegraphics[width=\textwidth,trim=0 0cm 0 1cm,clip]{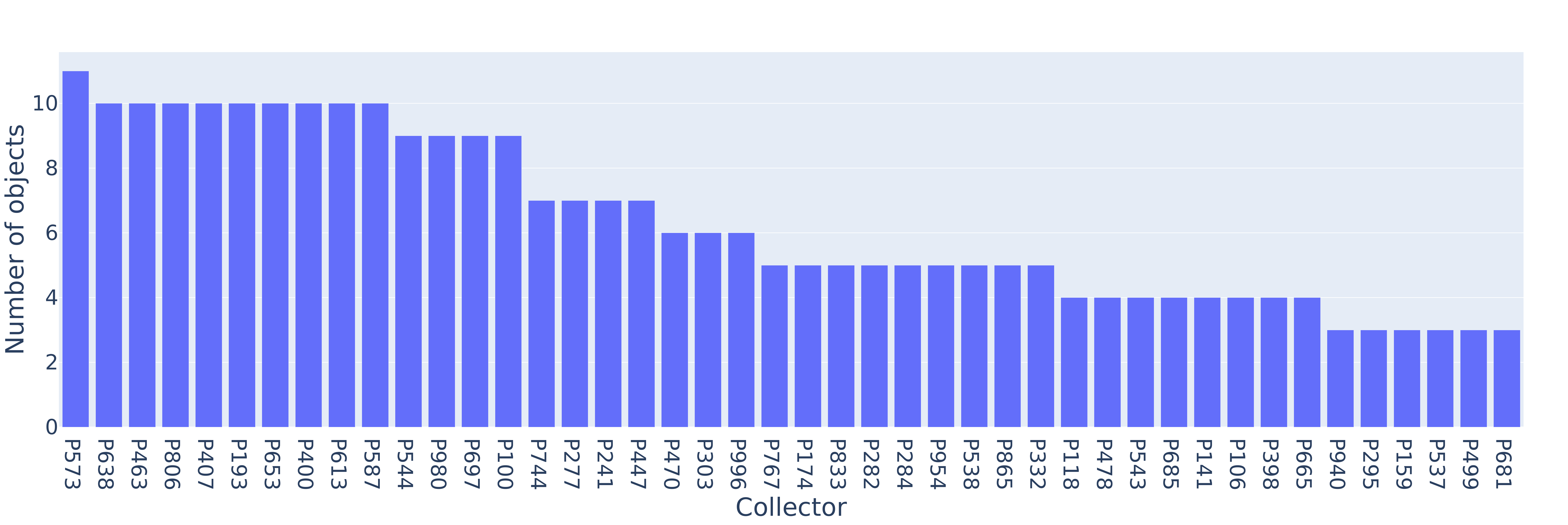}
        \vspace{-0.2em}
        \caption{Train users.}
        \vspace{-0.3em}
        \label{fig:benchmark-num-objects-train}
    \end{subfigure}
    \begin{subfigure}[t]{0.3\textwidth}
        \centering
        \hspace{-1.1em}
        \includegraphics[width=0.65\textwidth,trim=0 0cm 0 1cm,clip]{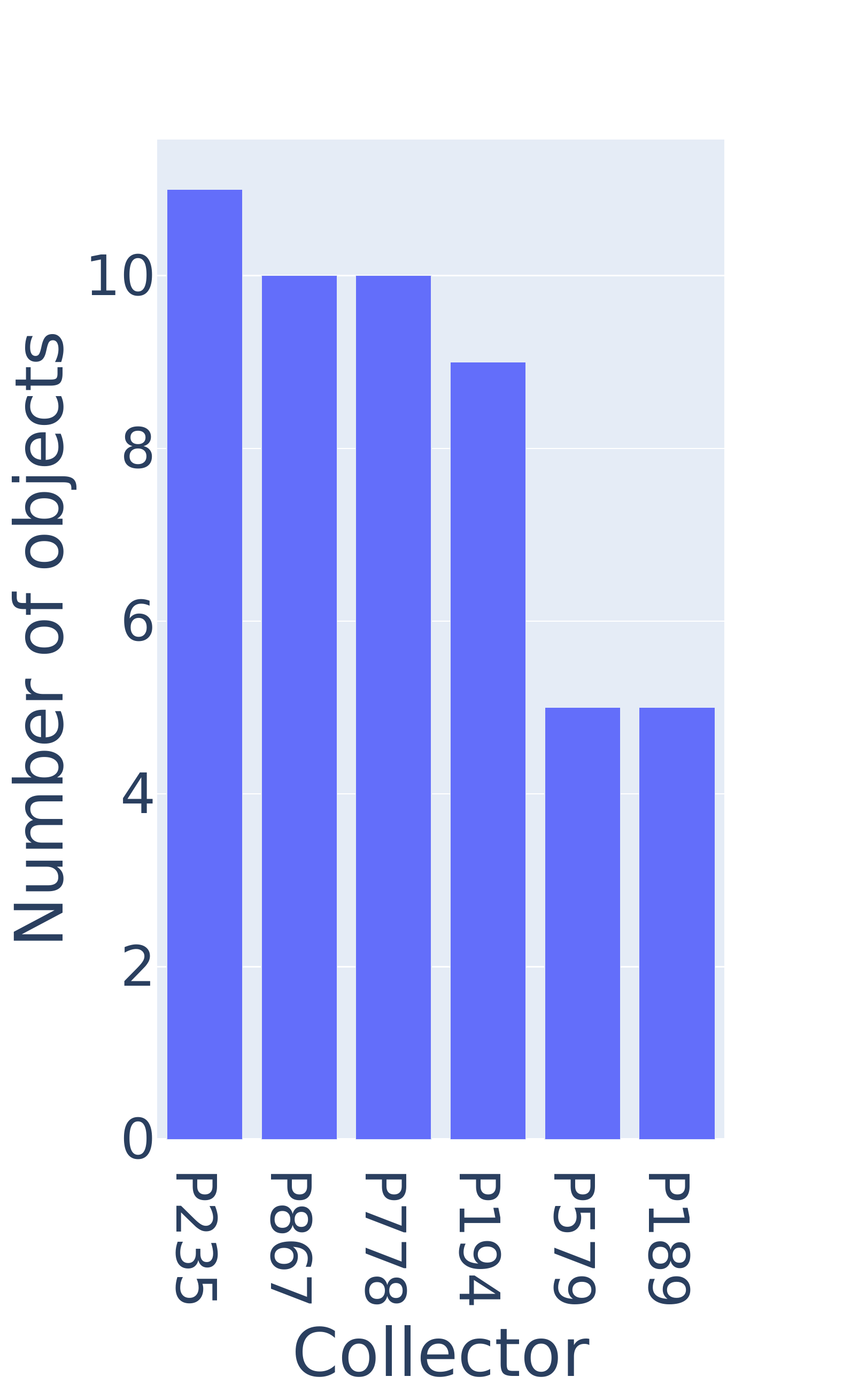}
        \vspace{-0.2em}
        \caption{Validation users.}
        \vspace{-0.3em}
        \label{fig:benchmark-num-objects-validation}
    \end{subfigure}
    \hspace{1em}
    \begin{subfigure}[t]{0.65\textwidth}
        \centering
        \hspace{-1em}
        \includegraphics[width=0.7\textwidth,trim=0 0cm 0 1cm,clip]{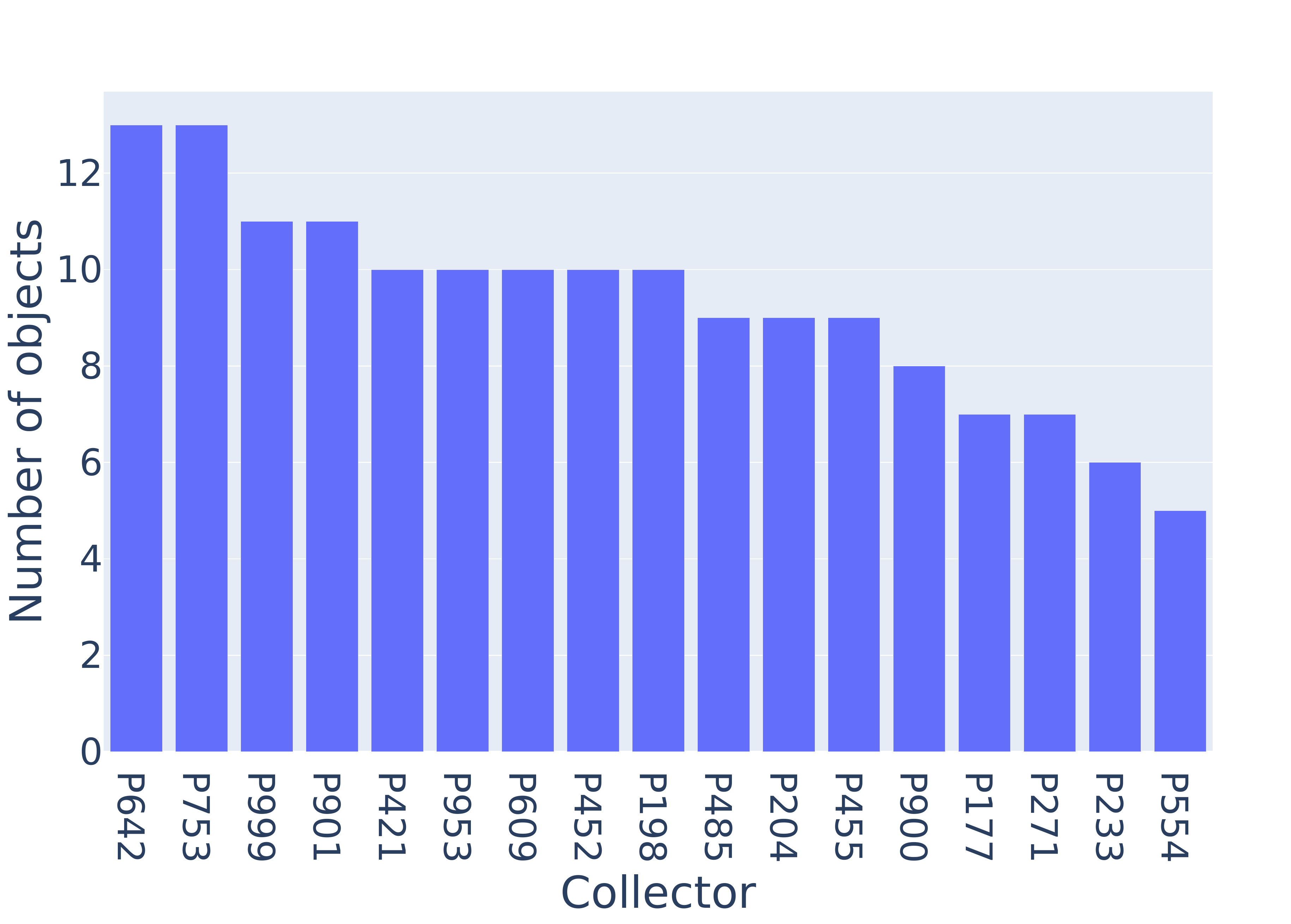}
        \vspace{-0.2em}
        \caption{Test users.}
        \vspace{-0.3em}
        \label{fig:benchmark-num-objects-test}
    \end{subfigure}
    \vspace{-0.4em}
    \caption{Number of objects per collector in the ORBIT benchmark dataset, grouped by train, validation, and test users.}
    \vspace{-1em}
    \label{fig:benchmark-num-objects-allsets}
\end{figure*}

\begin{figure*}[ht!]
    \centering
    \vspace{2em}
    \includegraphics[width=\textwidth,trim=0 21.2cm 0 0,clip]{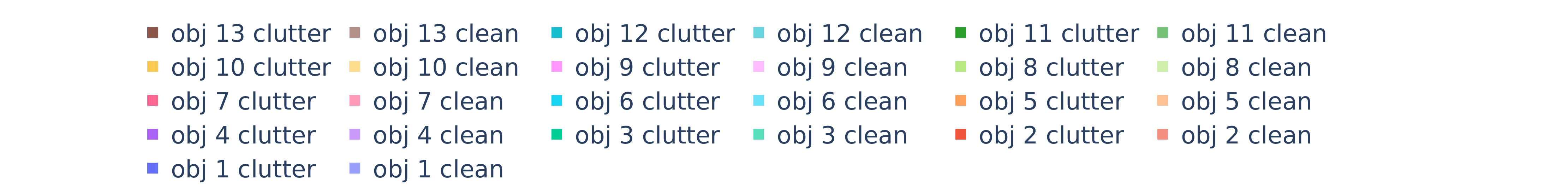}
    \begin{subfigure}[t]{\textwidth}
        \centering
        \hspace{-1.2em}
        \includegraphics[width=\textwidth]{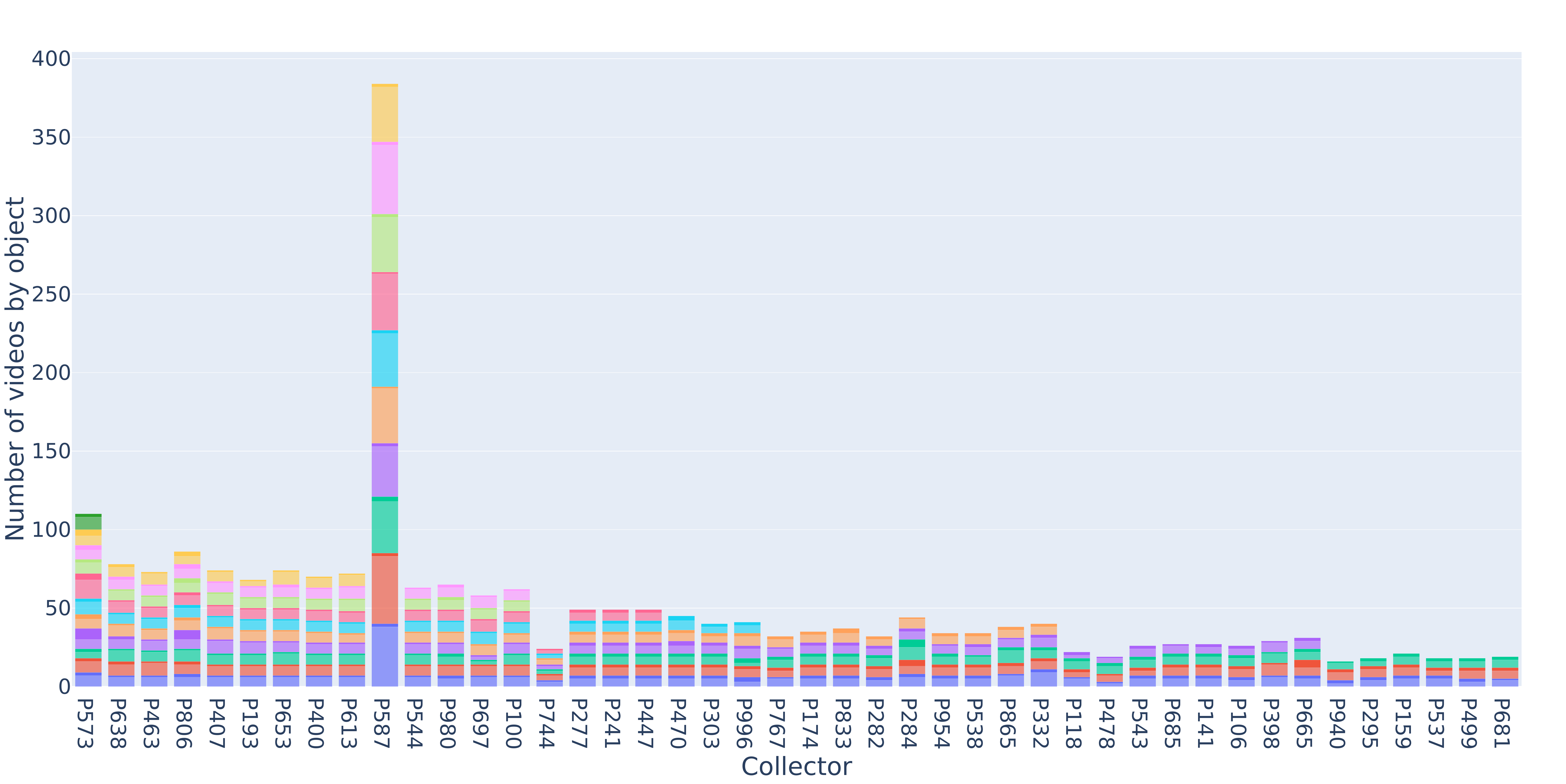}
        \vspace{-0.2em}
        \caption{Train users.}
        \vspace{-0.3em}
        \label{fig:benchmark-num-vids-by-object-train}
    \end{subfigure}
    \begin{subfigure}[t]{0.3\textwidth}
        \centering
        \hspace{-1.1em}
        \includegraphics[width=0.65\textwidth]{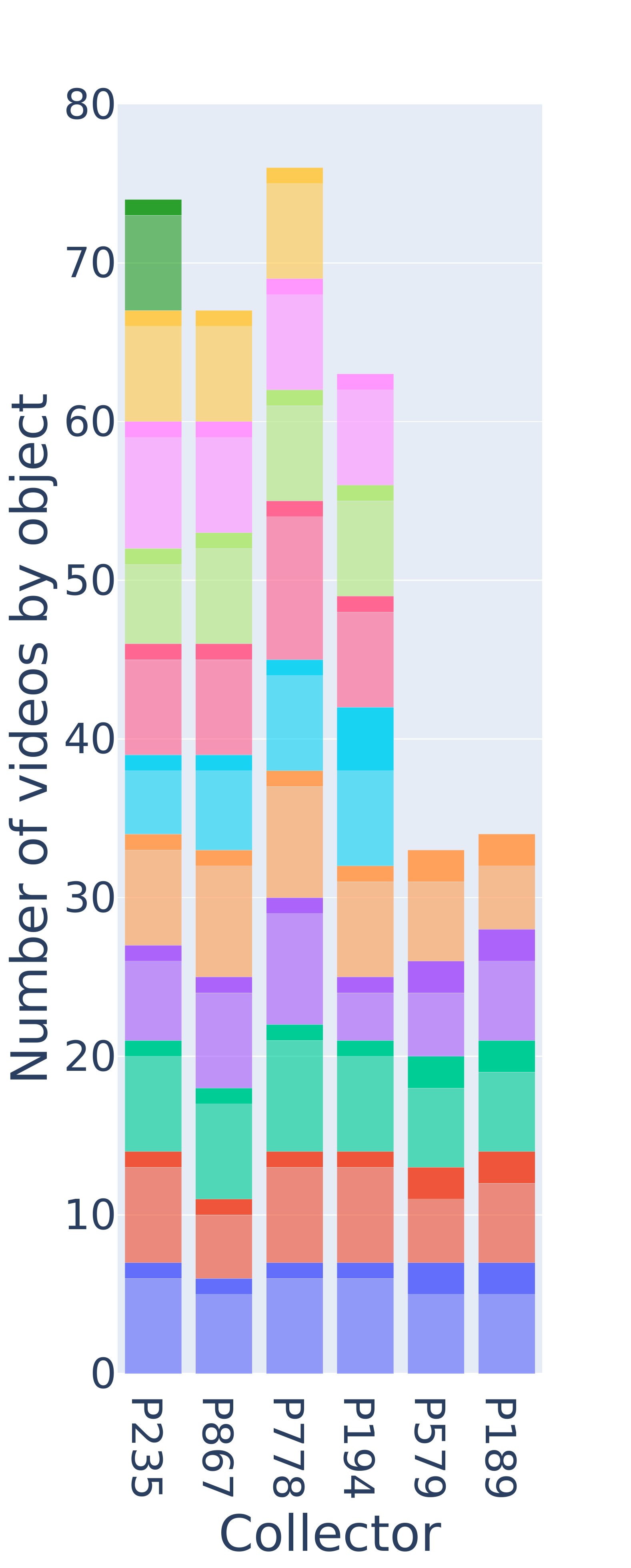}
        \vspace{-0.2em}
        \caption{Validation users.}
        \vspace{-0.4em}
        \label{fig:benchmark-num-vids-by-object-validation}
    \end{subfigure}
    \begin{subfigure}[t]{0.65\textwidth}
        \centering
        \hspace{-1.2em}
        \includegraphics[width=0.7\textwidth,trim=0 0 0 0,clip]{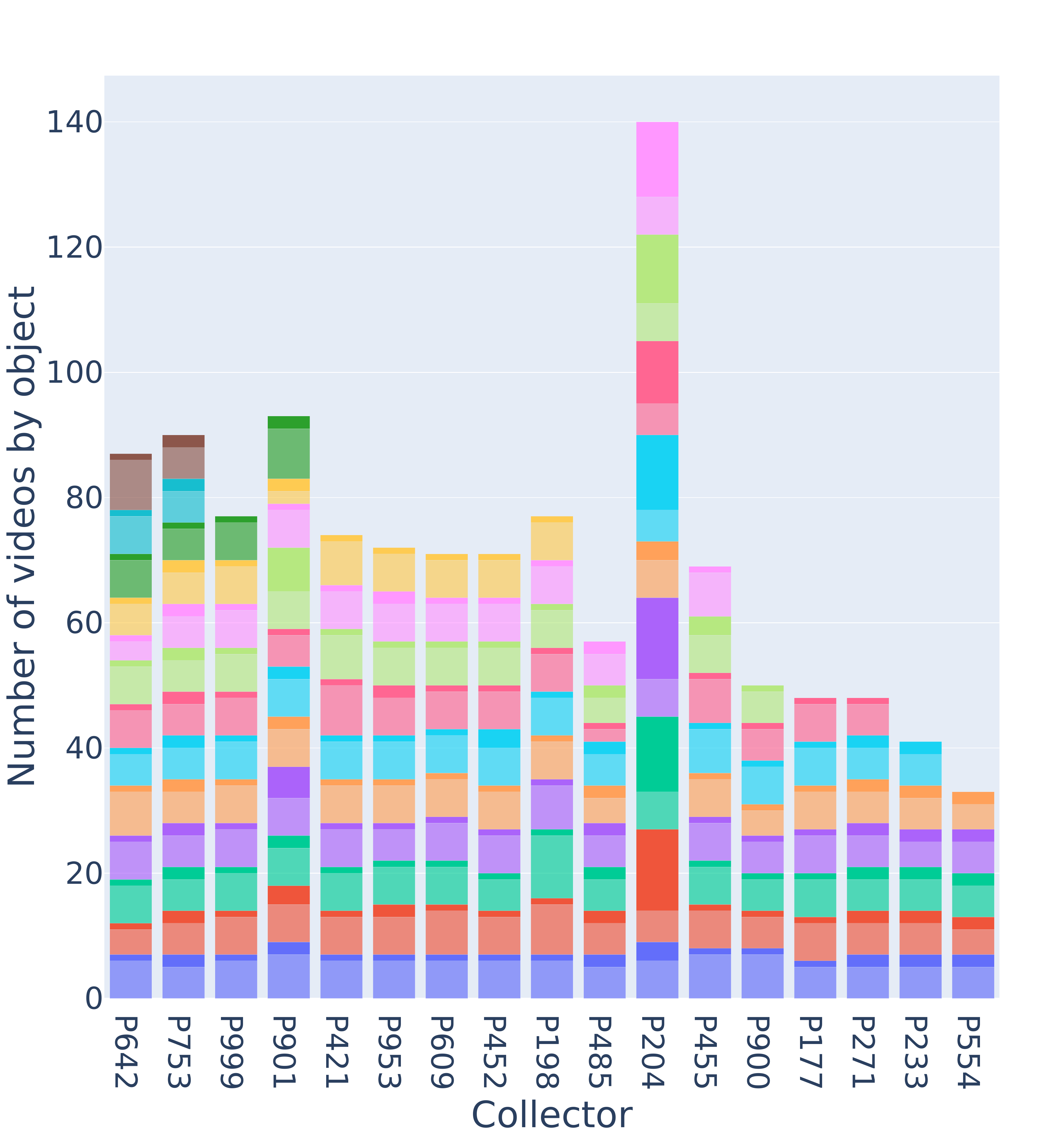}
        \caption{Test users.}
        \vspace{-0.4em}
        \label{fig:benchmark-num-vids-by-object-test}
    \end{subfigure}
    \vspace{0em}
    \caption{Number of clean and clutter videos (stacked by object) per collector in the ORBIT benchmark dataset, grouped by train, validation, and test users.}
    \label{fig:benchmark-num-vids-by-object-allsets}
    \vspace{-1em}
\end{figure*}

\begin{figure*}
    \centering
    \scalebox{0.95}{
    \mbox{\includegraphics[width=0.095\textwidth]{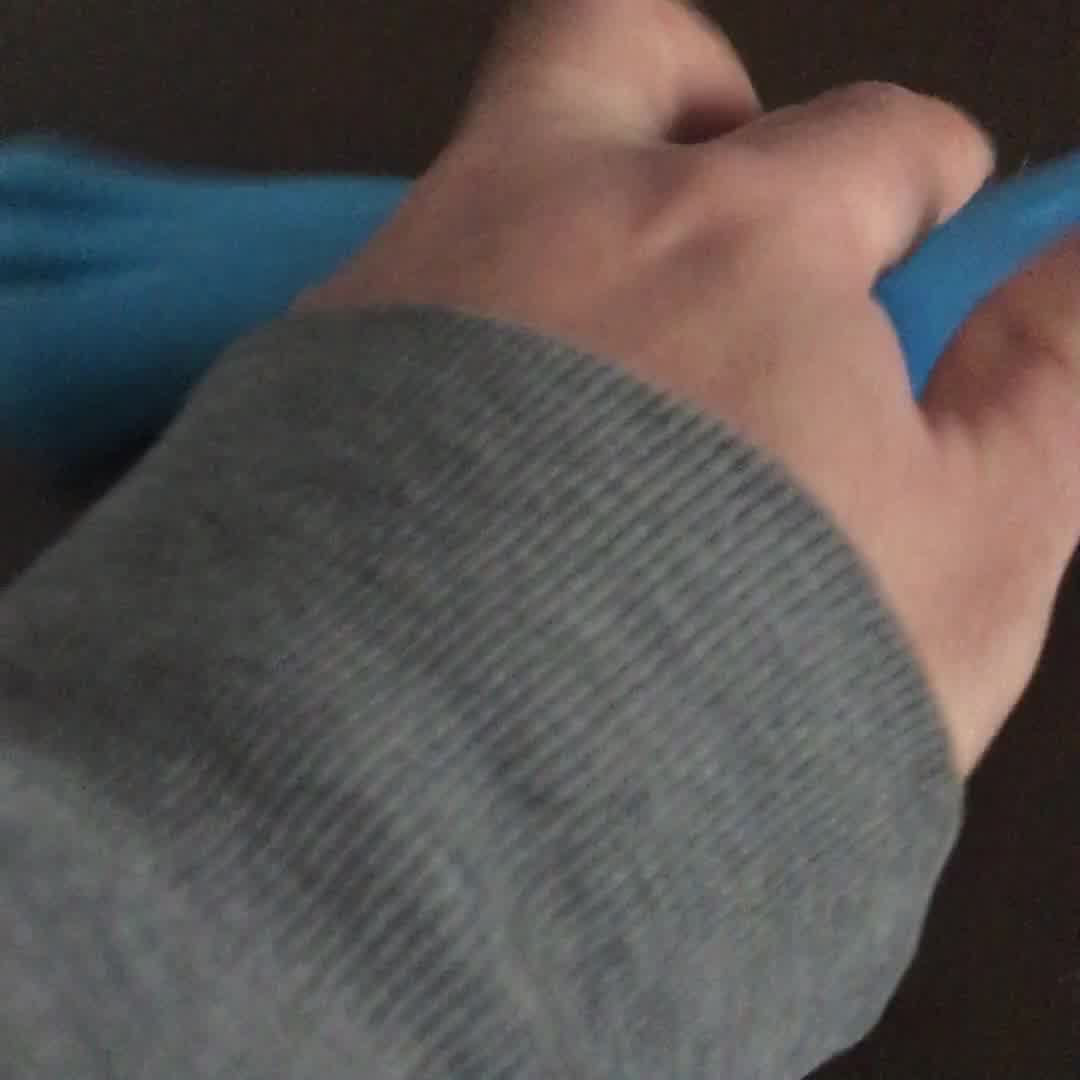}}
    \mbox{\includegraphics[width=0.095\textwidth]{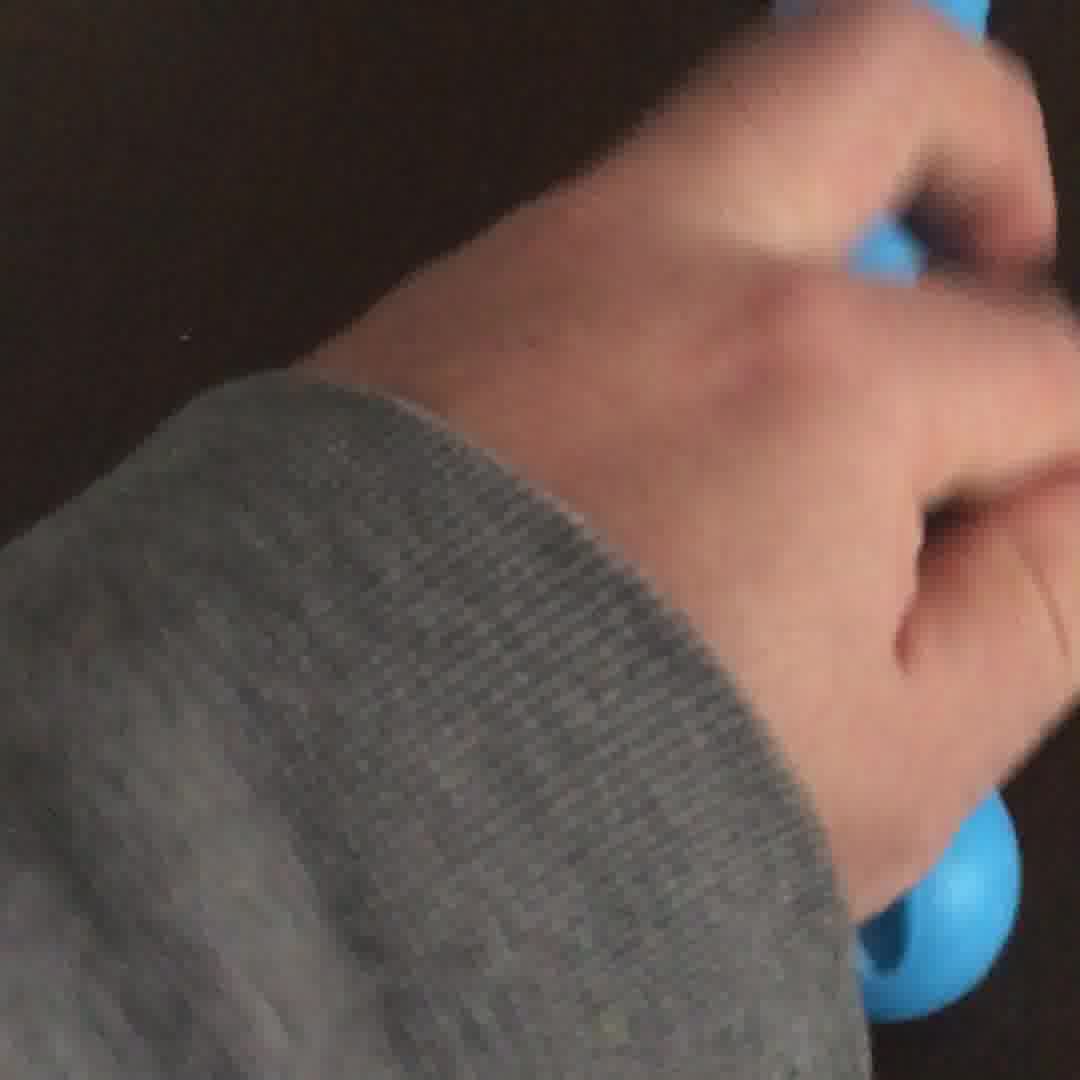}}
    \mbox{\includegraphics[width=0.095\textwidth]{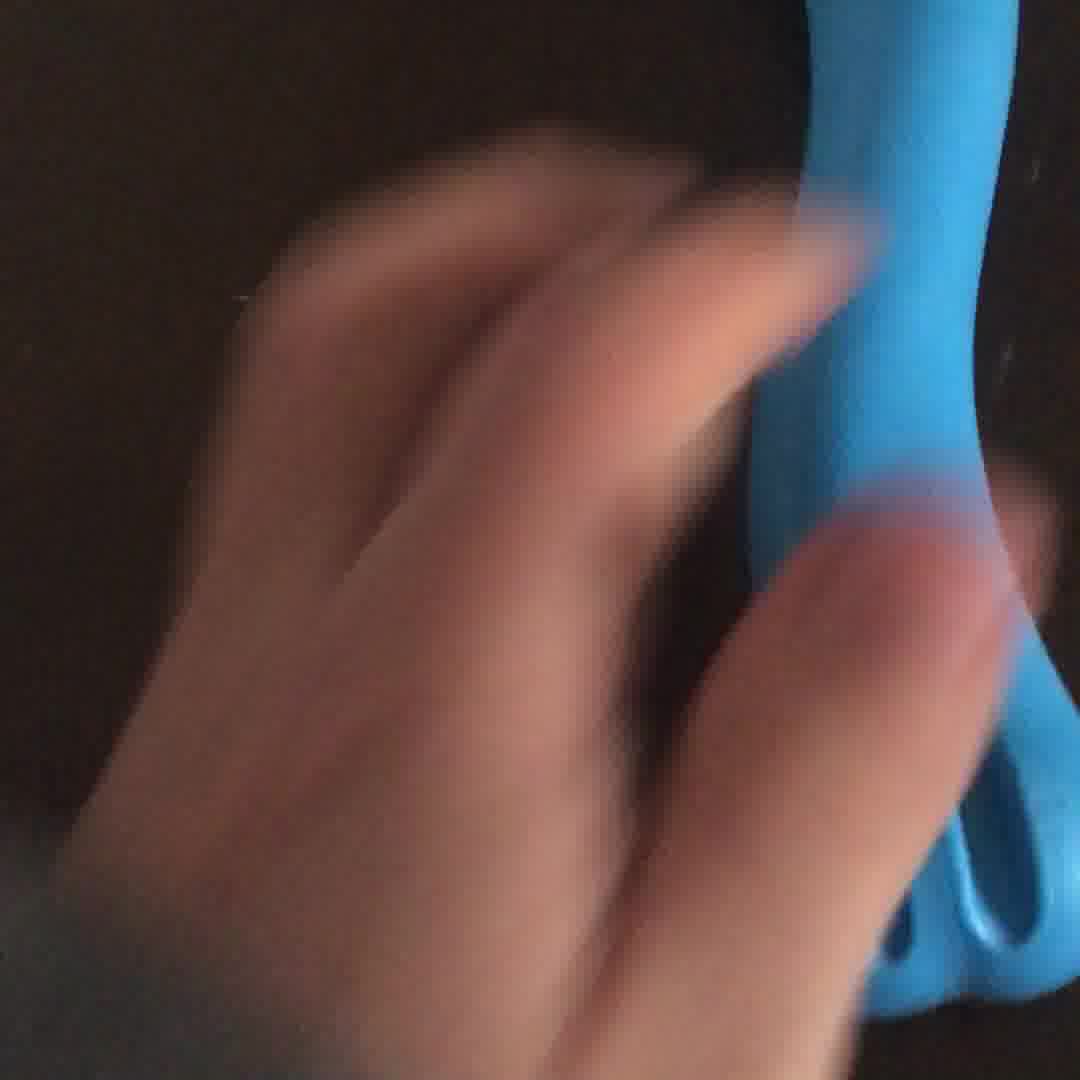}}
    \mbox{\includegraphics[width=0.095\textwidth]{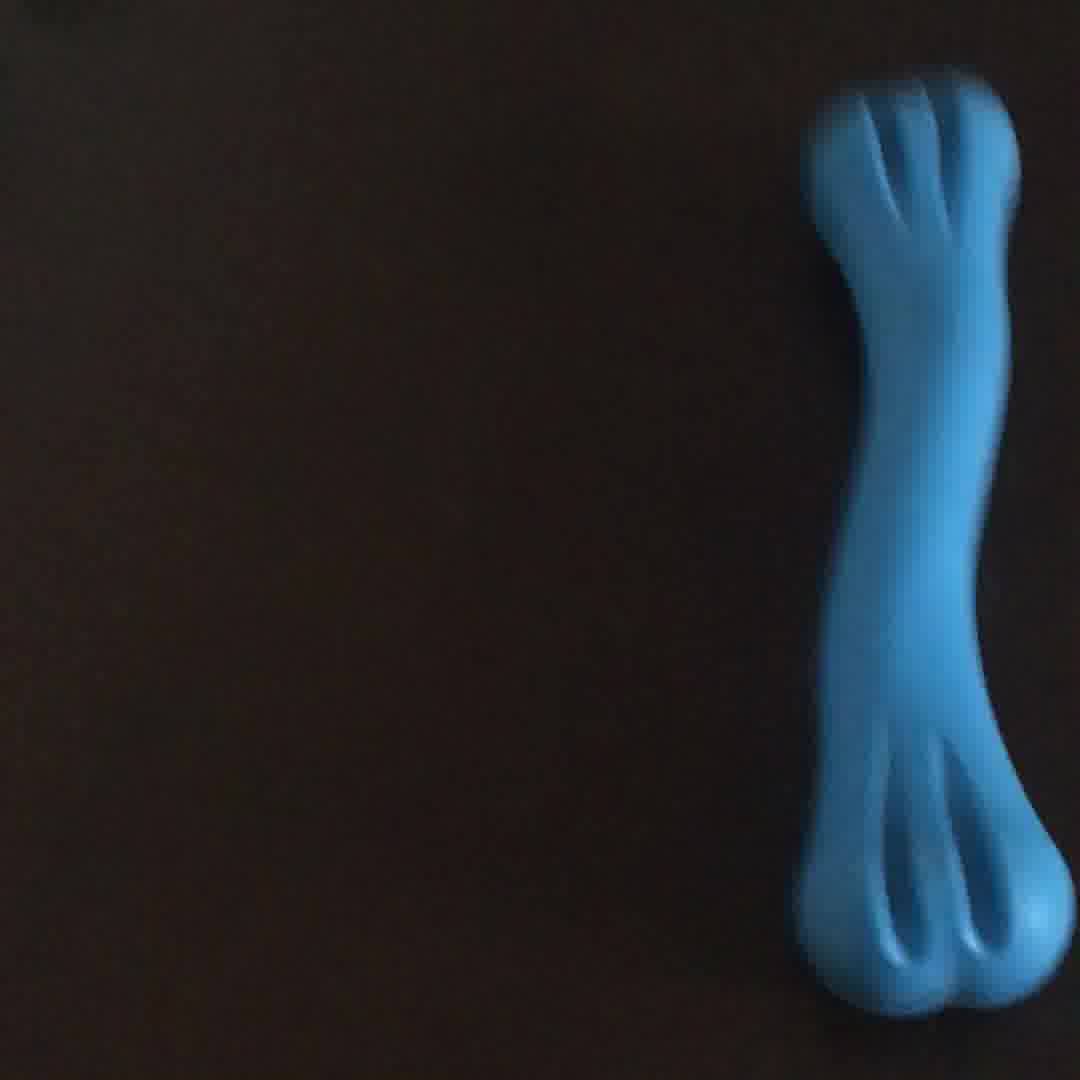}}
    \mbox{\includegraphics[width=0.095\textwidth]{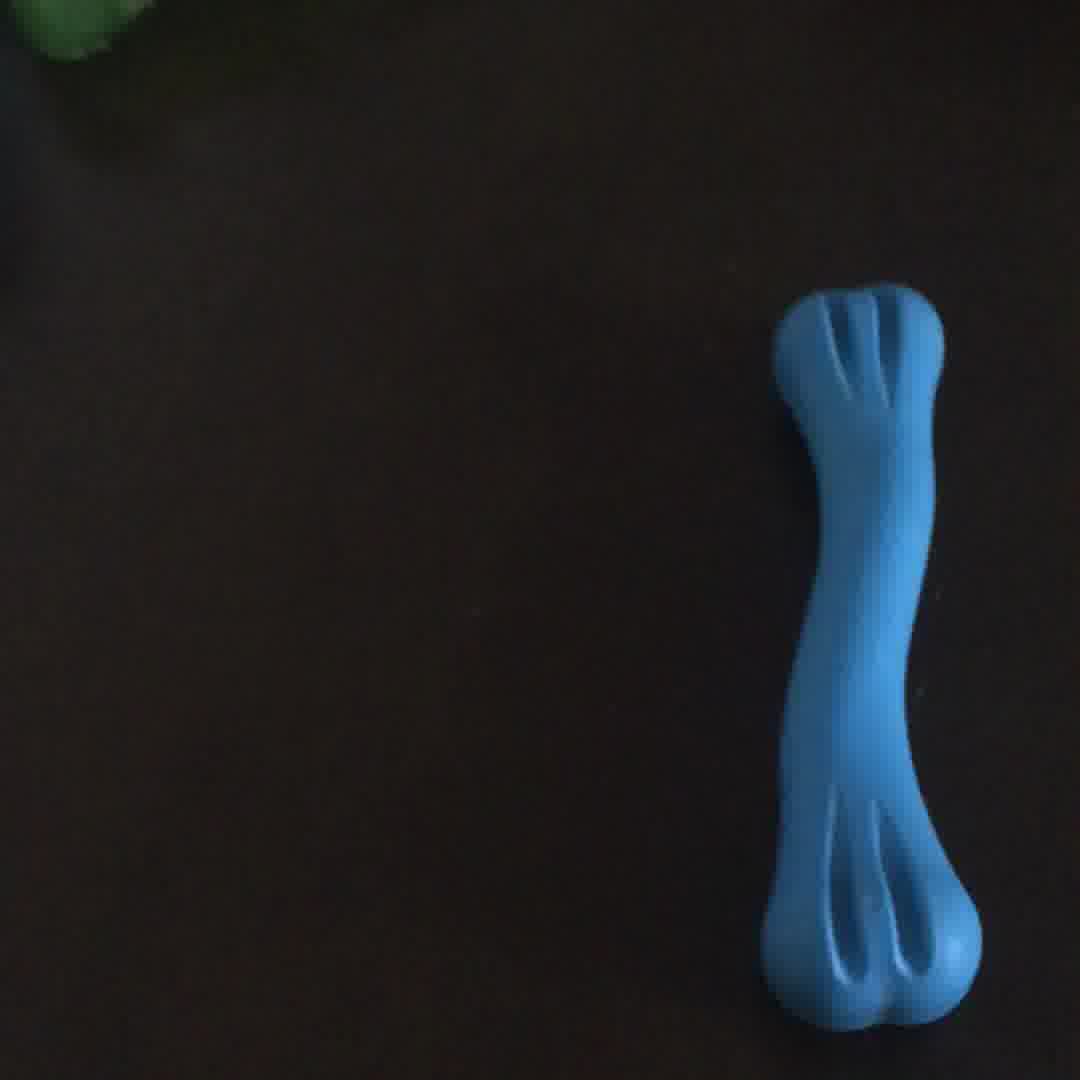}}
    \mbox{\includegraphics[width=0.095\textwidth]{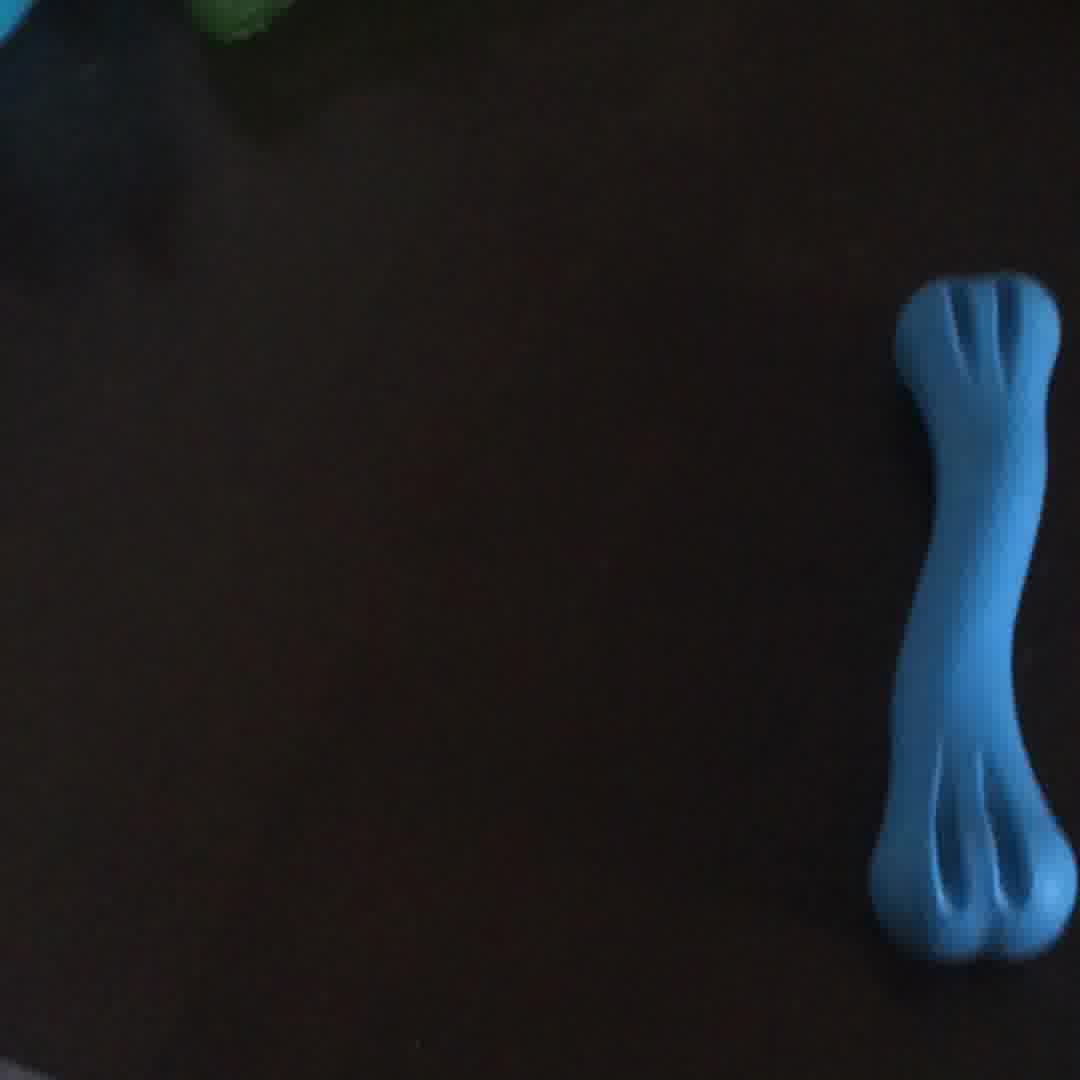}}
    \mbox{\includegraphics[width=0.095\textwidth]{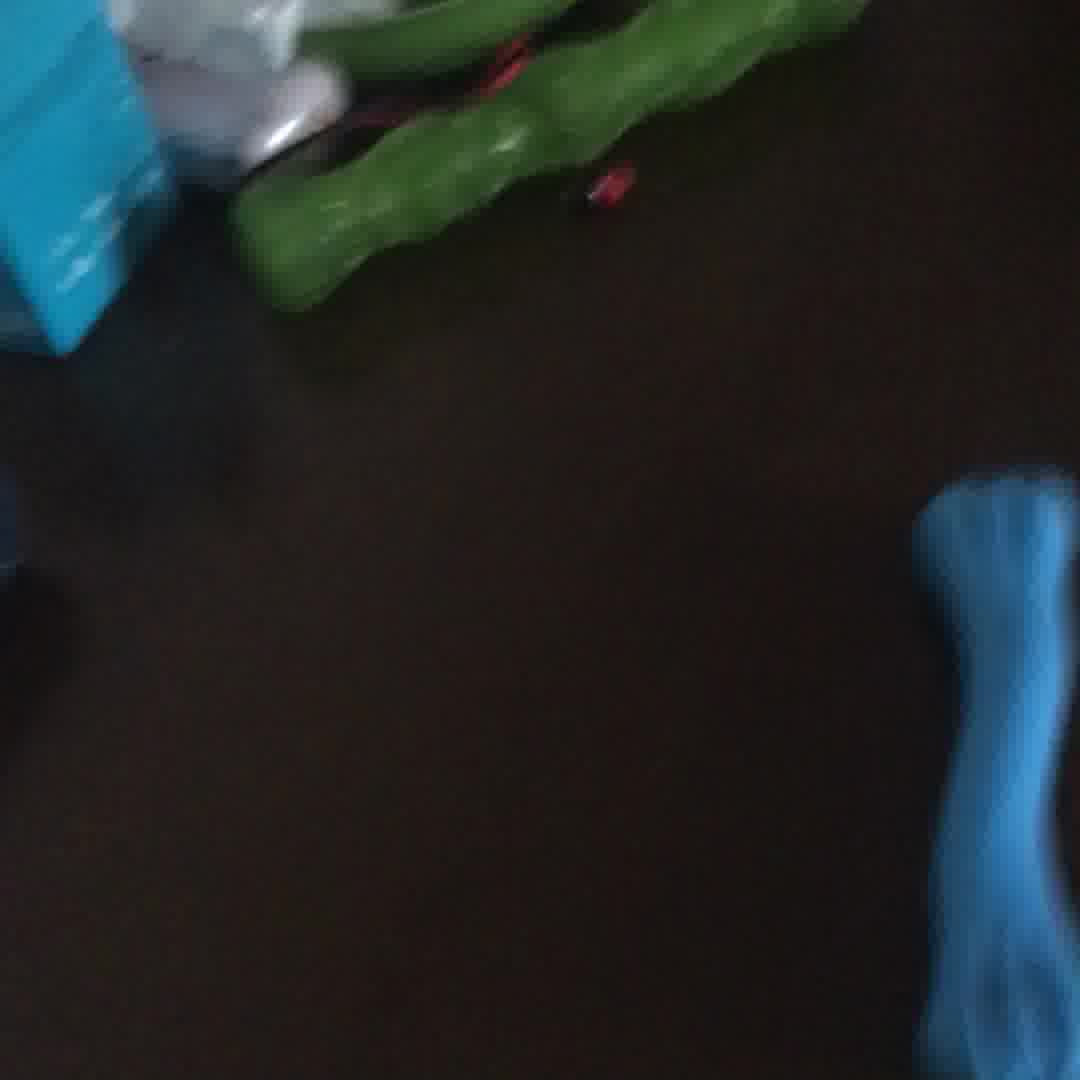}}
    \mbox{\includegraphics[width=0.095\textwidth]{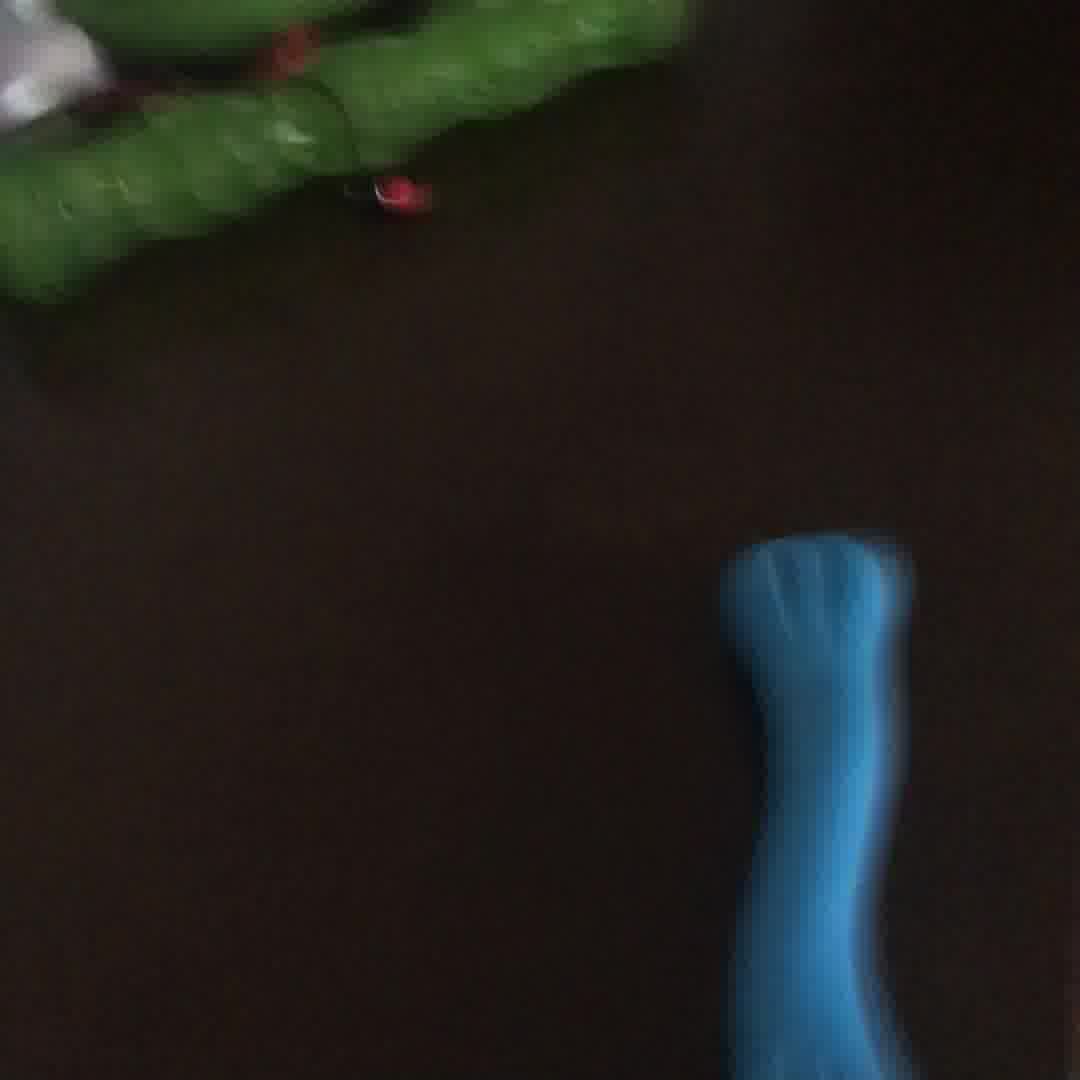}}
    \mbox{\includegraphics[width=0.095\textwidth]{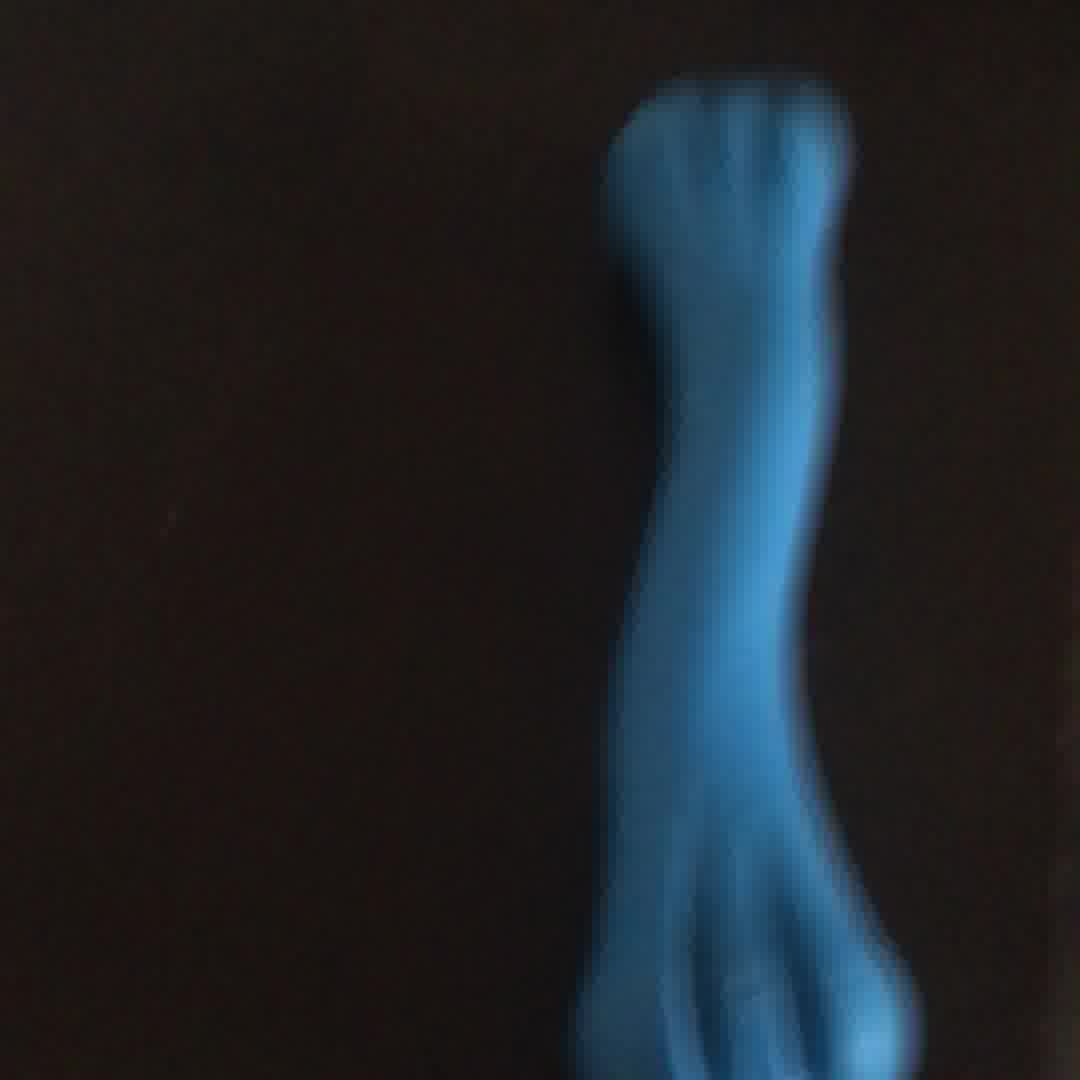}}
    \mbox{\includegraphics[width=0.095\textwidth]{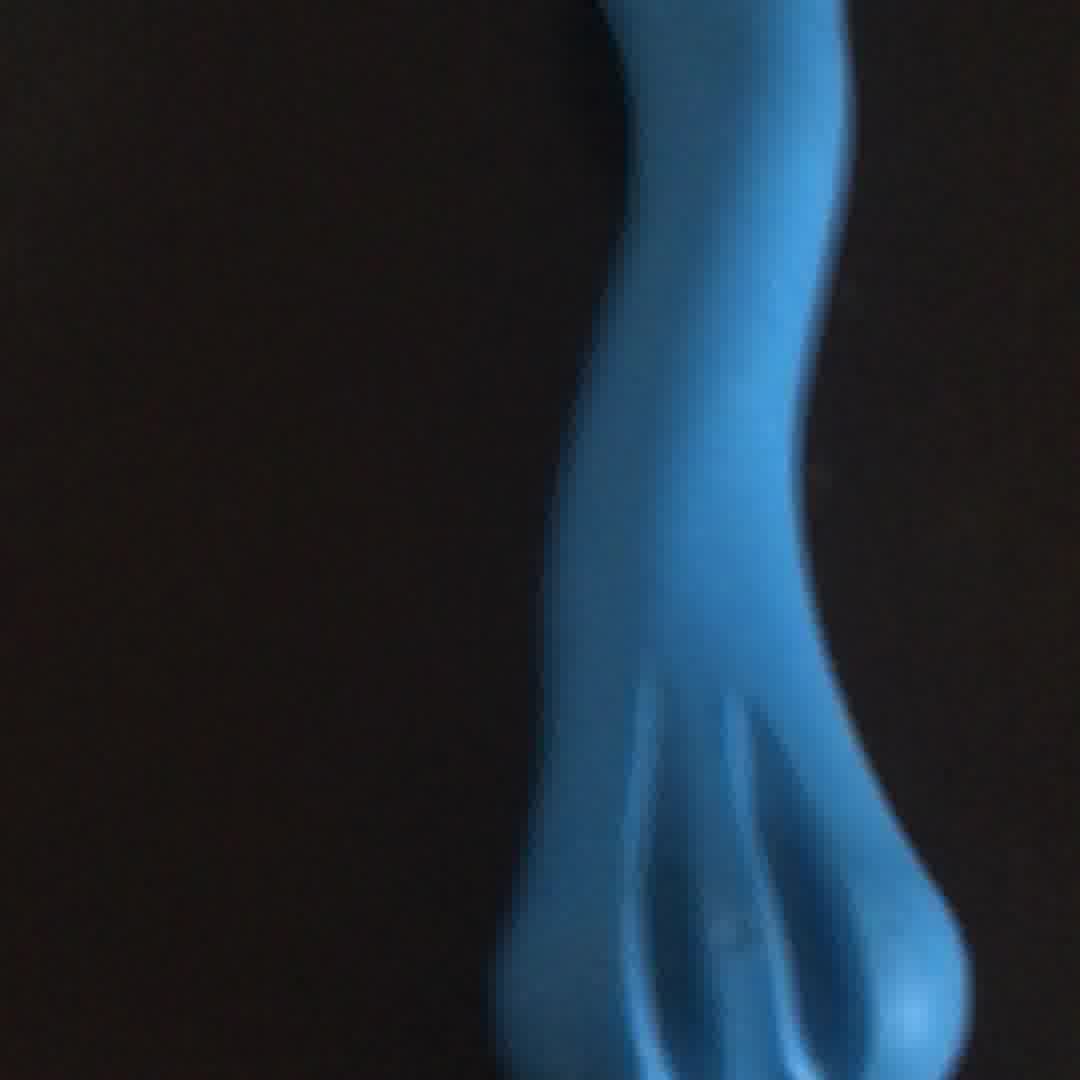}}}\\
    \vspace*{2px}
    \scalebox{0.95}{
    \mbox{\includegraphics[width=0.095\textwidth]{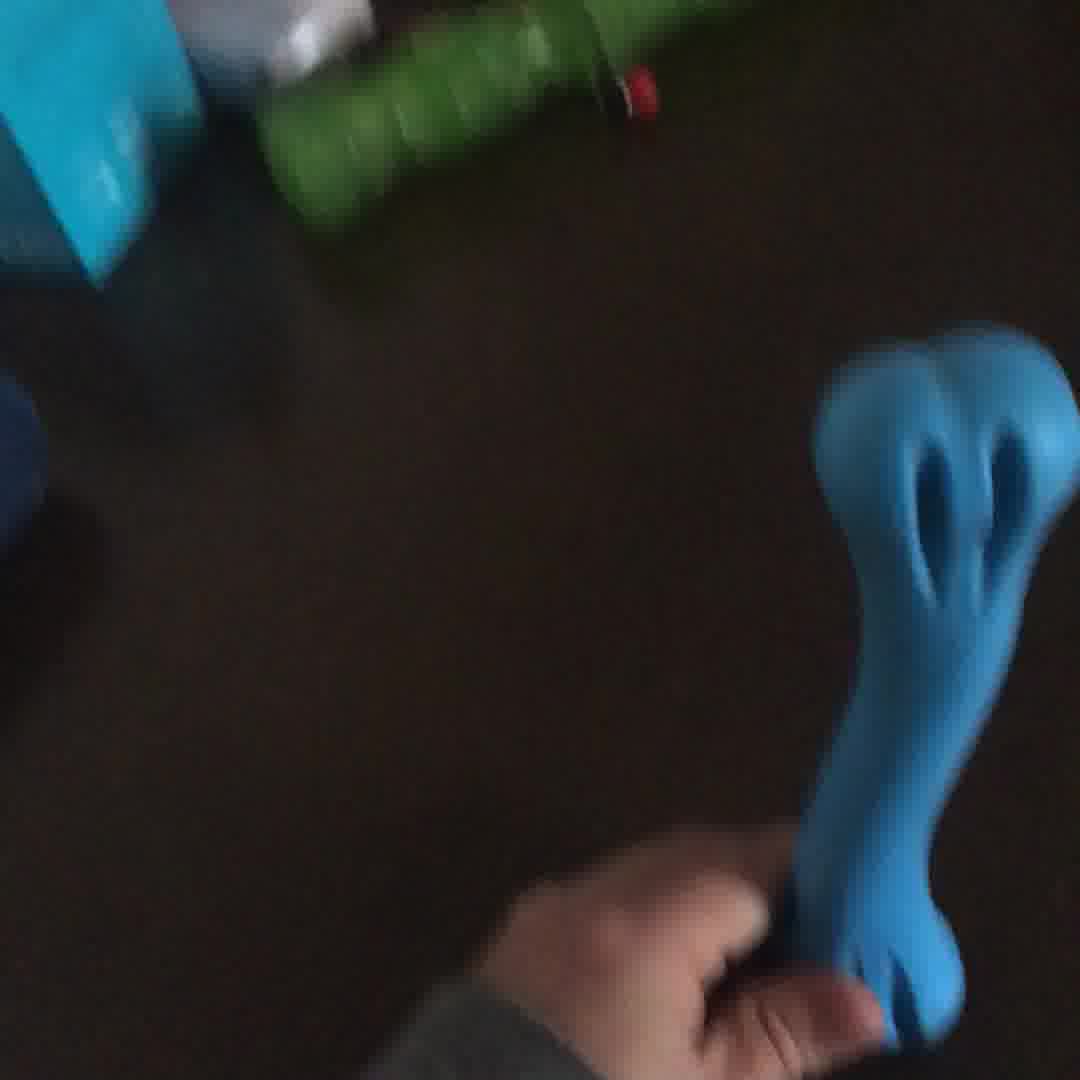}}
    \mbox{\includegraphics[width=0.095\textwidth]{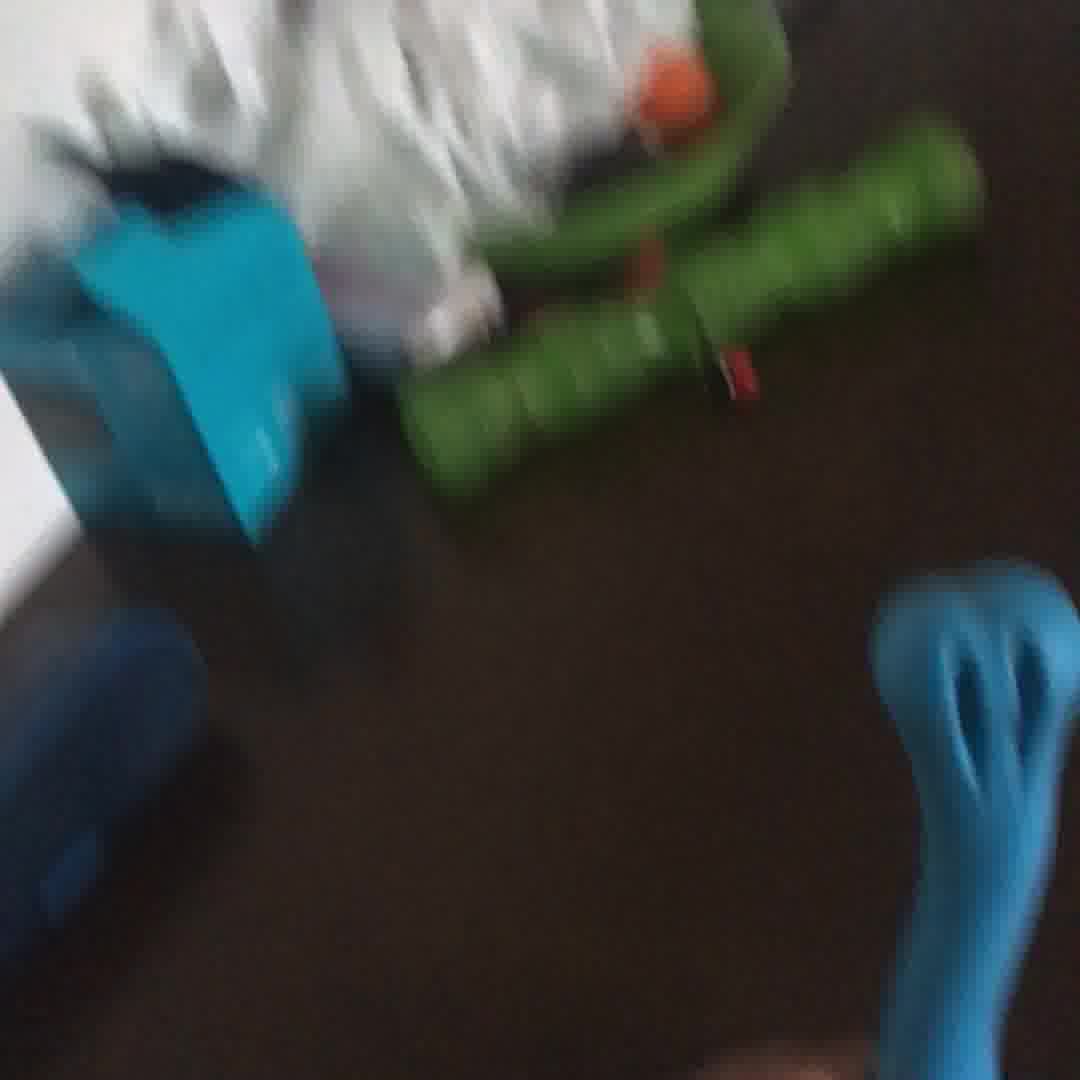}}
    \mbox{\includegraphics[width=0.095\textwidth]{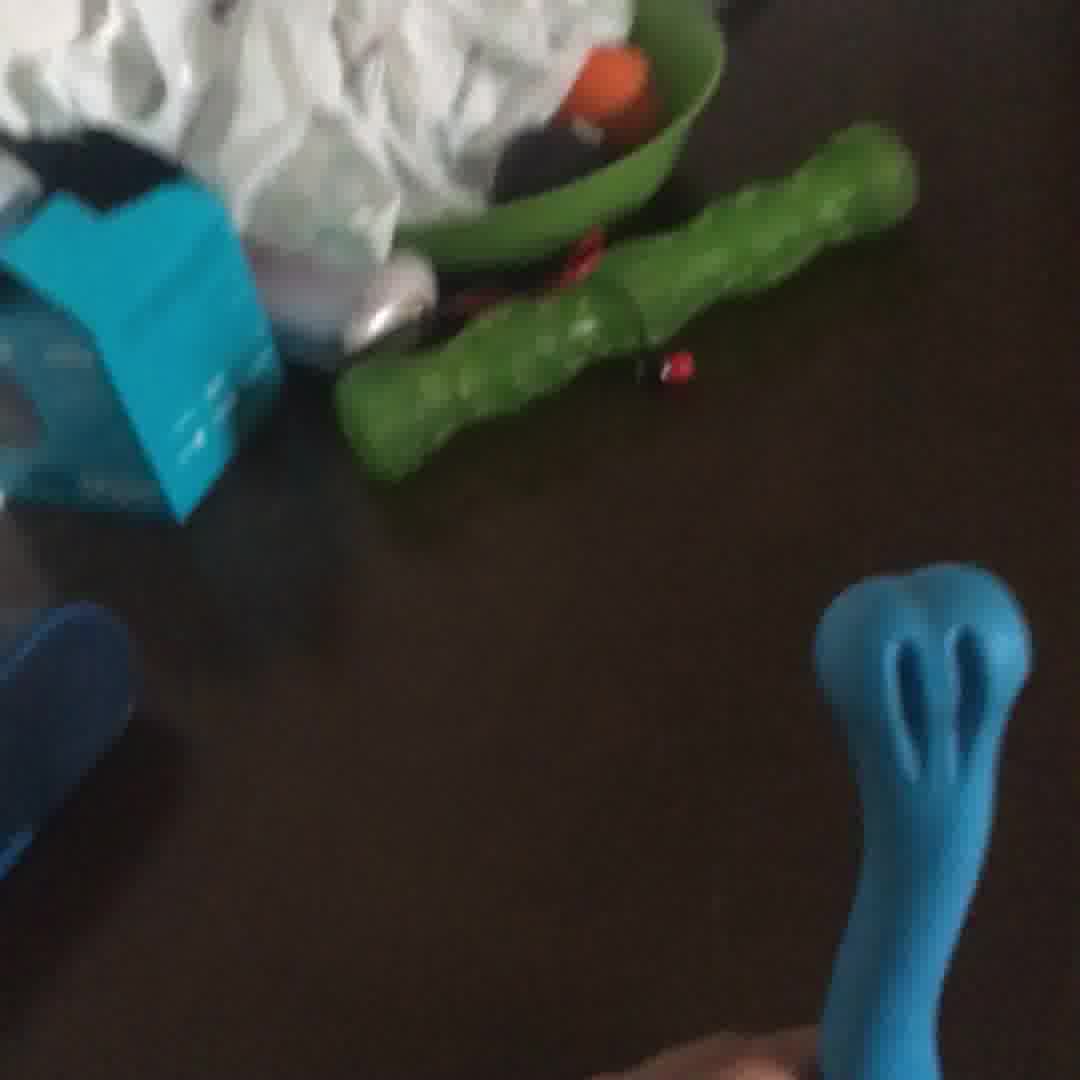}}
    \mbox{\includegraphics[width=0.095\textwidth]{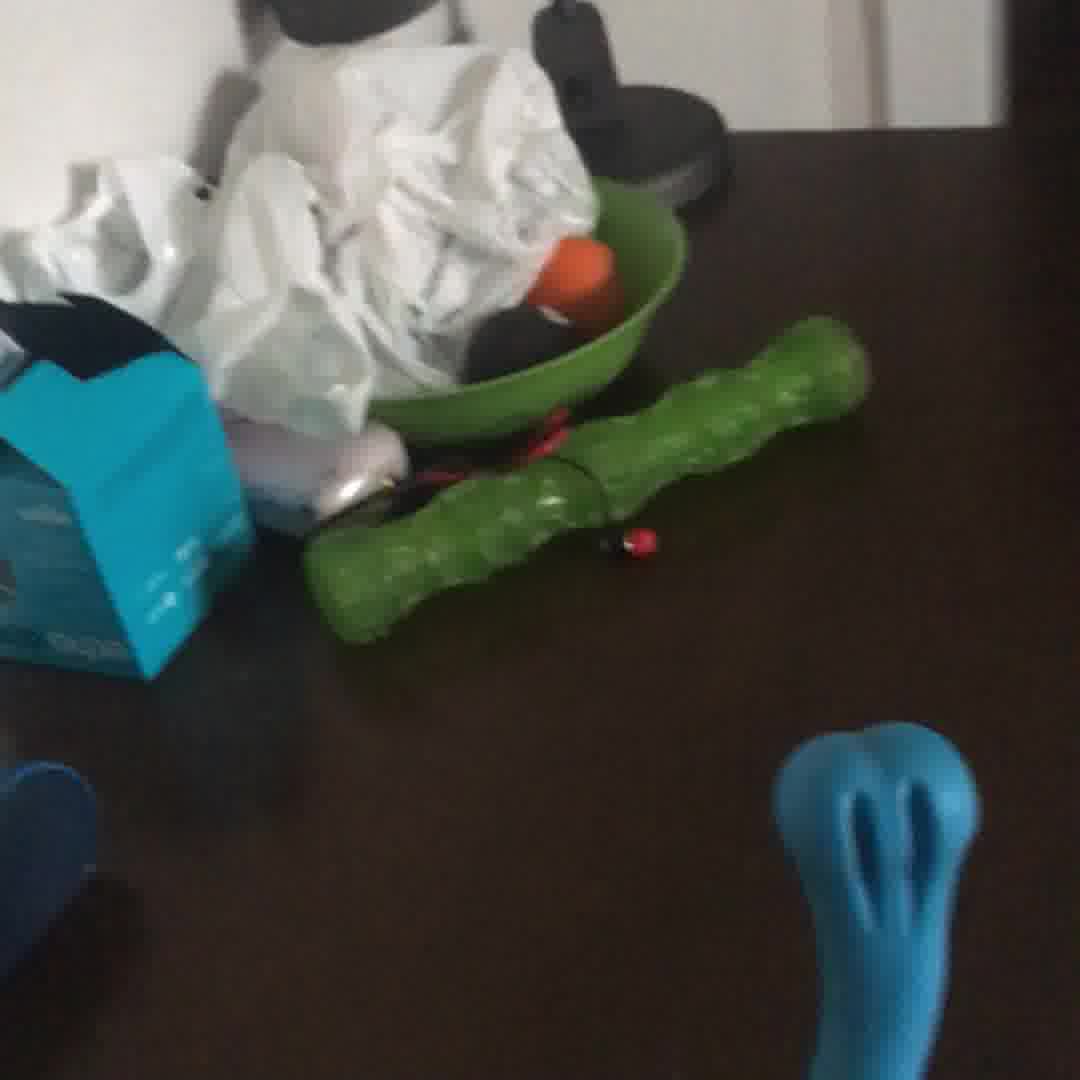}}
    \mbox{\includegraphics[width=0.095\textwidth]{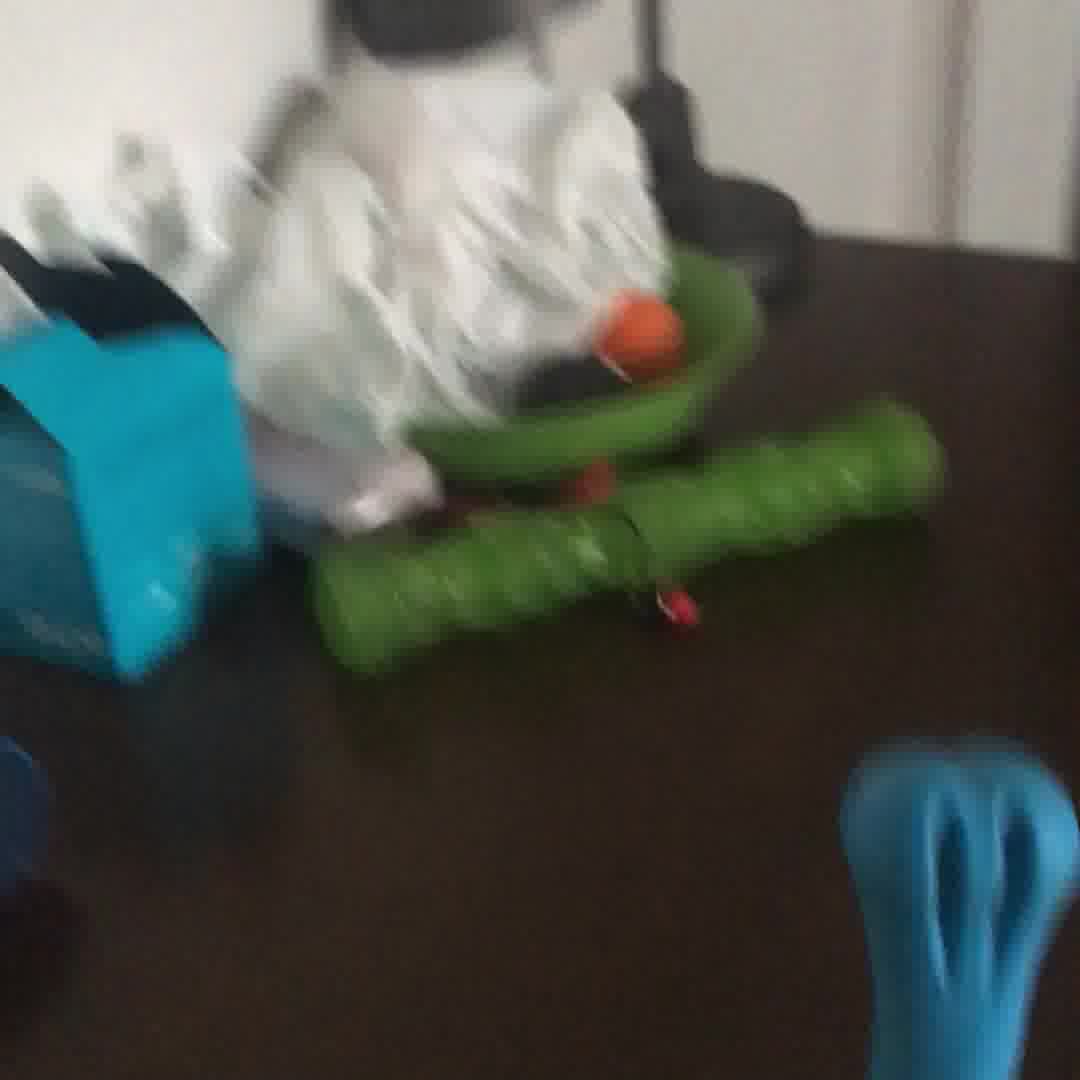}}
    \mbox{\includegraphics[width=0.095\textwidth]{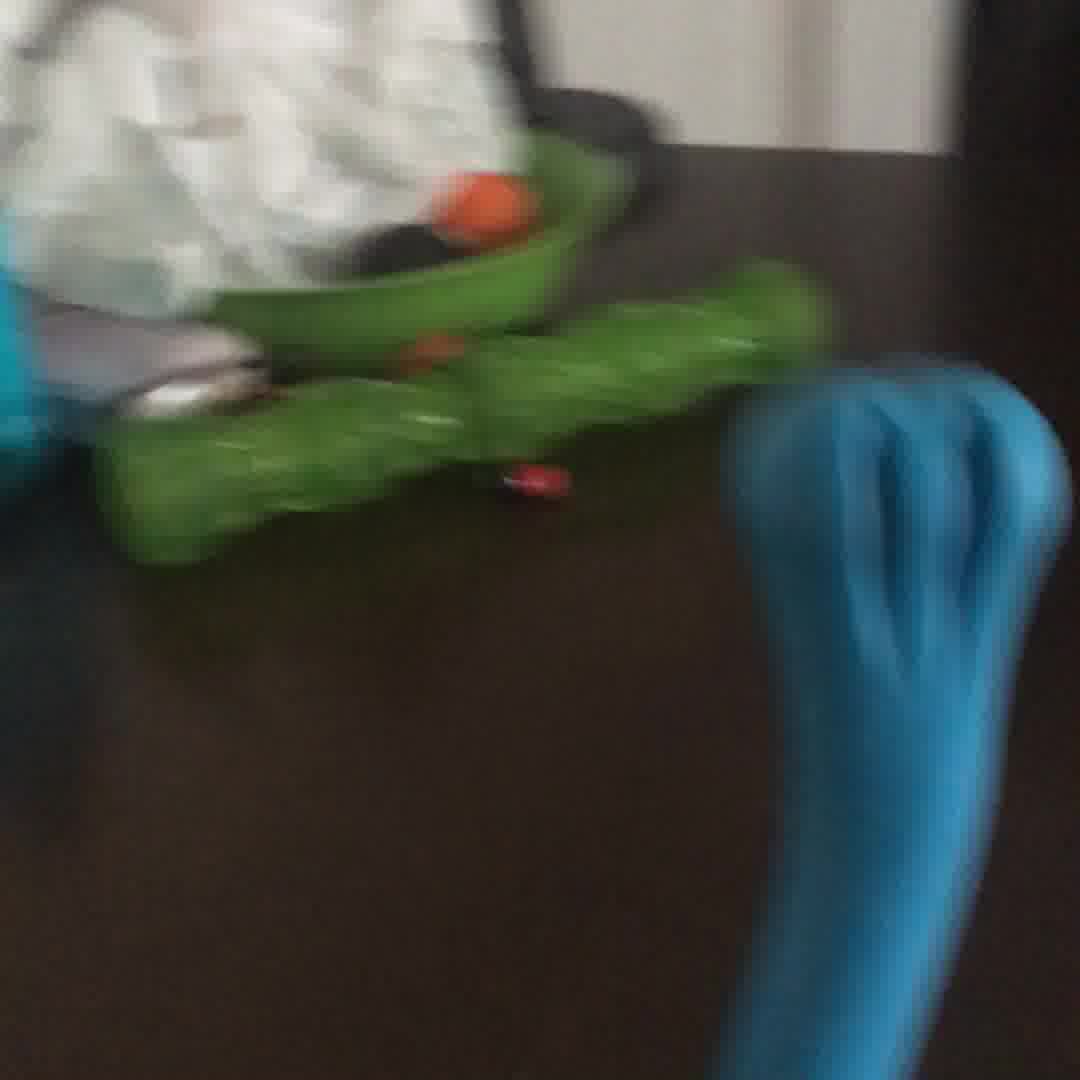}}
    \mbox{\includegraphics[width=0.095\textwidth]{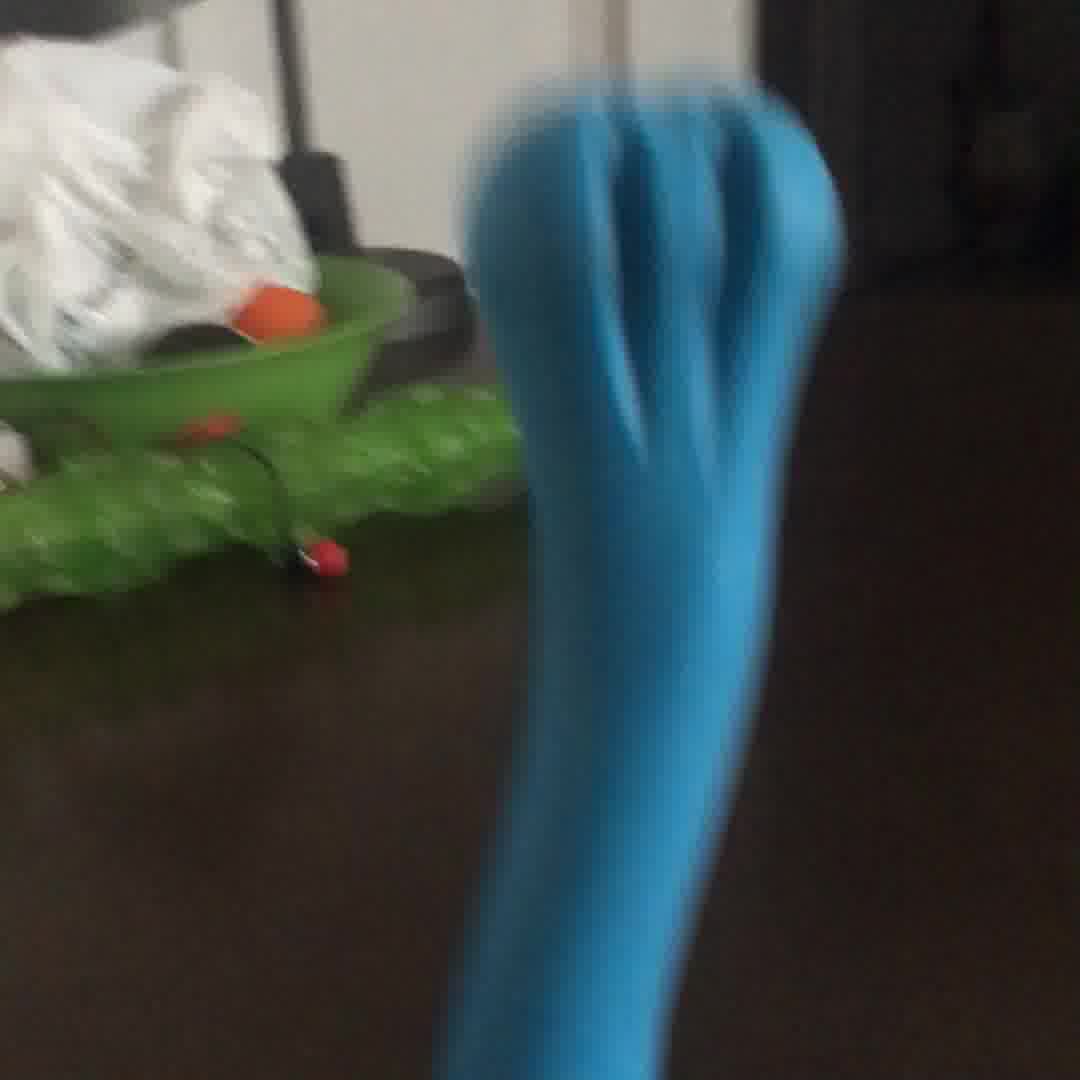}}
    \mbox{\includegraphics[width=0.095\textwidth]{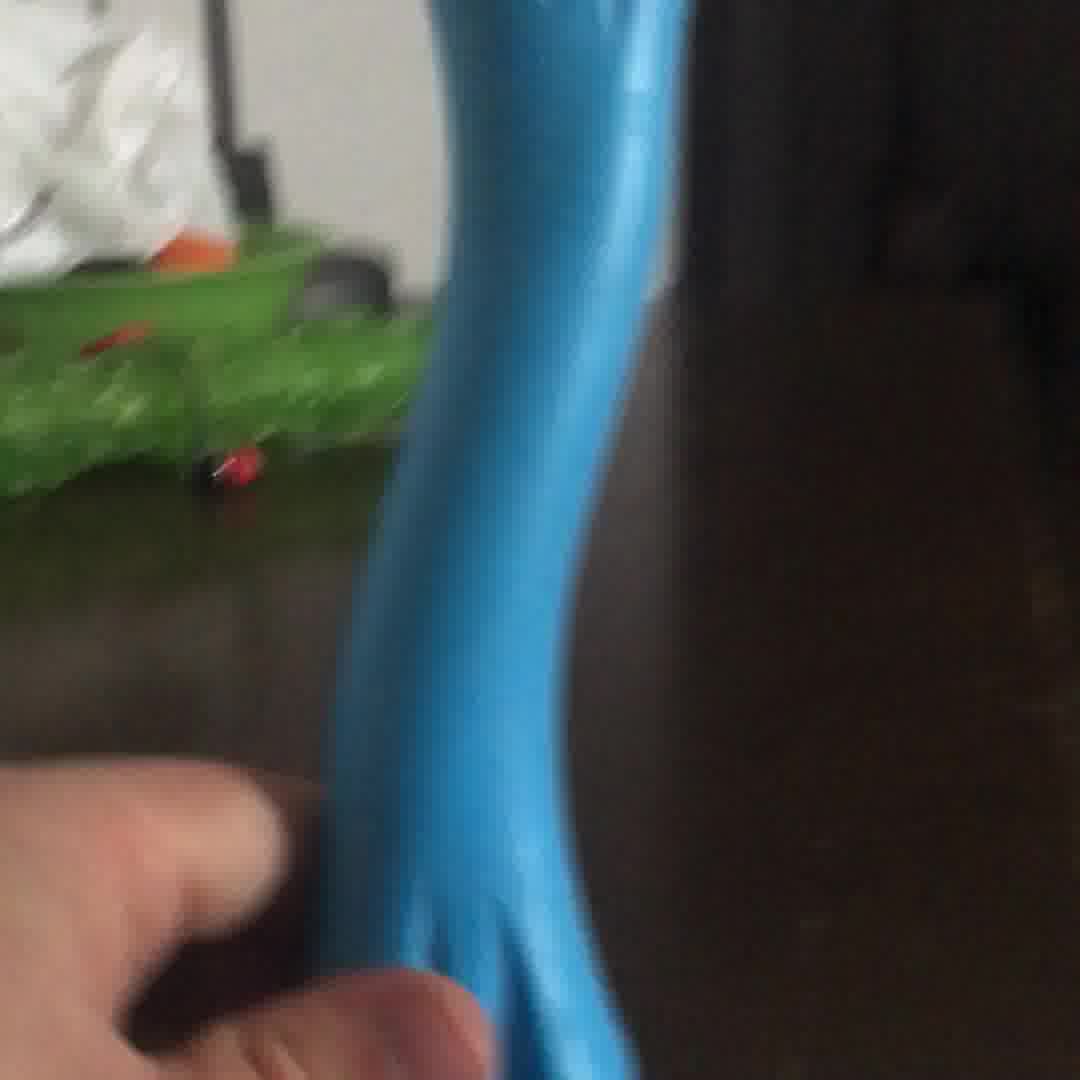}}
    \mbox{\includegraphics[width=0.095\textwidth]{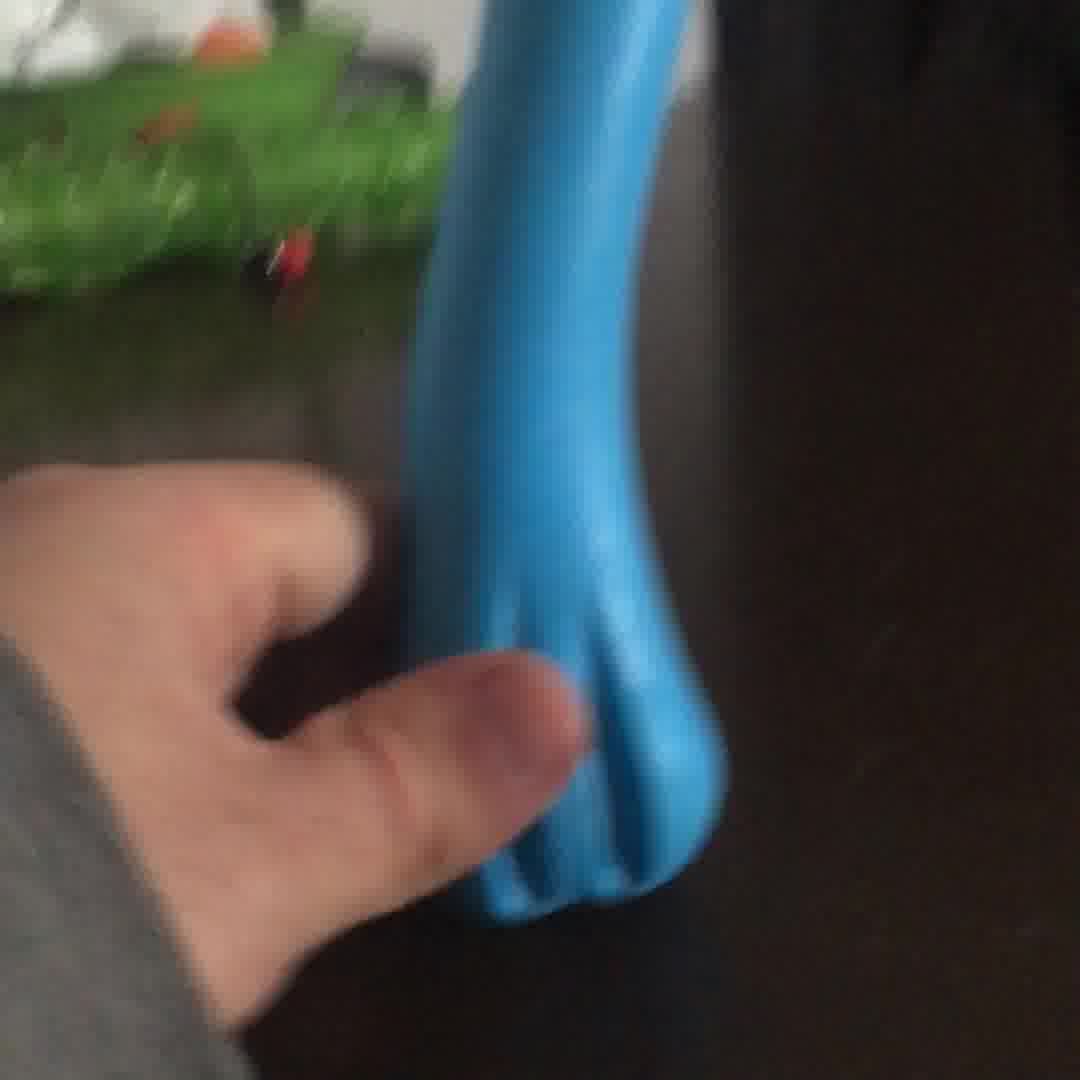}}
    \mbox{\includegraphics[width=0.095\textwidth]{figures/P106-dog-toy/P106--dog-toy--clean--8Fy0uPGuN7MW_btoysF4oQdoS4kb2wCBst_0LeHTJGU-00091.jpg}}}\\
    \vspace*{2px}
    \scalebox{0.95}{
    \mbox{\includegraphics[width=0.095\textwidth]{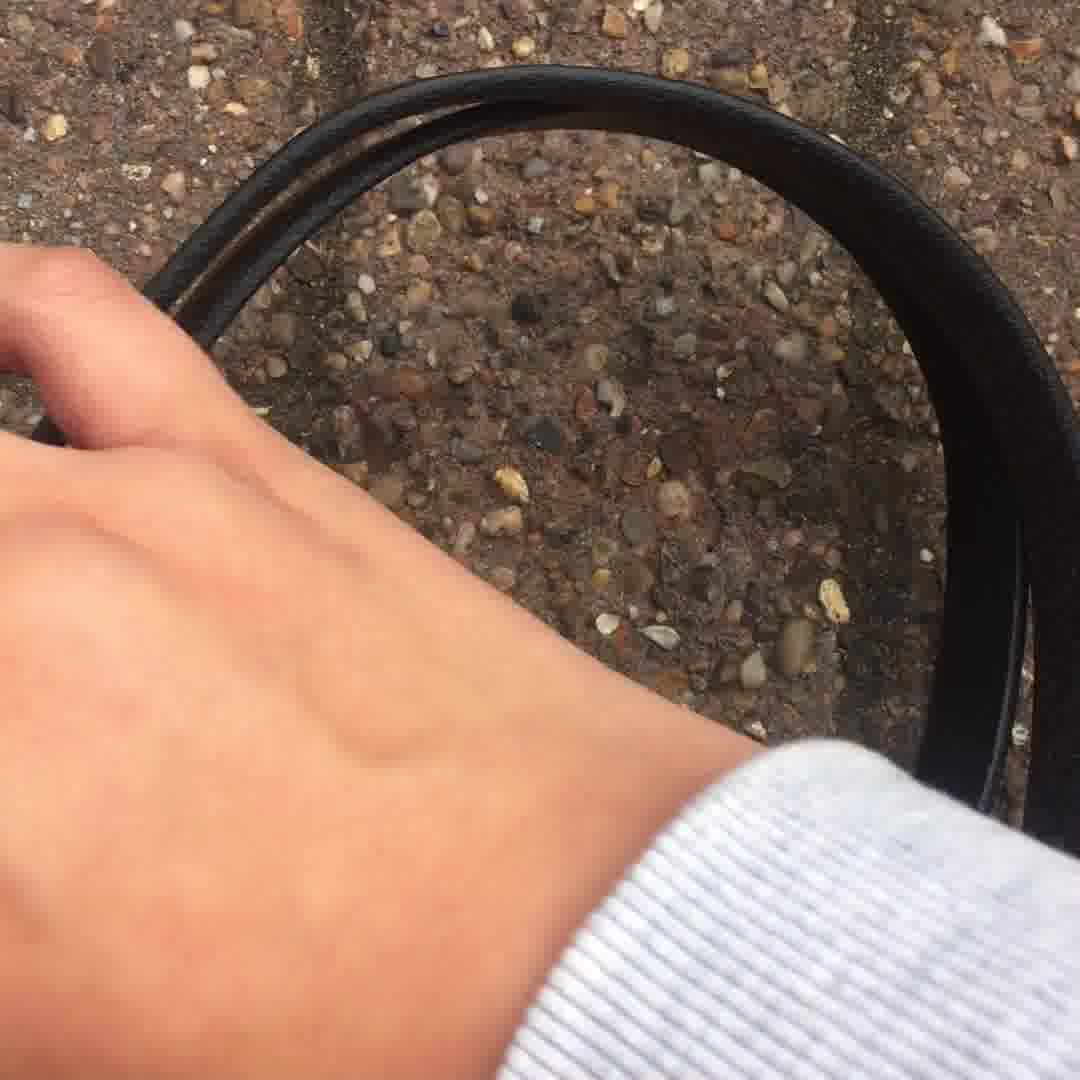}}
    \mbox{\includegraphics[width=0.095\textwidth]{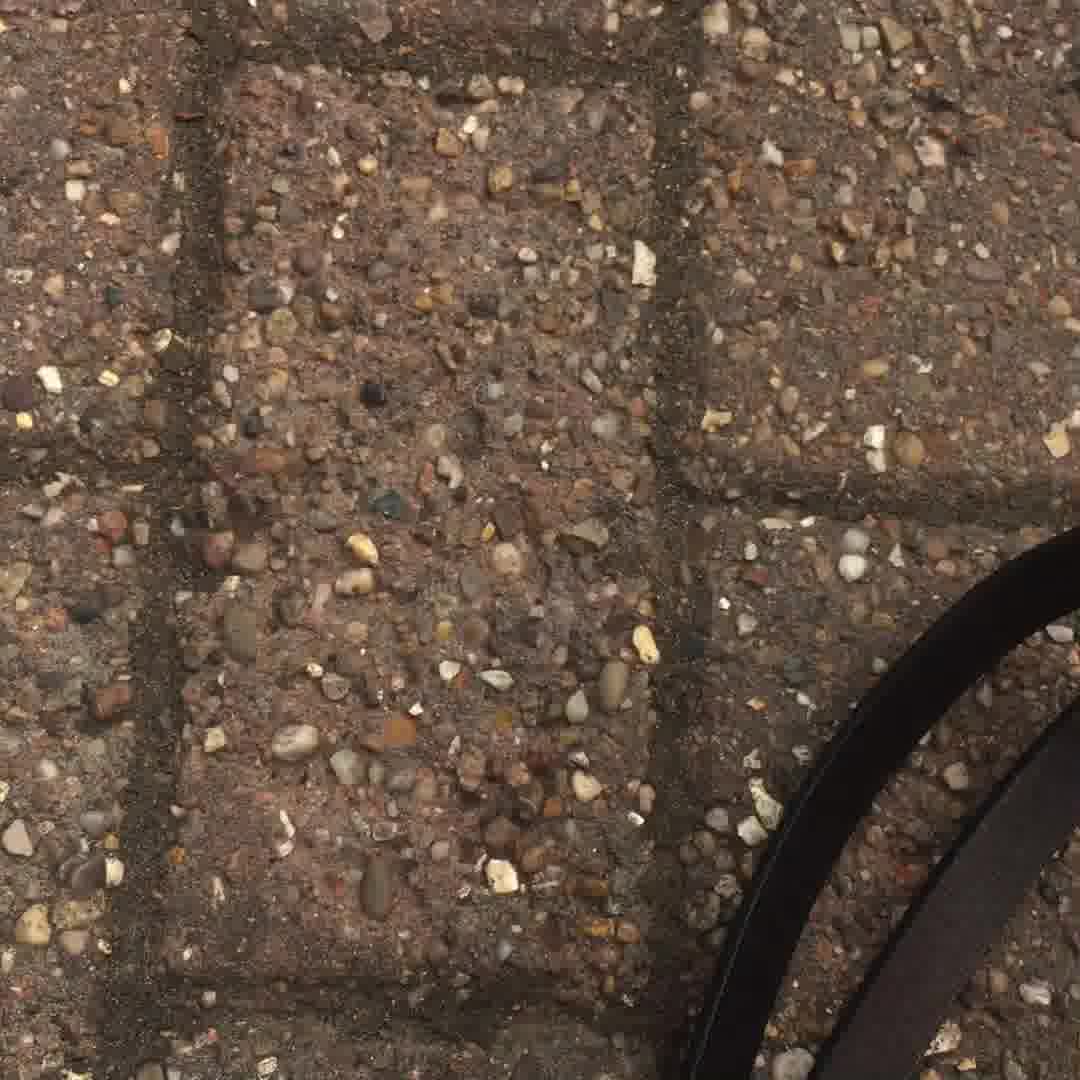}}
    \mbox{\includegraphics[width=0.095\textwidth]{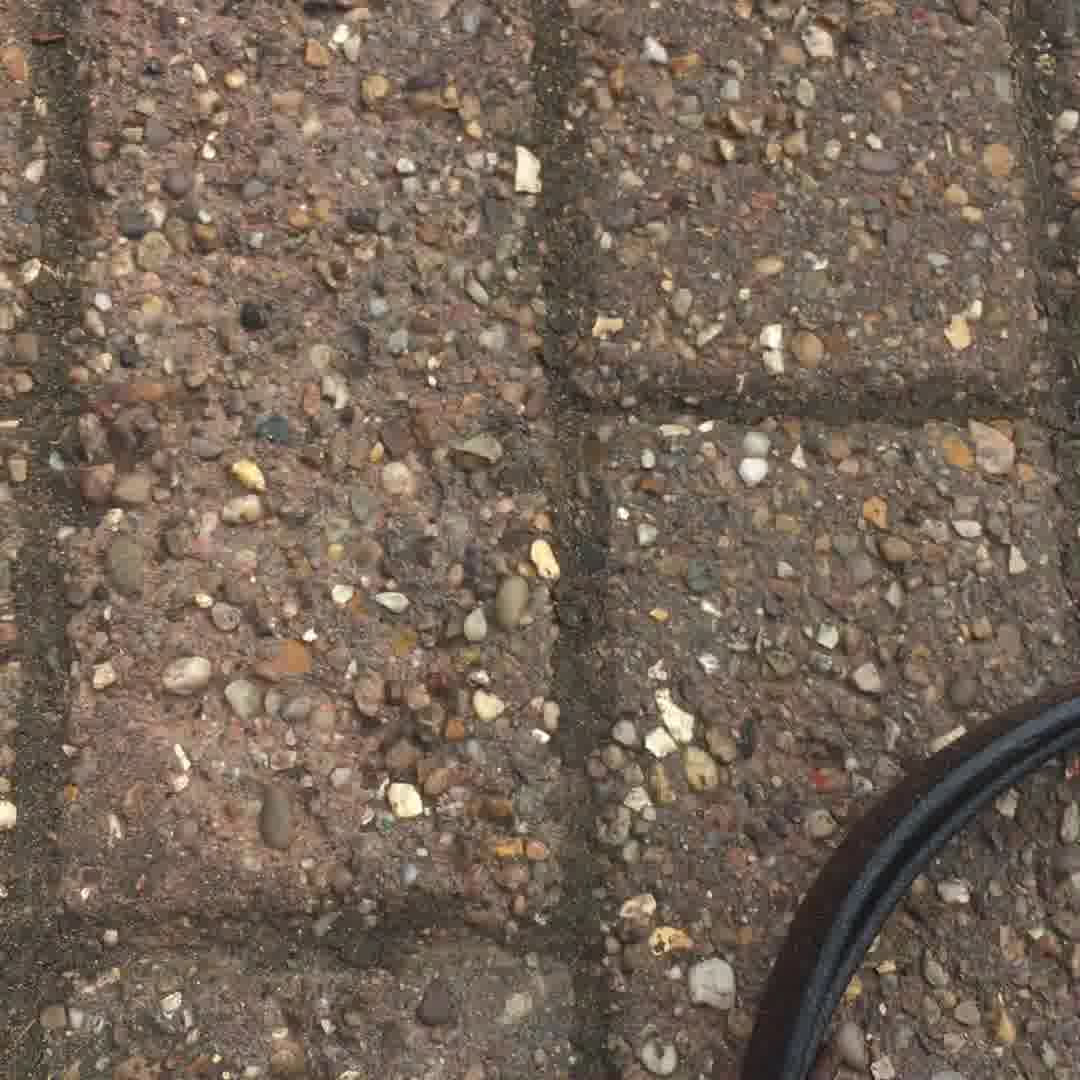}}
    \mbox{\includegraphics[width=0.095\textwidth]{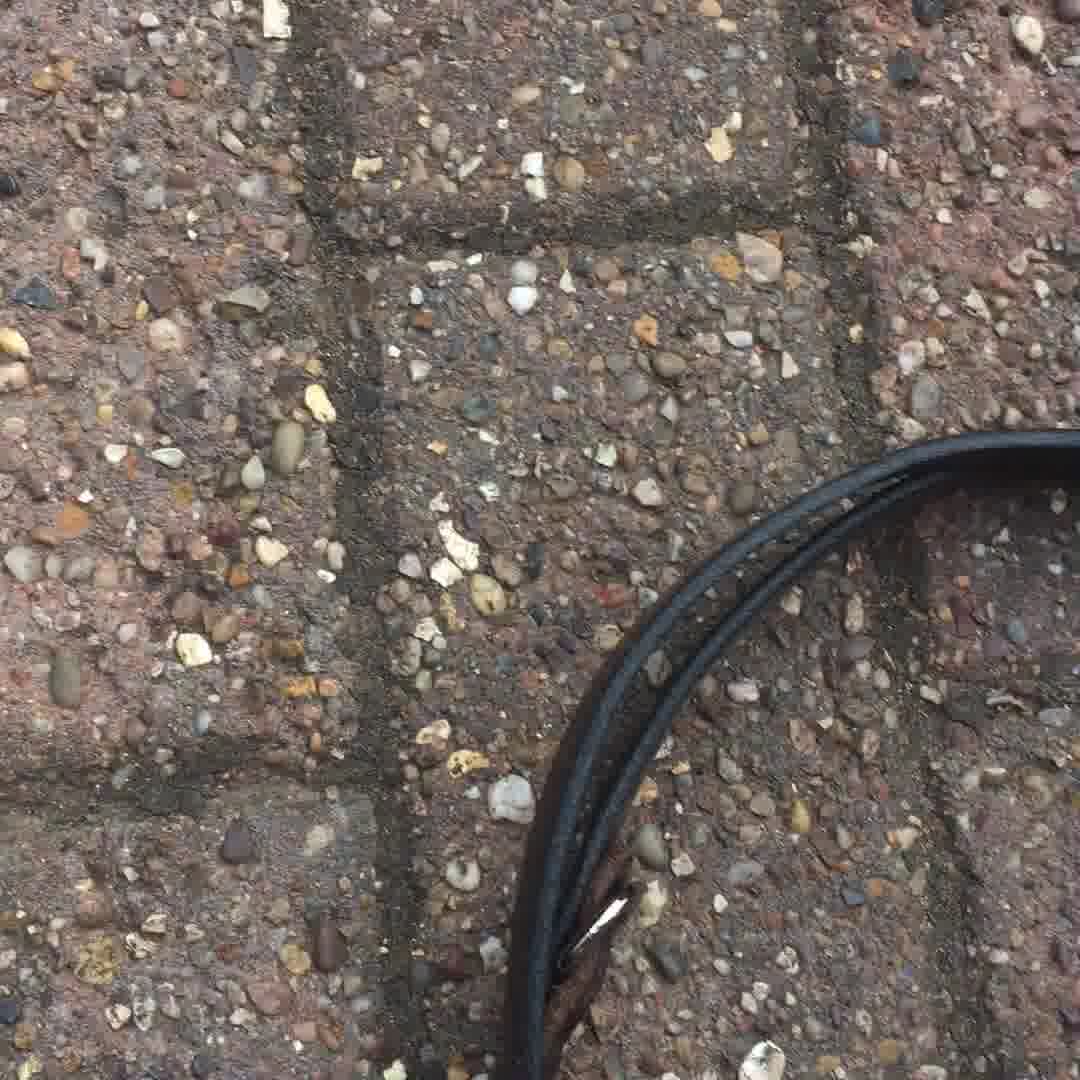}}
    \mbox{\includegraphics[width=0.095\textwidth]{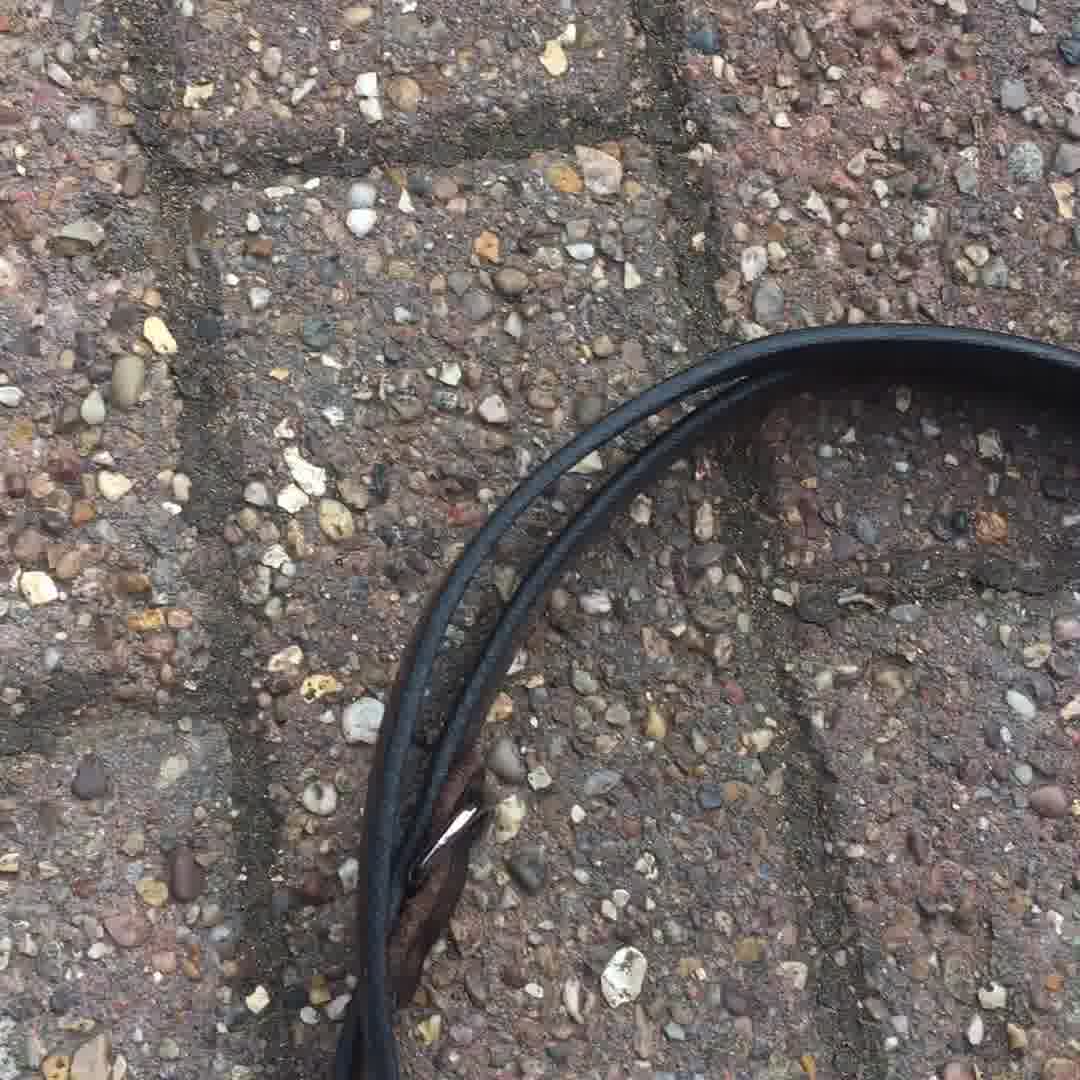}}
    \mbox{\includegraphics[width=0.095\textwidth]{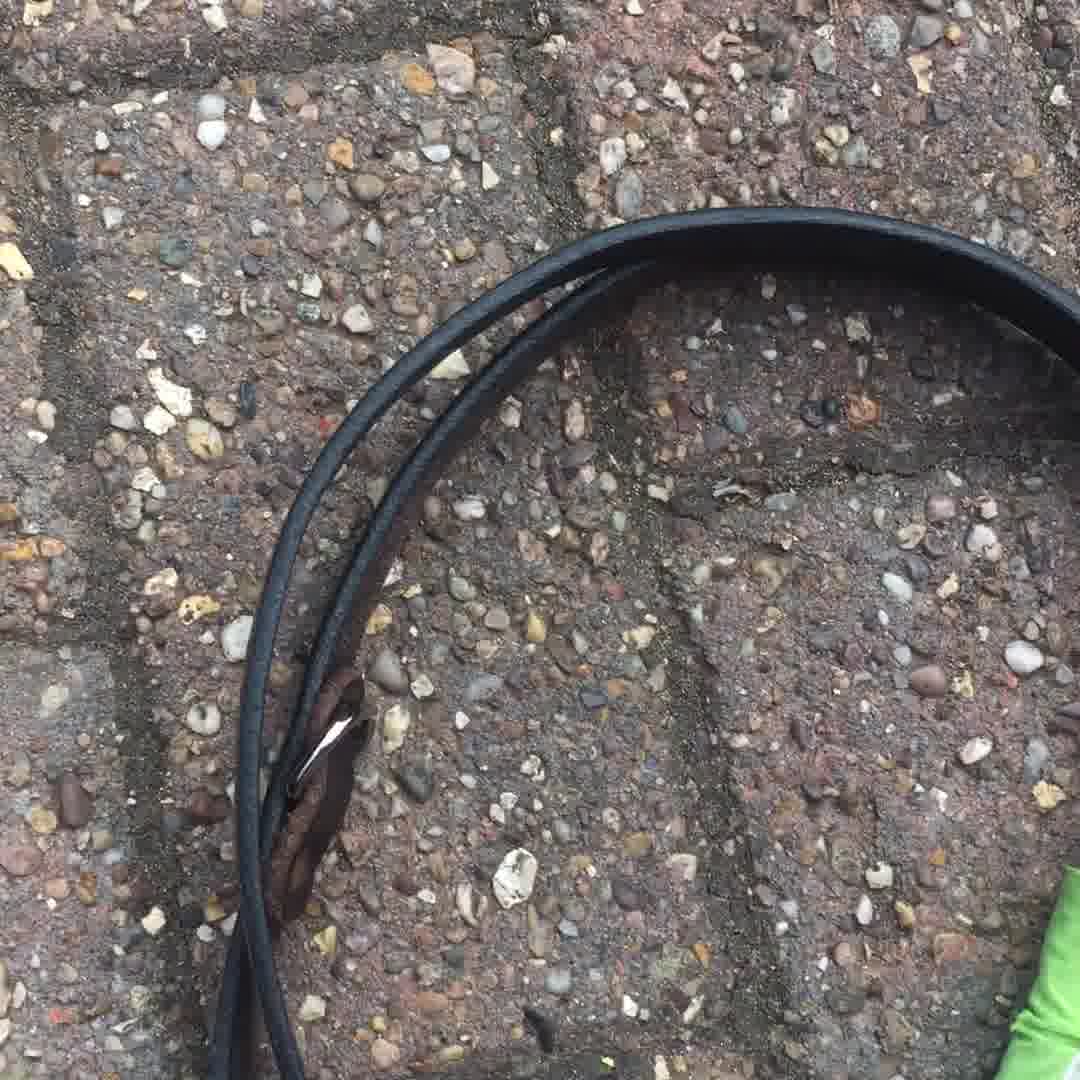}}
    \mbox{\includegraphics[width=0.095\textwidth]{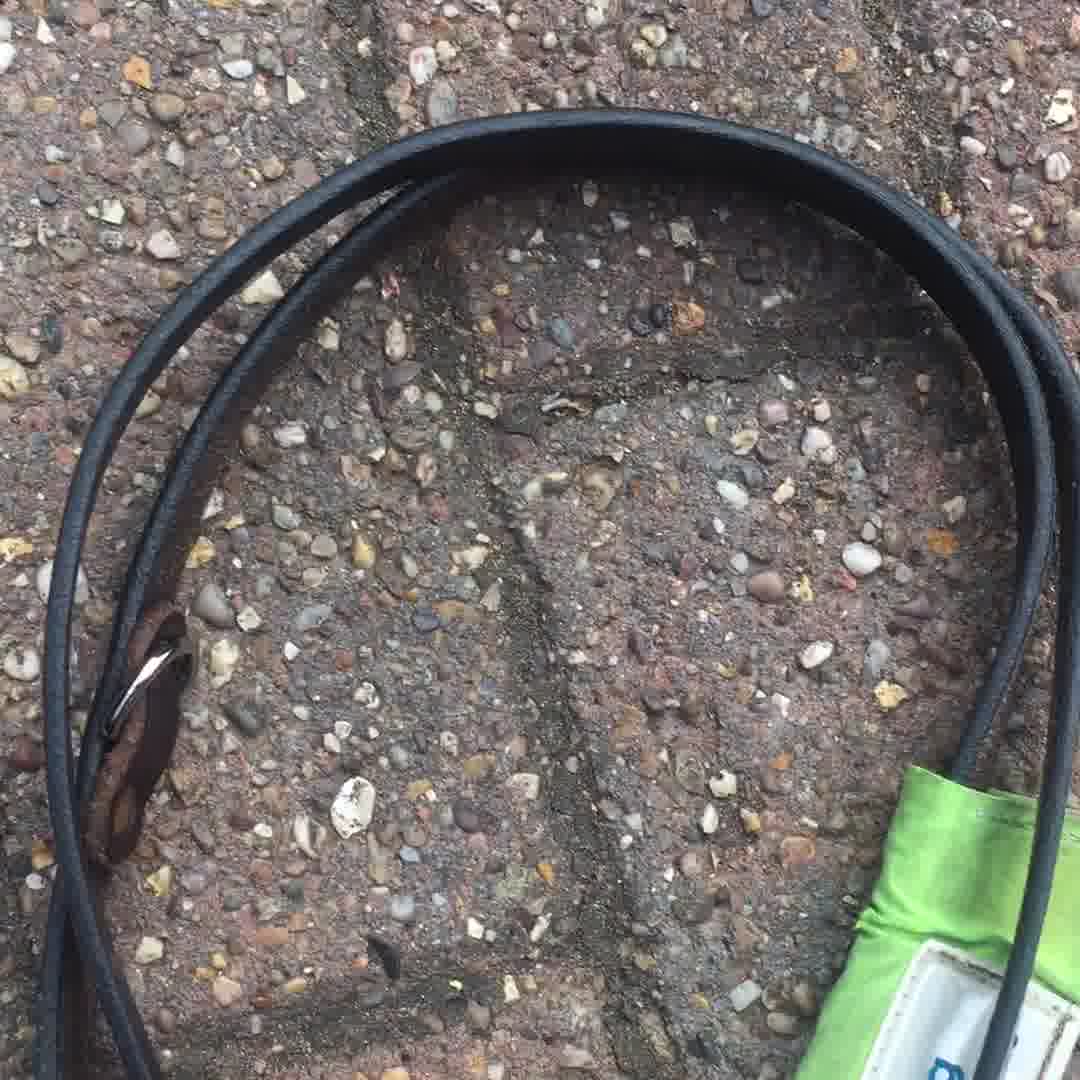}}
    \mbox{\includegraphics[width=0.095\textwidth]{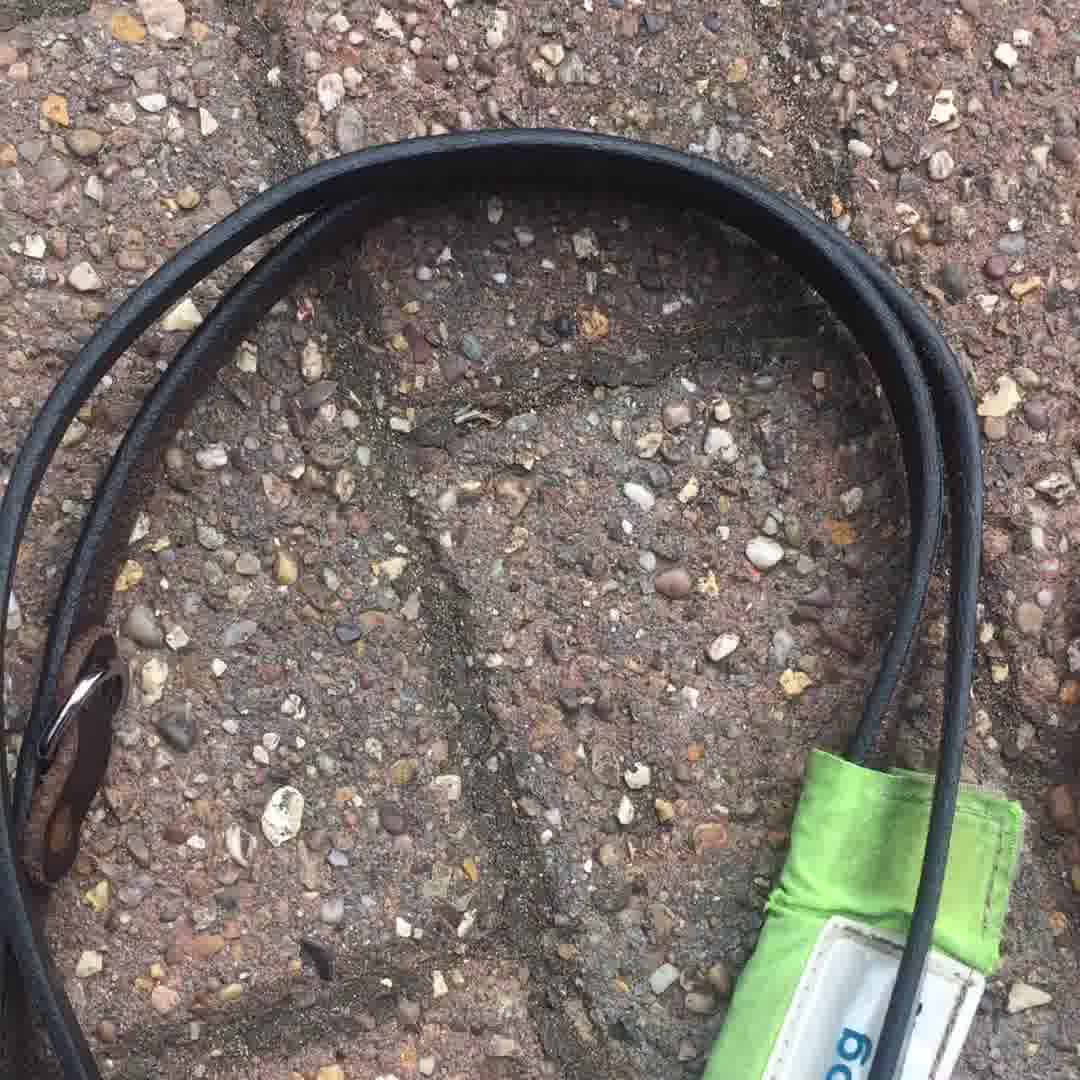}}
    \mbox{\includegraphics[width=0.095\textwidth]{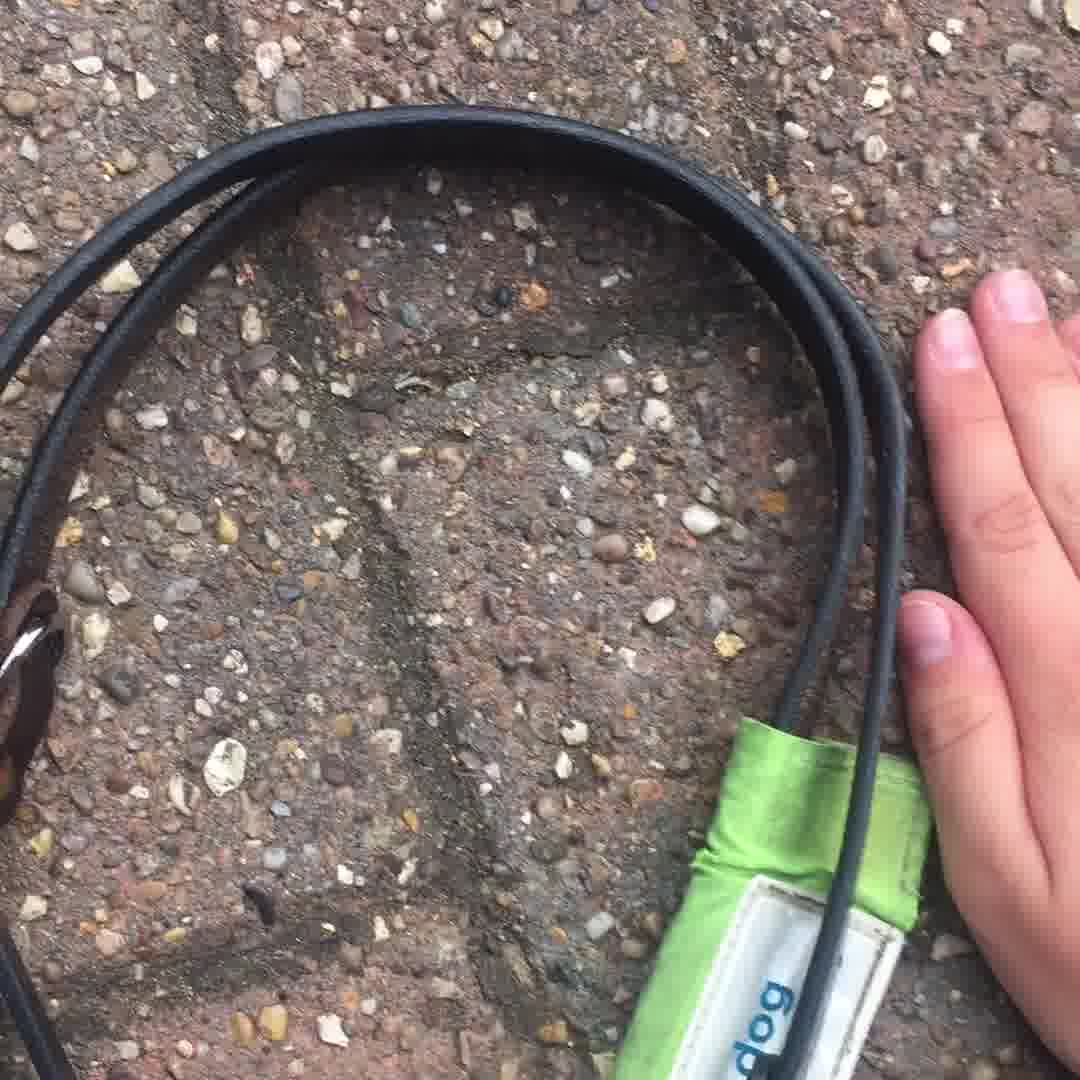}}
    \mbox{\includegraphics[width=0.095\textwidth]{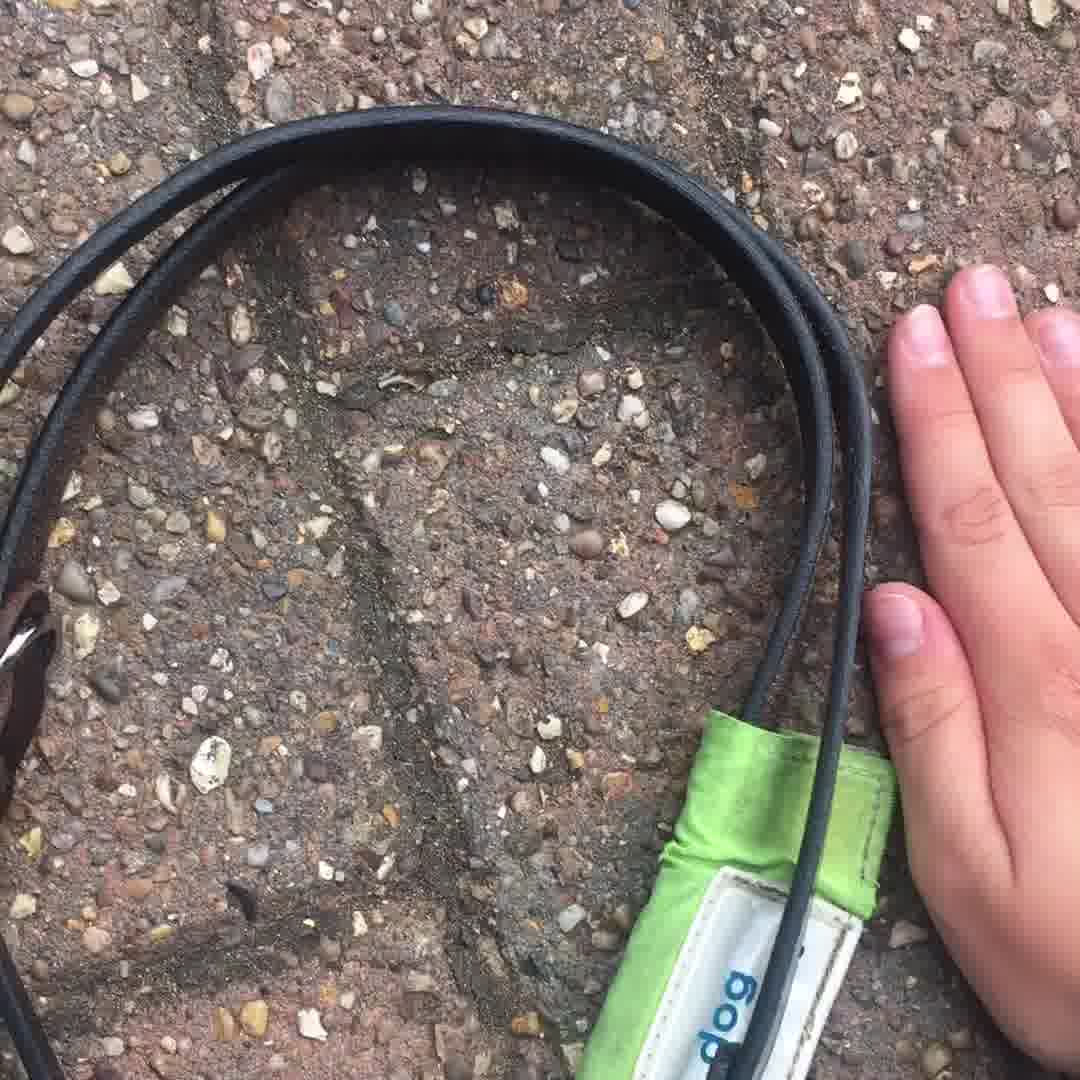}}}\\
    \vspace*{2px}
    \scalebox{0.95}{
    \mbox{\includegraphics[width=0.095\textwidth]{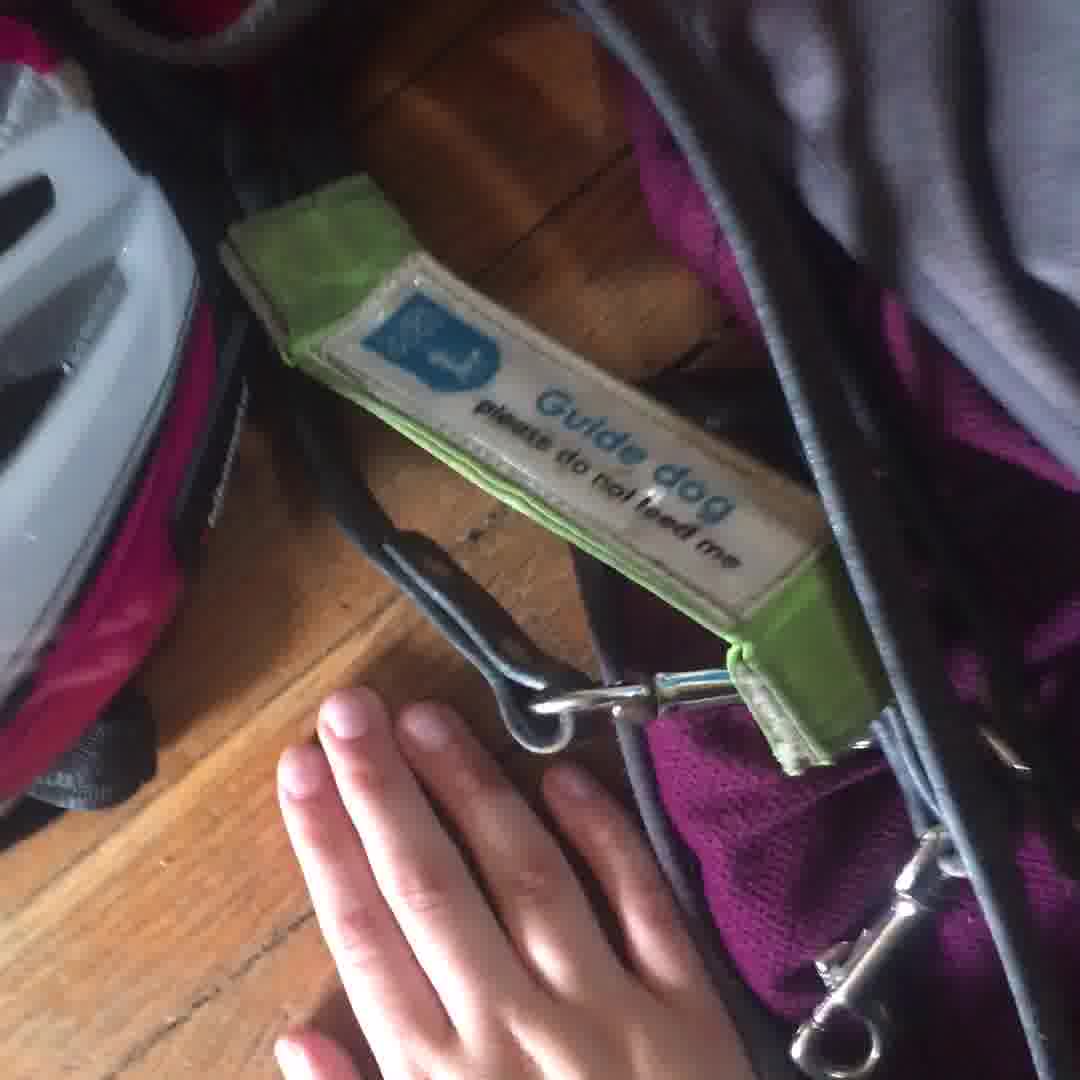}}
    \mbox{\includegraphics[width=0.095\textwidth]{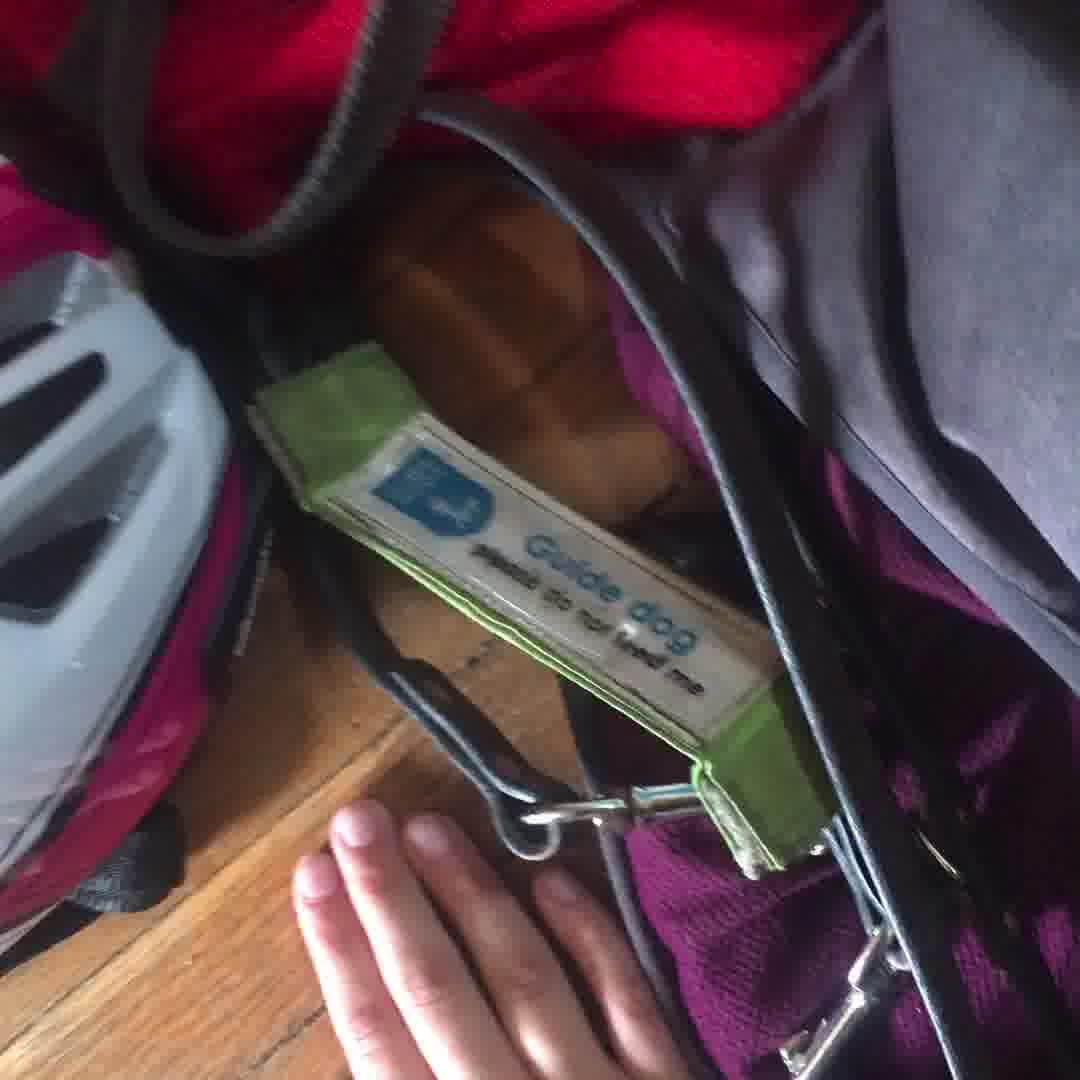}}
    \mbox{\includegraphics[width=0.095\textwidth]{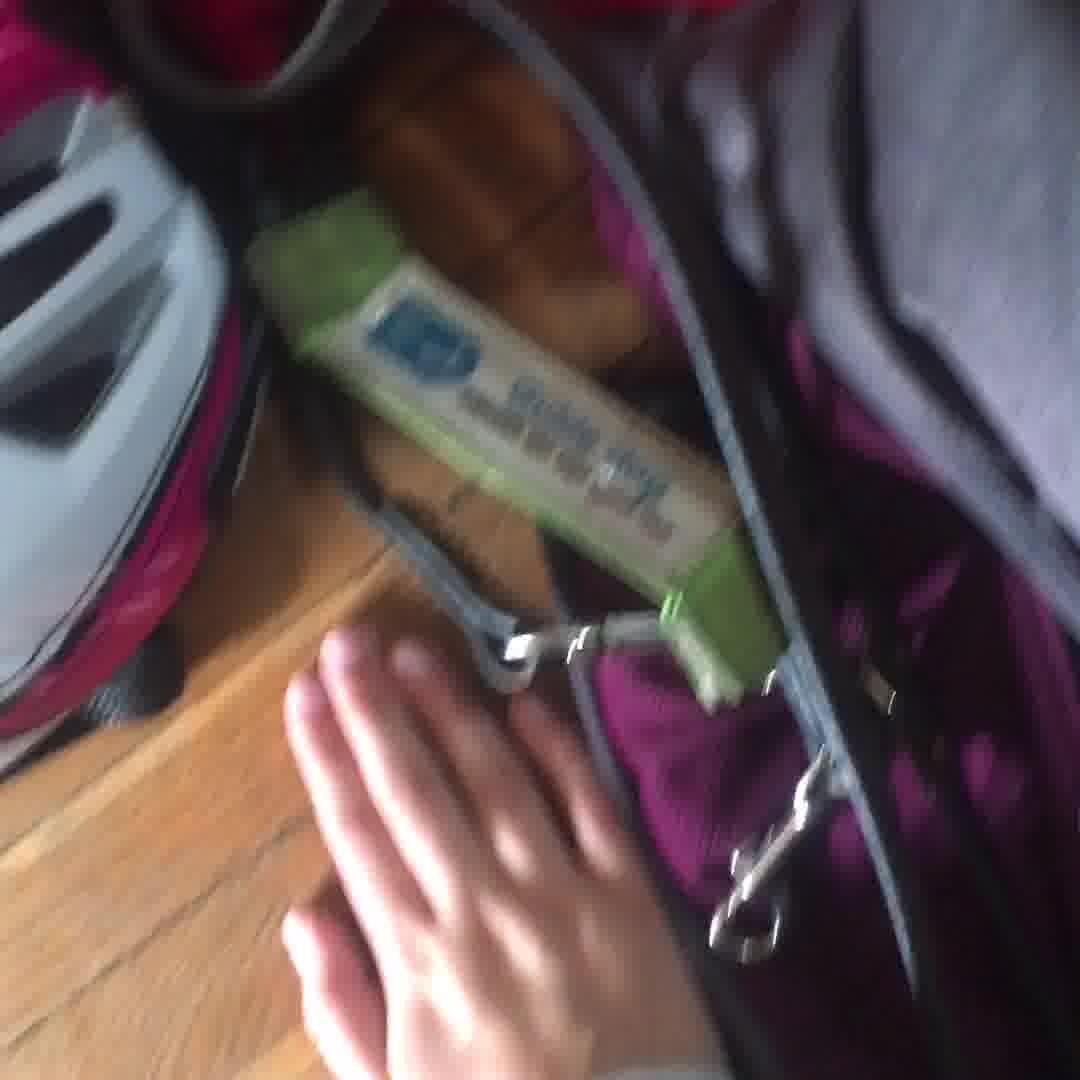}}
    \mbox{\includegraphics[width=0.095\textwidth]{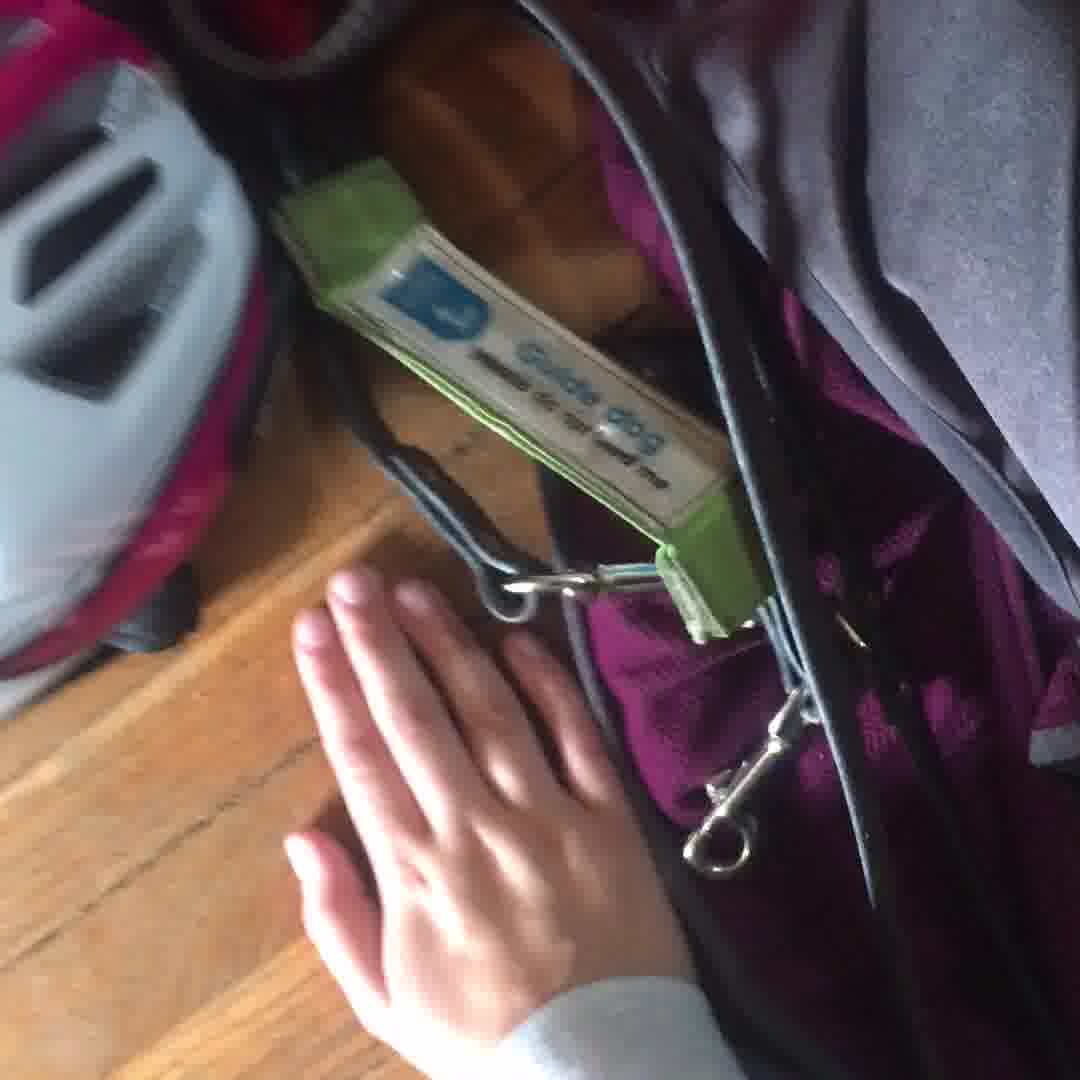}}
    \mbox{\includegraphics[width=0.095\textwidth]{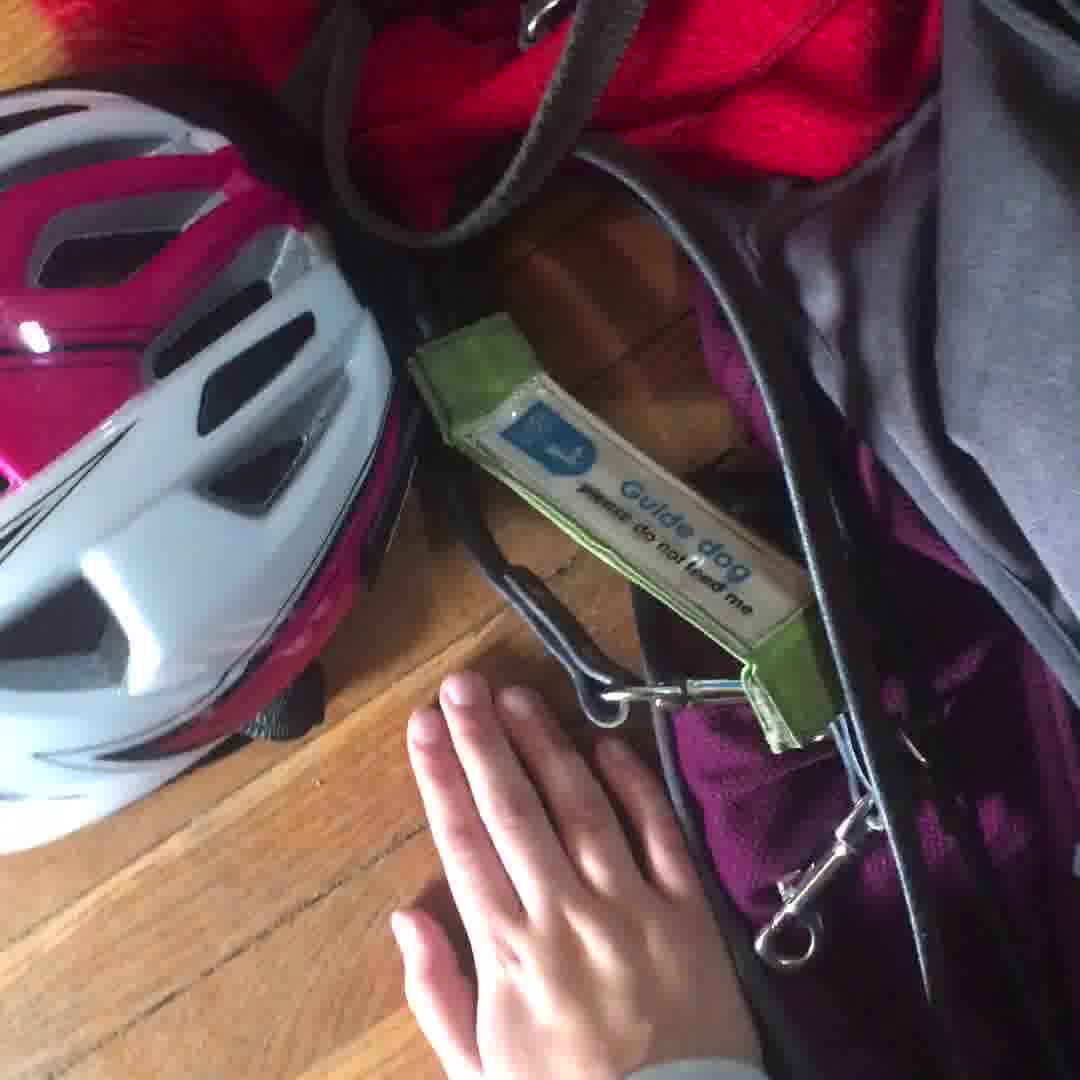}}
    \mbox{\includegraphics[width=0.095\textwidth]{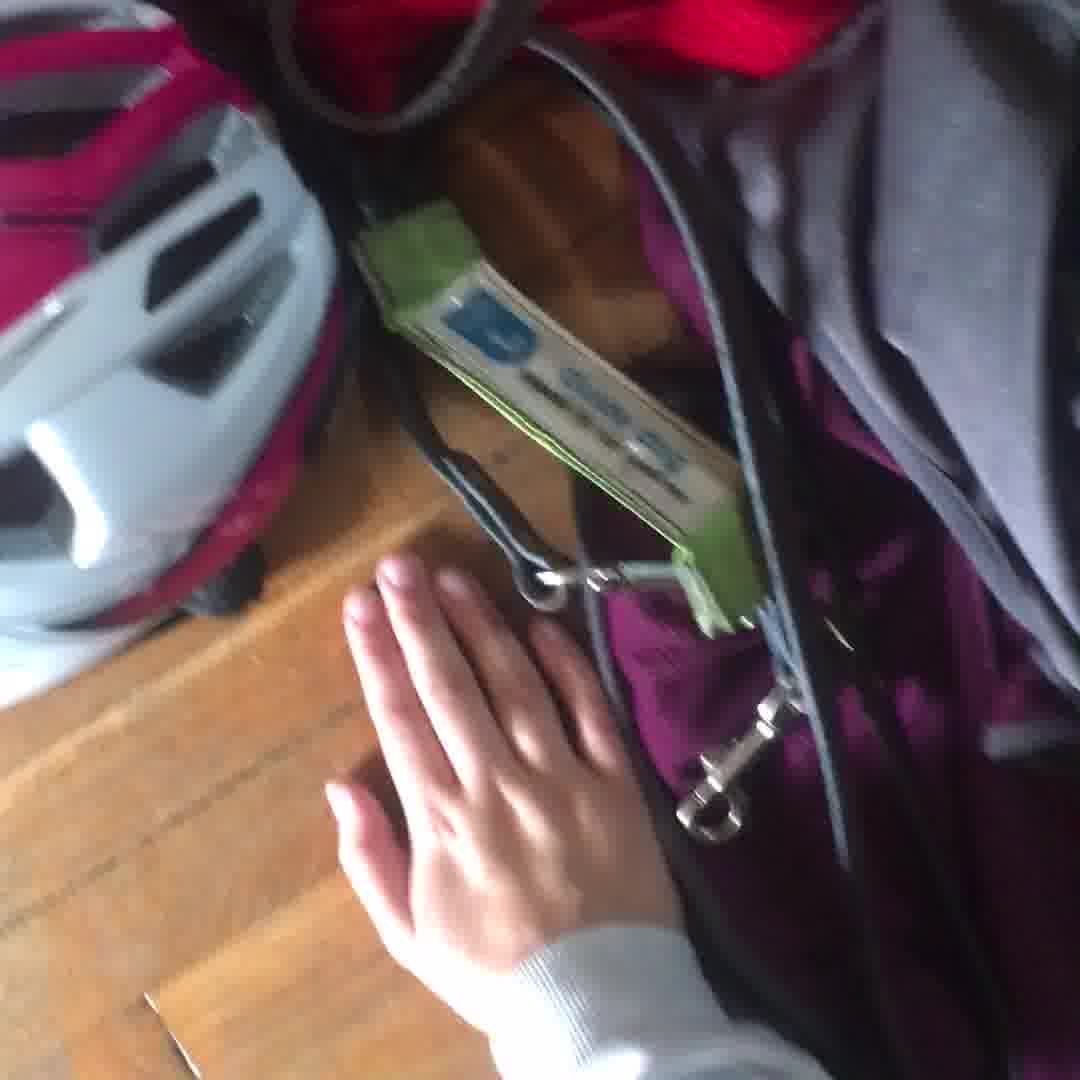}}
    \mbox{\includegraphics[width=0.095\textwidth]{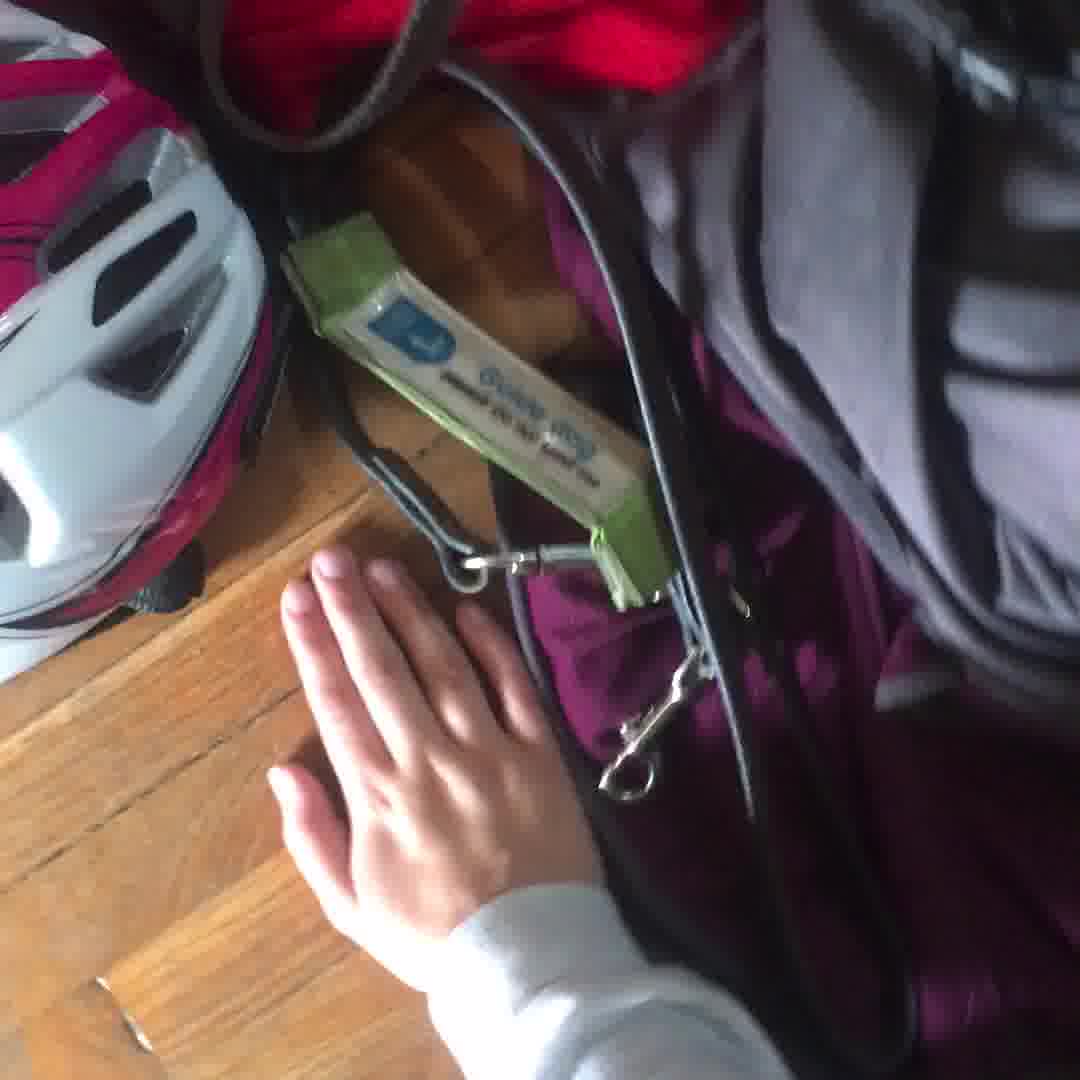}}
    \mbox{\includegraphics[width=0.095\textwidth]{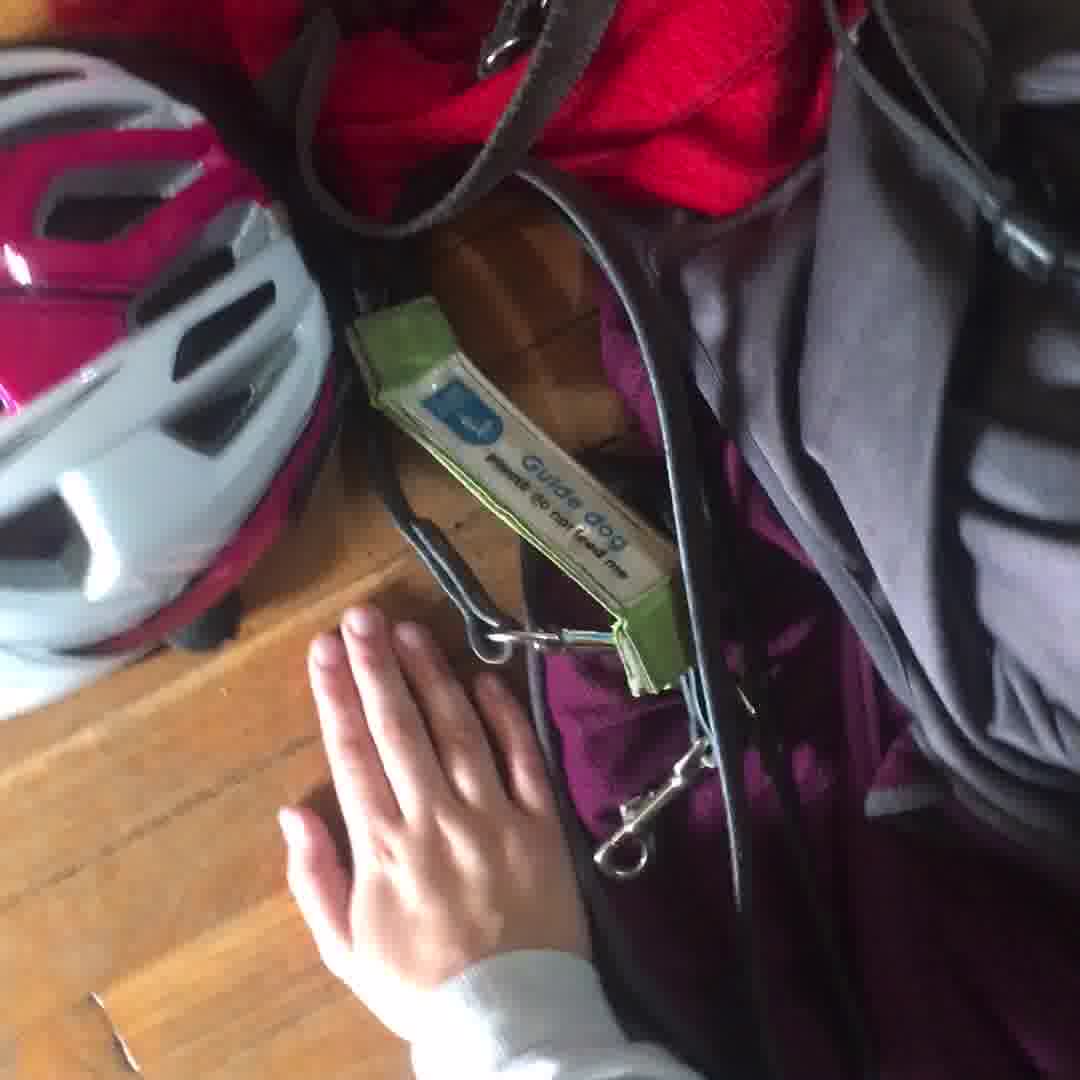}}
    \mbox{\includegraphics[width=0.095\textwidth]{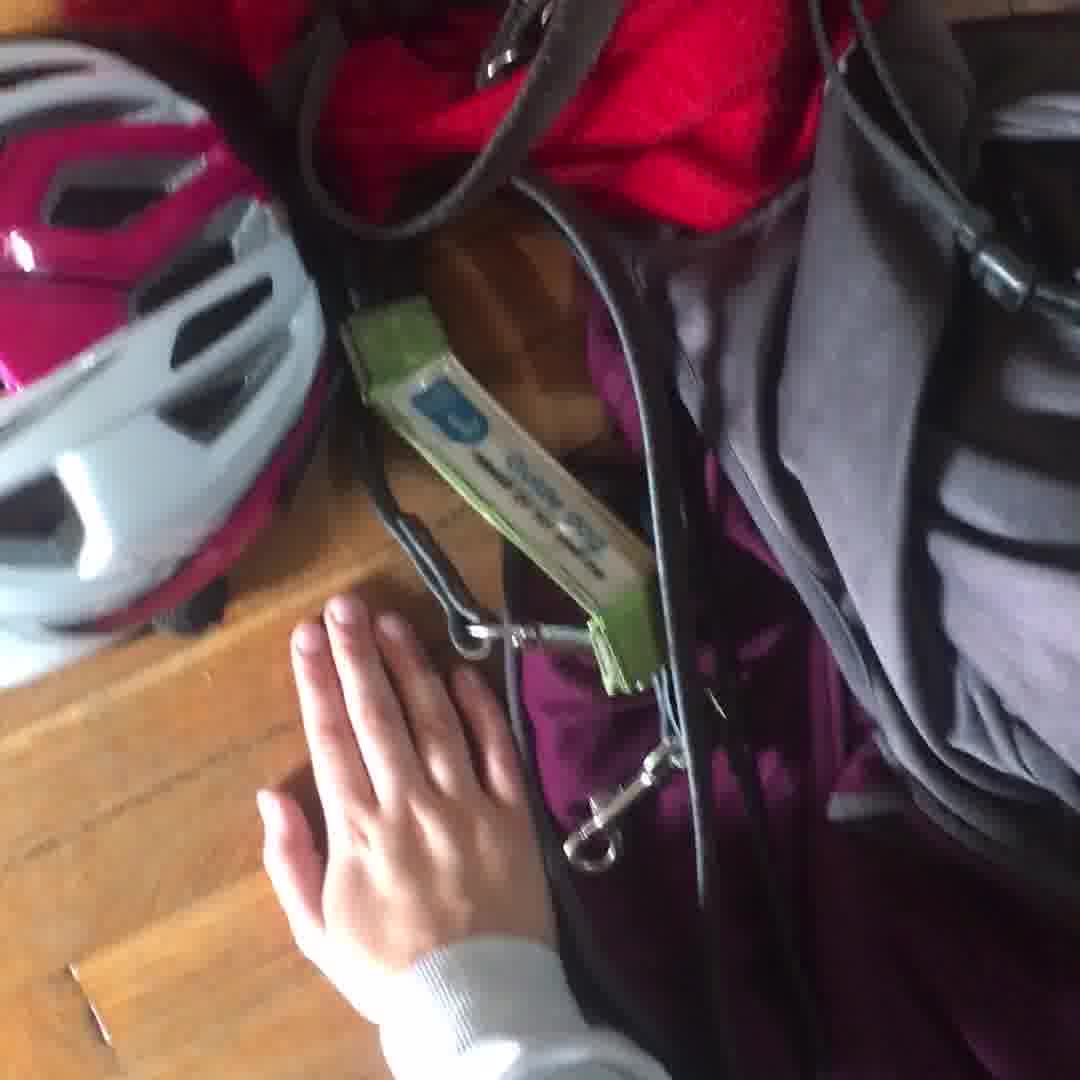}}
    \mbox{\includegraphics[width=0.095\textwidth]{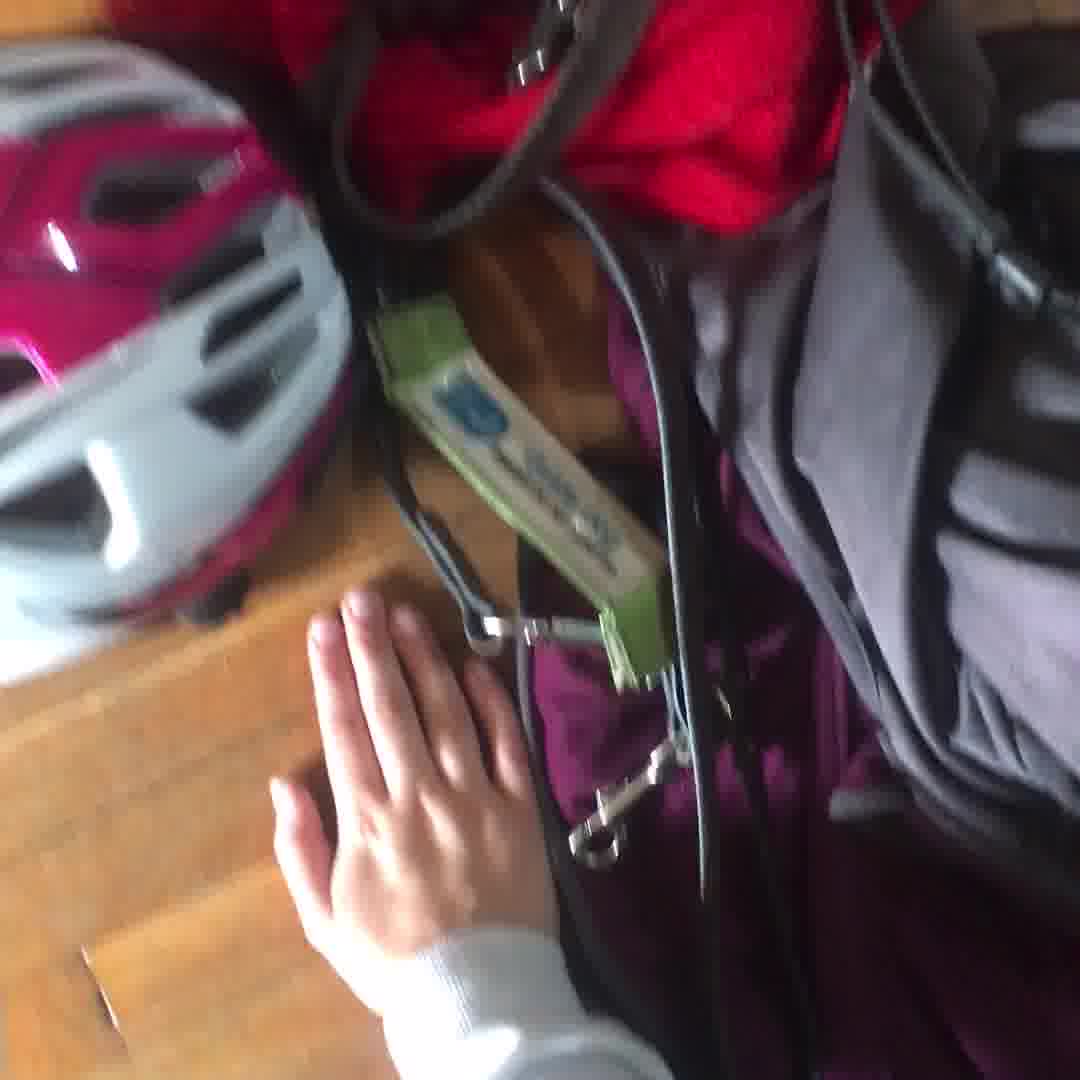}}}\\
    \vspace*{2px}
    \scalebox{0.95}{
    \mbox{\includegraphics[width=0.095\textwidth]{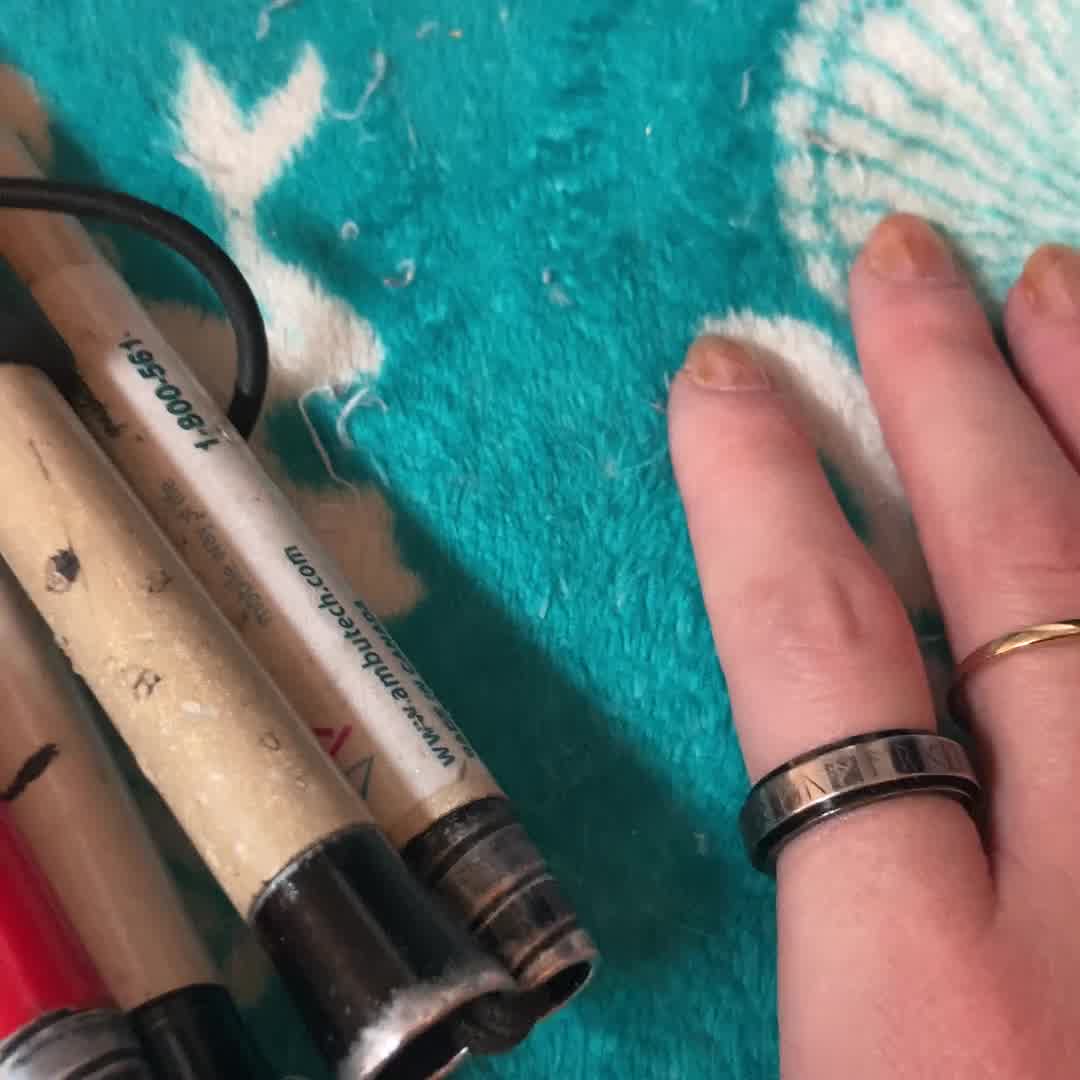}}
    \mbox{\includegraphics[width=0.095\textwidth]{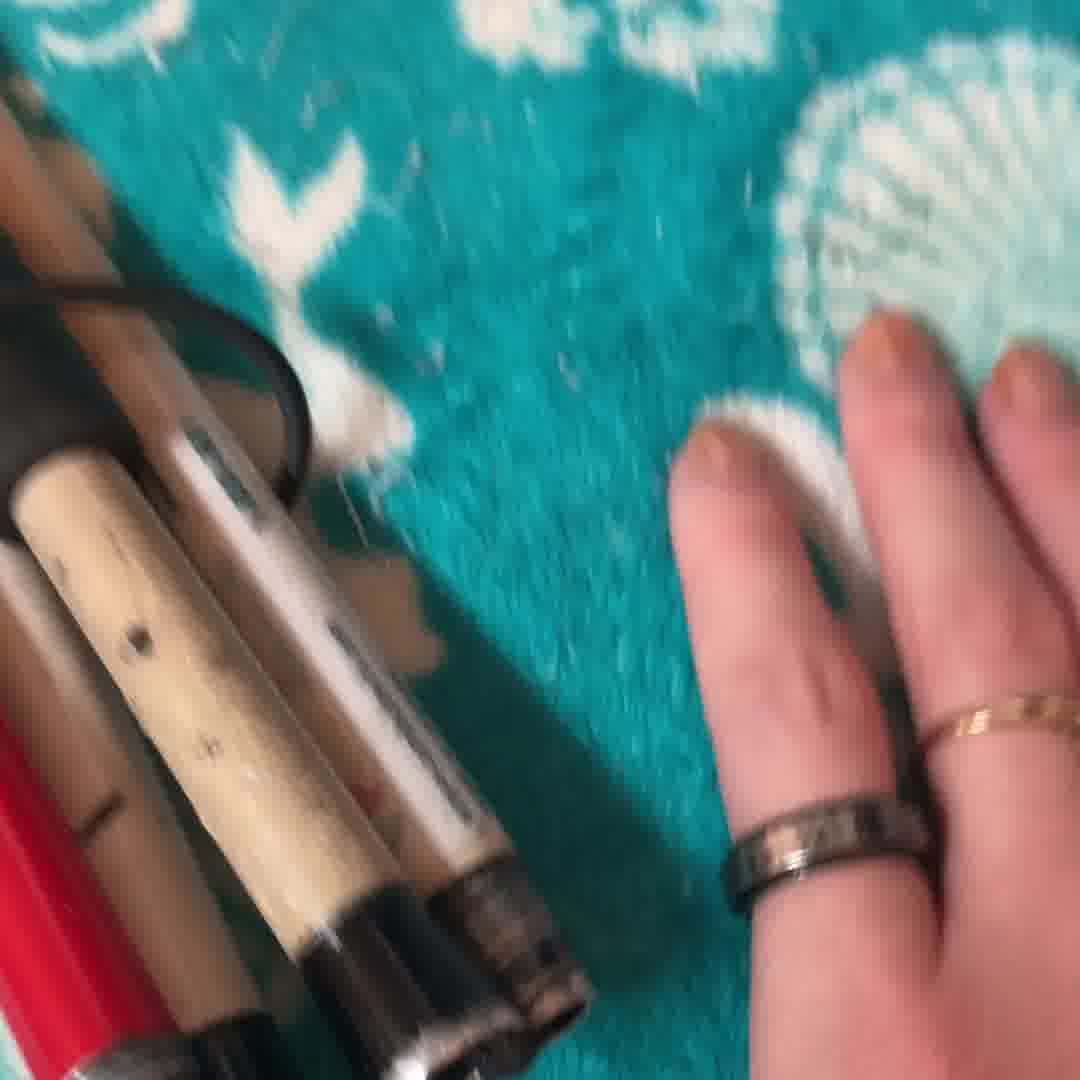}}
    \mbox{\includegraphics[width=0.095\textwidth]{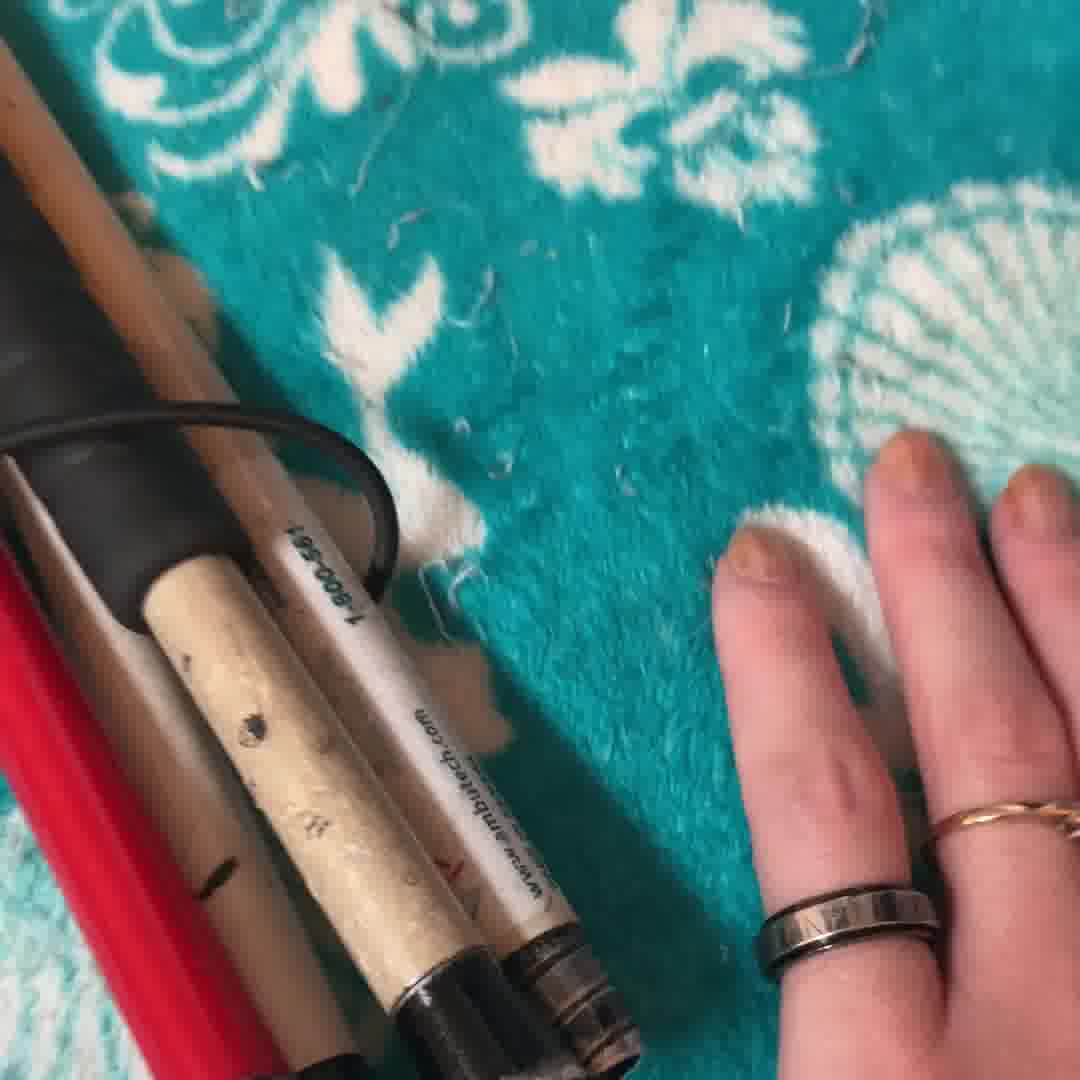}}
    \mbox{\includegraphics[width=0.095\textwidth]{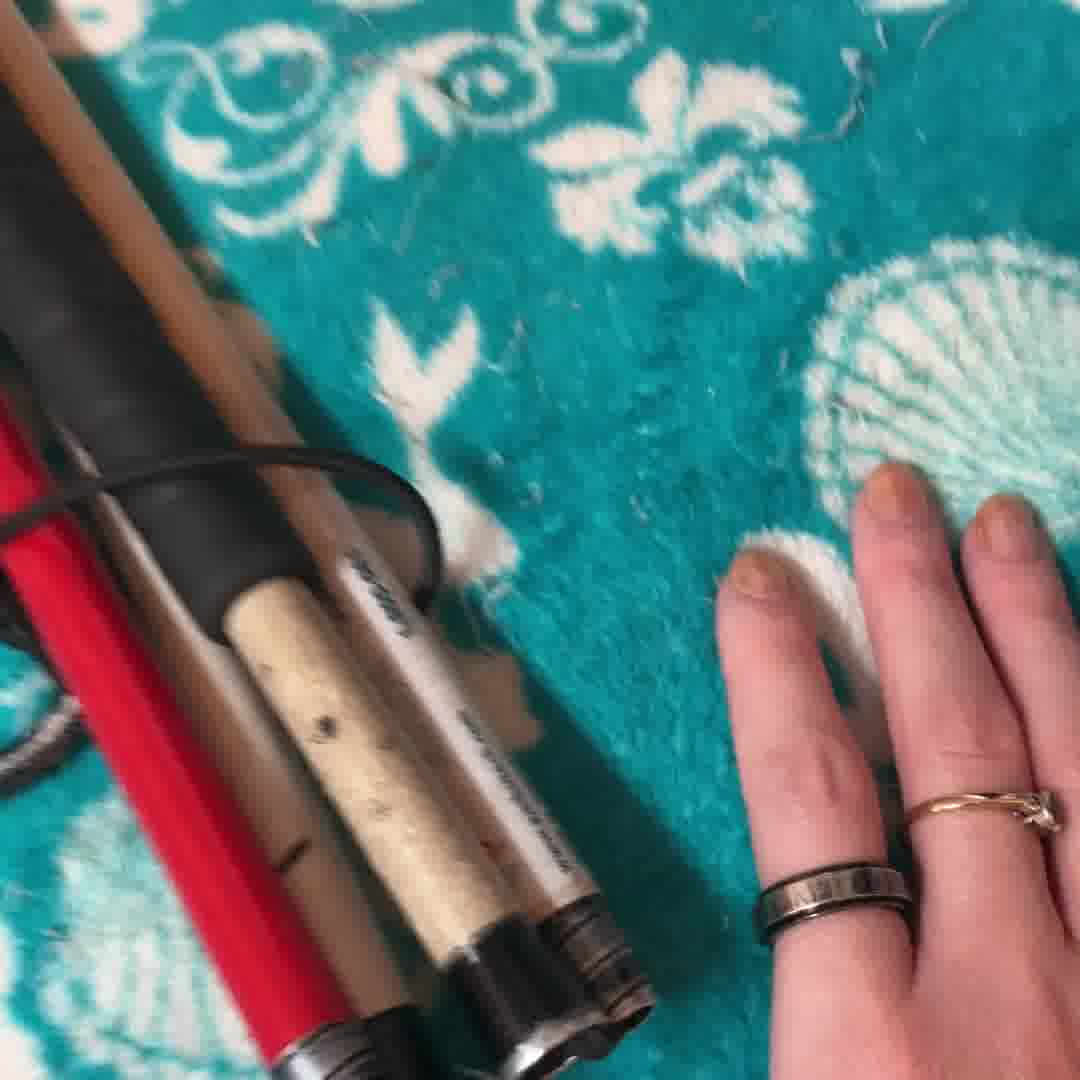}}
    \mbox{\includegraphics[width=0.095\textwidth]{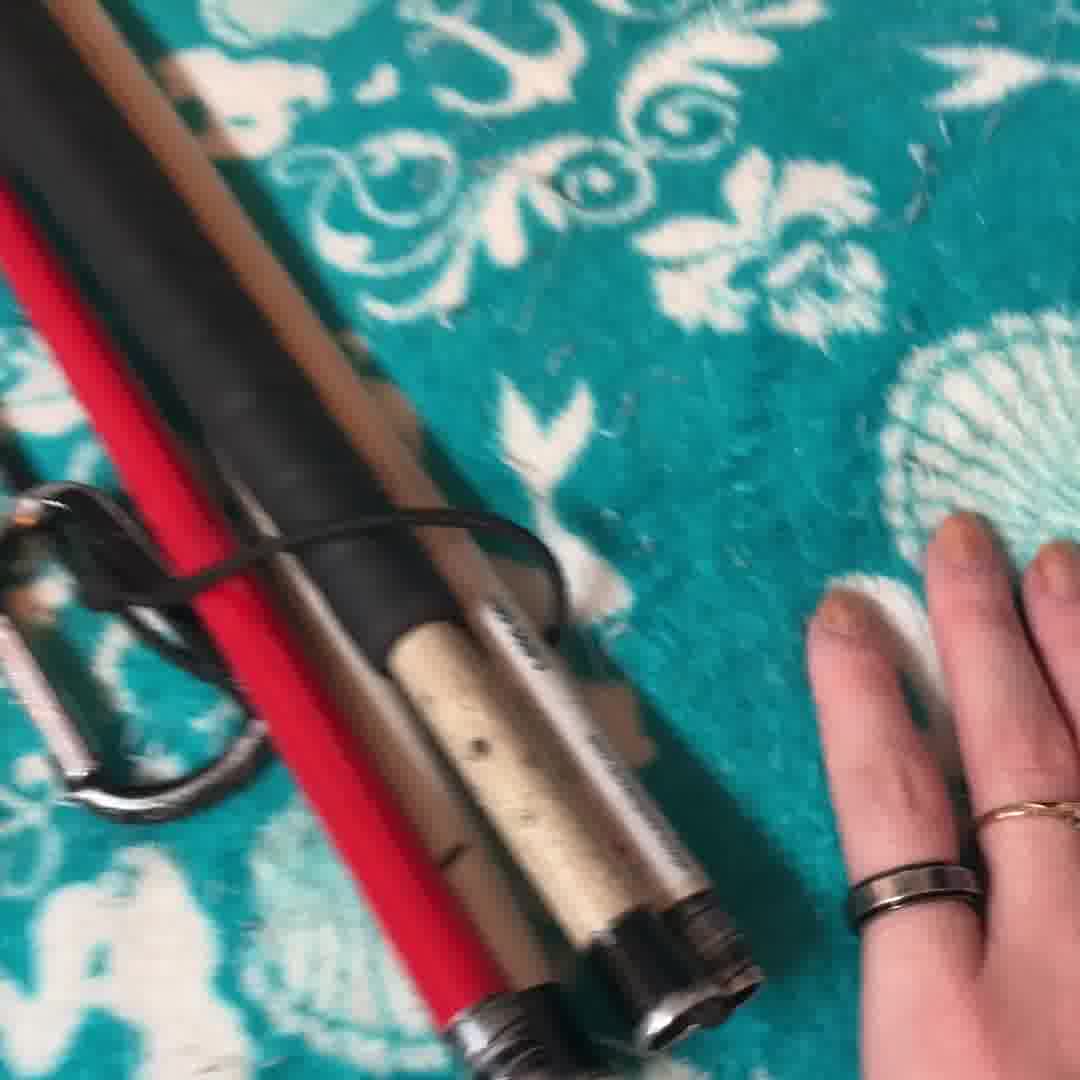}}
    \mbox{\includegraphics[width=0.095\textwidth]{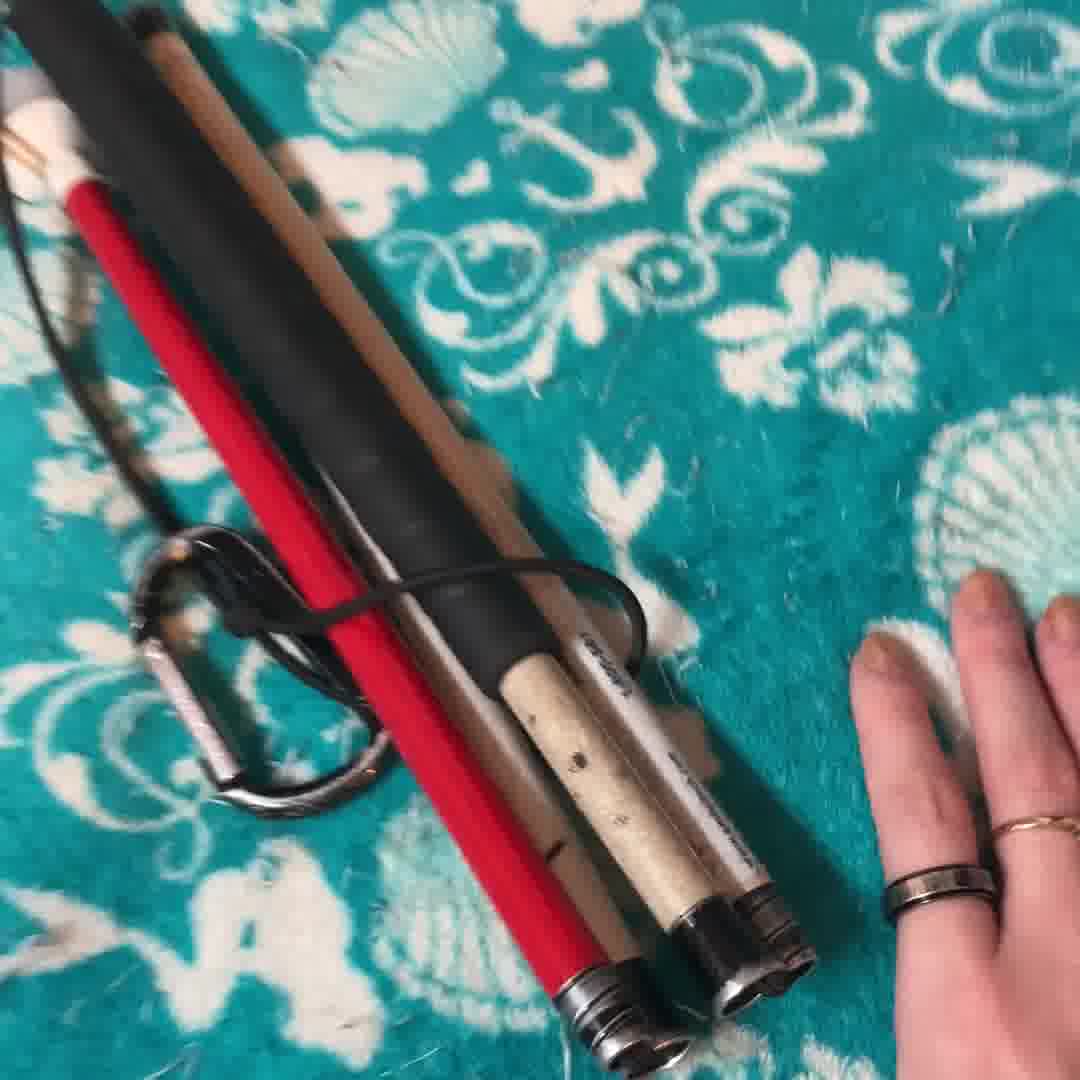}}
    \mbox{\includegraphics[width=0.095\textwidth]{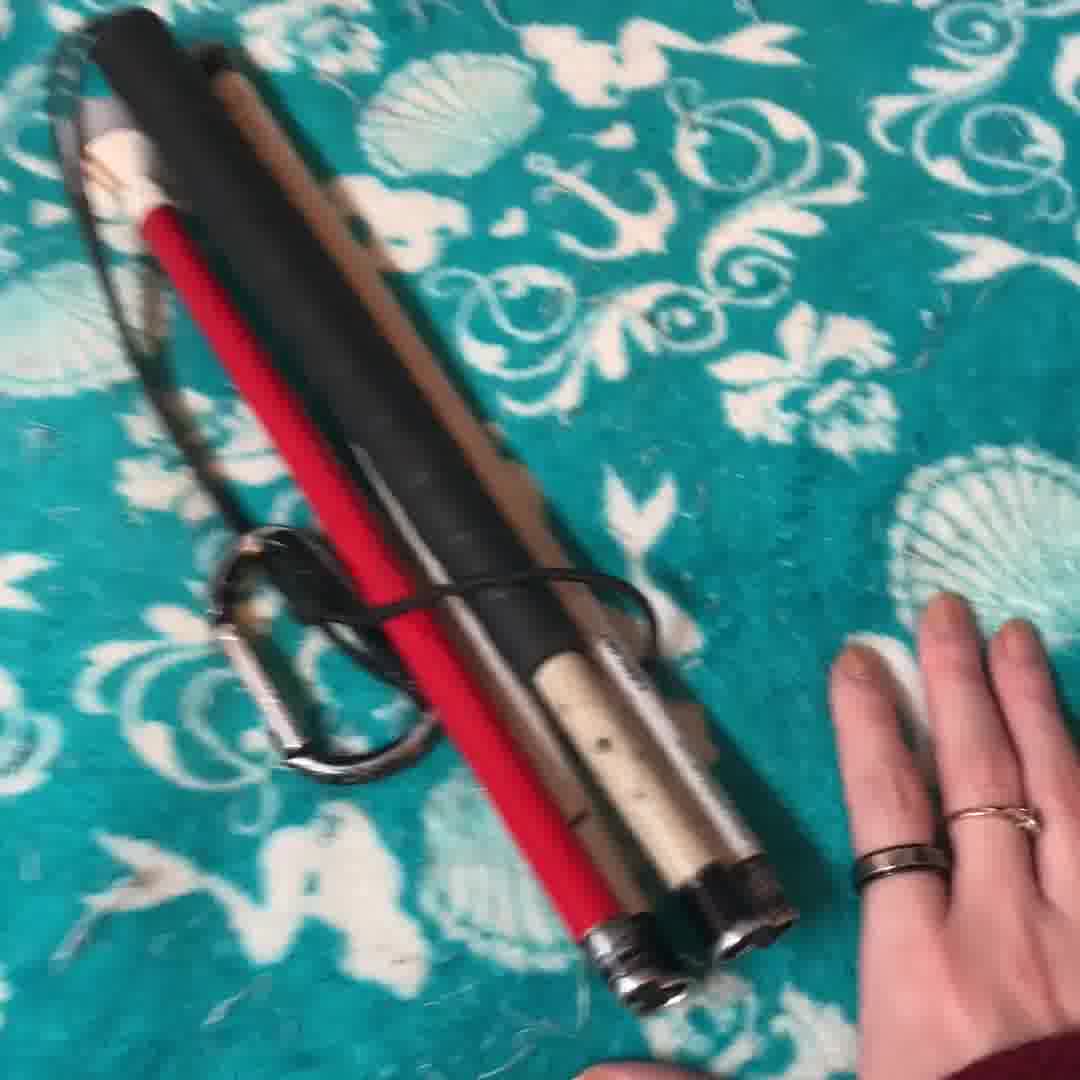}}
    \mbox{\includegraphics[width=0.095\textwidth]{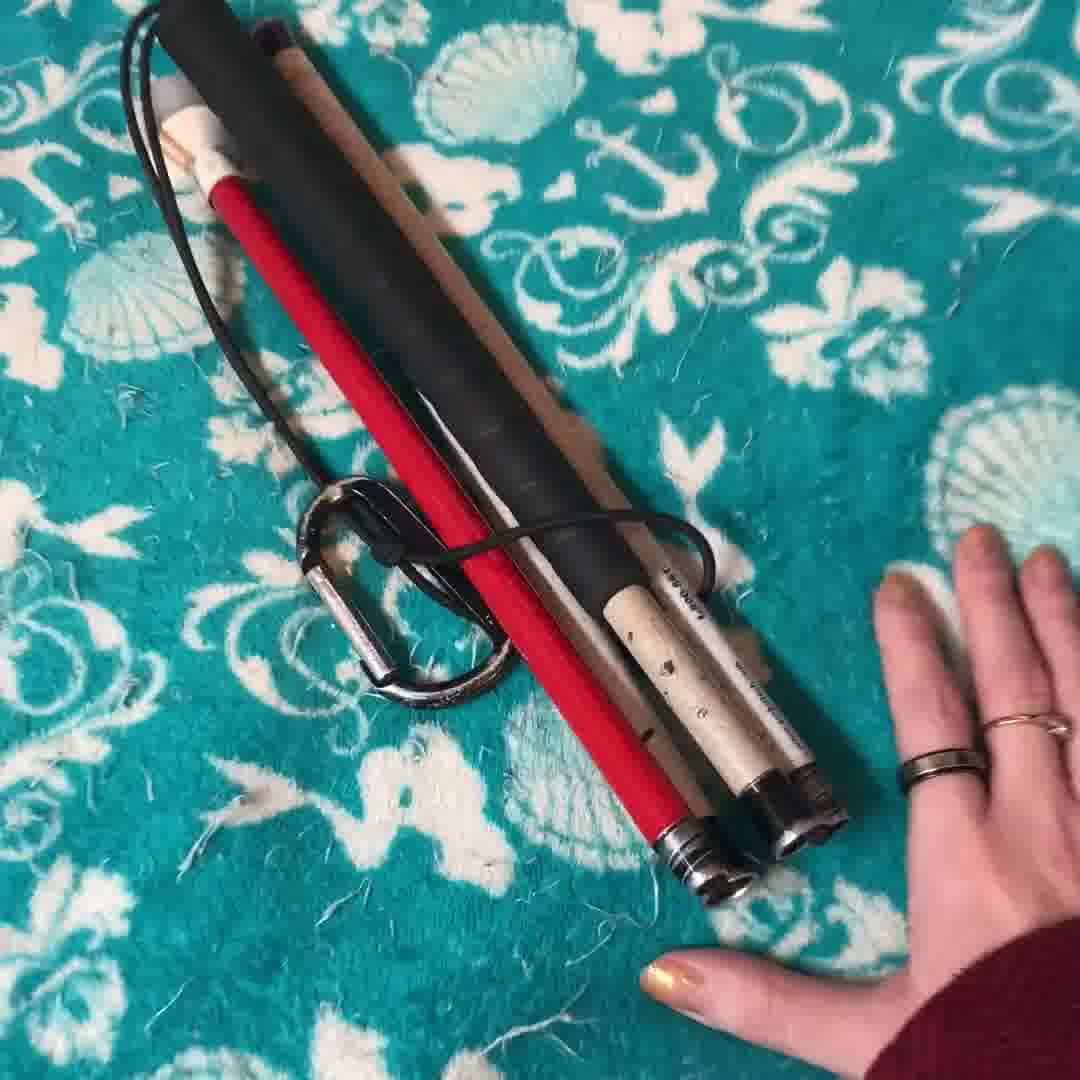}}
    \mbox{\includegraphics[width=0.095\textwidth]{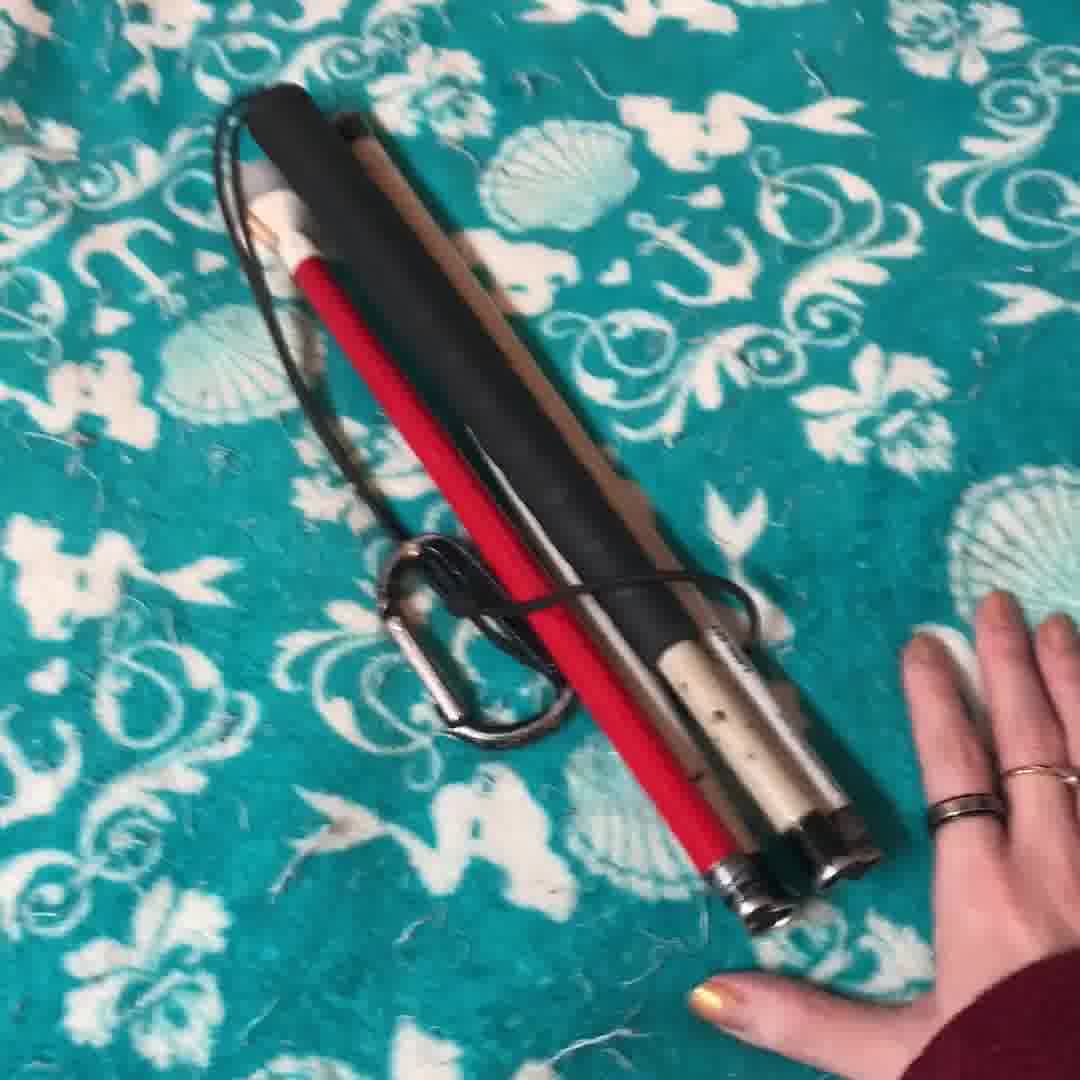}}
    \mbox{\includegraphics[width=0.095\textwidth]{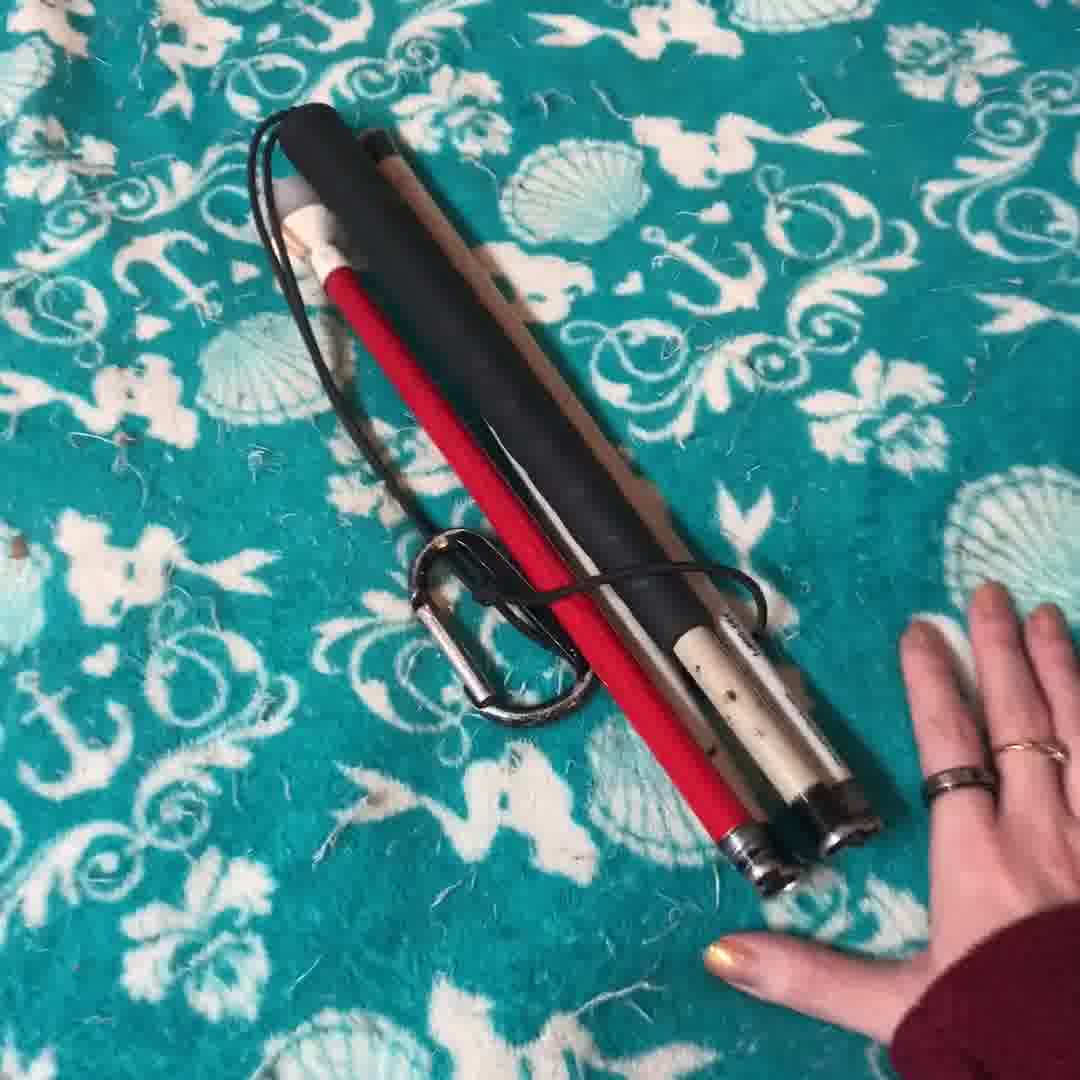}}}\\
    \vspace*{2px}
    \scalebox{0.95}{
    \mbox{\includegraphics[width=0.095\textwidth]{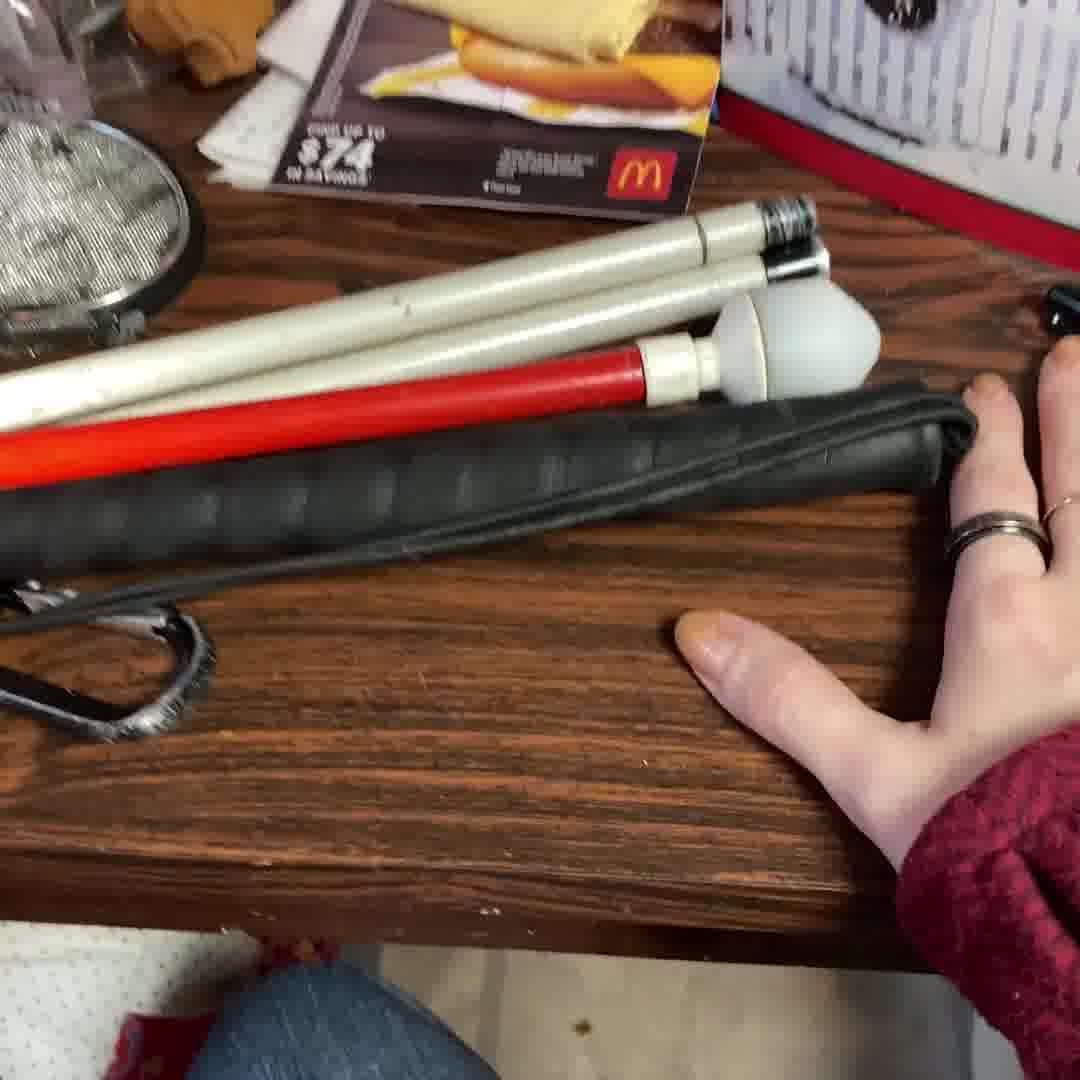}}
    \mbox{\includegraphics[width=0.095\textwidth]{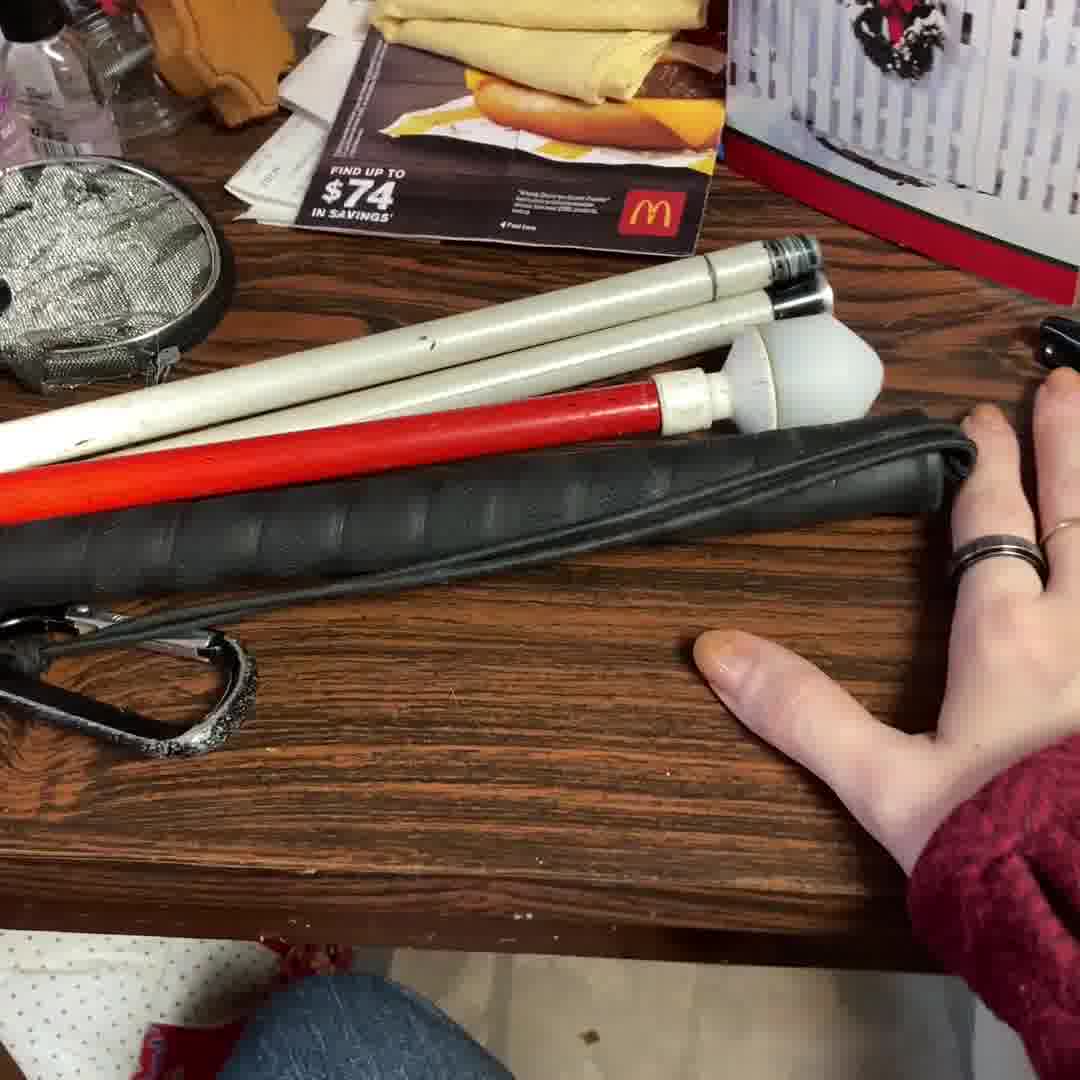}}
    \mbox{\includegraphics[width=0.095\textwidth]{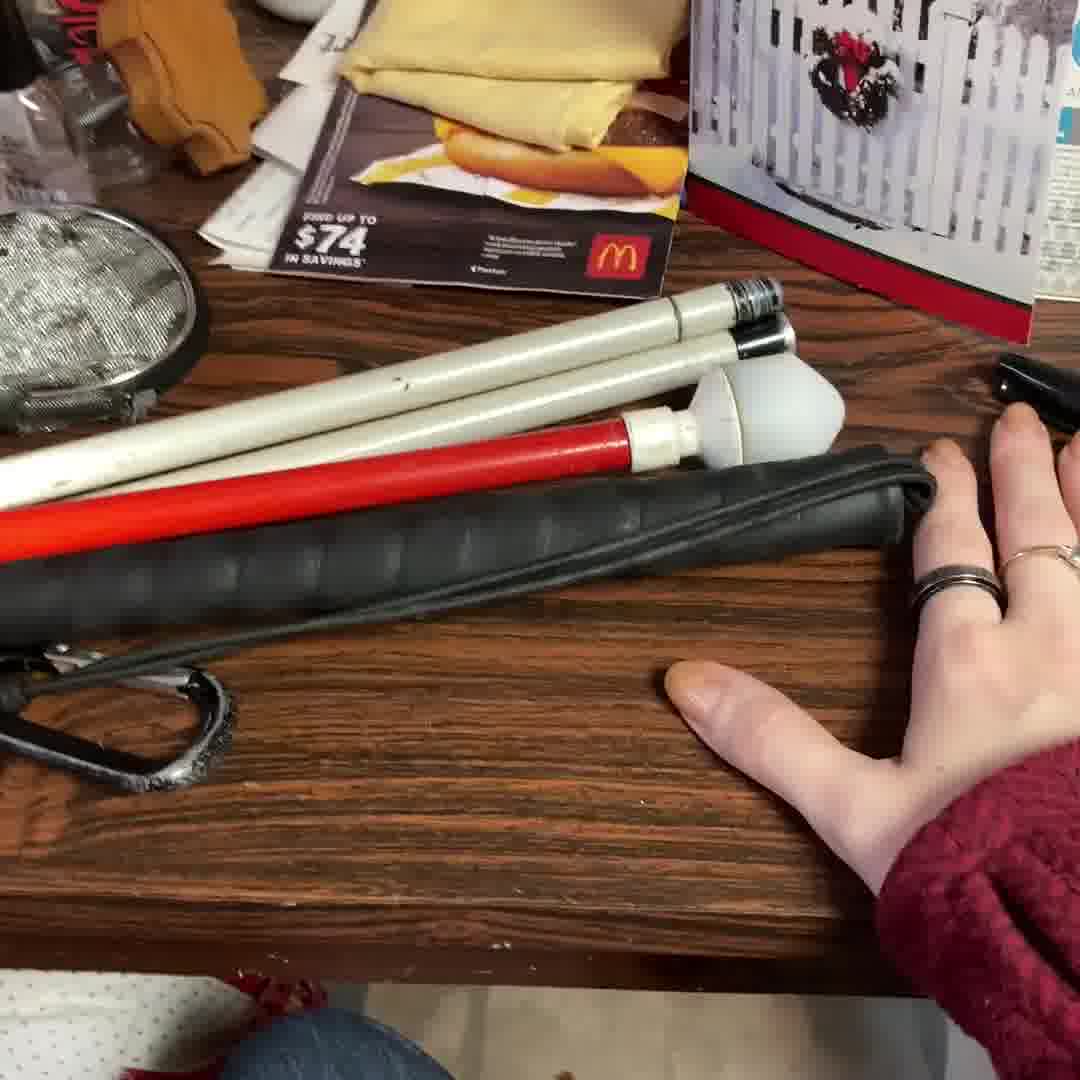}}
    \mbox{\includegraphics[width=0.095\textwidth]{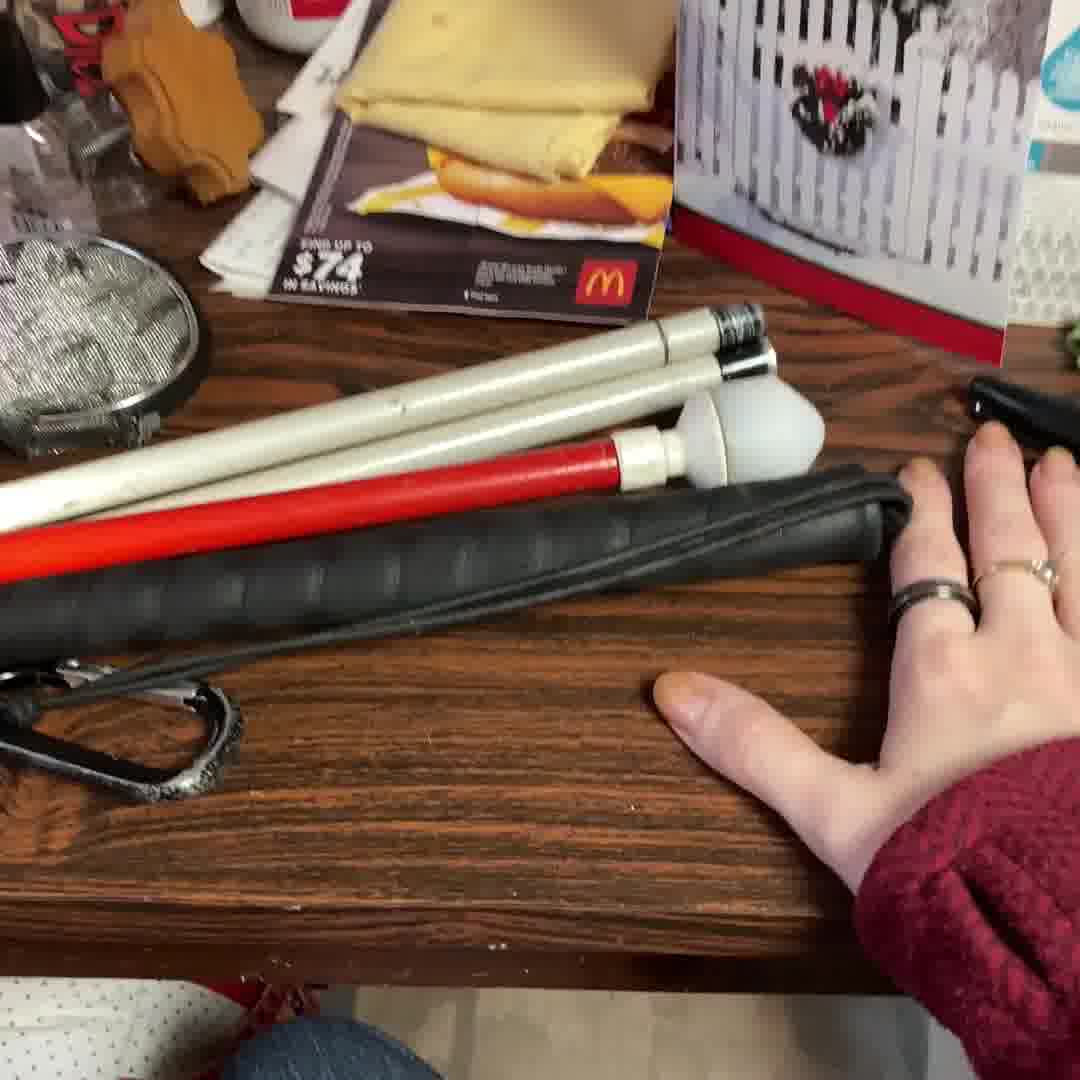}}
    \mbox{\includegraphics[width=0.095\textwidth]{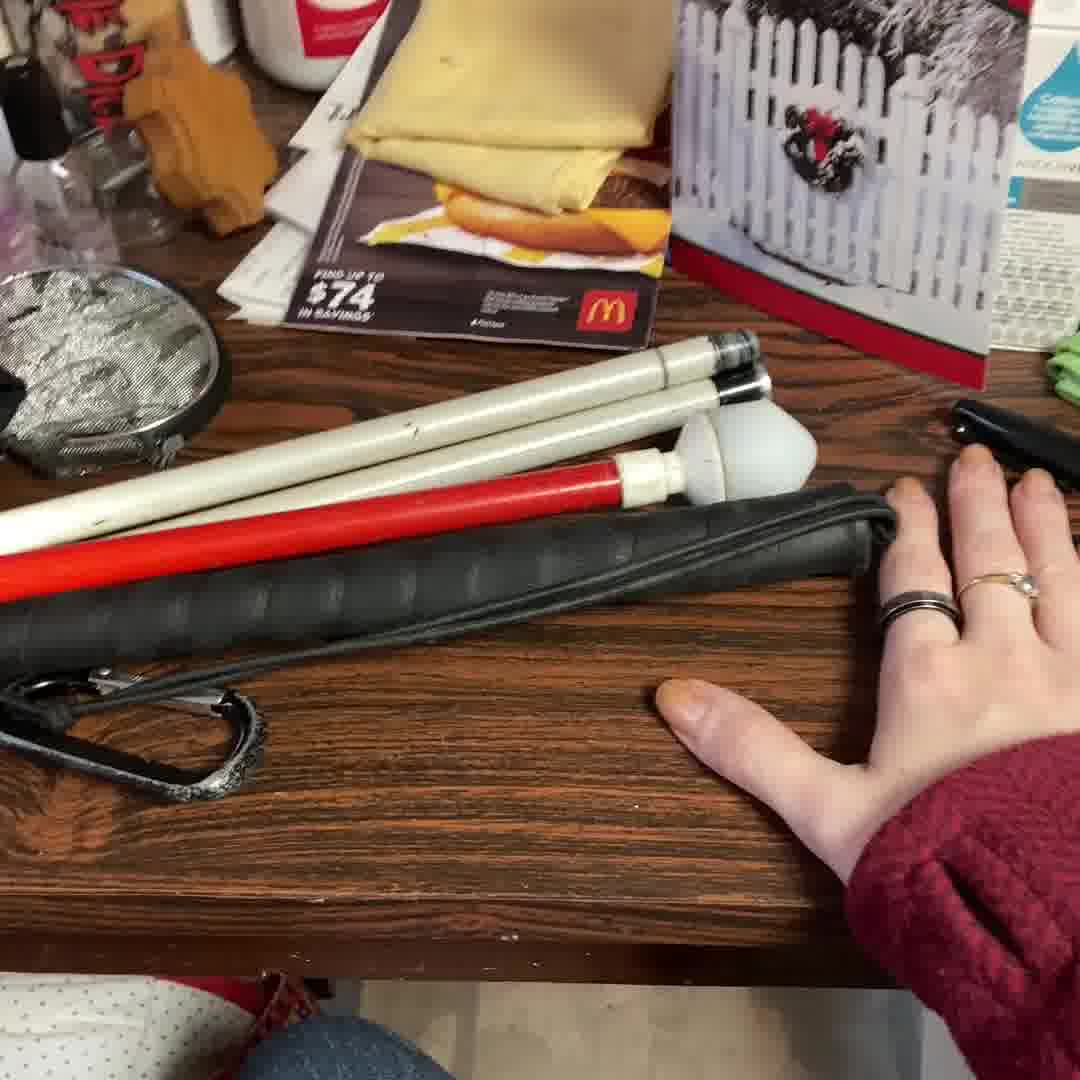}}
    \mbox{\includegraphics[width=0.095\textwidth]{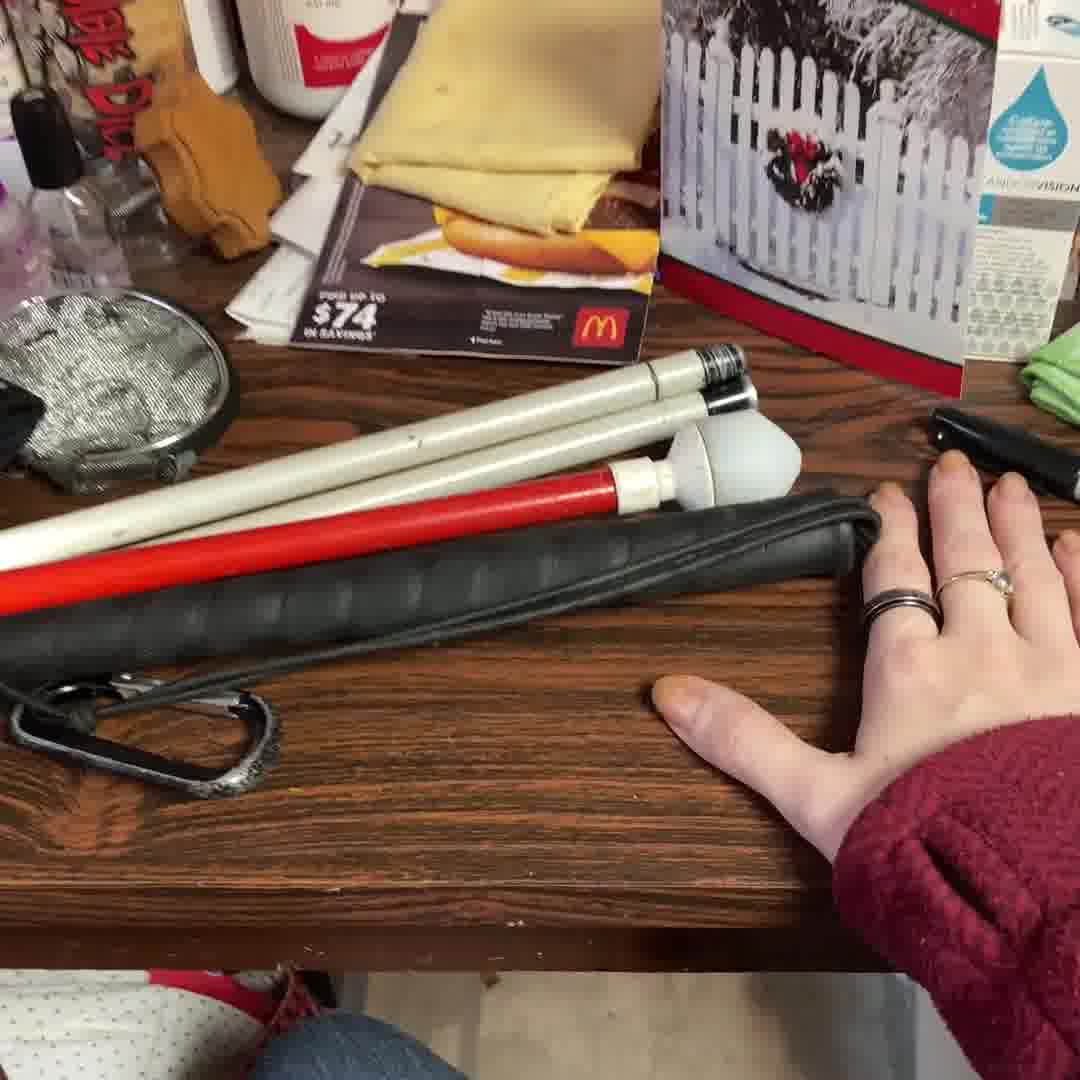}}
    \mbox{\includegraphics[width=0.095\textwidth]{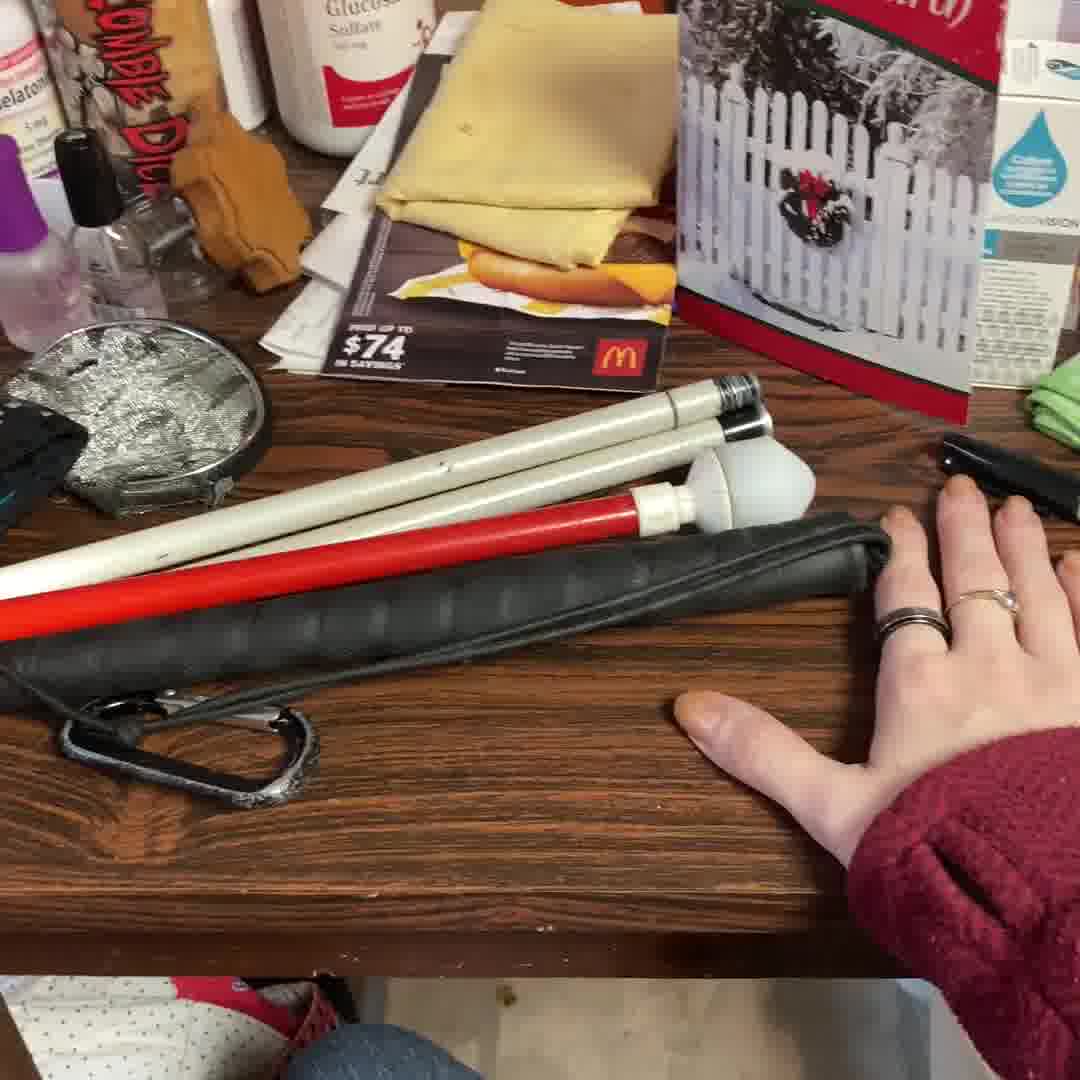}}
    \mbox{\includegraphics[width=0.095\textwidth]{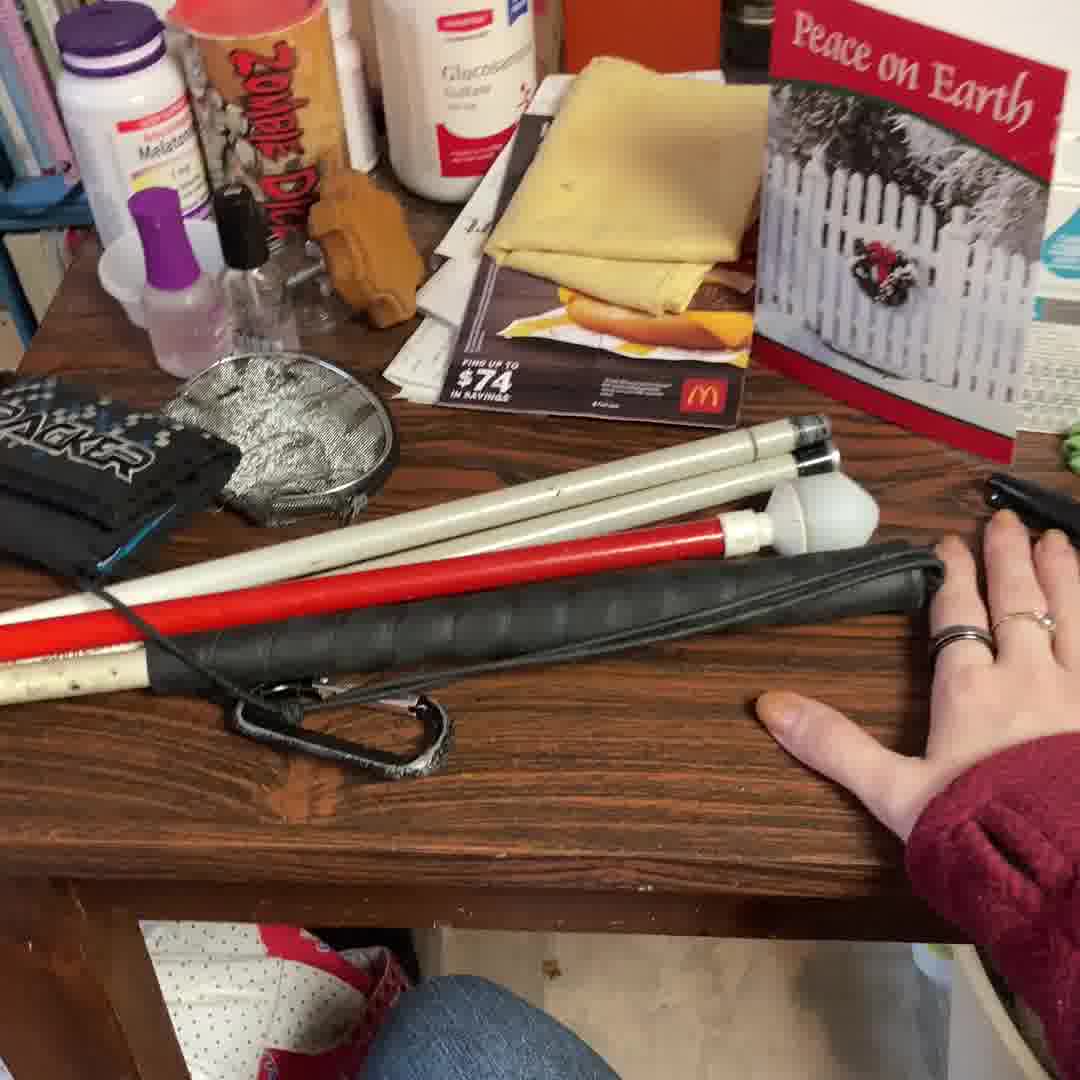}}
    \mbox{\includegraphics[width=0.095\textwidth]{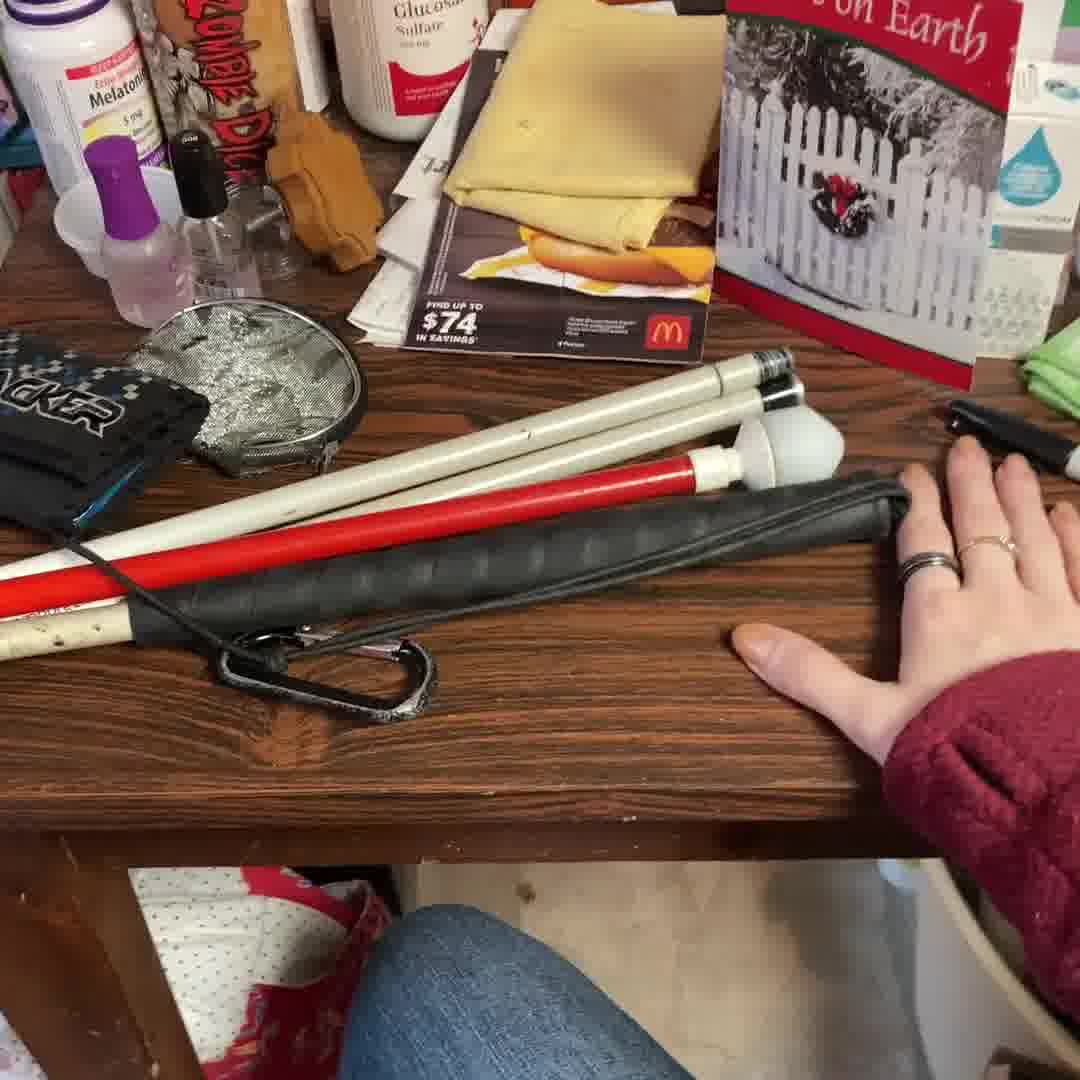}}
    \mbox{\includegraphics[width=0.095\textwidth]{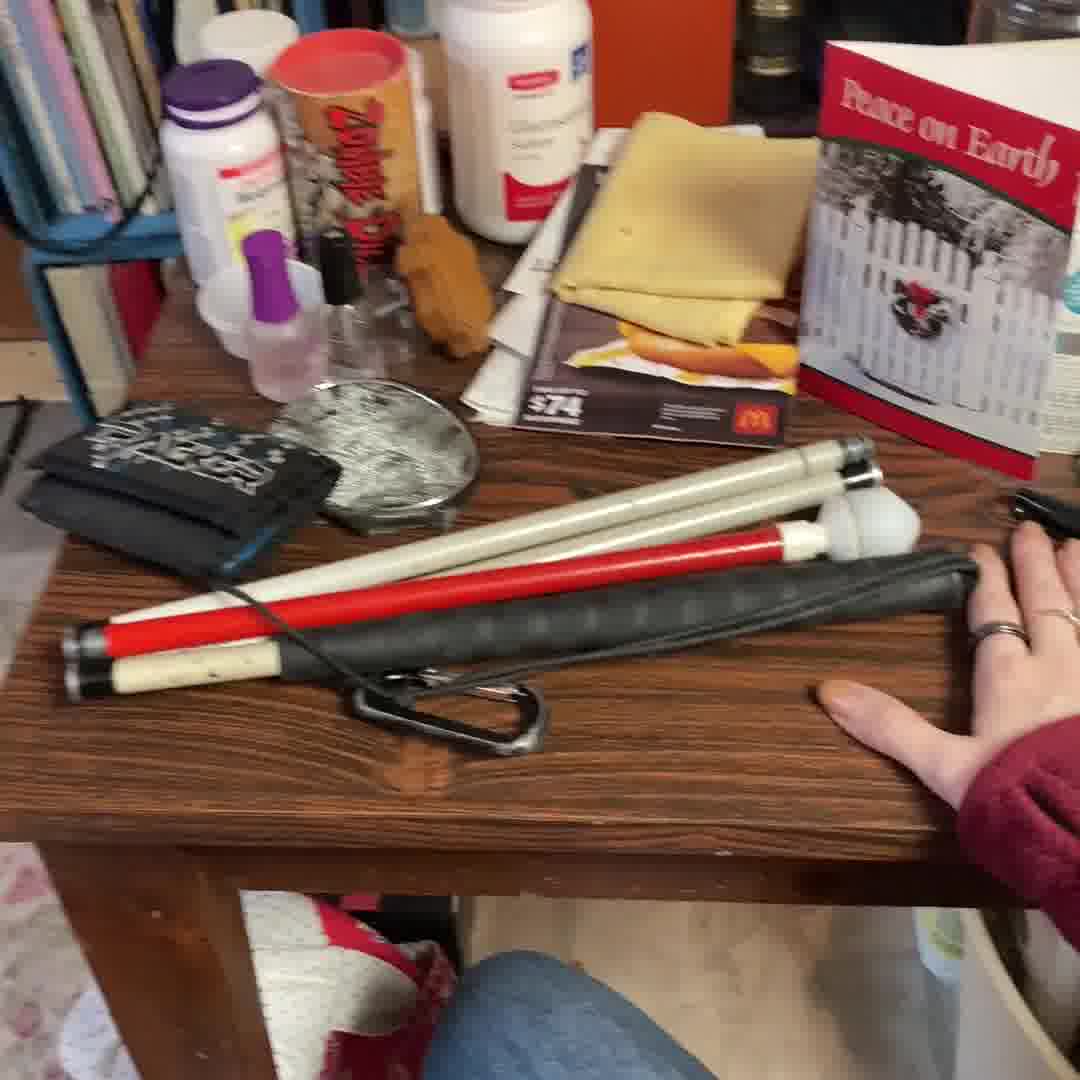}}}\\
    \vspace*{2px}
    \scalebox{0.95}{
    \mbox{\includegraphics[width=0.095\textwidth]{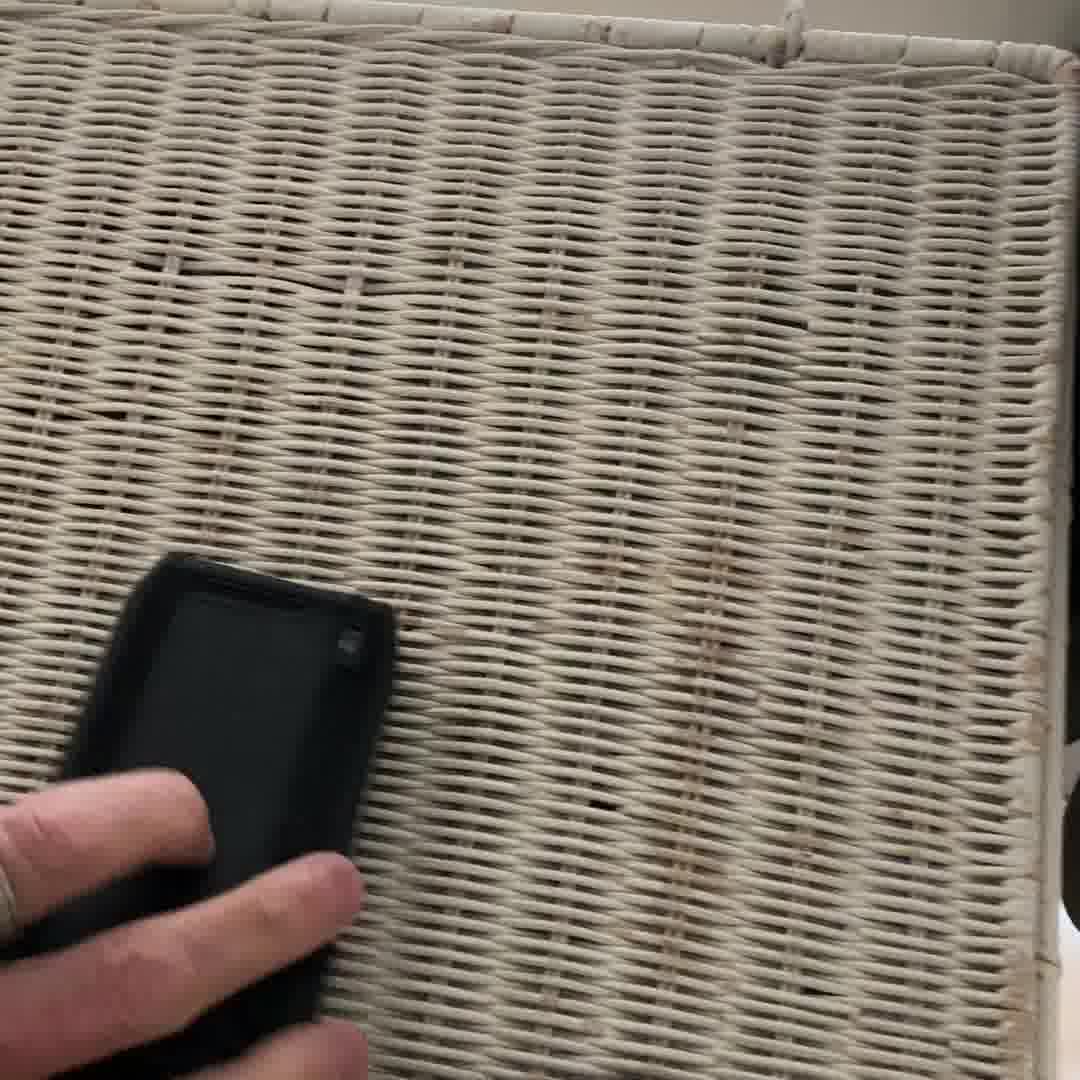}}
    \mbox{\includegraphics[width=0.095\textwidth]{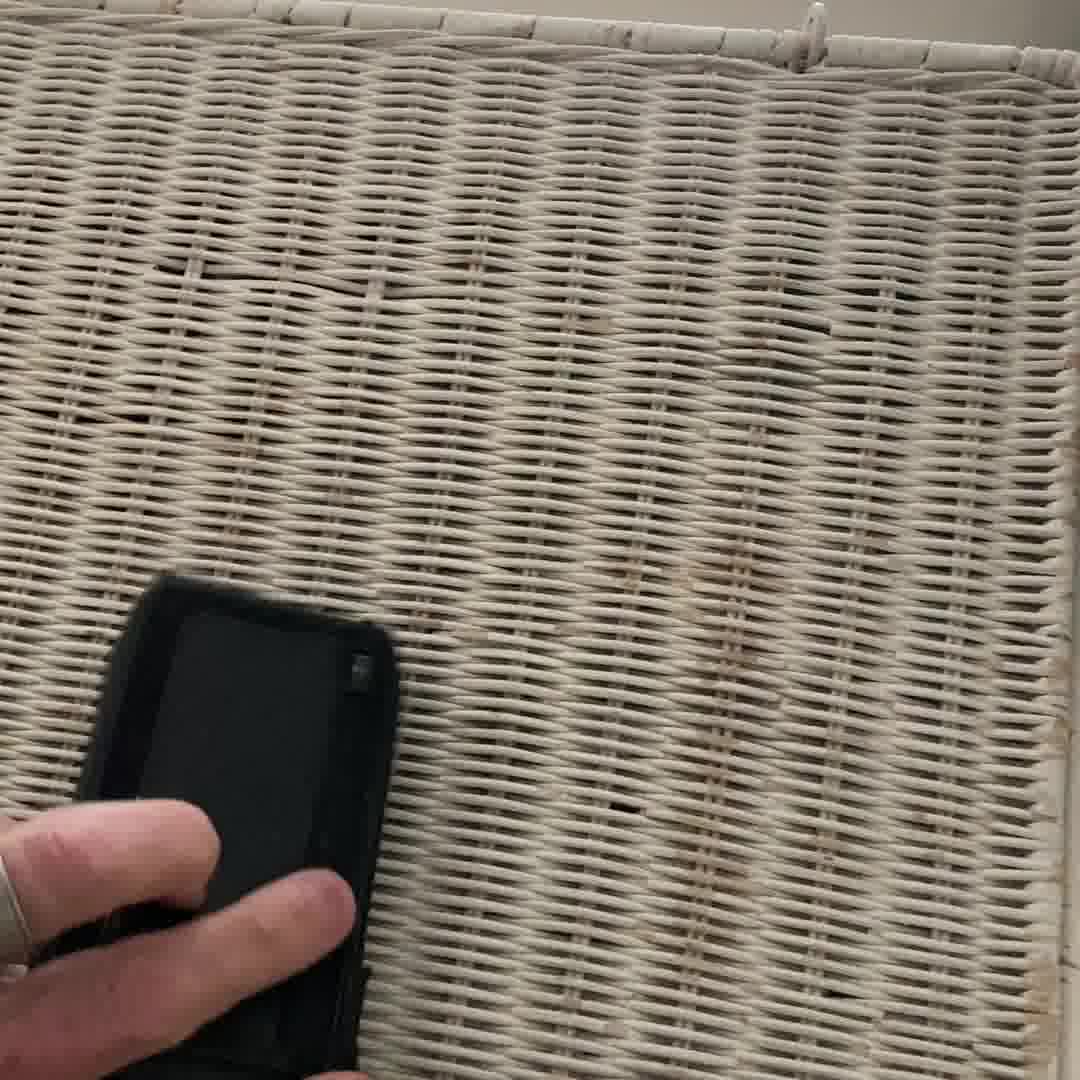}}
    \mbox{\includegraphics[width=0.095\textwidth]{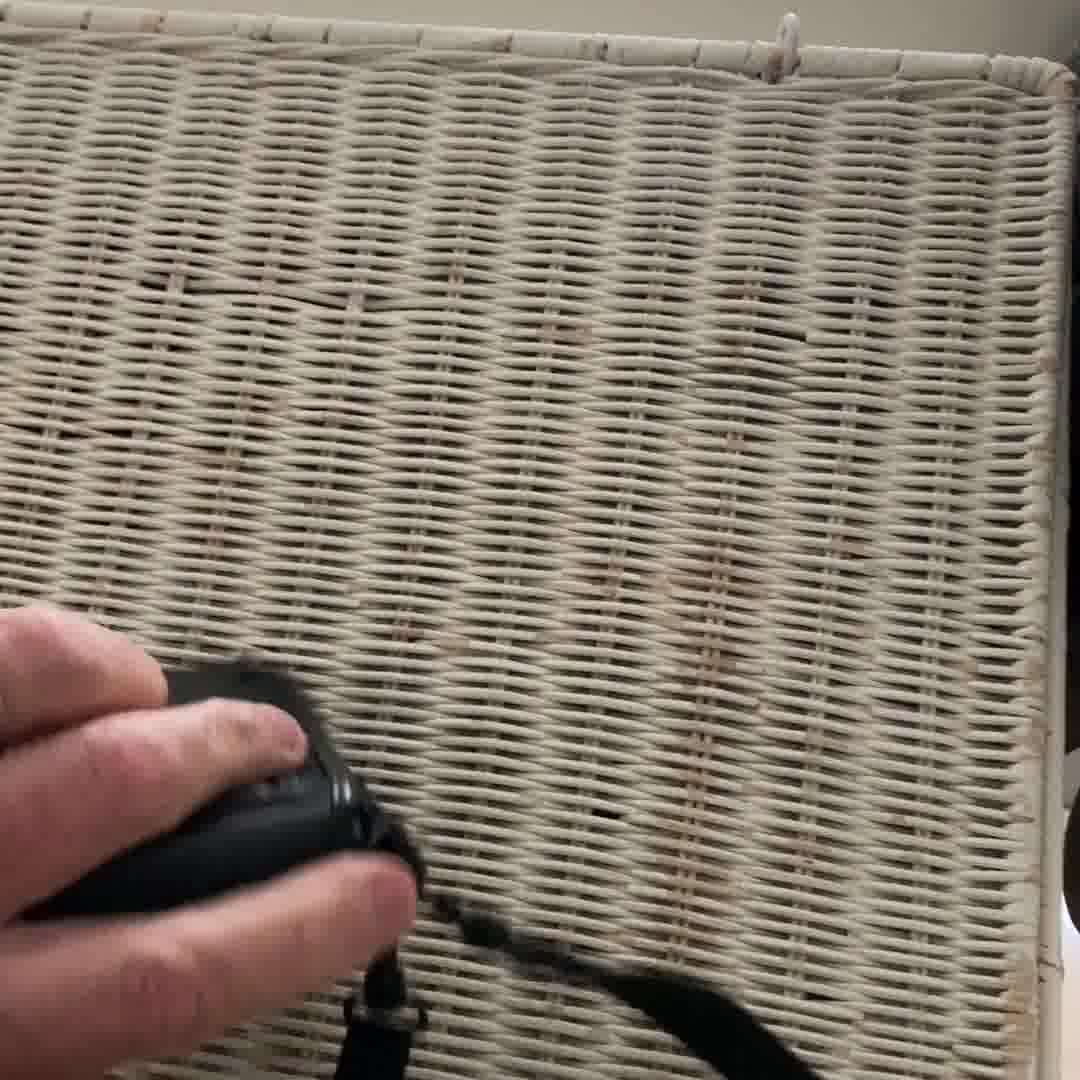}}
    \mbox{\includegraphics[width=0.095\textwidth]{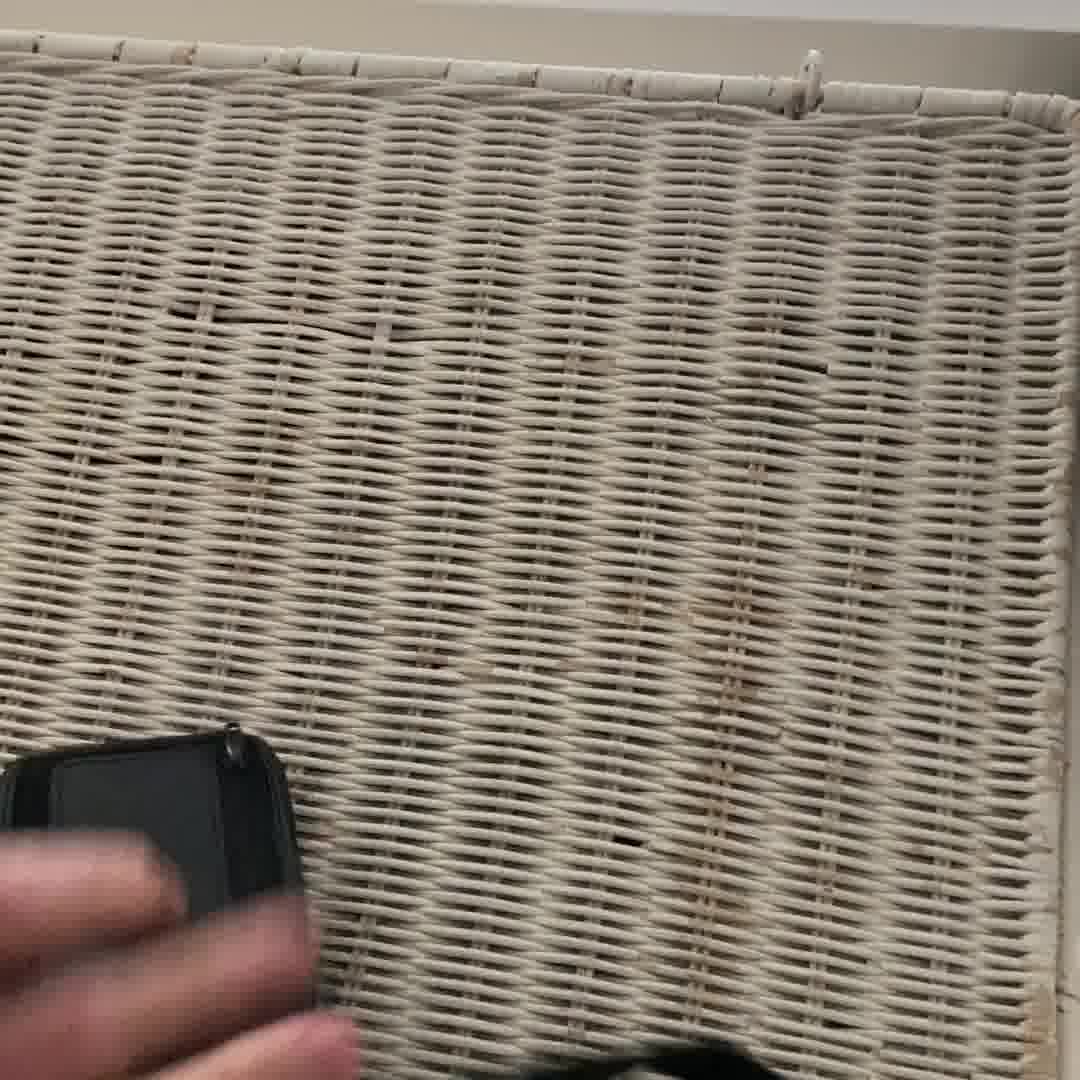}}
    \mbox{\includegraphics[width=0.095\textwidth]{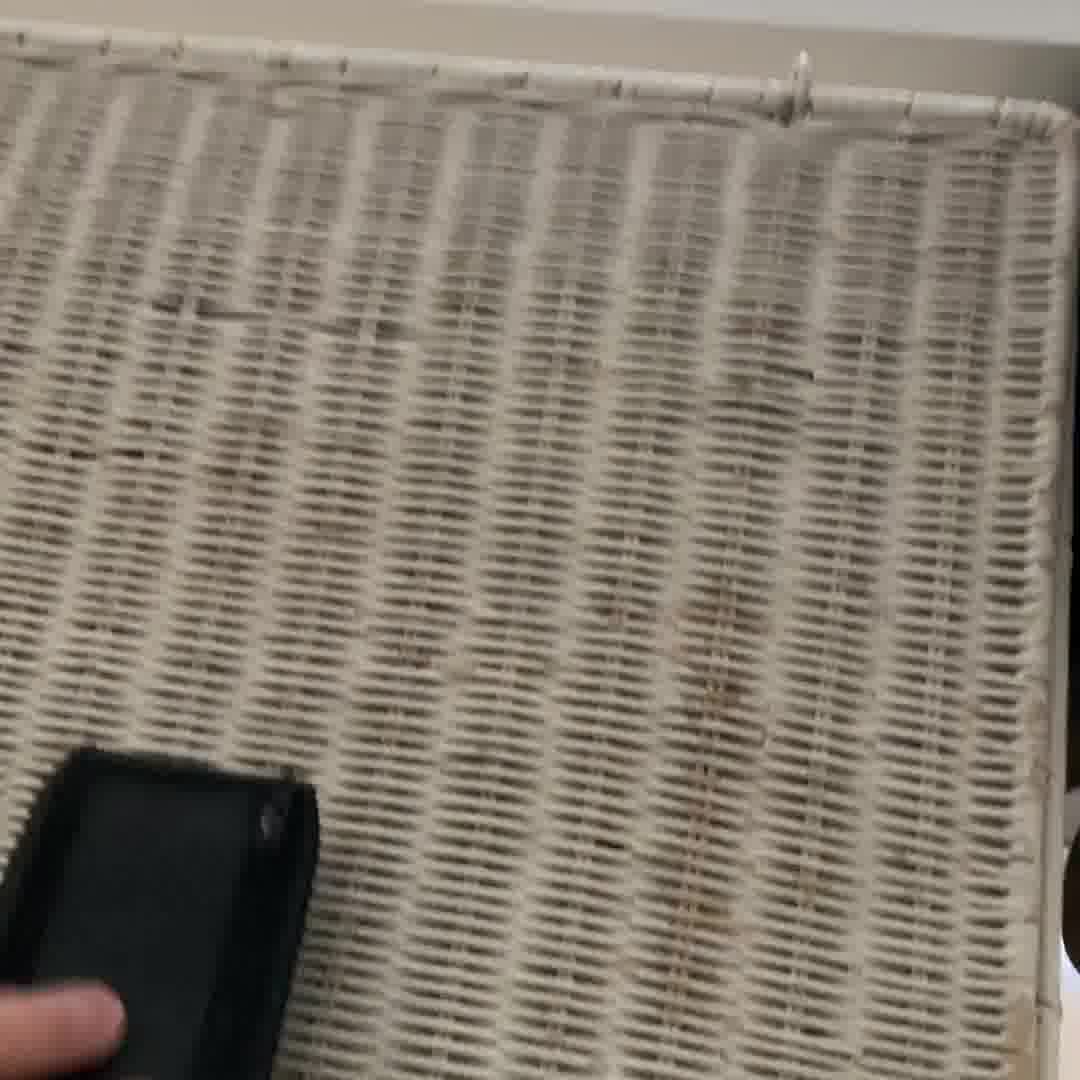}}
    \mbox{\includegraphics[width=0.095\textwidth]{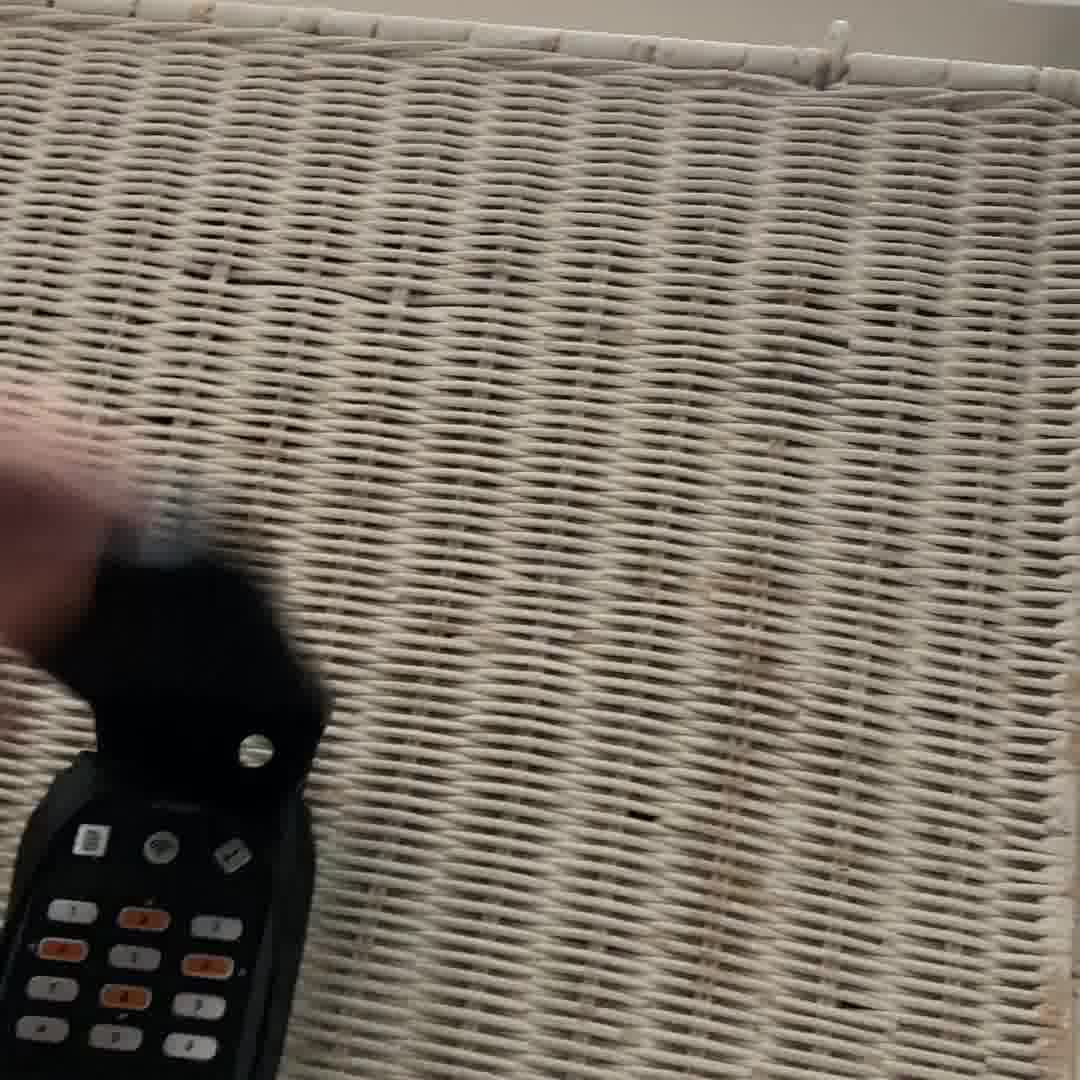}}
    \mbox{\includegraphics[width=0.095\textwidth]{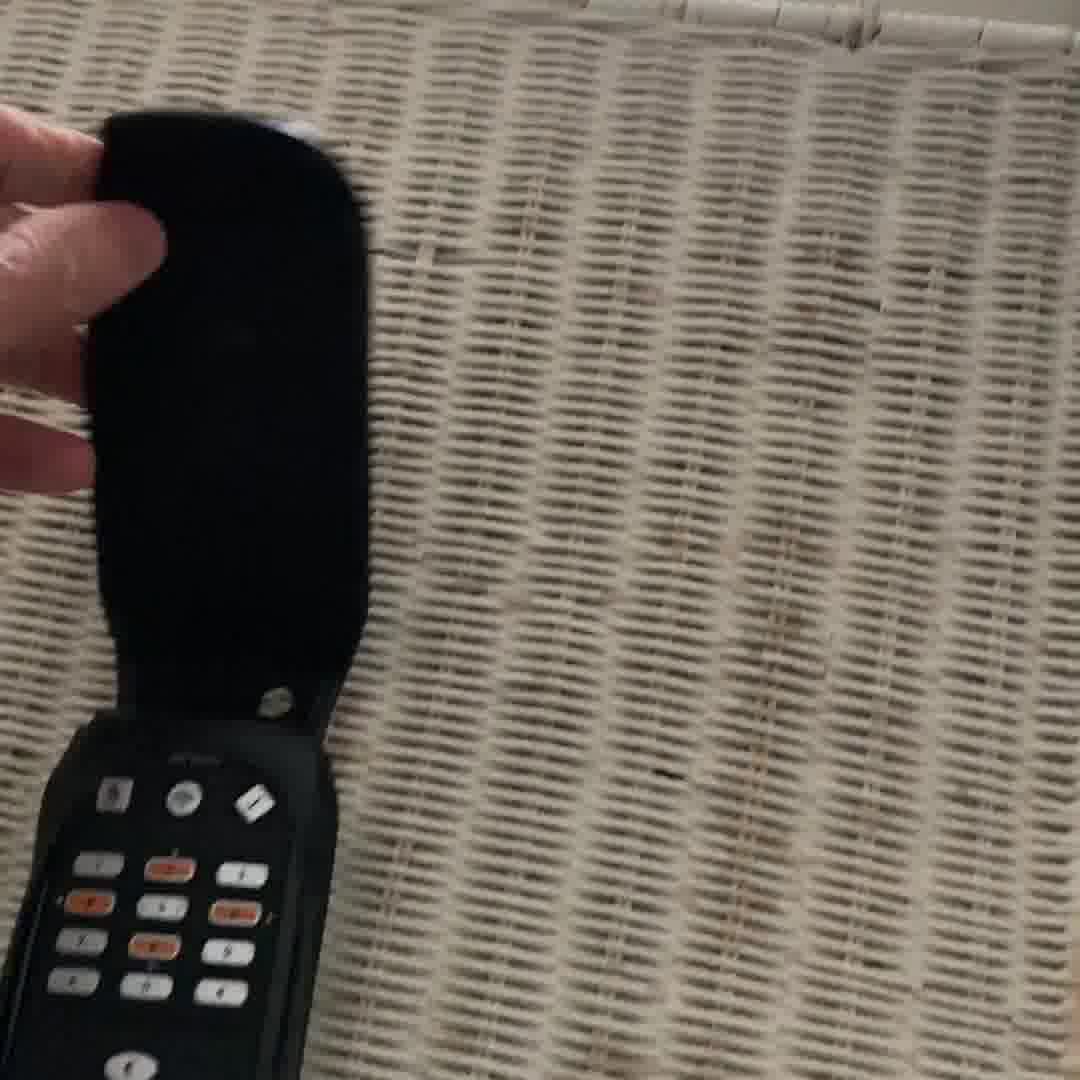}}
    \mbox{\includegraphics[width=0.095\textwidth]{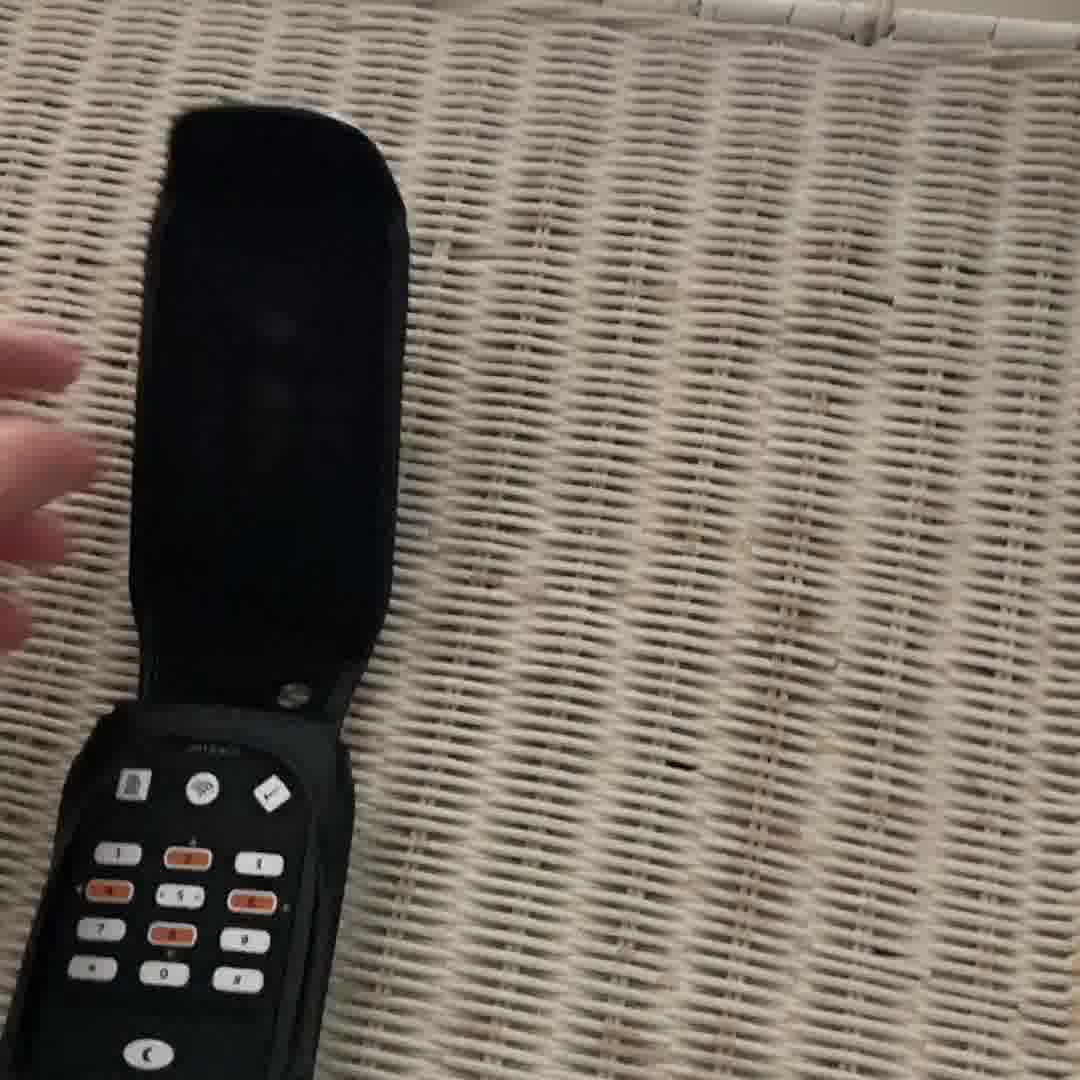}}
    \mbox{\includegraphics[width=0.095\textwidth]{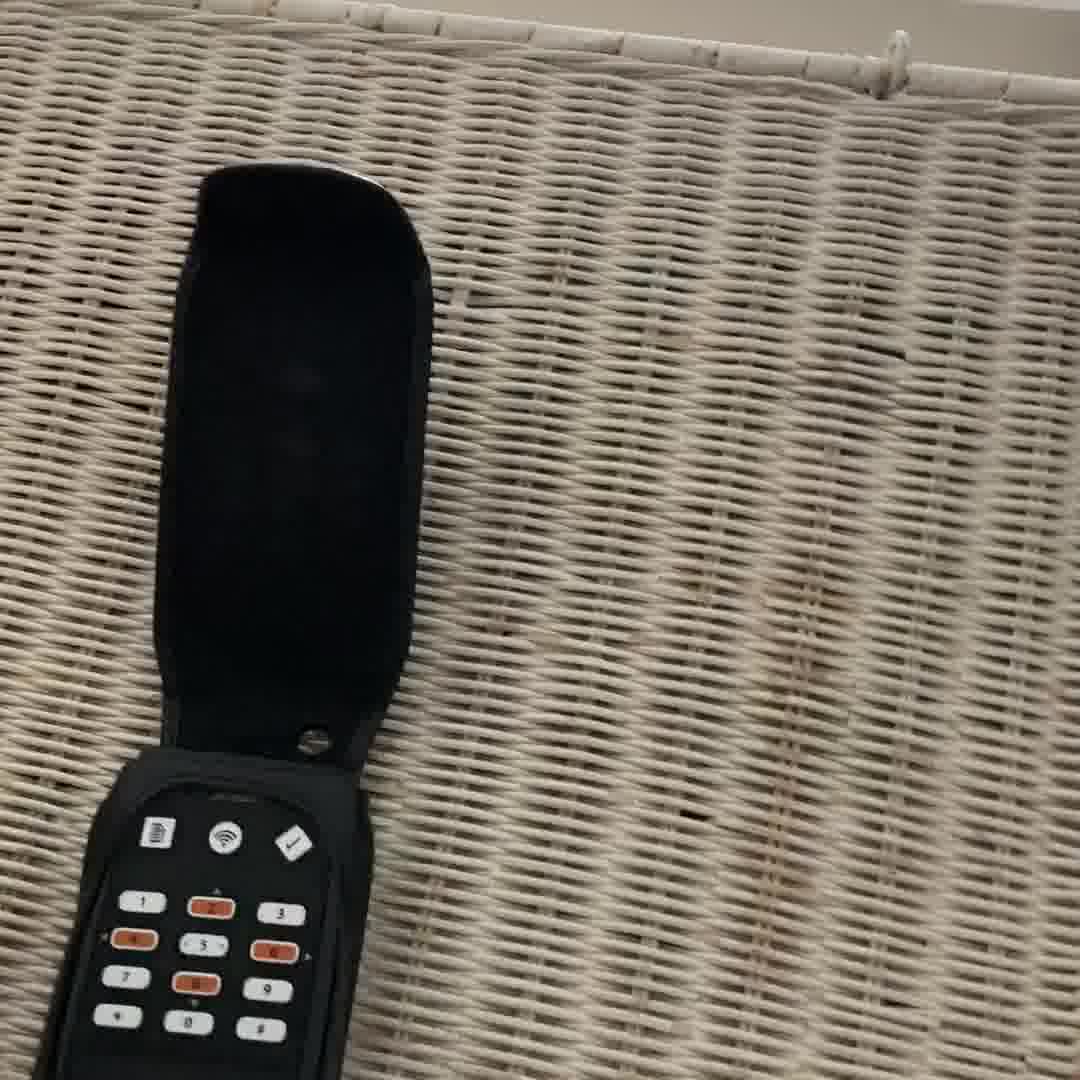}}
    \mbox{\includegraphics[width=0.095\textwidth]{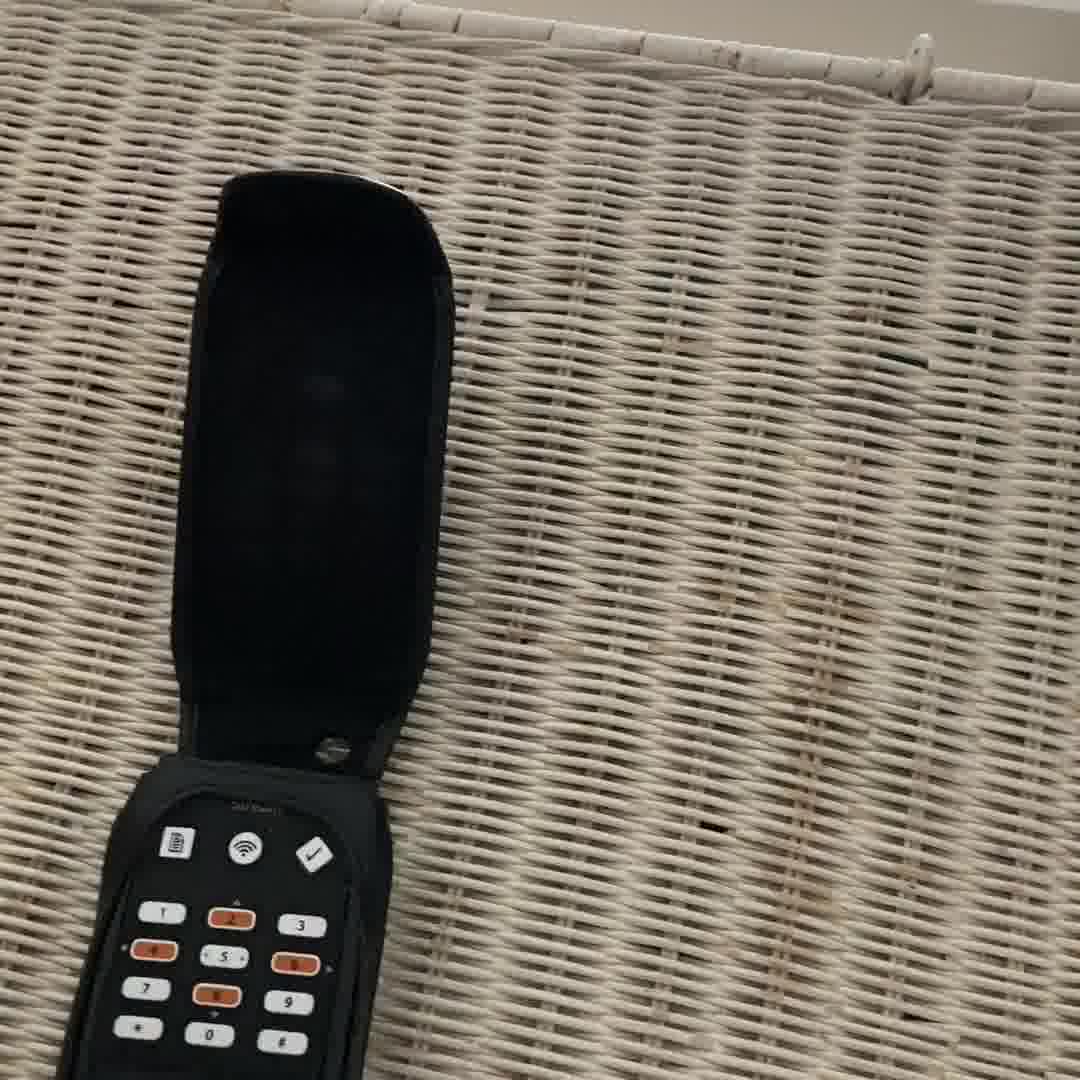}}}\\
    \vspace*{2px}
    \scalebox{0.95}{
    \mbox{\includegraphics[width=0.095\textwidth]{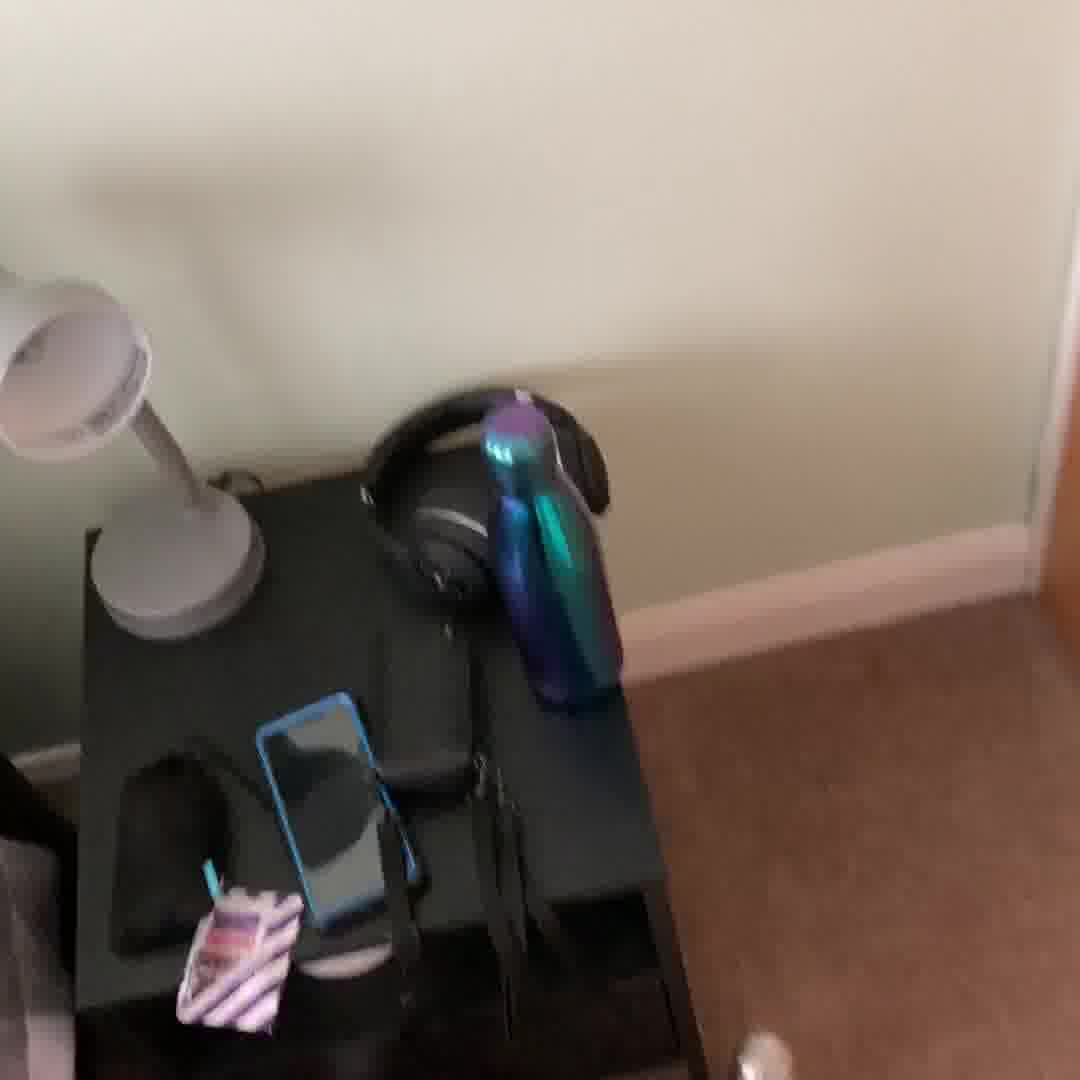}}
    \mbox{\includegraphics[width=0.095\textwidth]{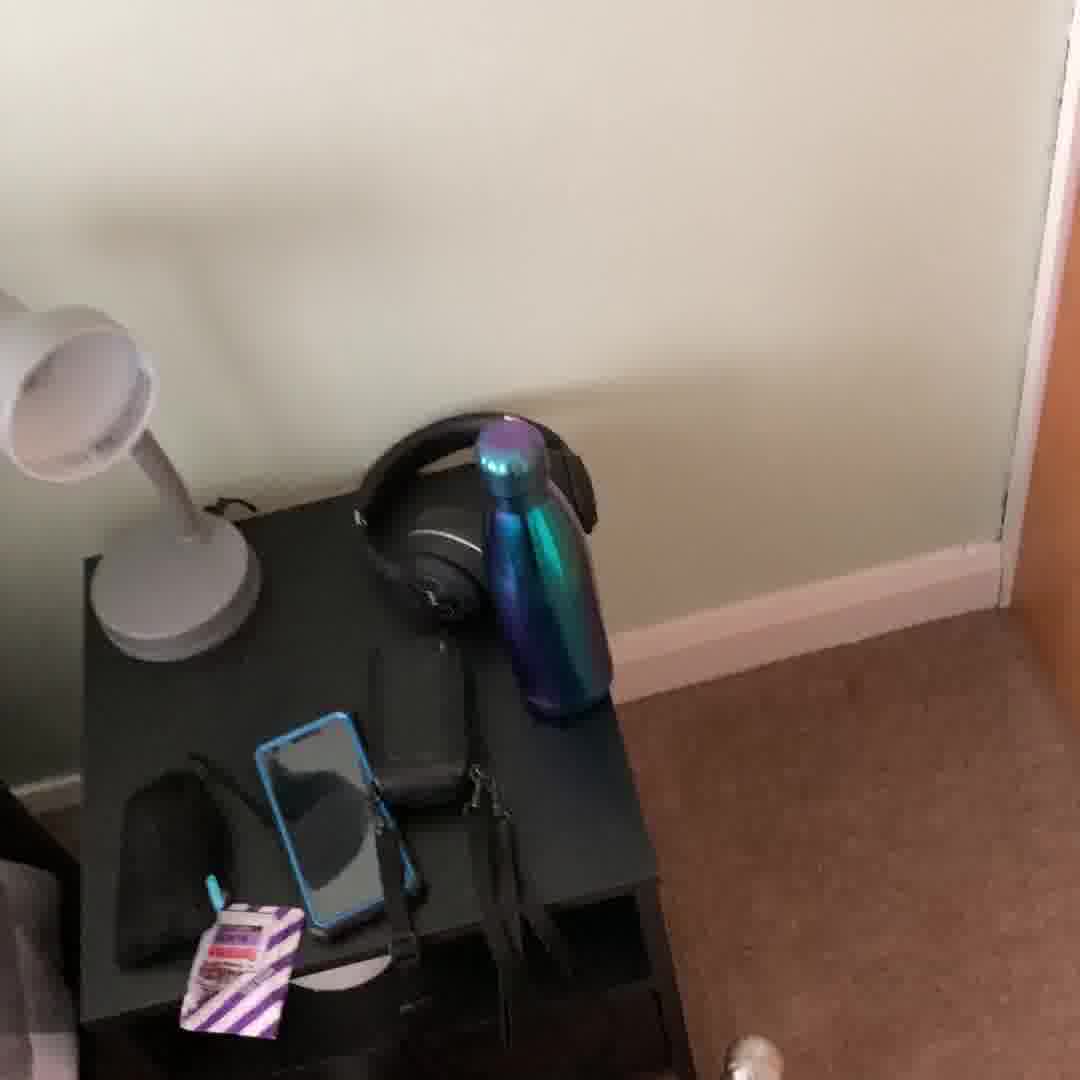}}
    \mbox{\includegraphics[width=0.095\textwidth]{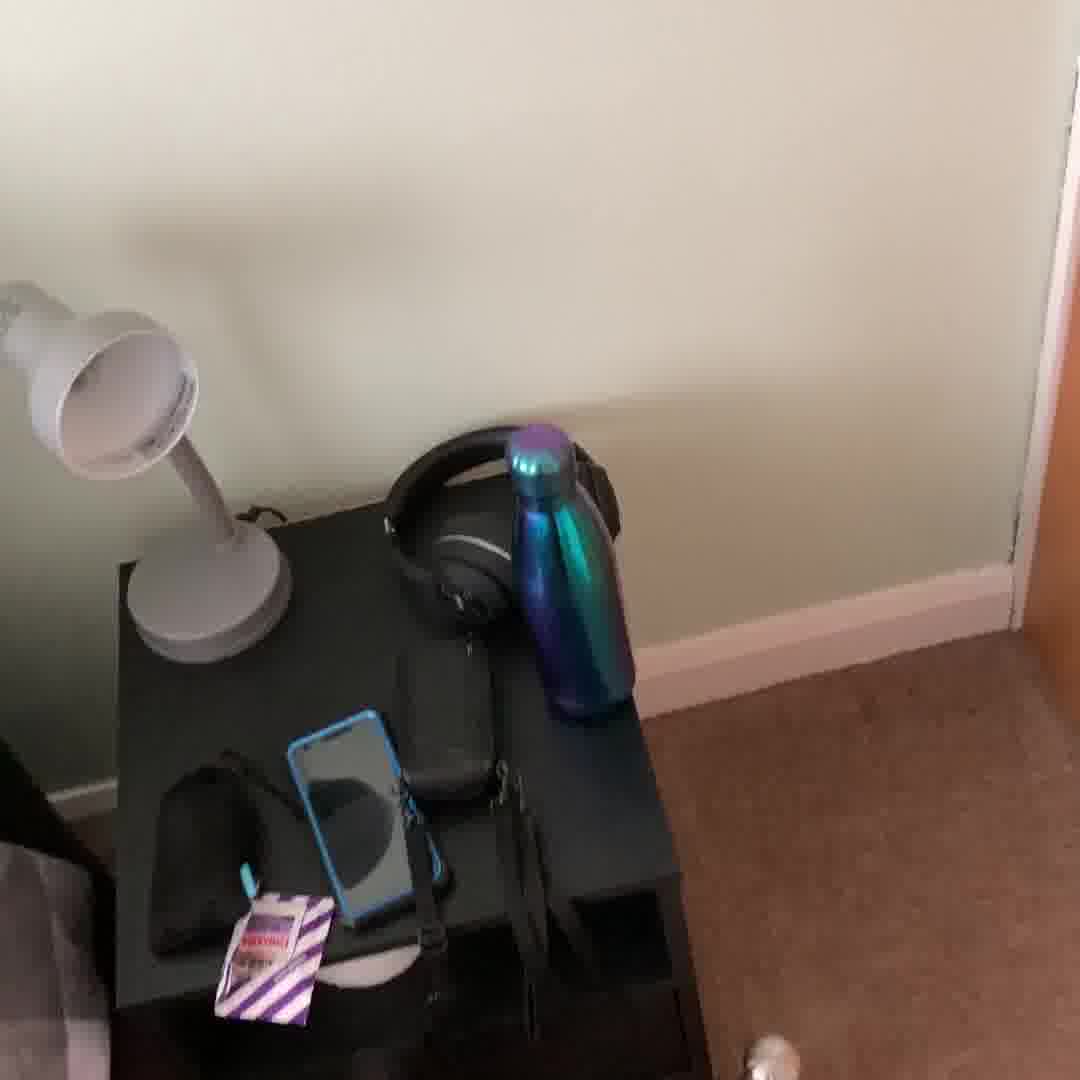}}
    \mbox{\includegraphics[width=0.095\textwidth]{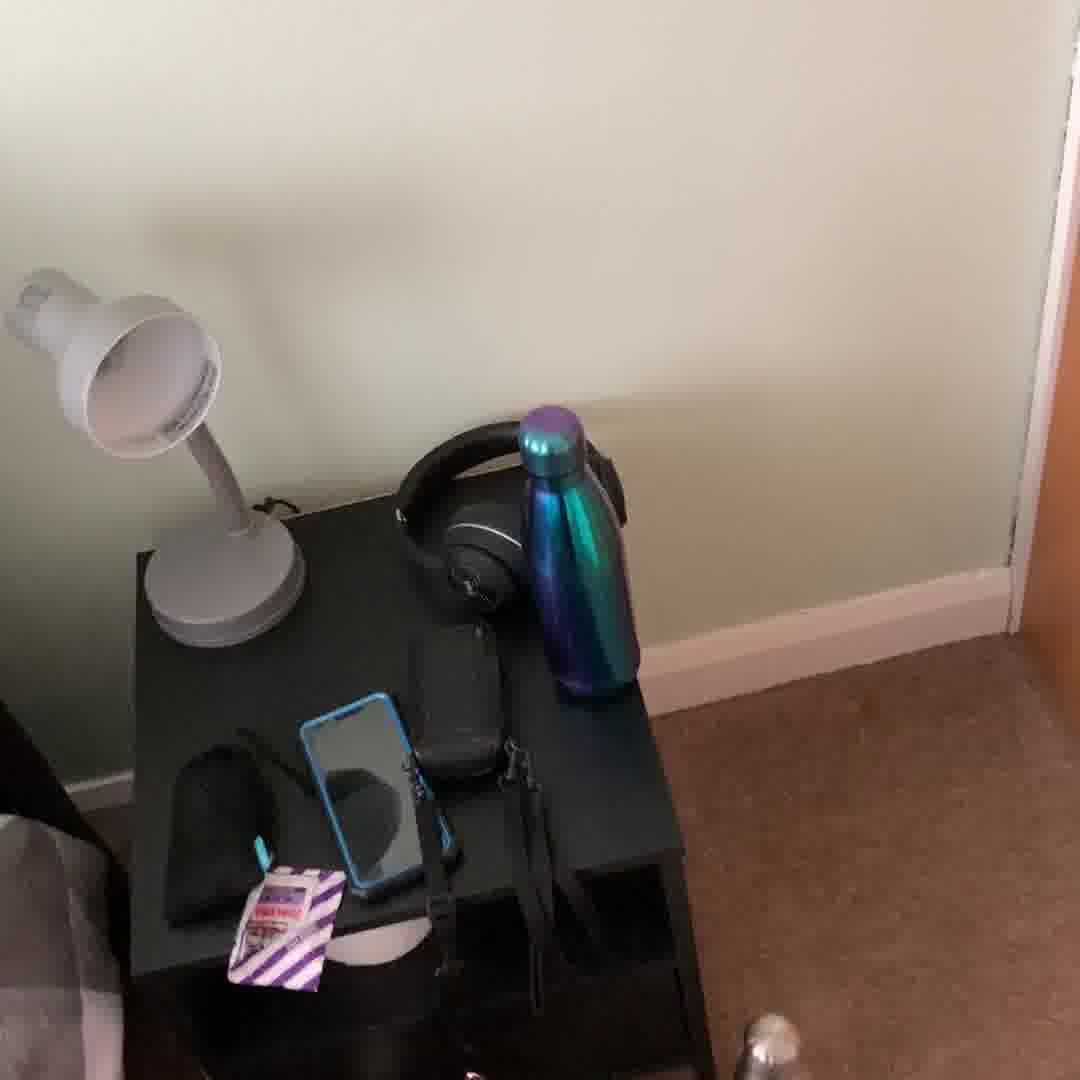}}
    \mbox{\includegraphics[width=0.095\textwidth]{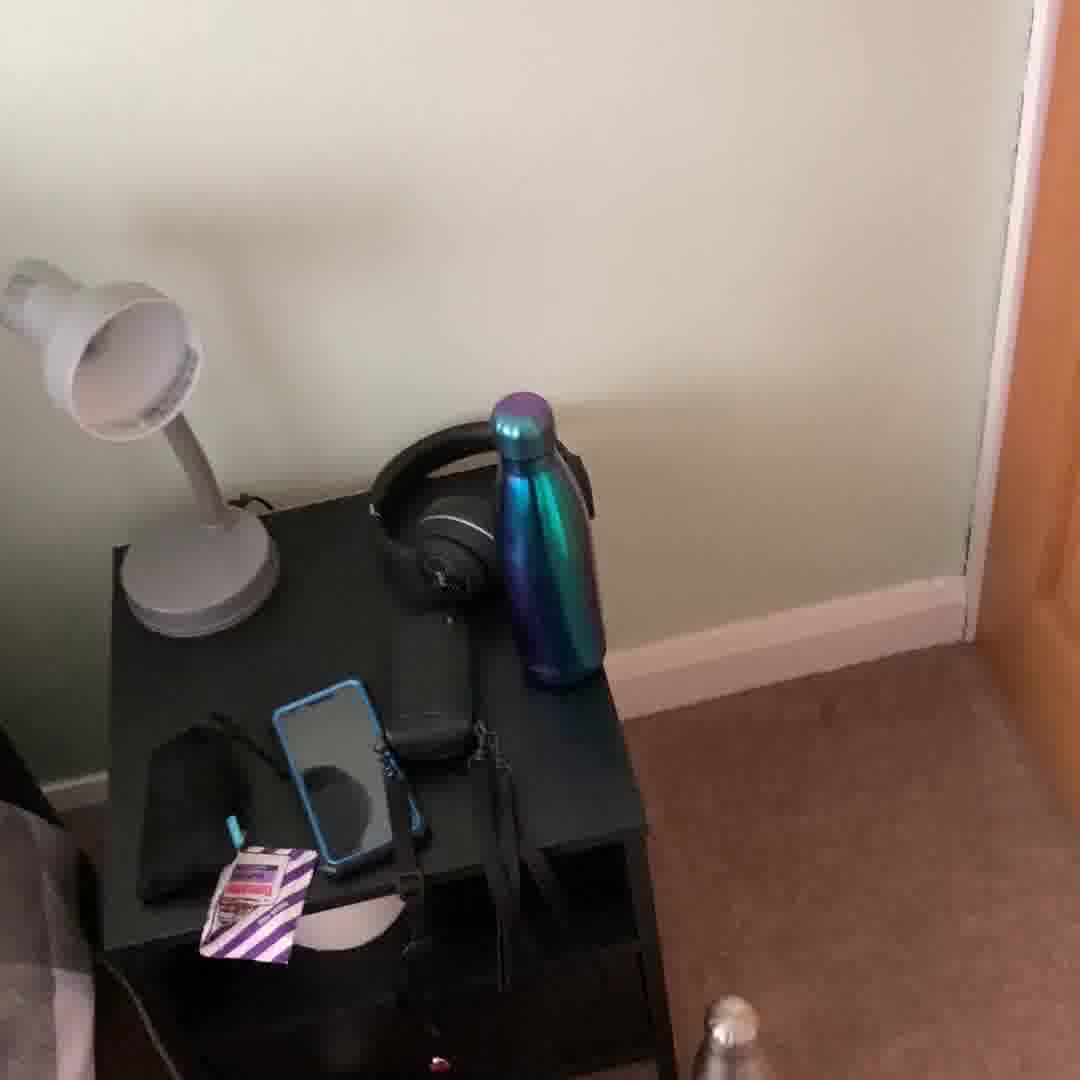}}
    \mbox{\includegraphics[width=0.095\textwidth]{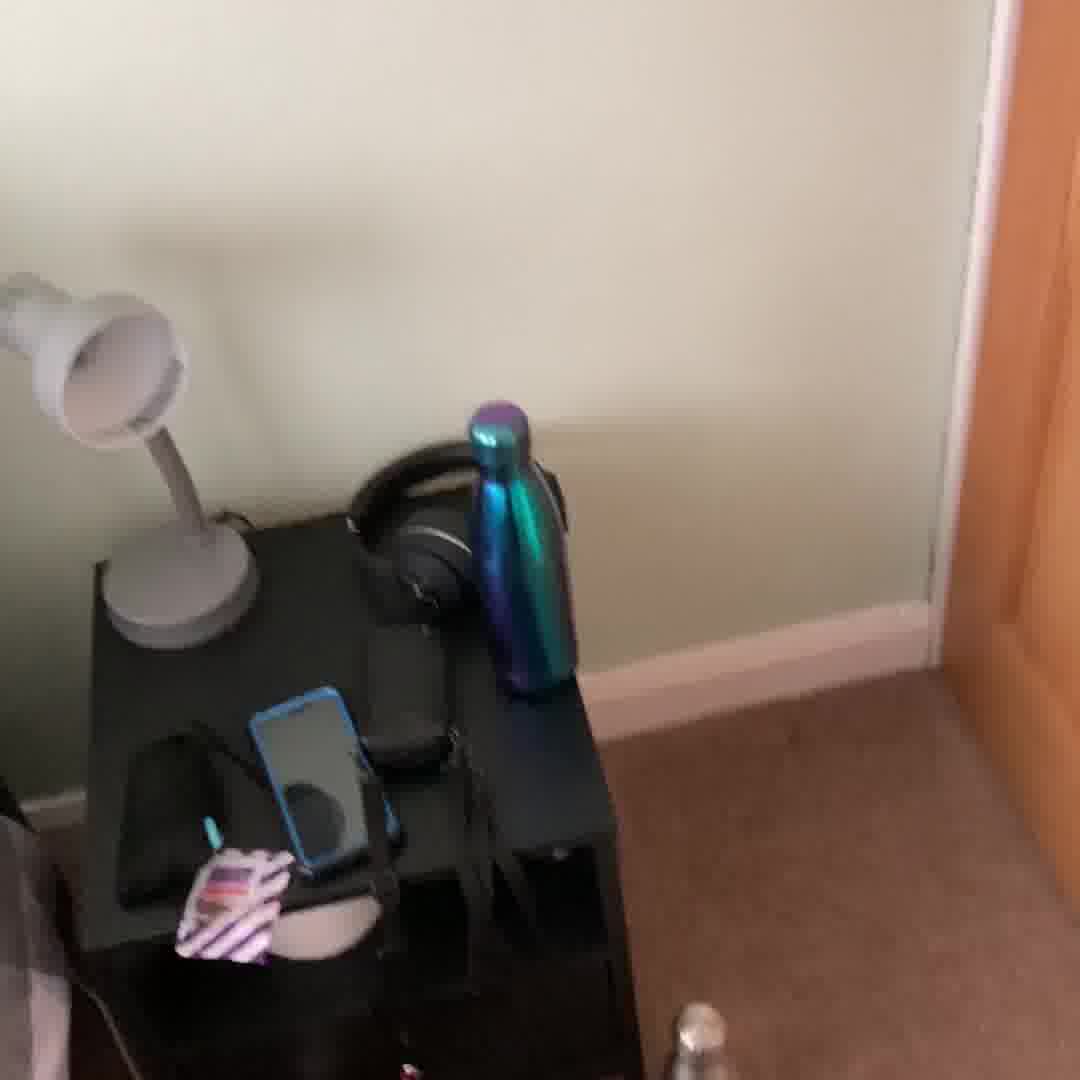}}
    \mbox{\includegraphics[width=0.095\textwidth]{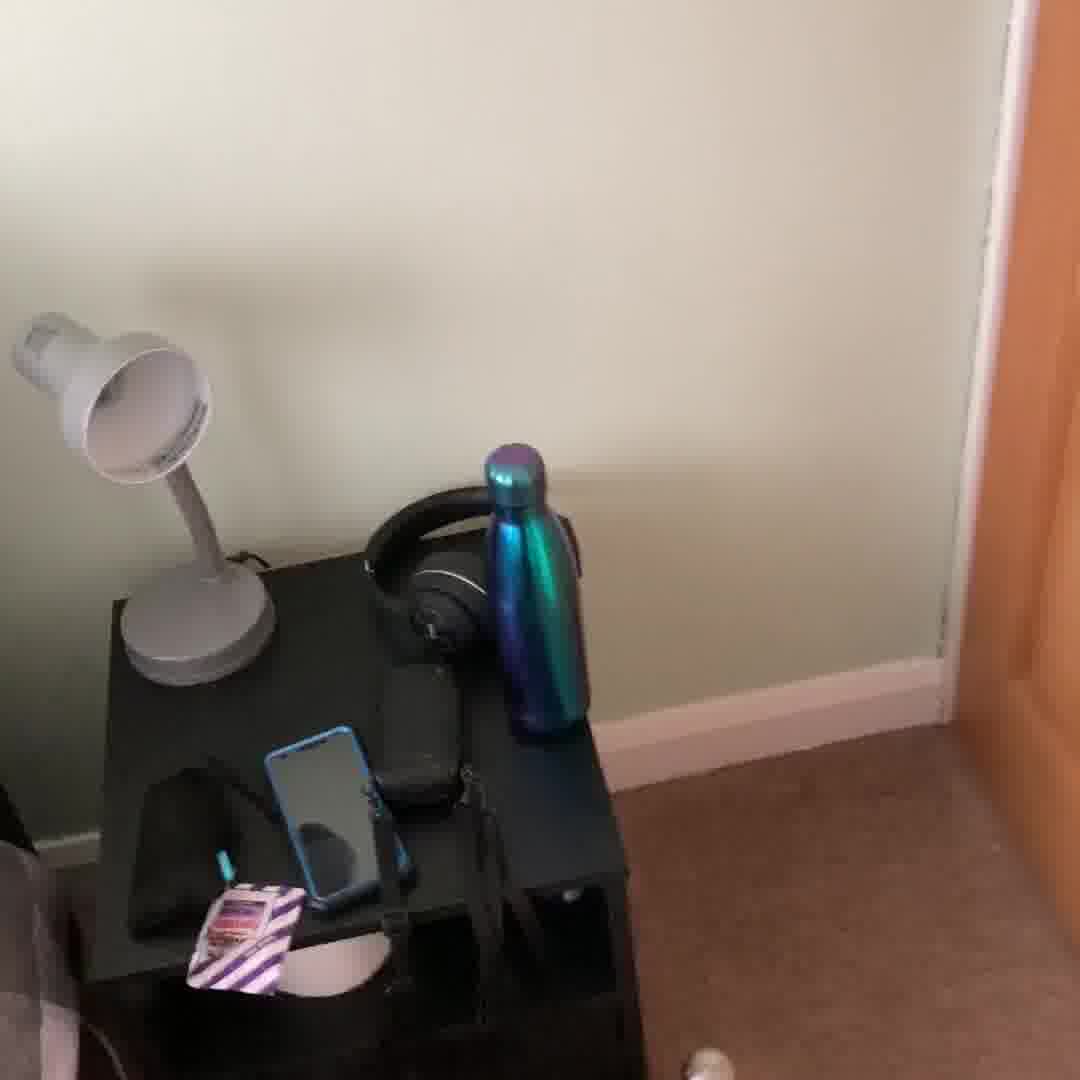}}
    \mbox{\includegraphics[width=0.095\textwidth]{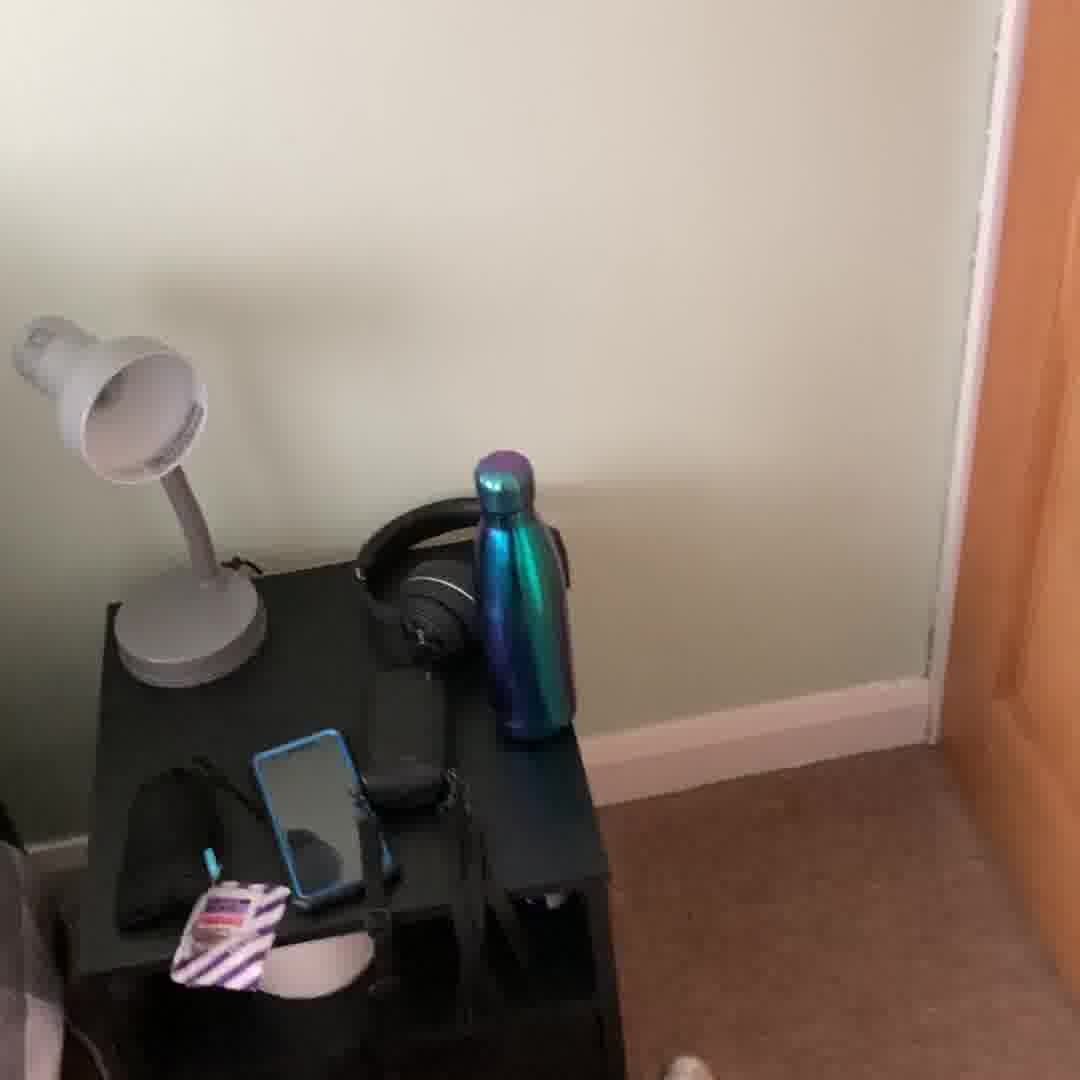}}
    \mbox{\includegraphics[width=0.095\textwidth]{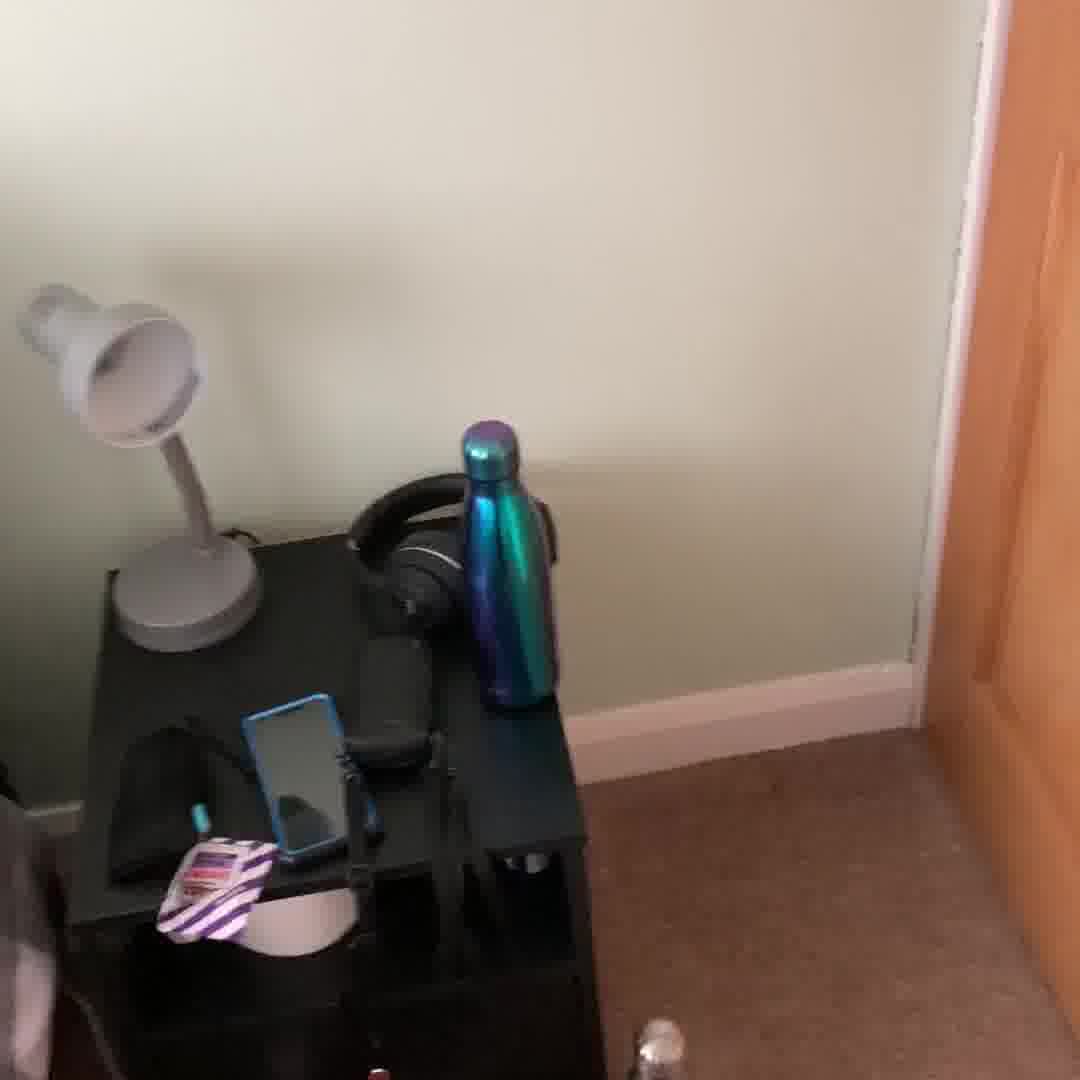}}
    \mbox{\includegraphics[width=0.095\textwidth]{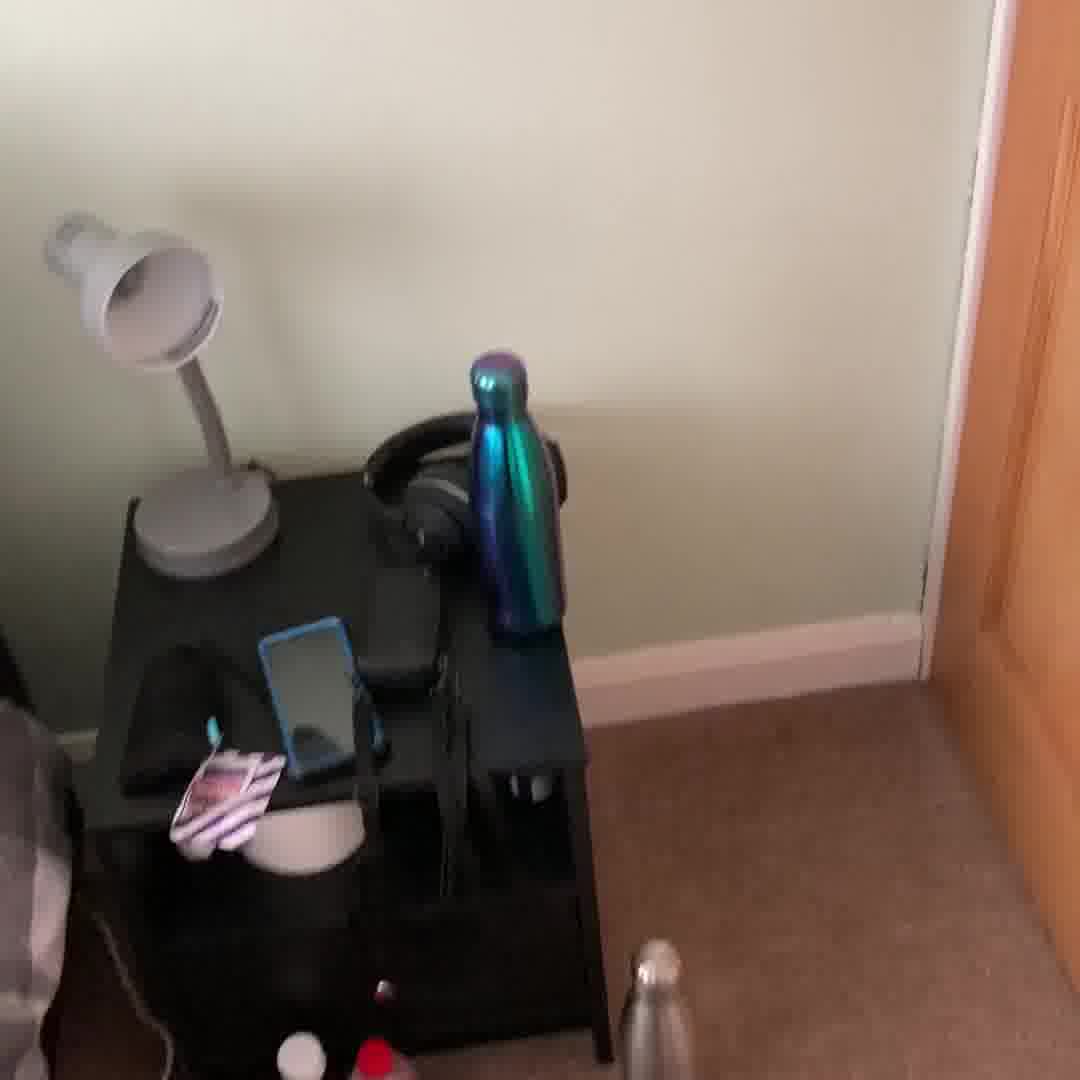}}}\\
        \vspace*{2px}
    \scalebox{0.95}{
    \mbox{\includegraphics[width=0.095\textwidth]{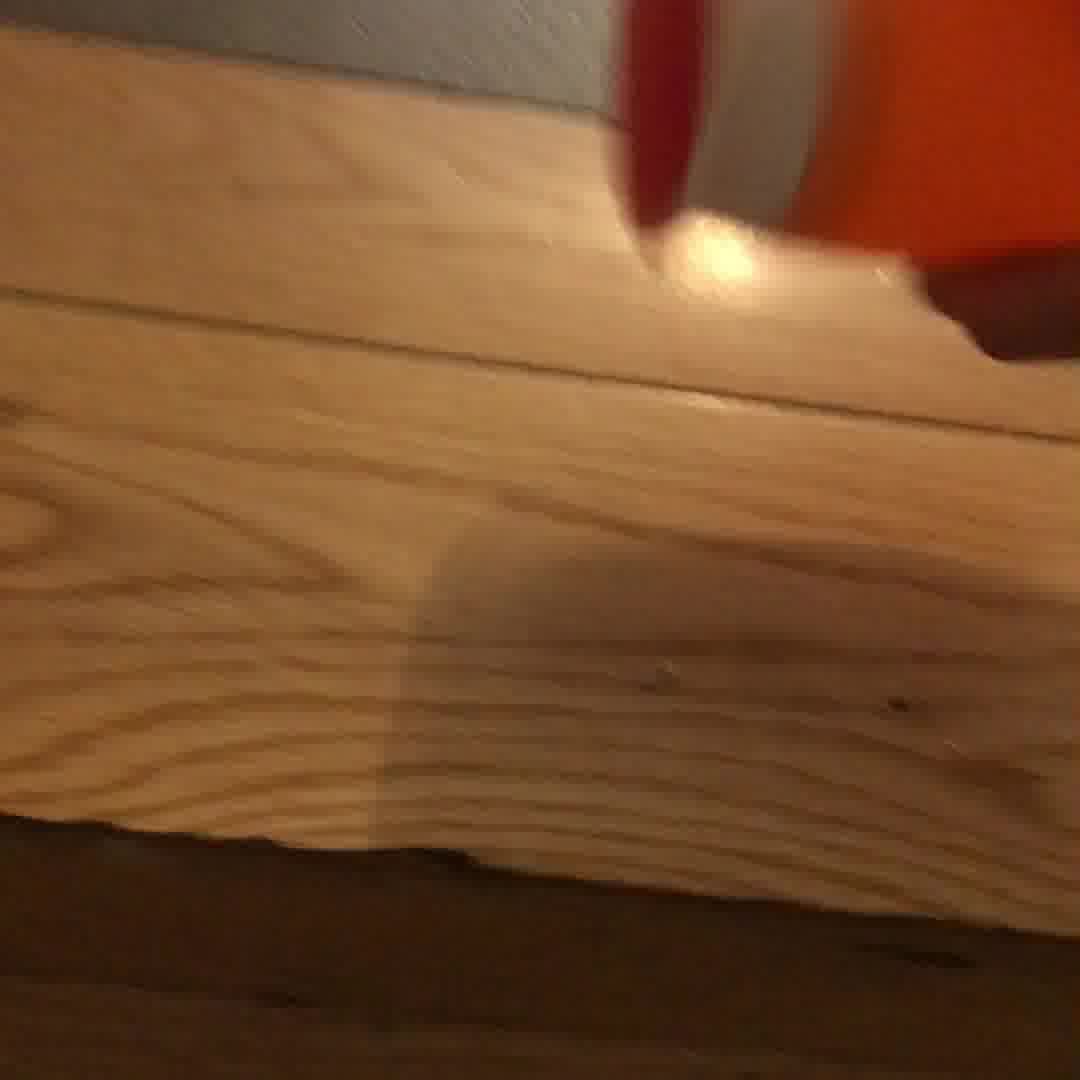}}
    \mbox{\includegraphics[width=0.095\textwidth]{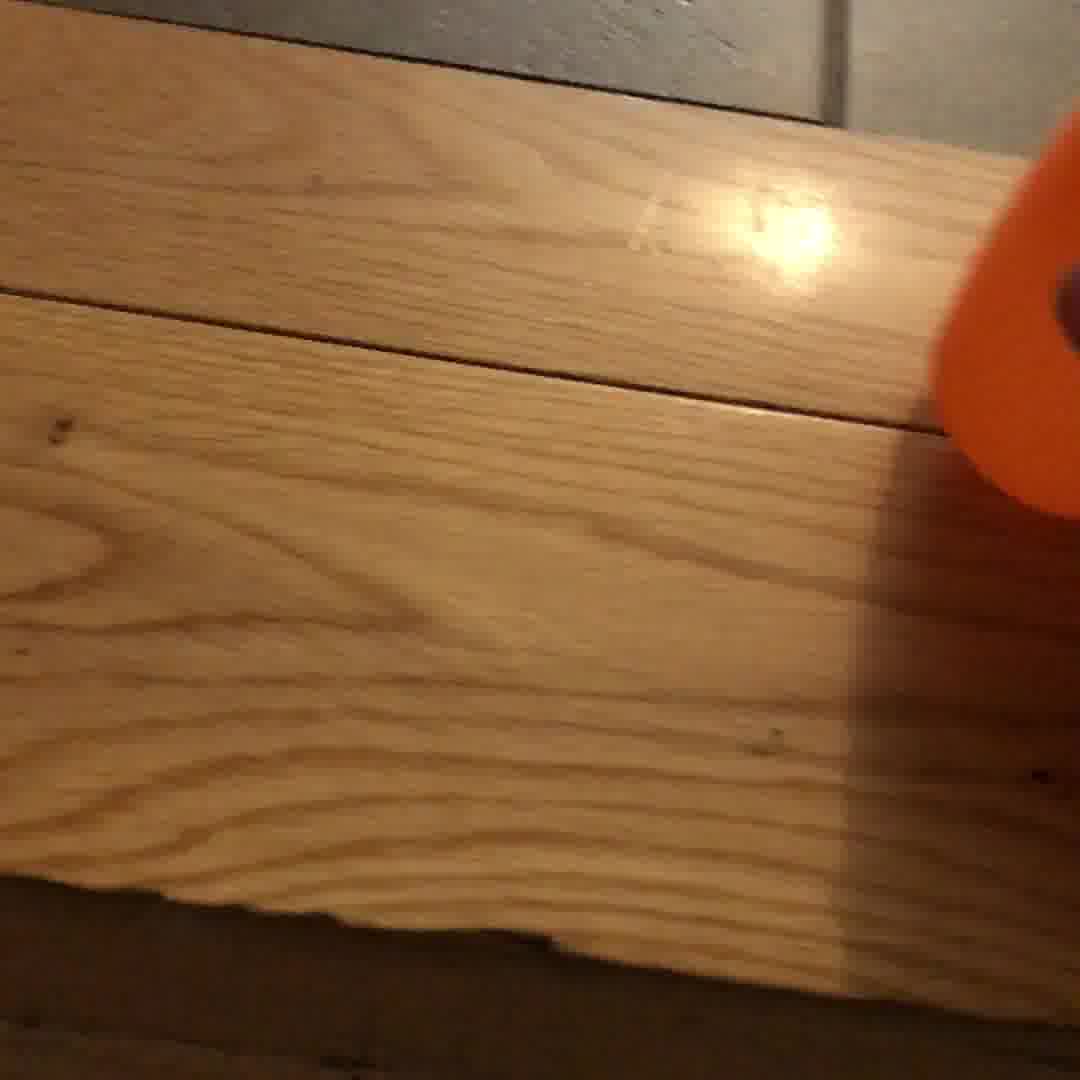}}
    \mbox{\includegraphics[width=0.095\textwidth]{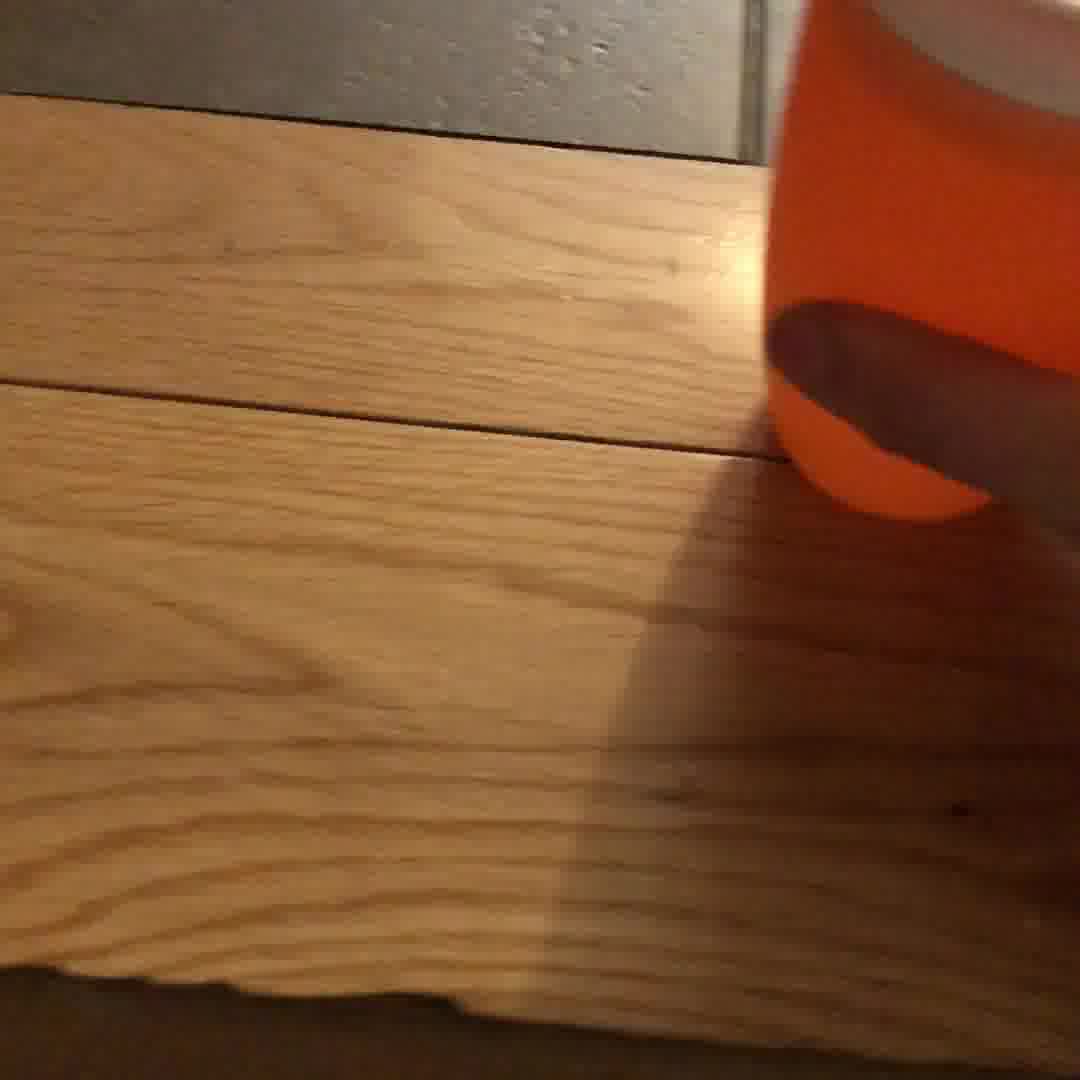}}
    \mbox{\includegraphics[width=0.095\textwidth]{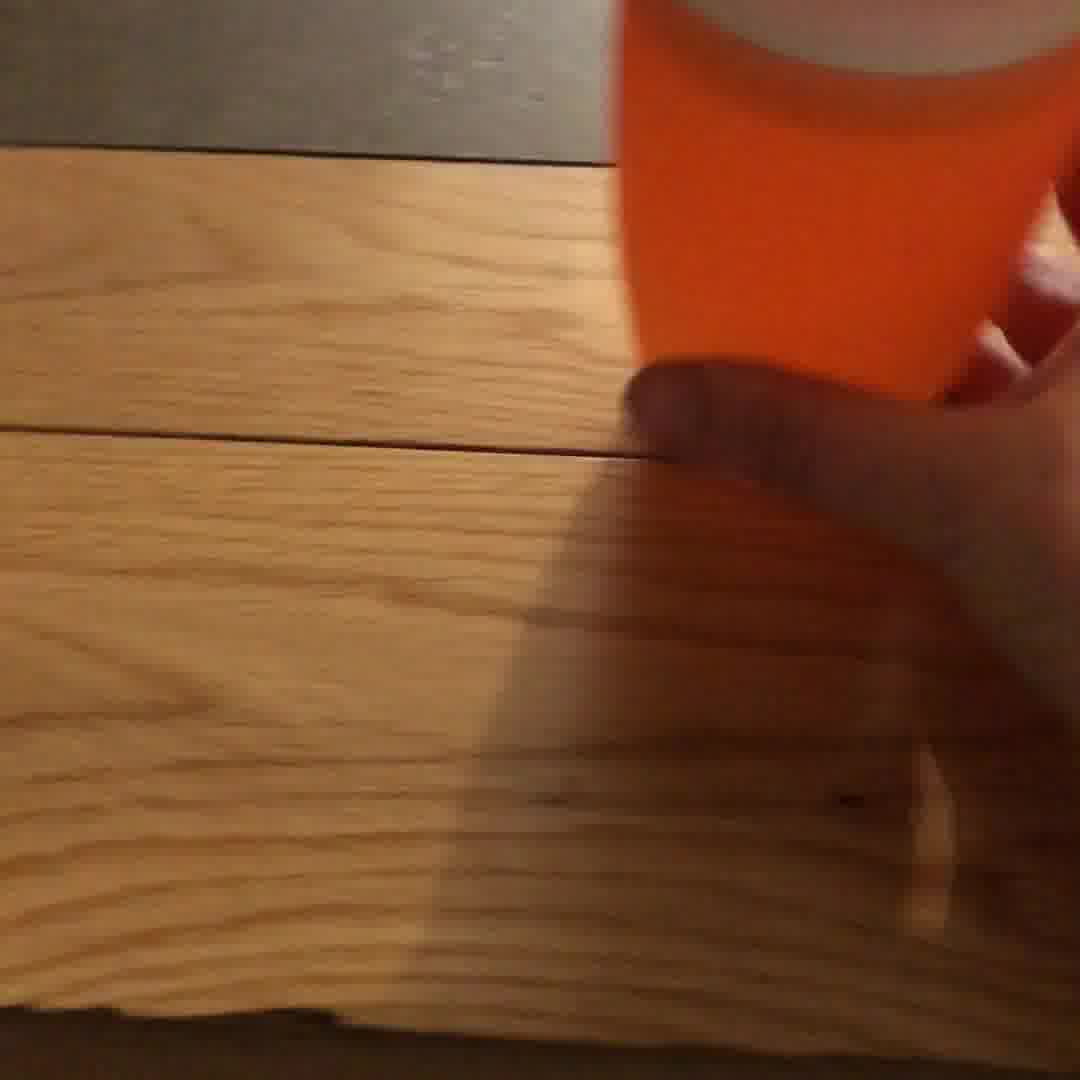}}
    \mbox{\includegraphics[width=0.095\textwidth]{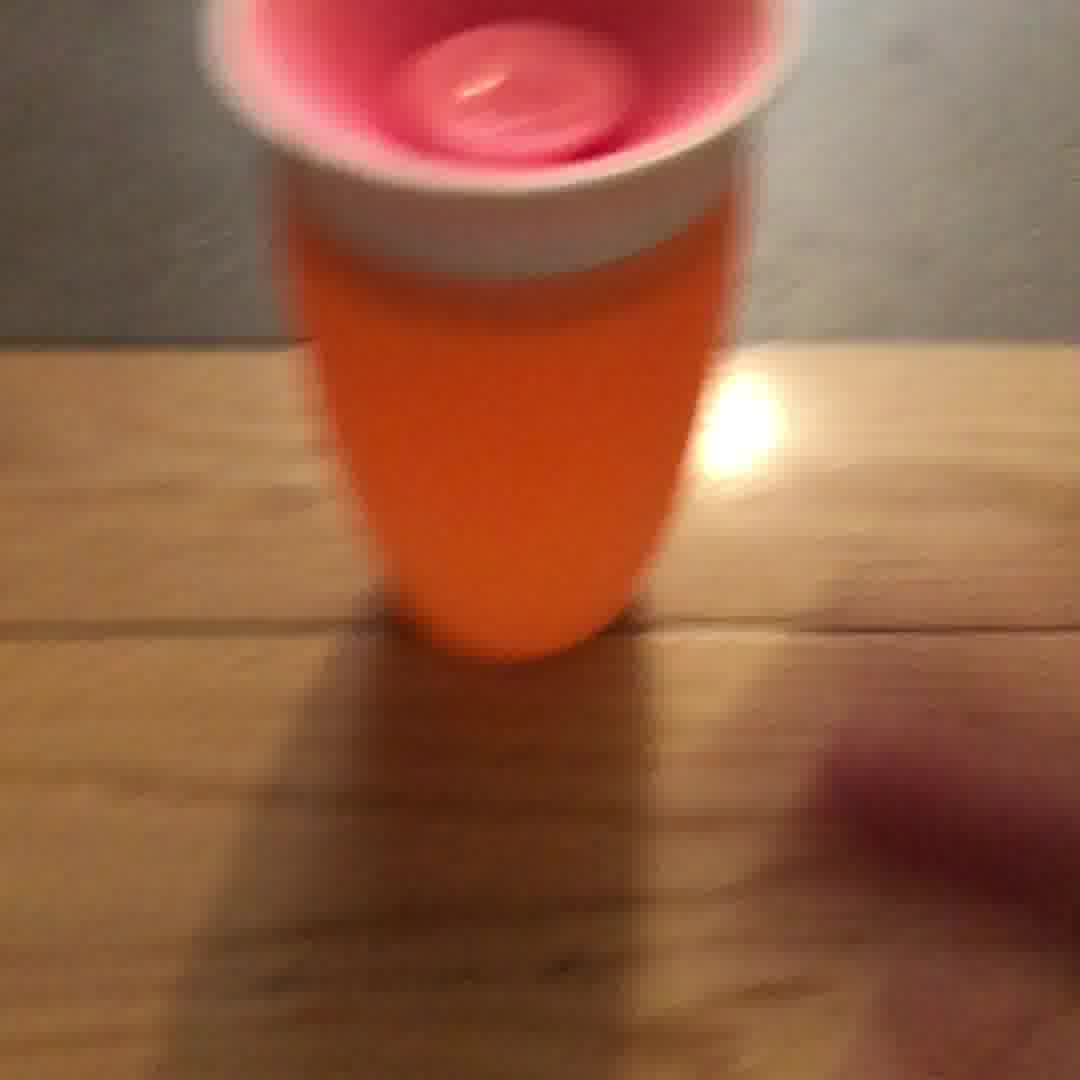}}
    \mbox{\includegraphics[width=0.095\textwidth]{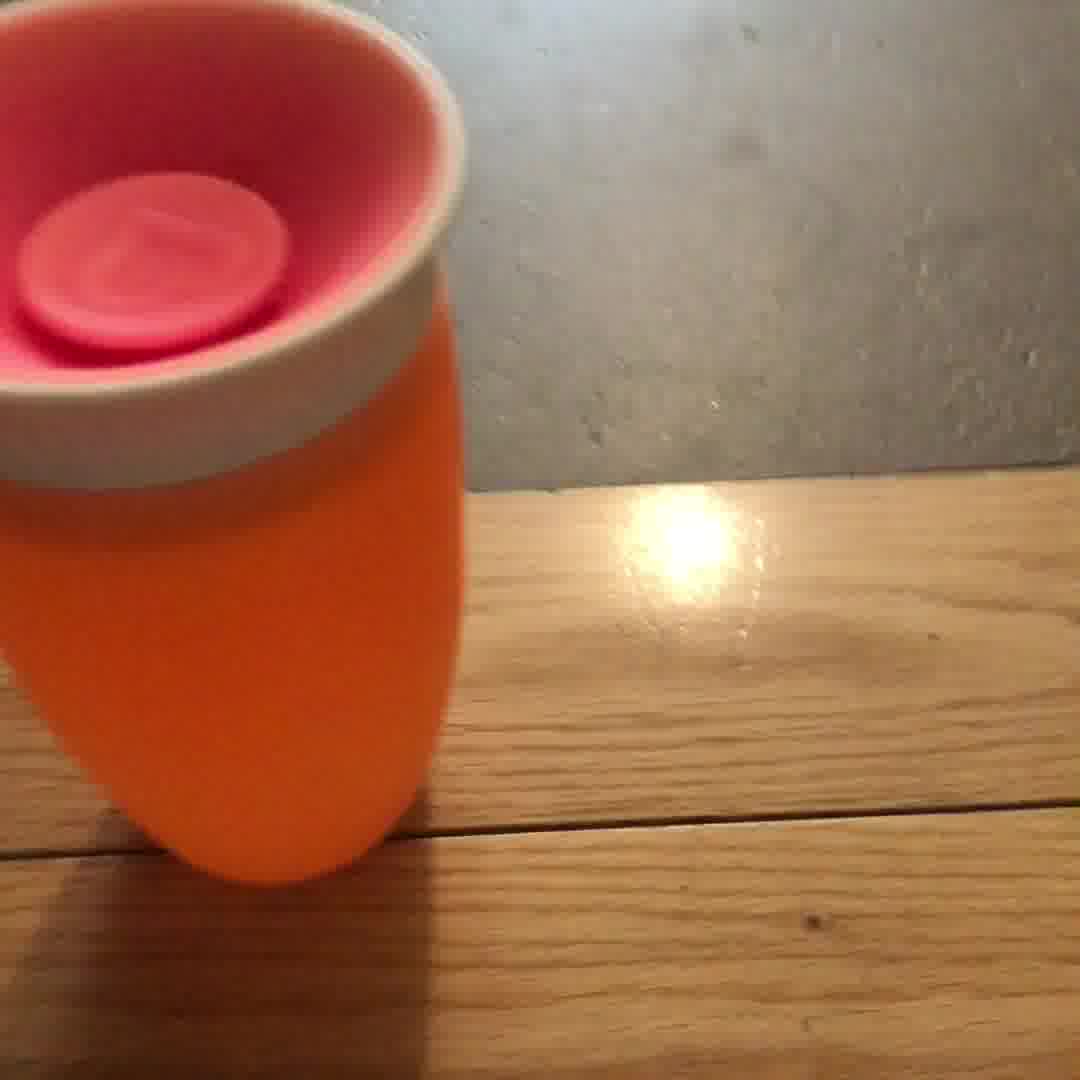}}
    \mbox{\includegraphics[width=0.095\textwidth]{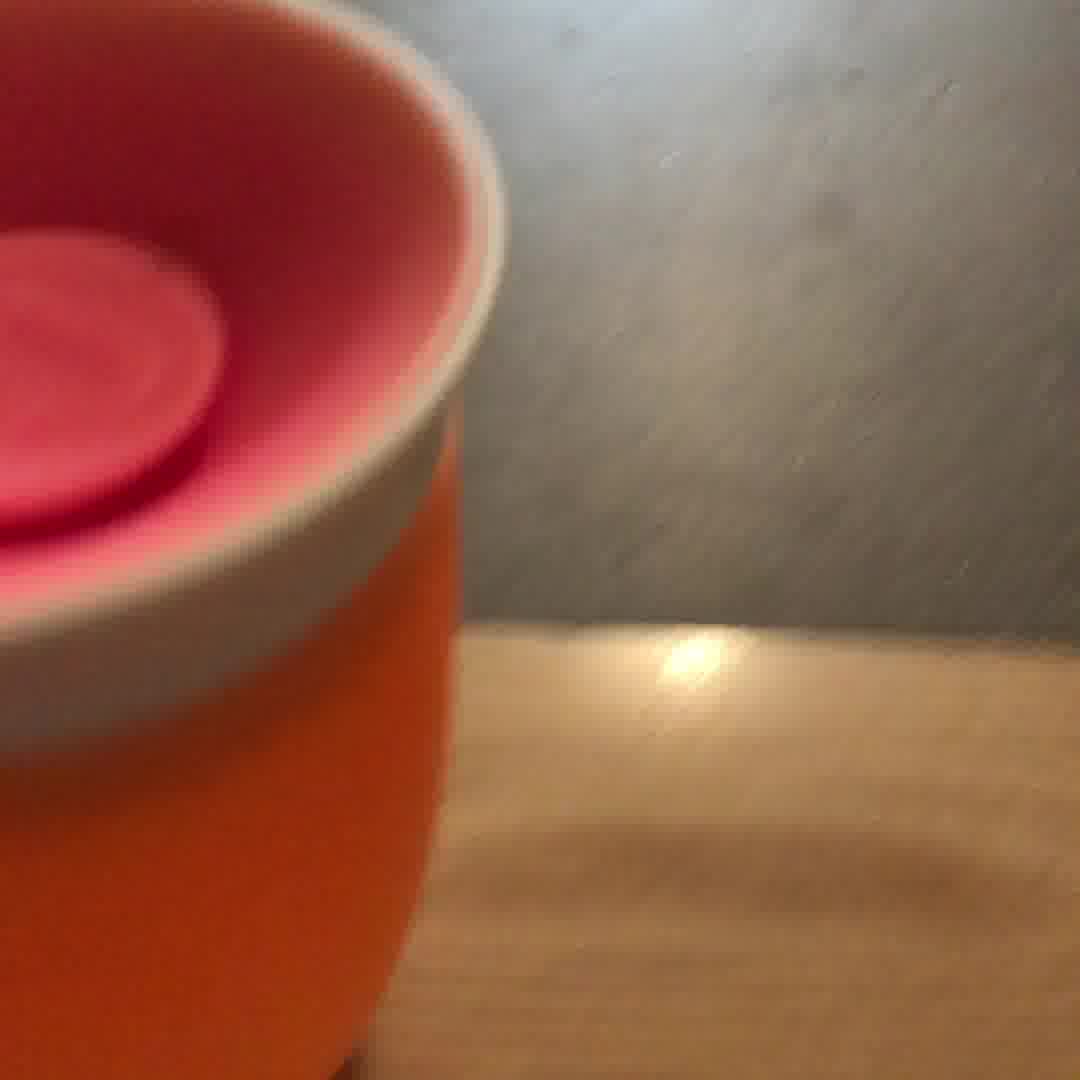}}
    \mbox{\includegraphics[width=0.095\textwidth]{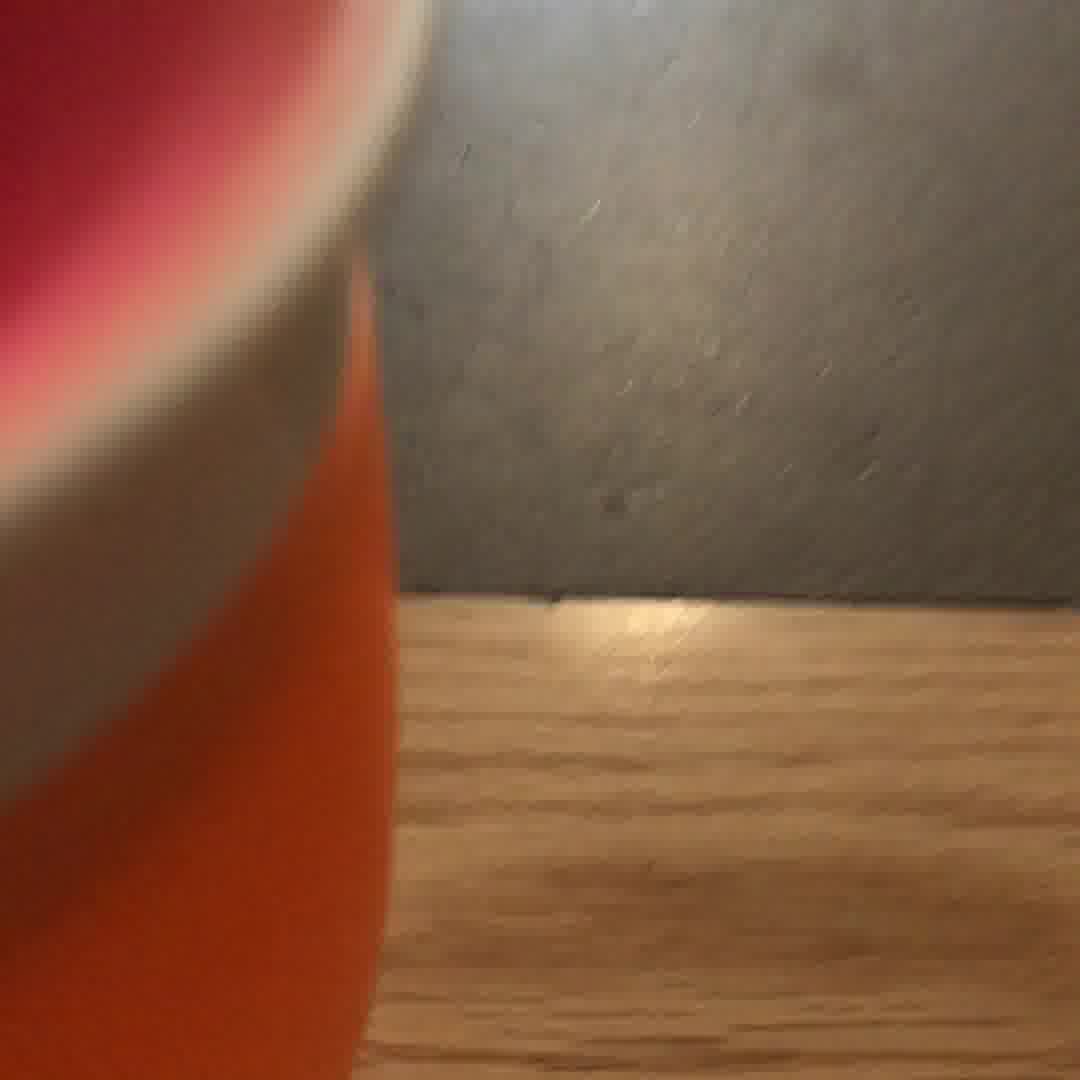}}
    \mbox{\includegraphics[width=0.095\textwidth]{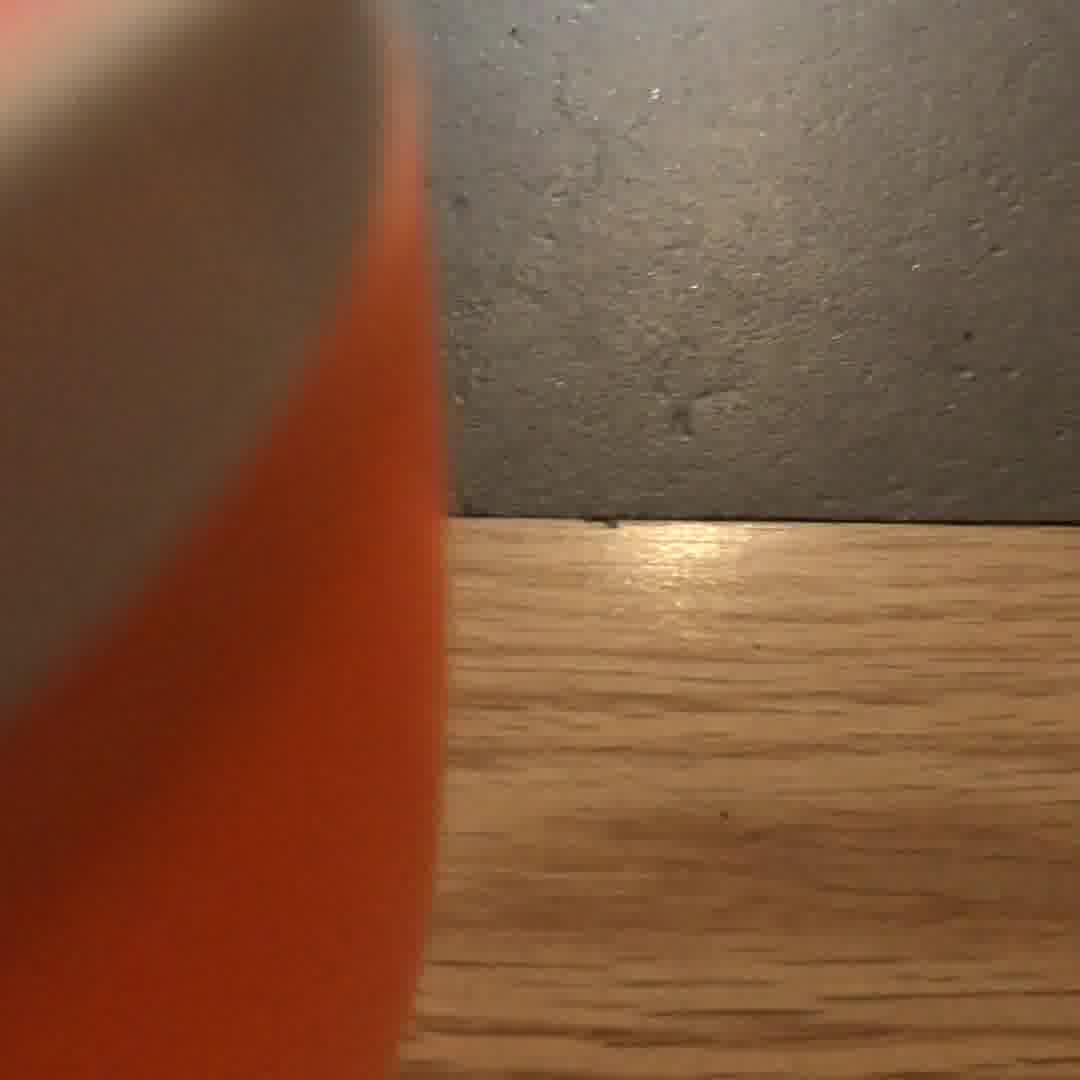}}
    \mbox{\includegraphics[width=0.095\textwidth]{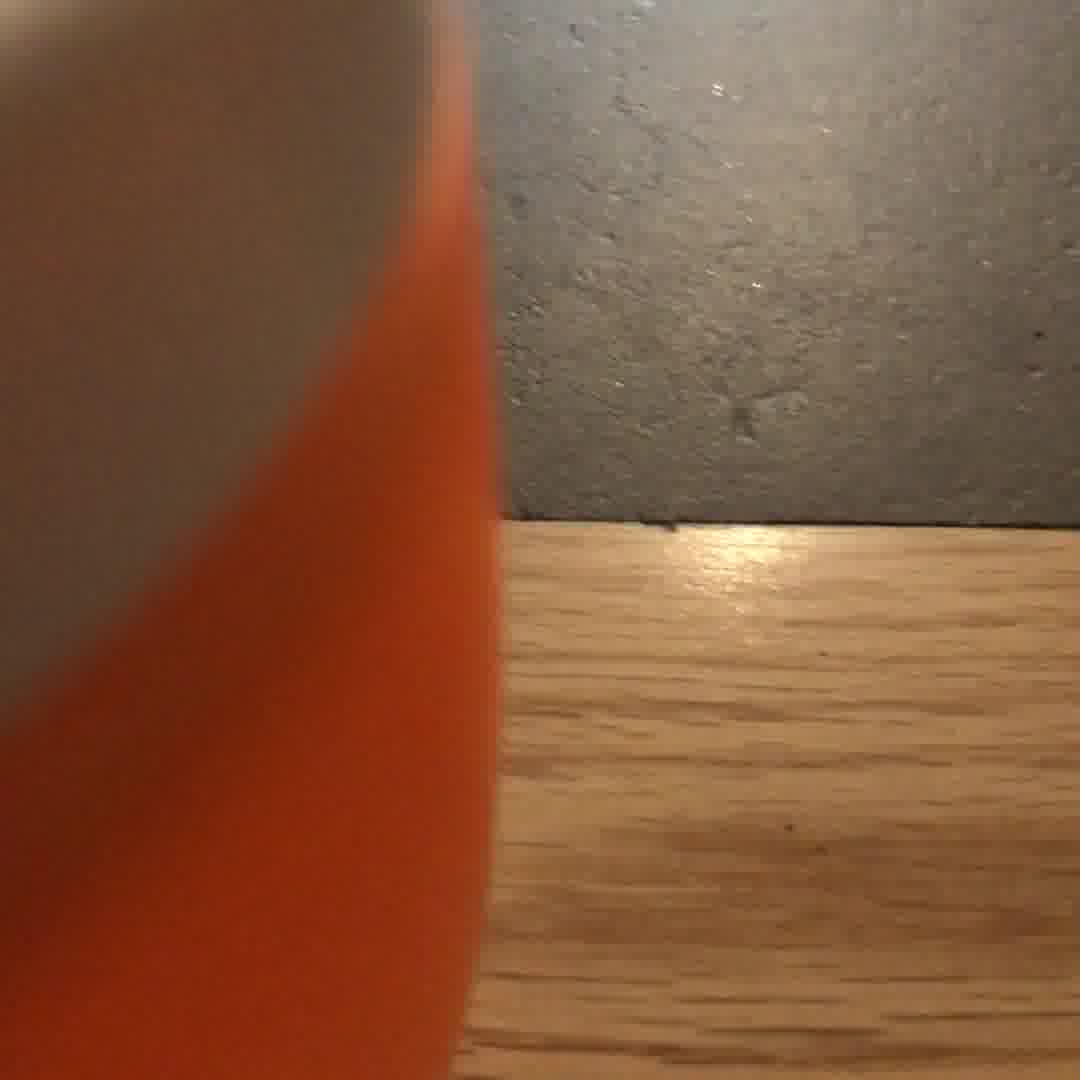}}}\\
    \vspace*{2px}
    \scalebox{0.95}{
    \mbox{\includegraphics[width=0.095\textwidth]{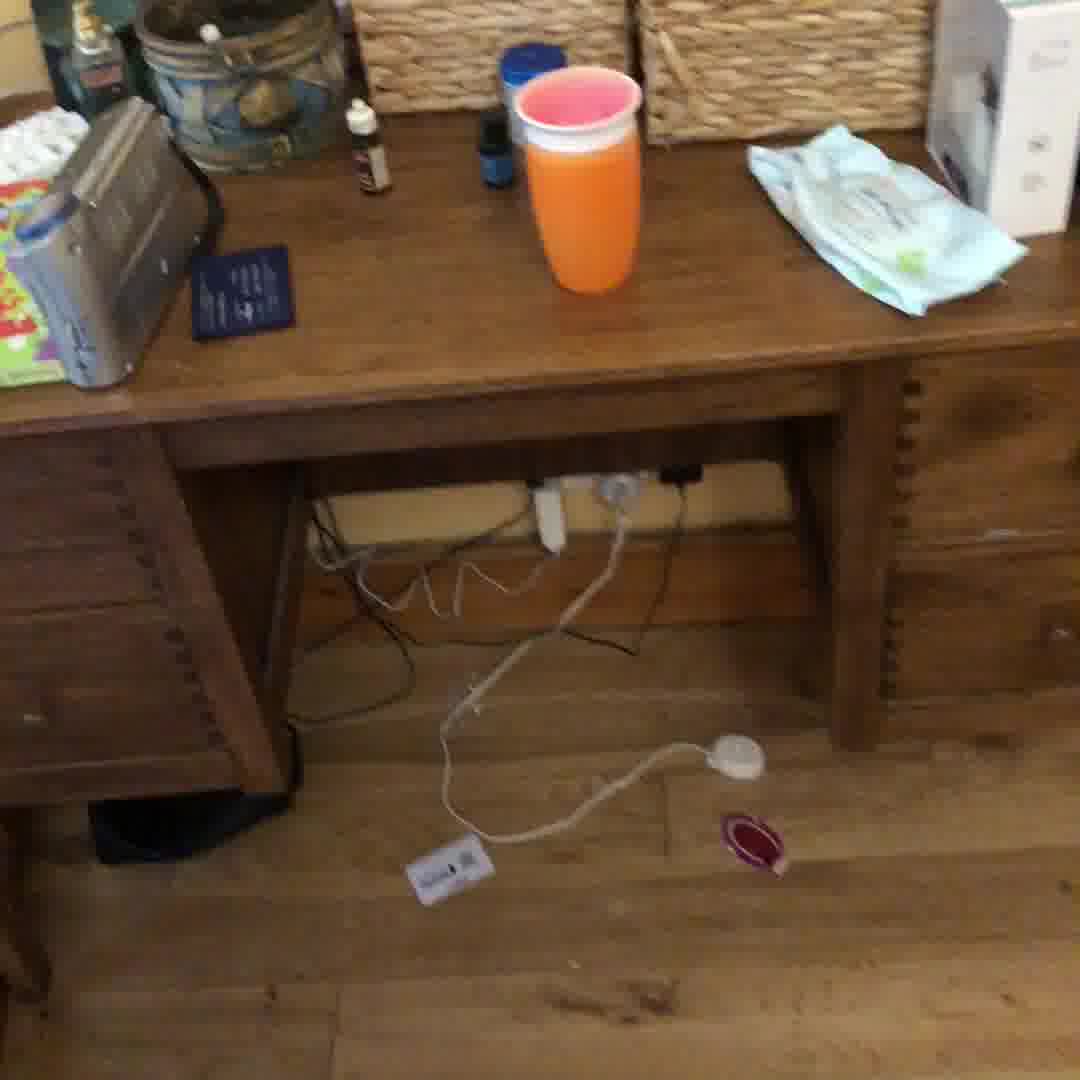}}
    \mbox{\includegraphics[width=0.095\textwidth]{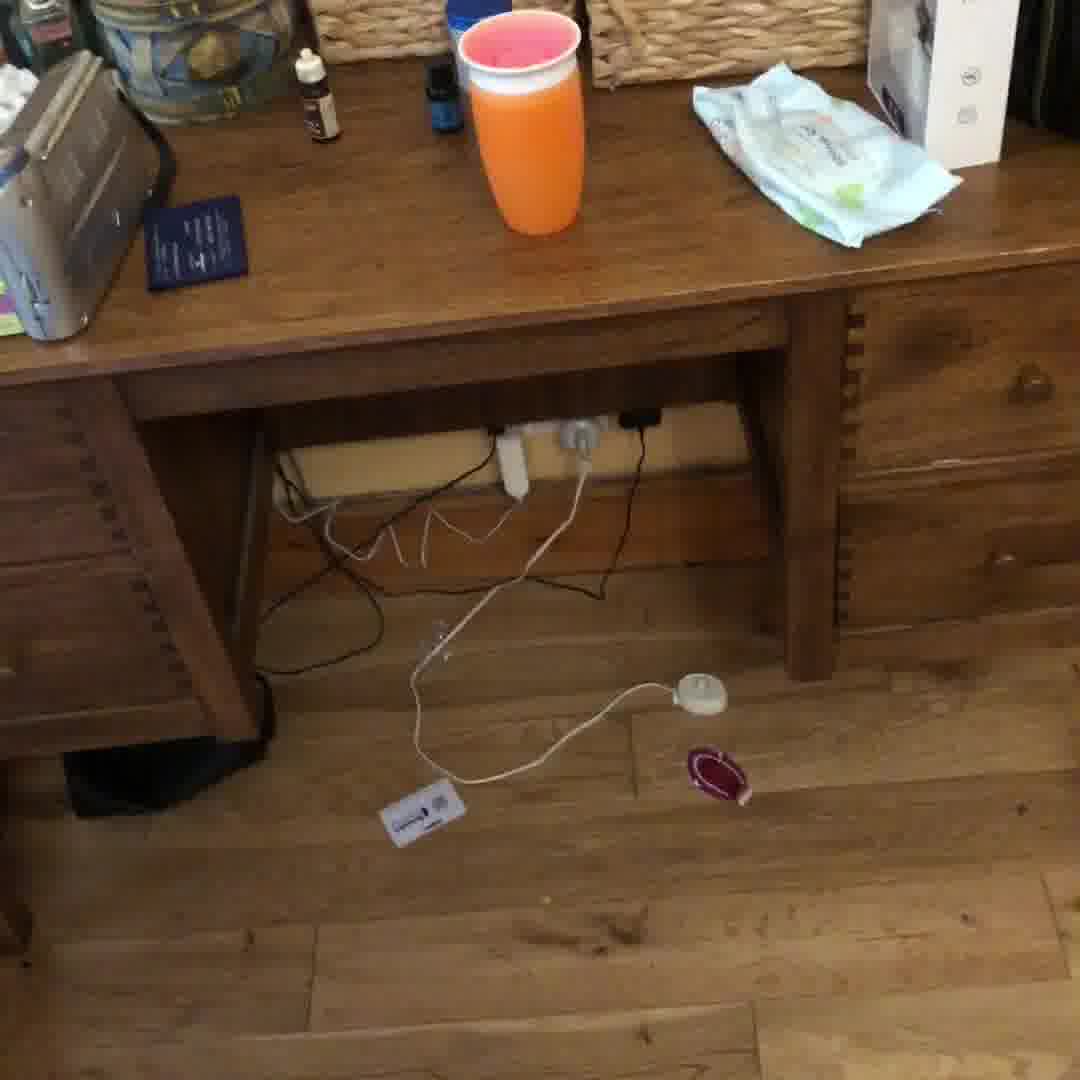}}
    \mbox{\includegraphics[width=0.095\textwidth]{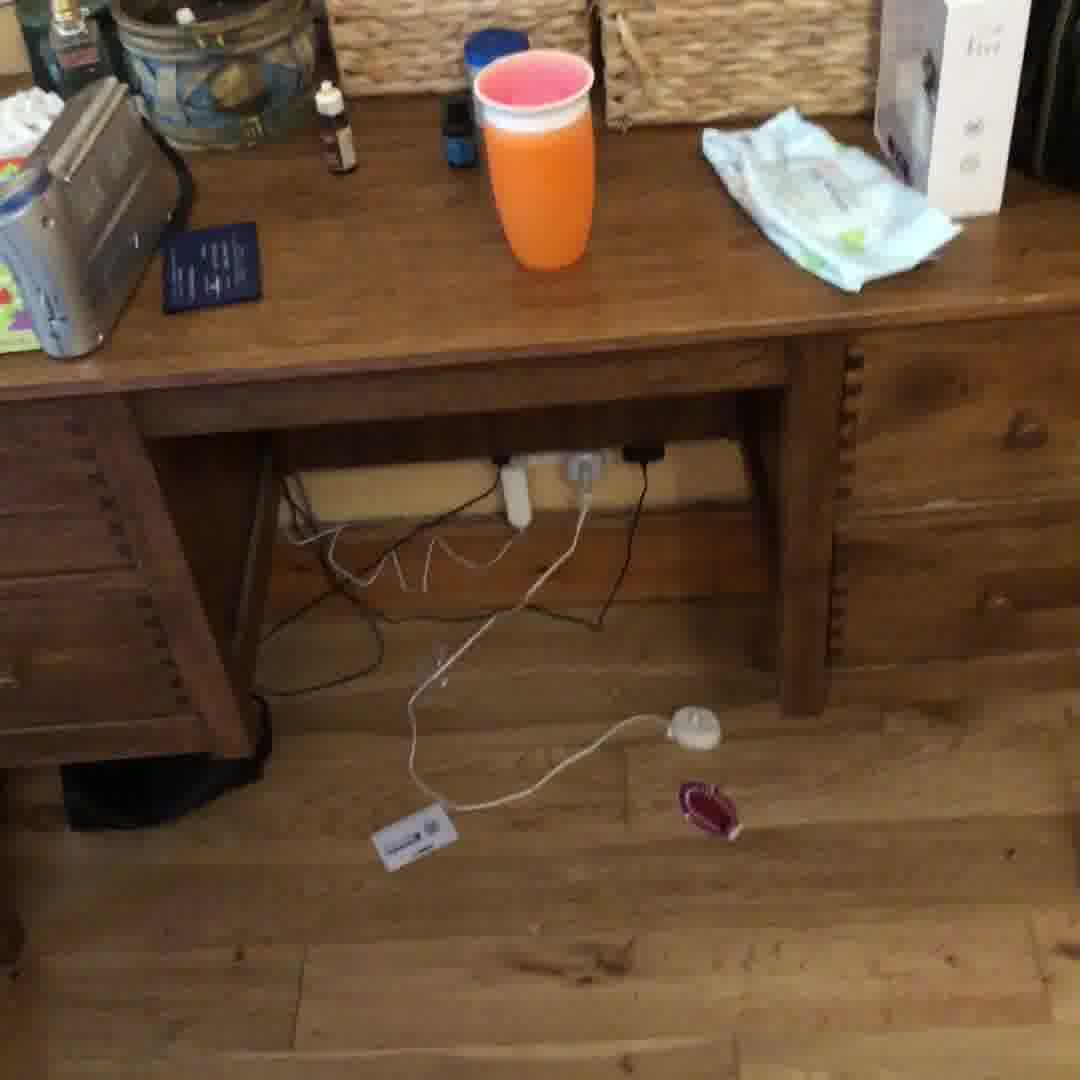}}
    \mbox{\includegraphics[width=0.095\textwidth]{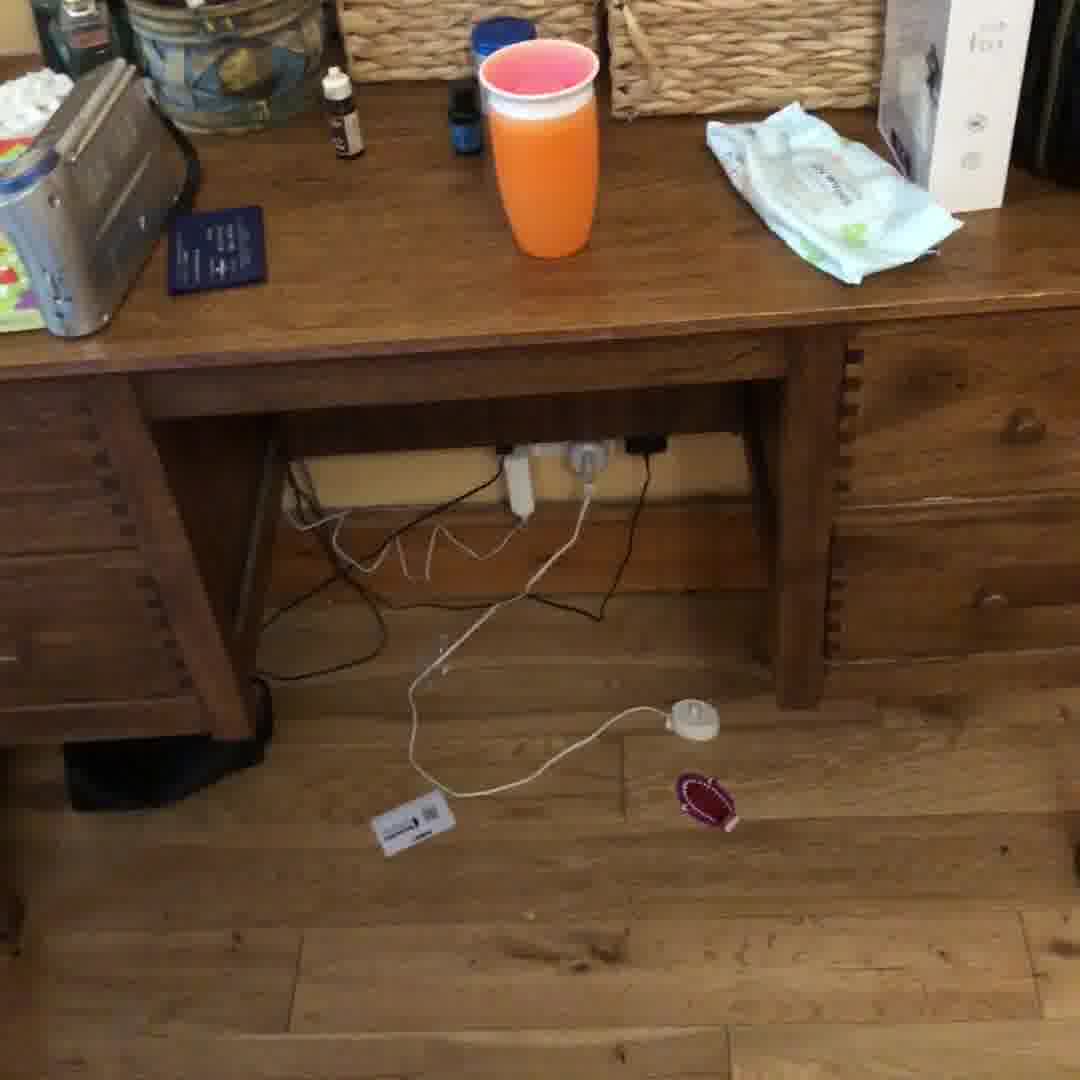}}
    \mbox{\includegraphics[width=0.095\textwidth]{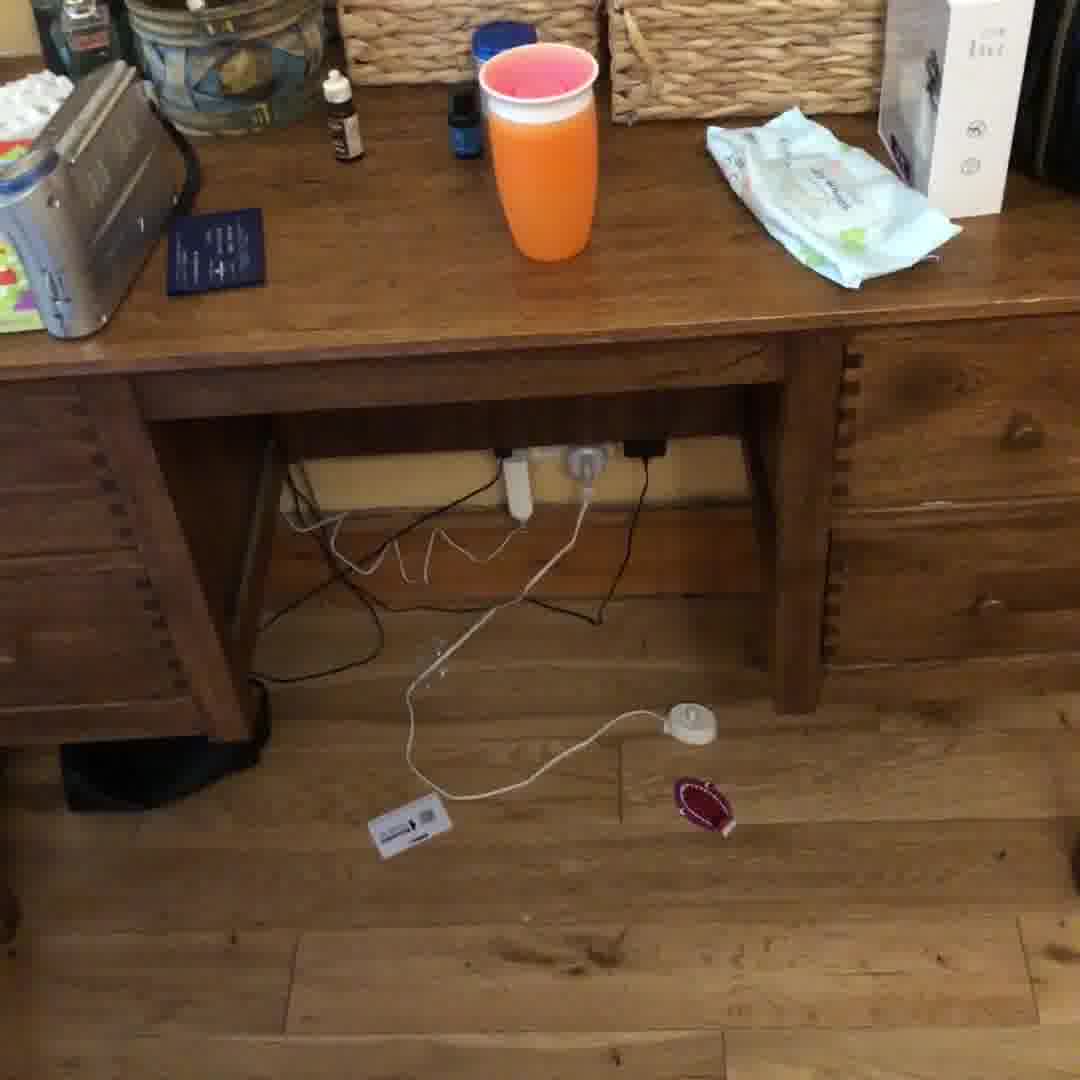}}
    \mbox{\includegraphics[width=0.095\textwidth]{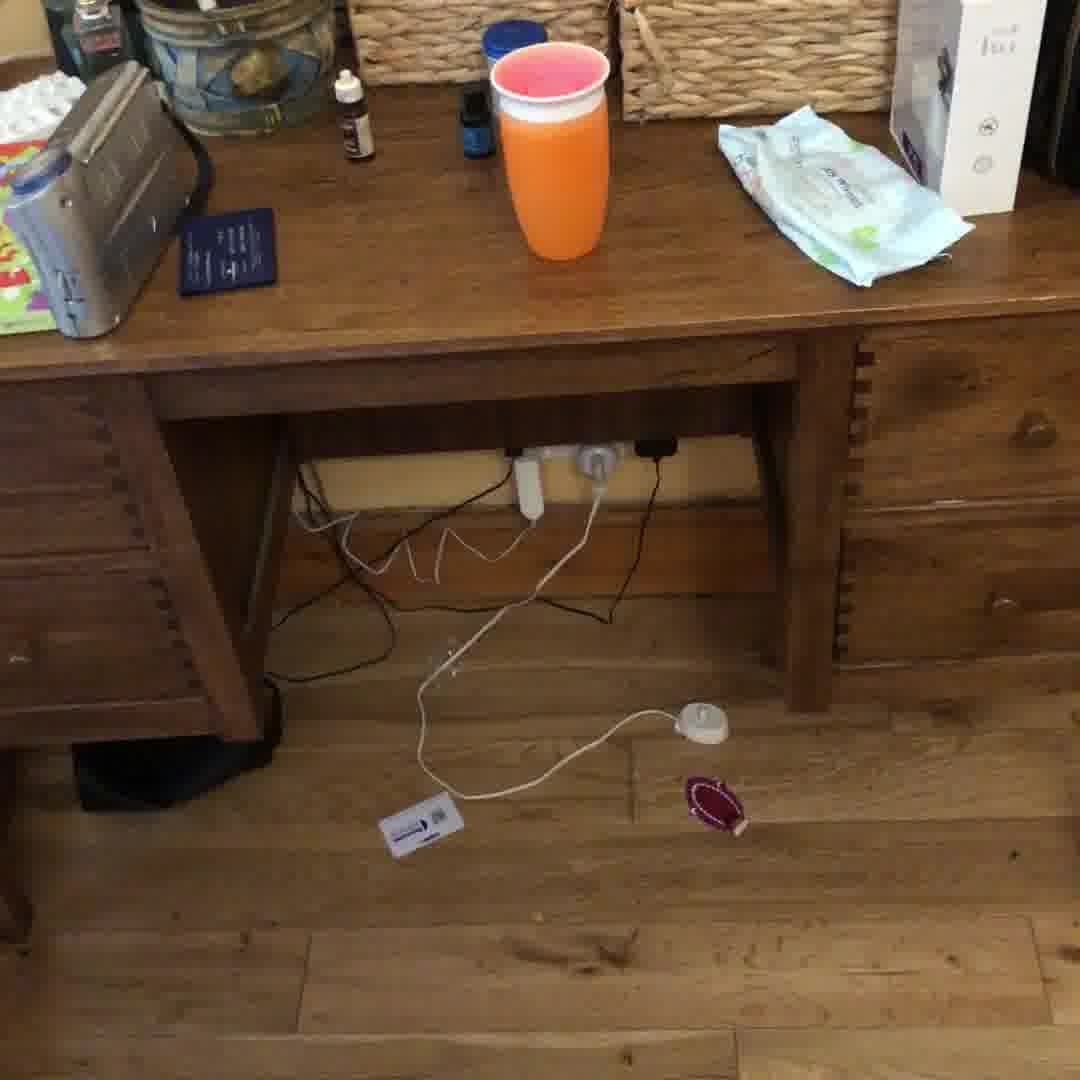}}
    \mbox{\includegraphics[width=0.095\textwidth]{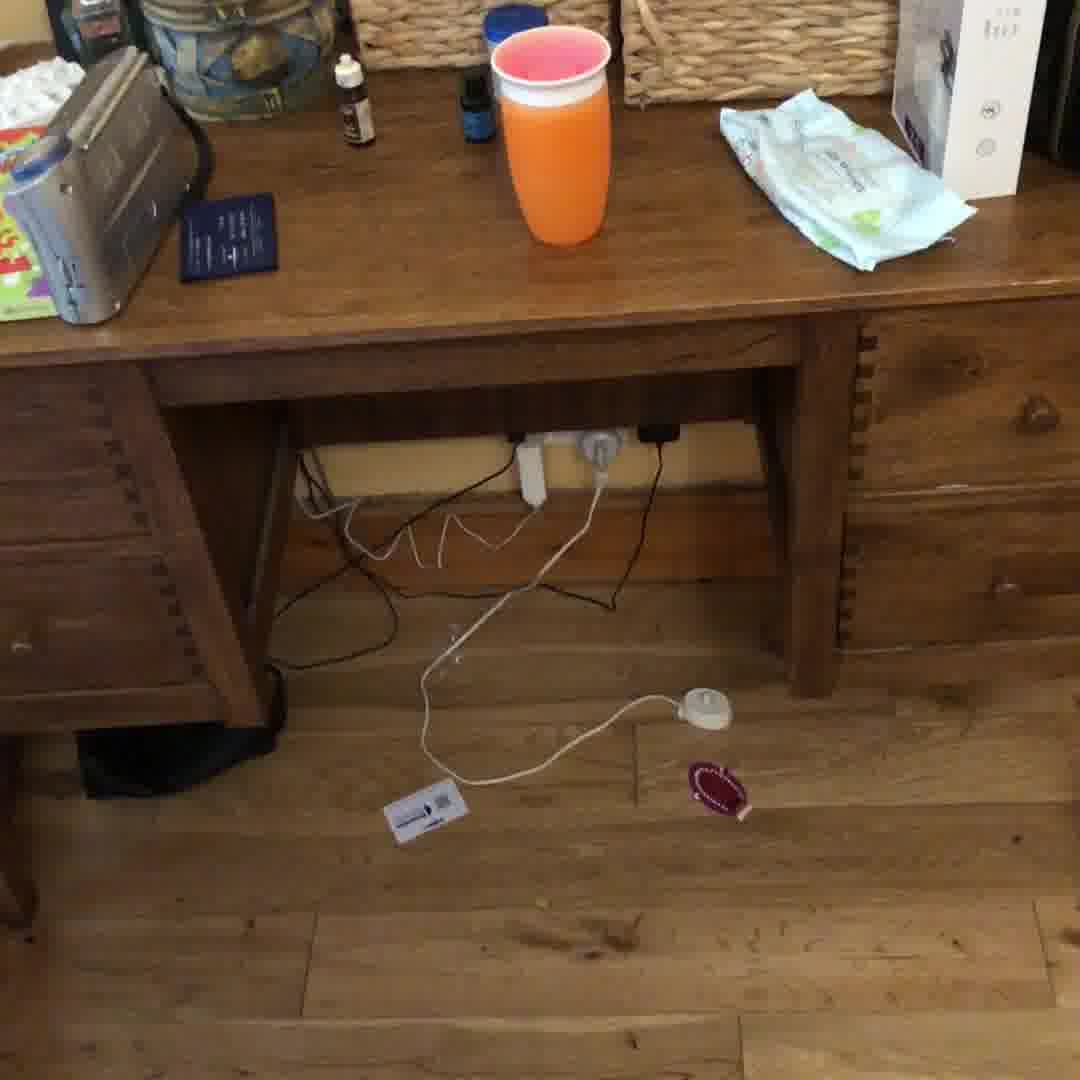}}
    \mbox{\includegraphics[width=0.095\textwidth]{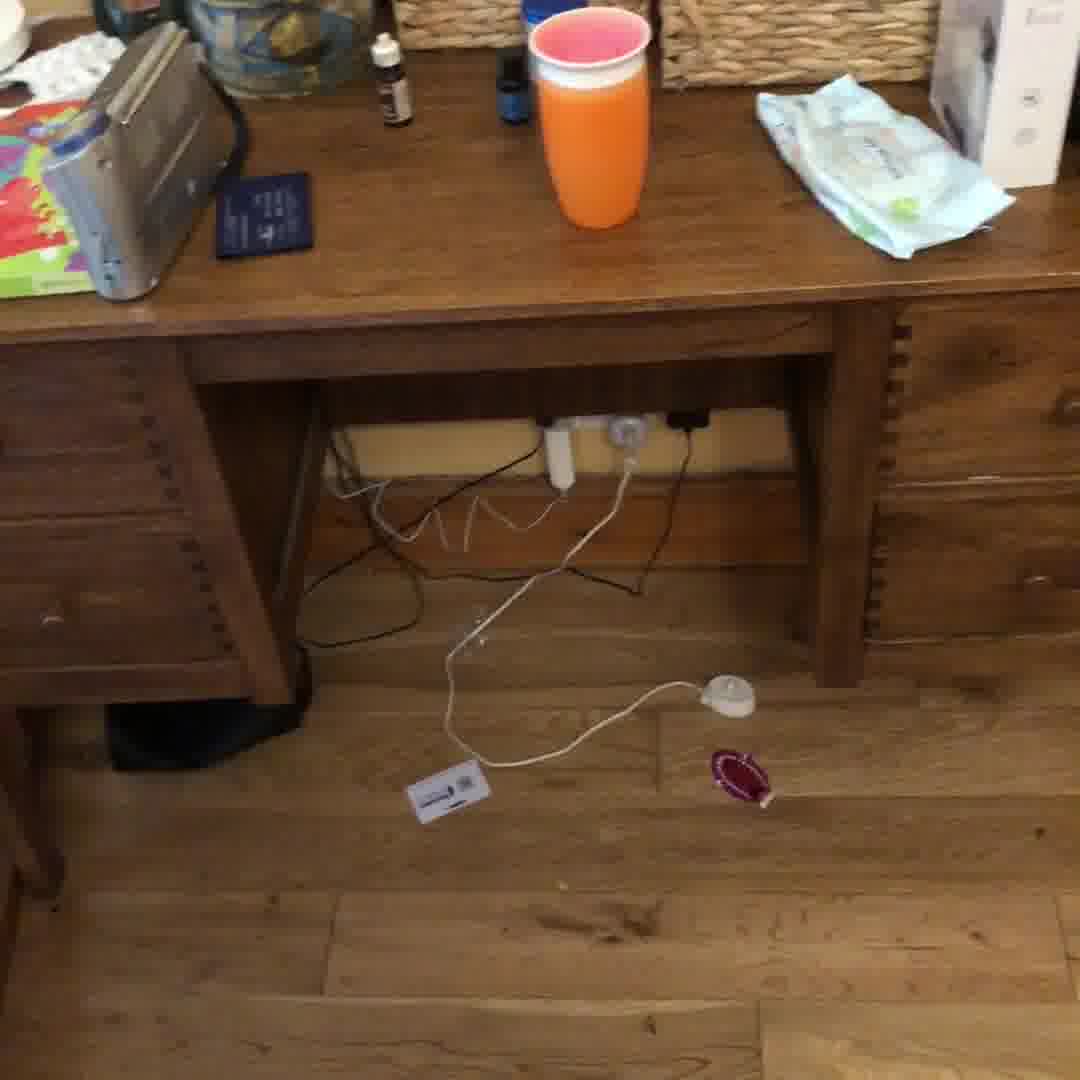}}
    \mbox{\includegraphics[width=0.095\textwidth]{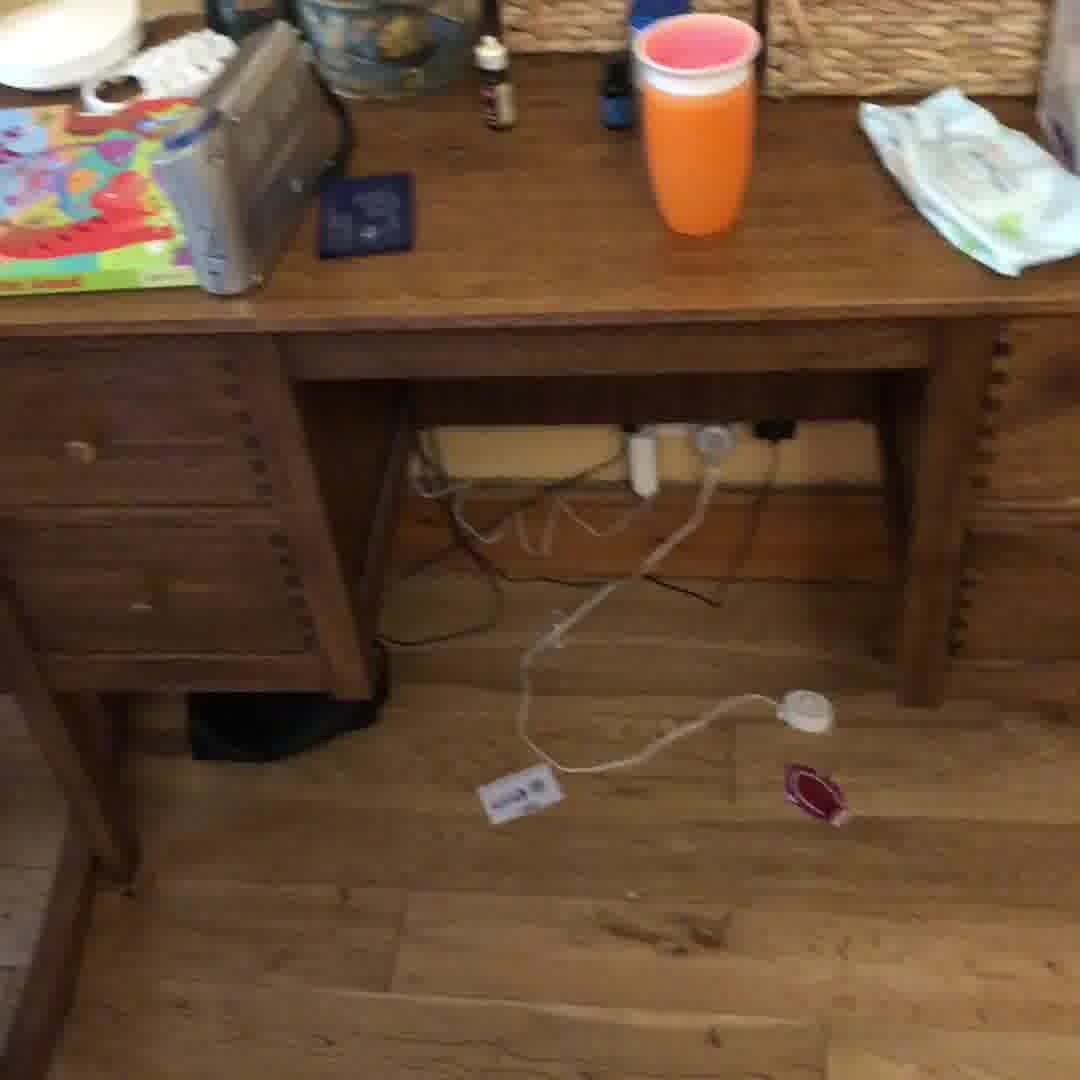}}
    \mbox{\includegraphics[width=0.095\textwidth]{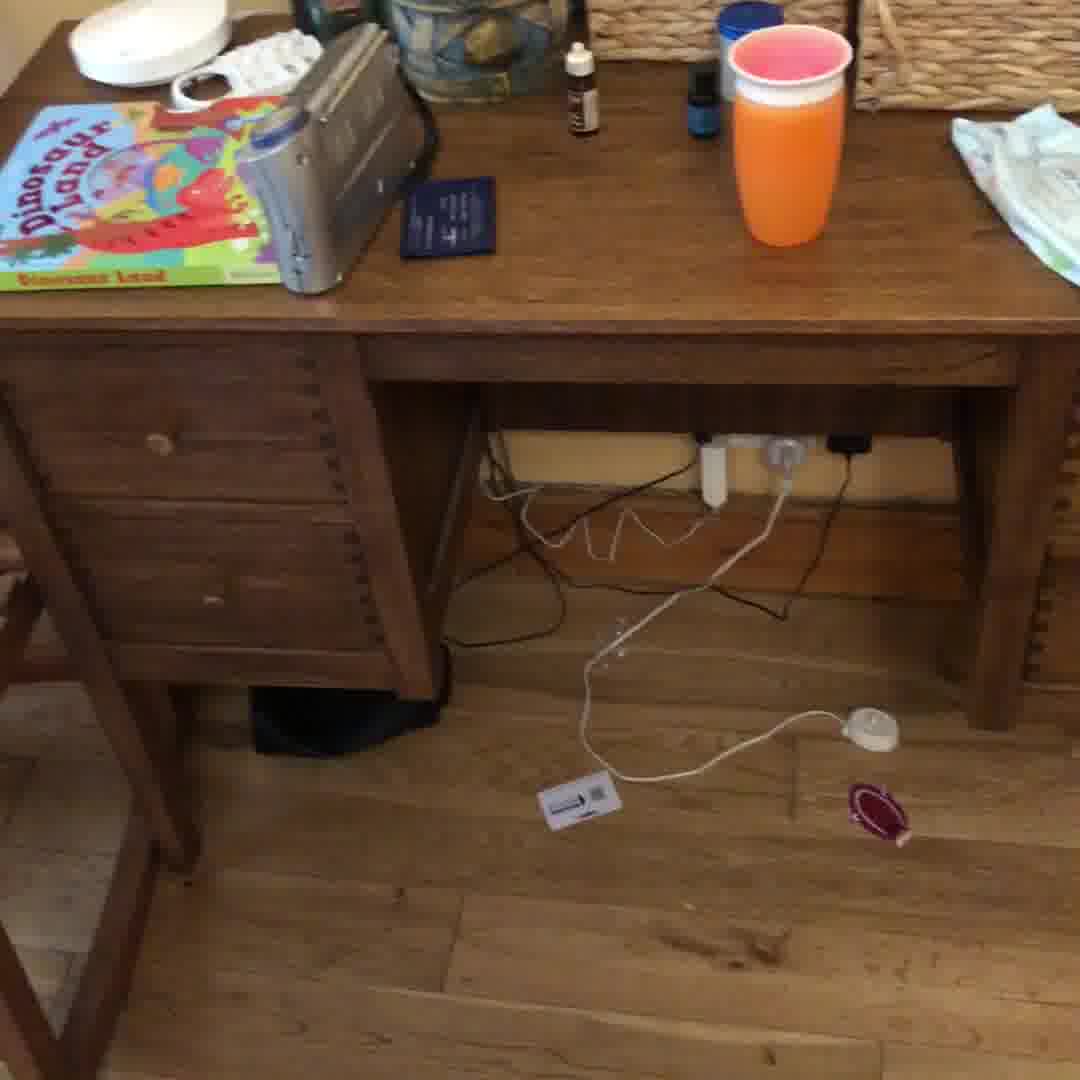}}}\\
    \vspace*{2px}
    \scalebox{0.95}{
    \mbox{\includegraphics[width=0.095\textwidth]{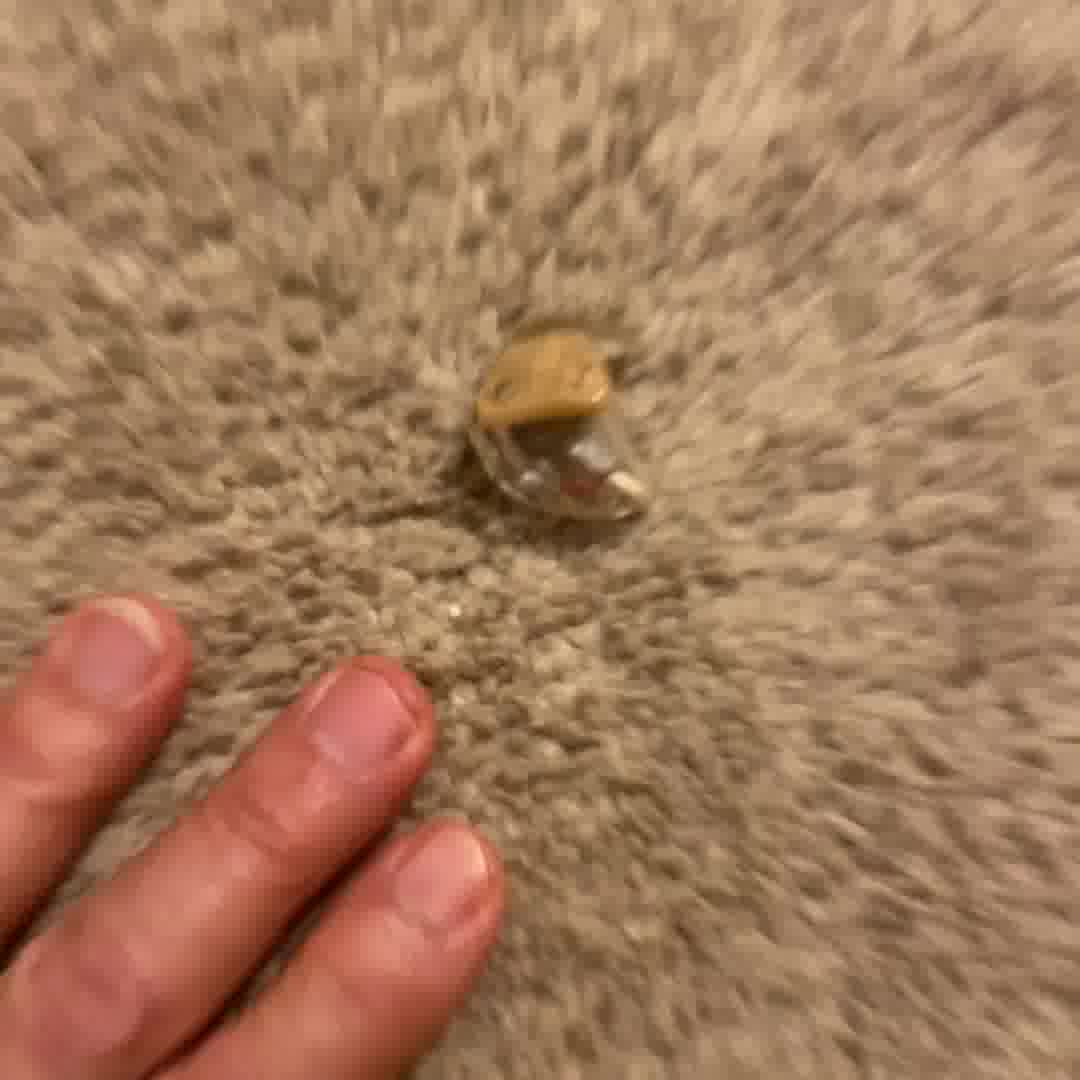}}
    \mbox{\includegraphics[width=0.095\textwidth]{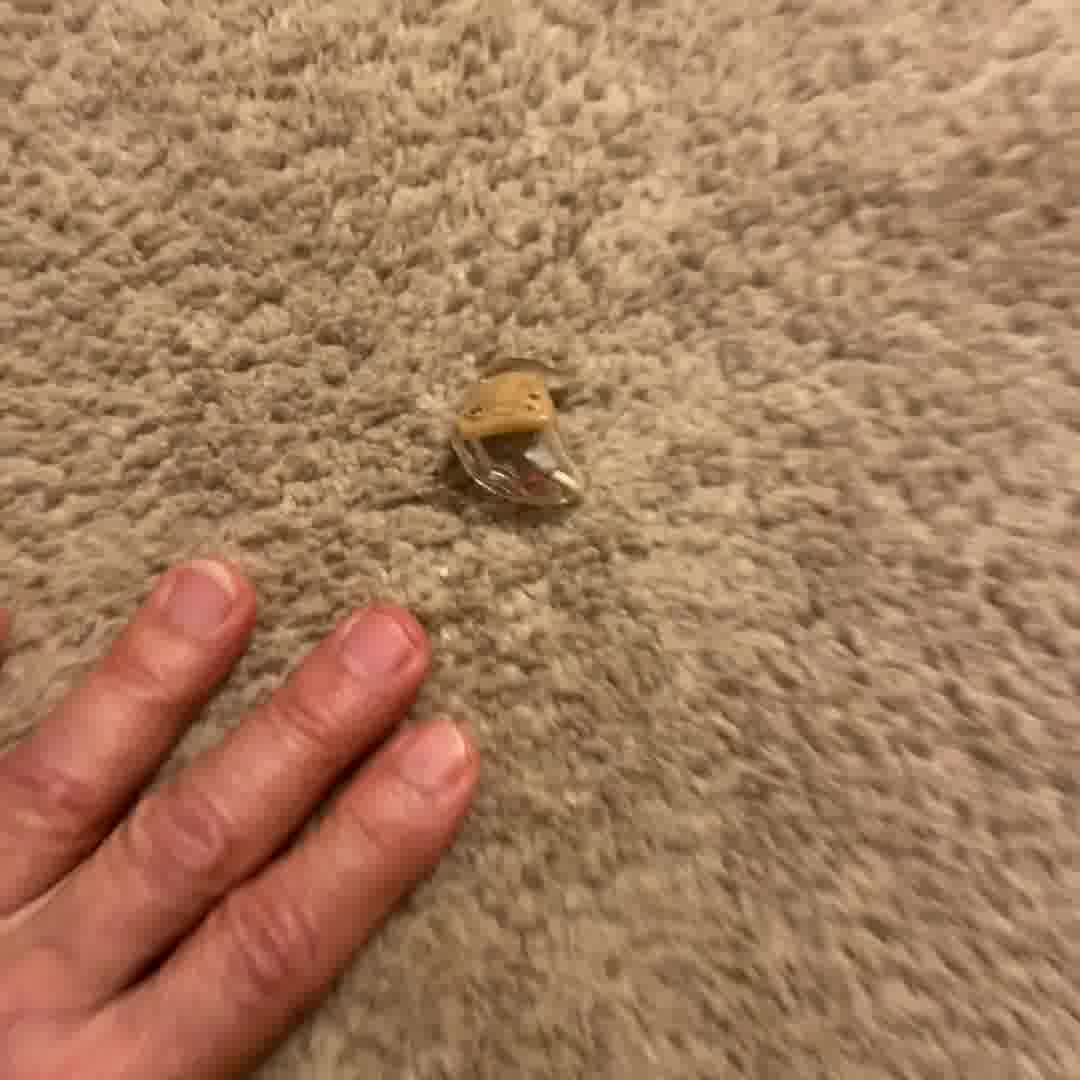}}
    \mbox{\includegraphics[width=0.095\textwidth]{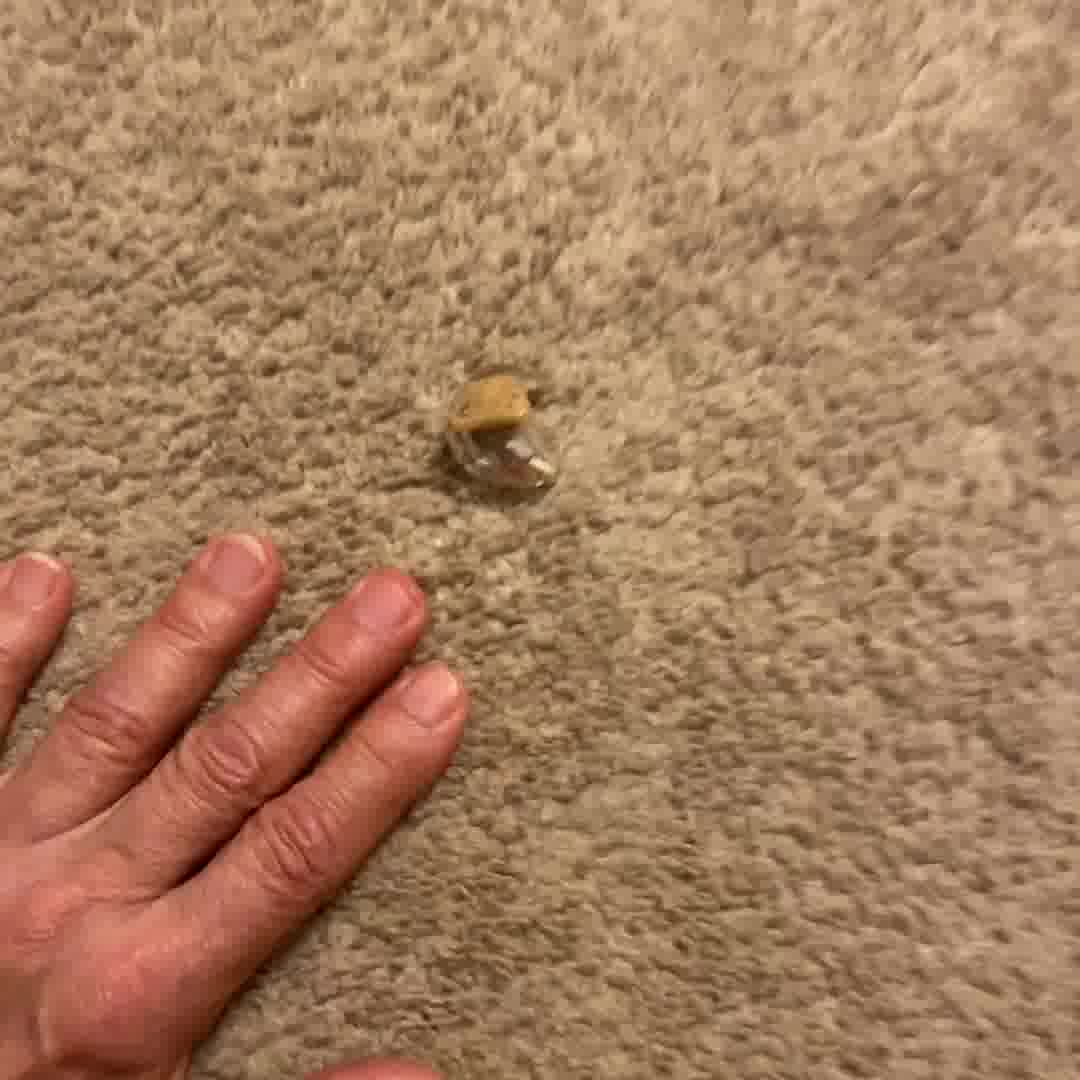}}
    \mbox{\includegraphics[width=0.095\textwidth]{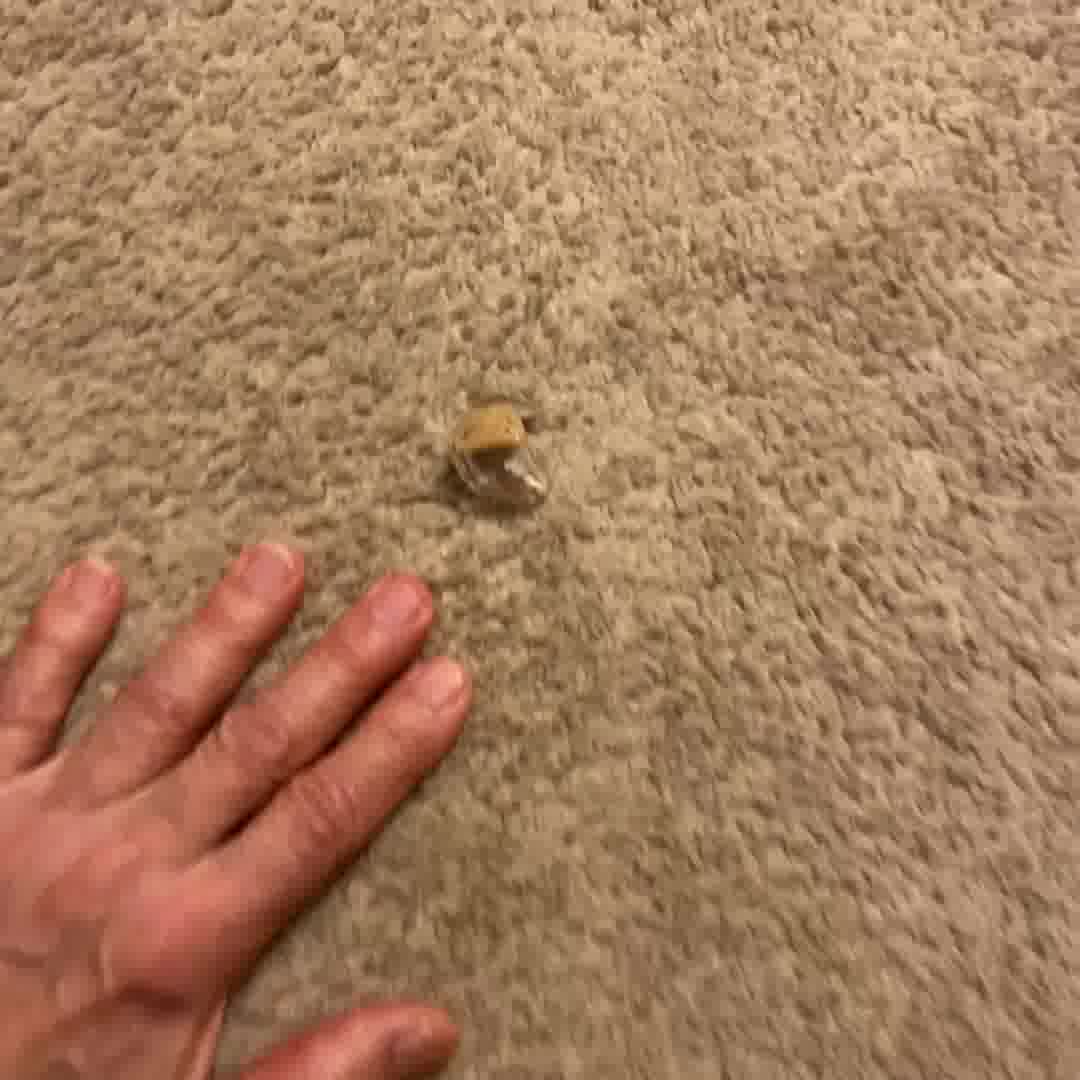}}
    \mbox{\includegraphics[width=0.095\textwidth]{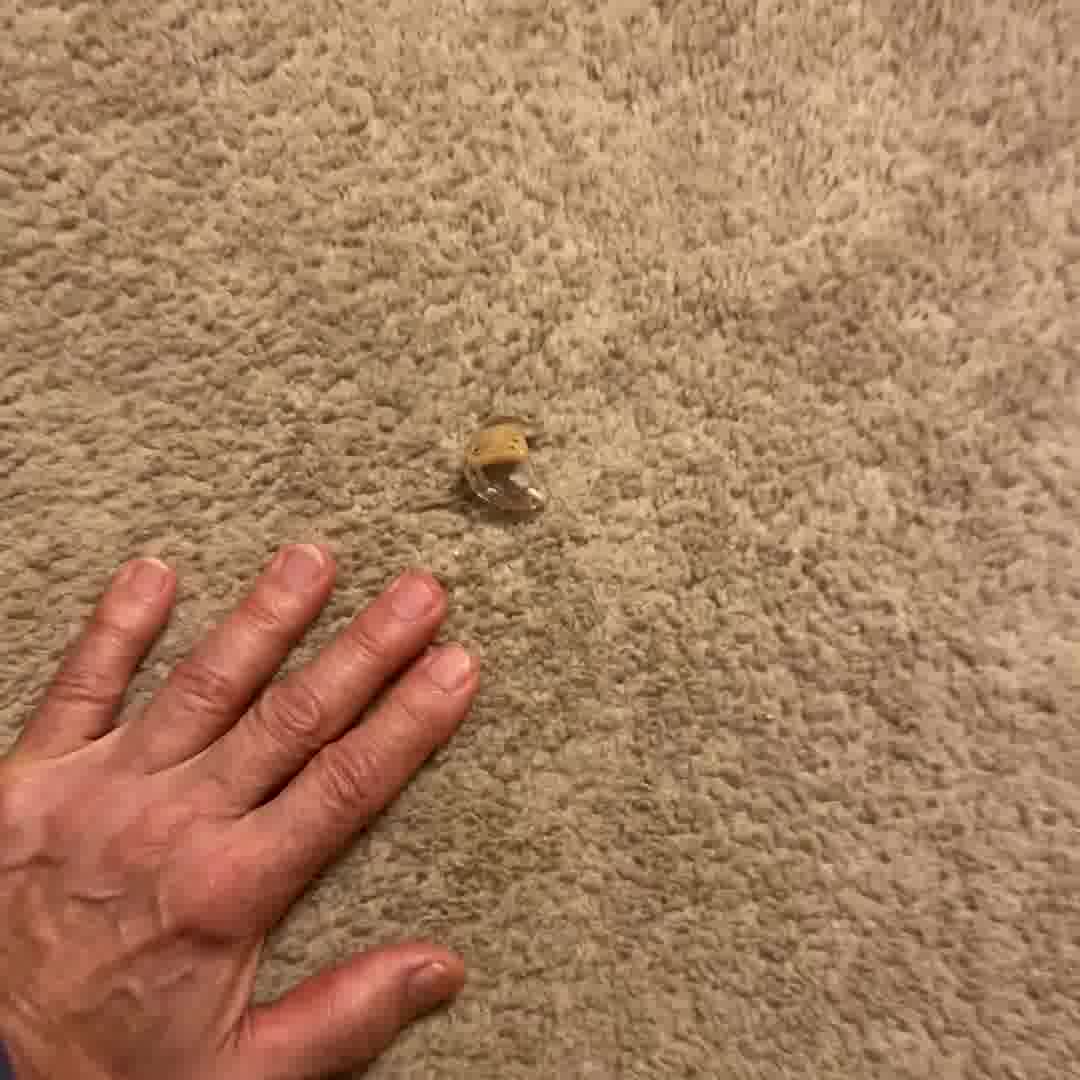}}
    \mbox{\includegraphics[width=0.095\textwidth]{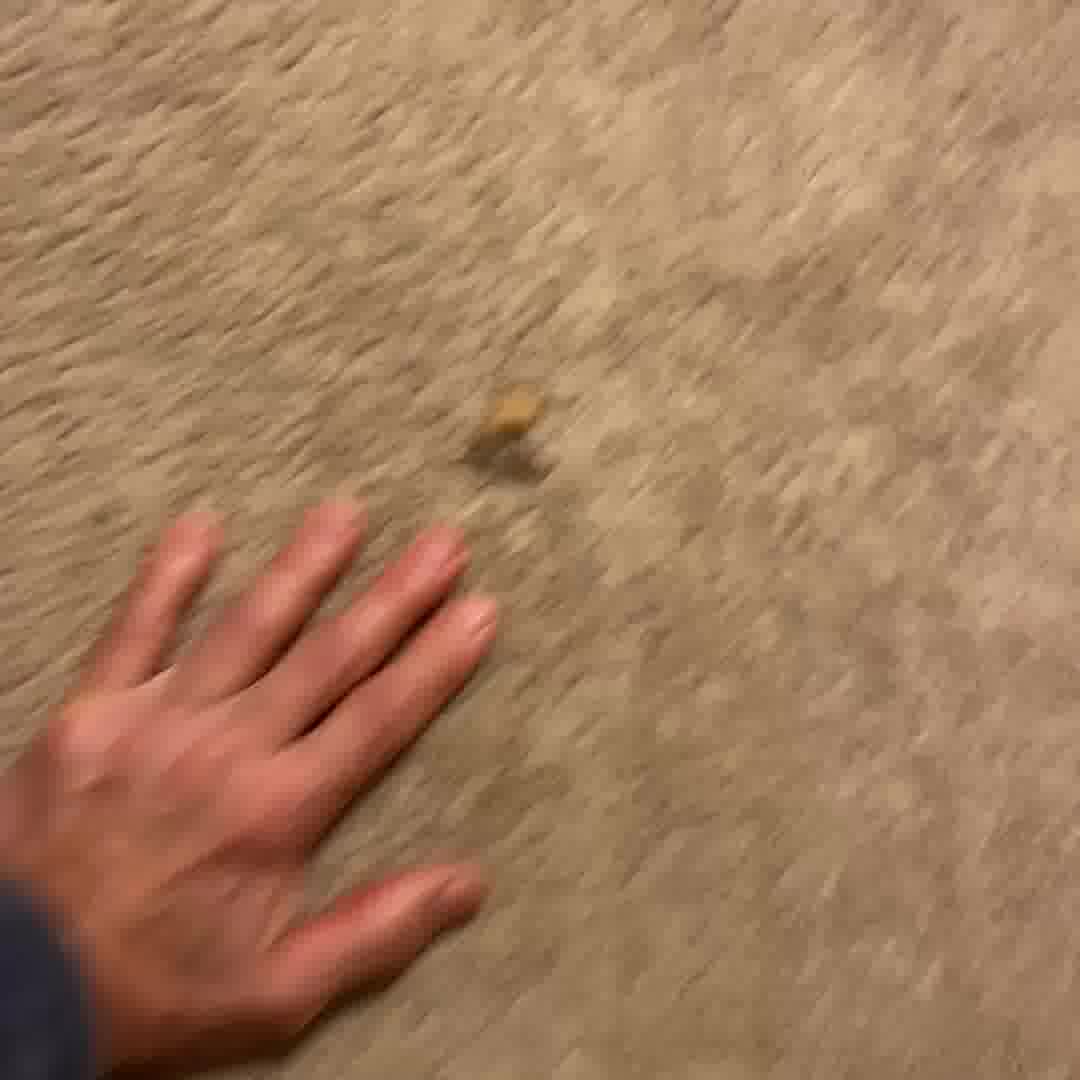}}
    \mbox{\includegraphics[width=0.095\textwidth]{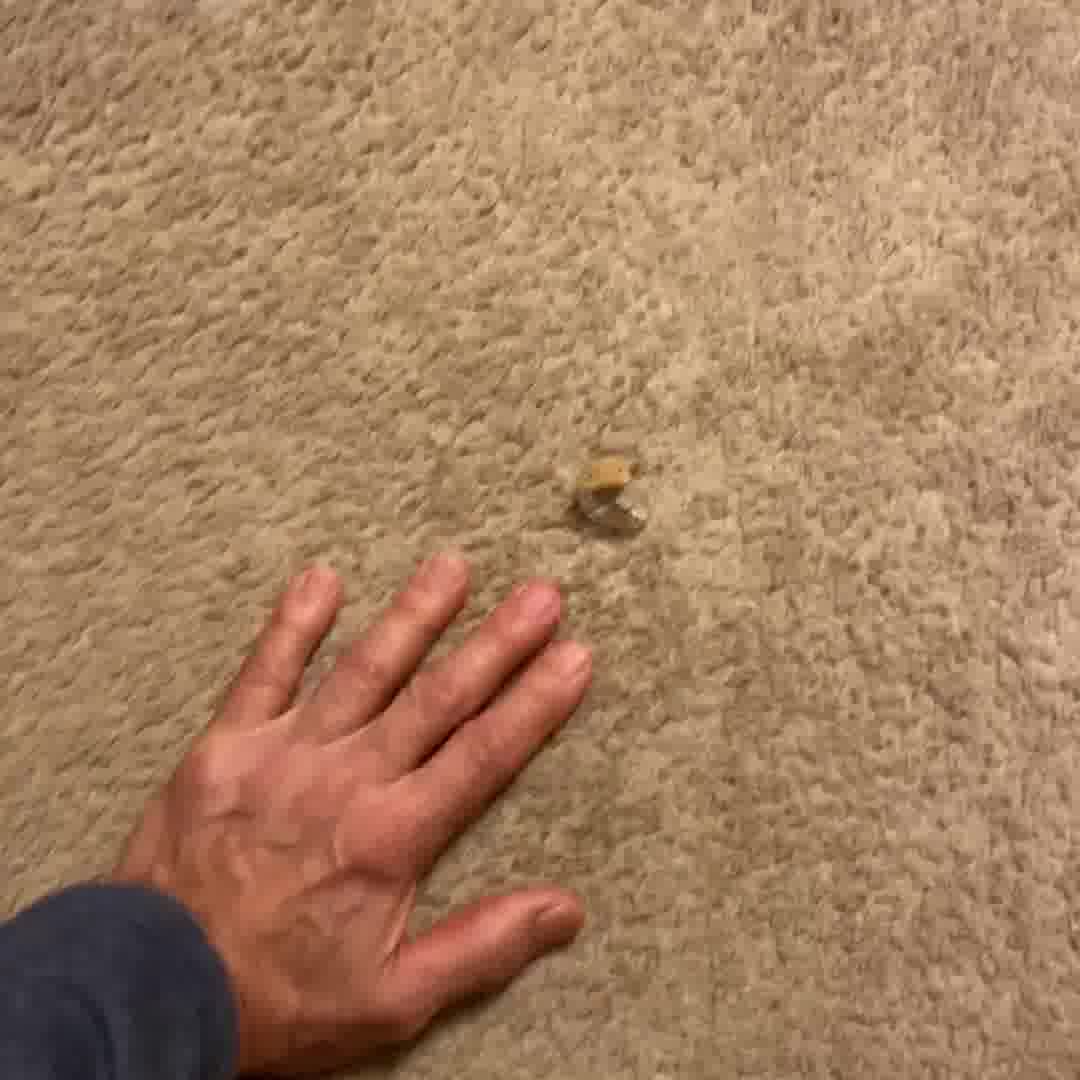}}
    \mbox{\includegraphics[width=0.095\textwidth]{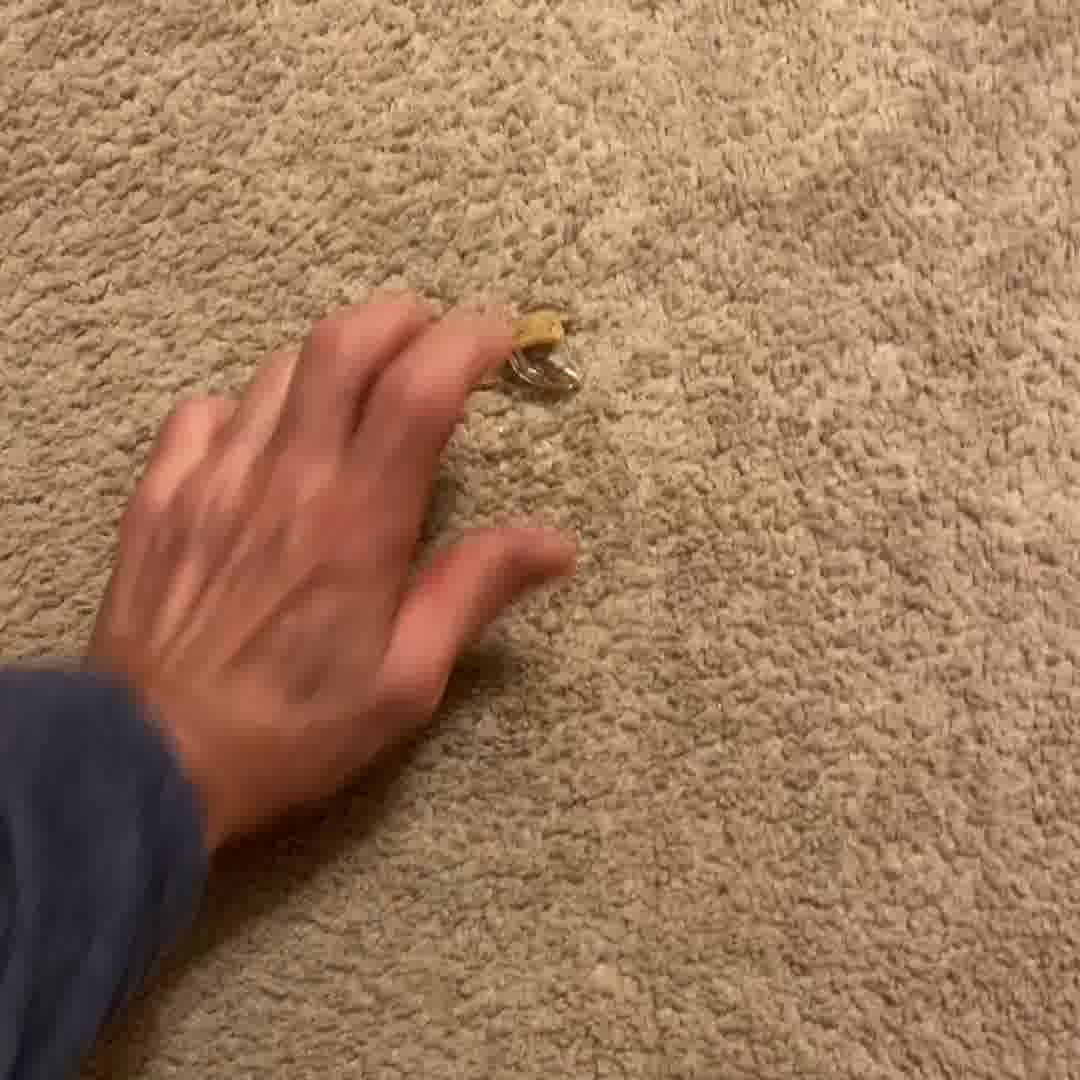}}
    \mbox{\includegraphics[width=0.095\textwidth]{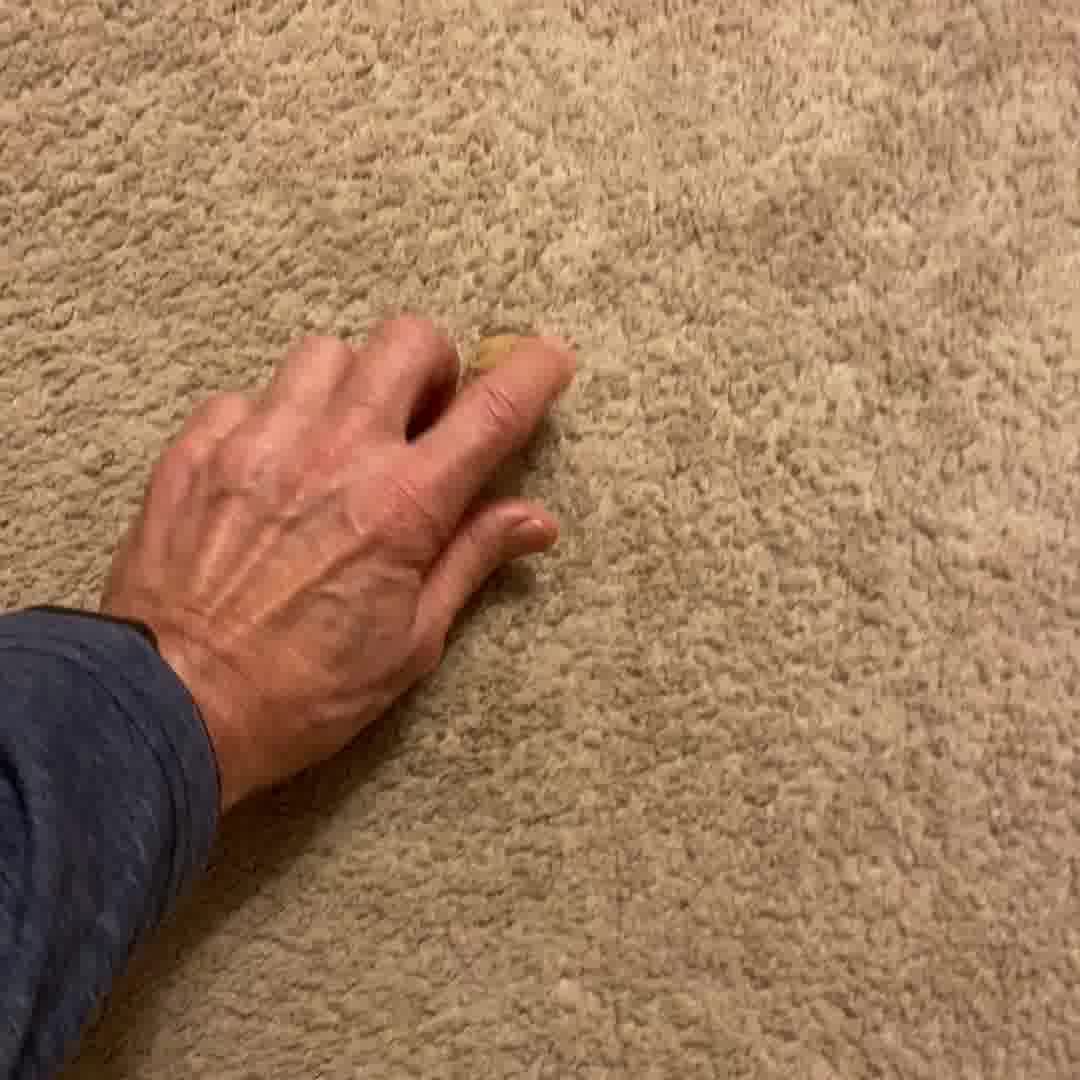}}
    \mbox{\includegraphics[width=0.095\textwidth]{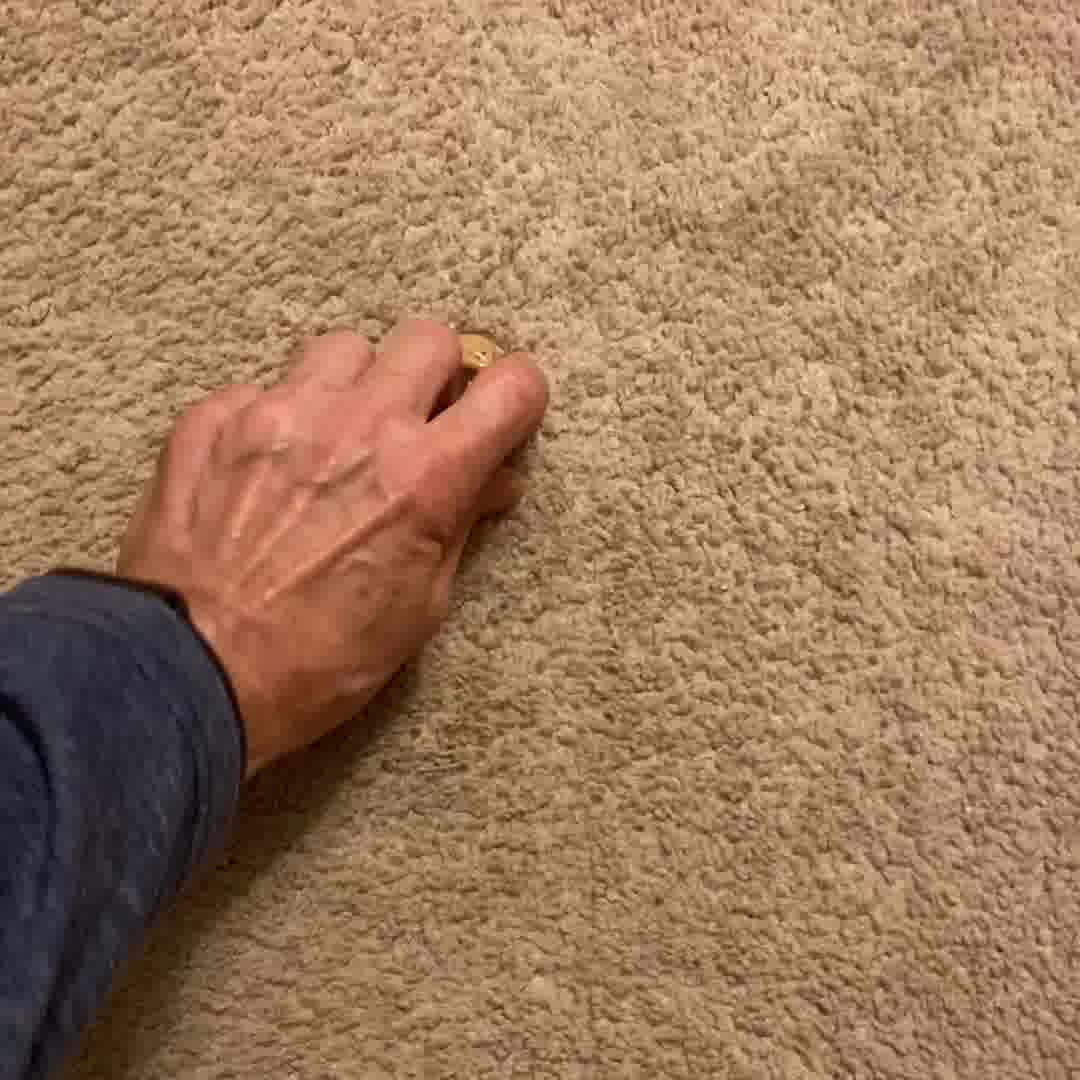}}}\\
    \vspace*{2px}
    \scalebox{0.95}{
    \mbox{\includegraphics[width=0.095\textwidth]{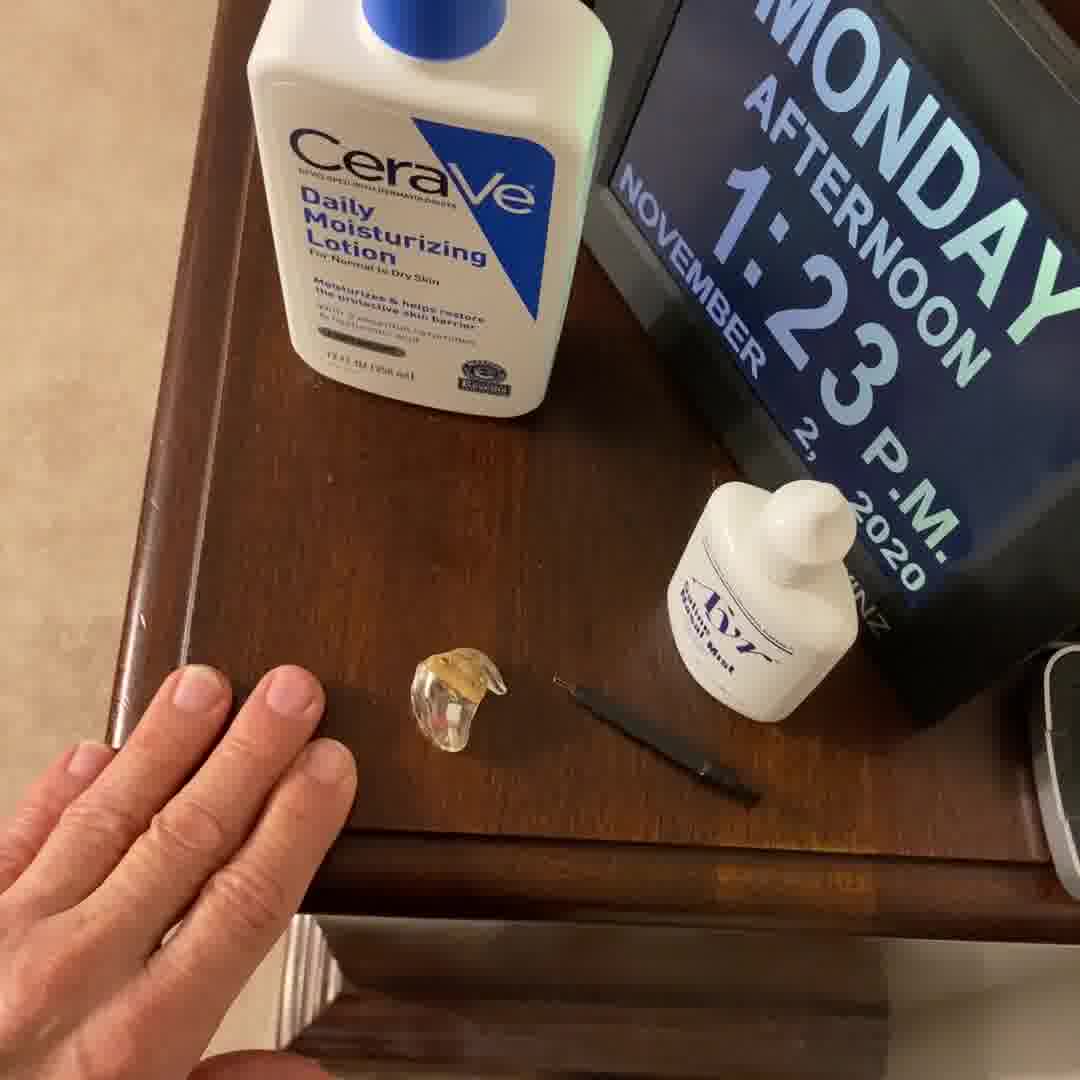}}
    \mbox{\includegraphics[width=0.095\textwidth]{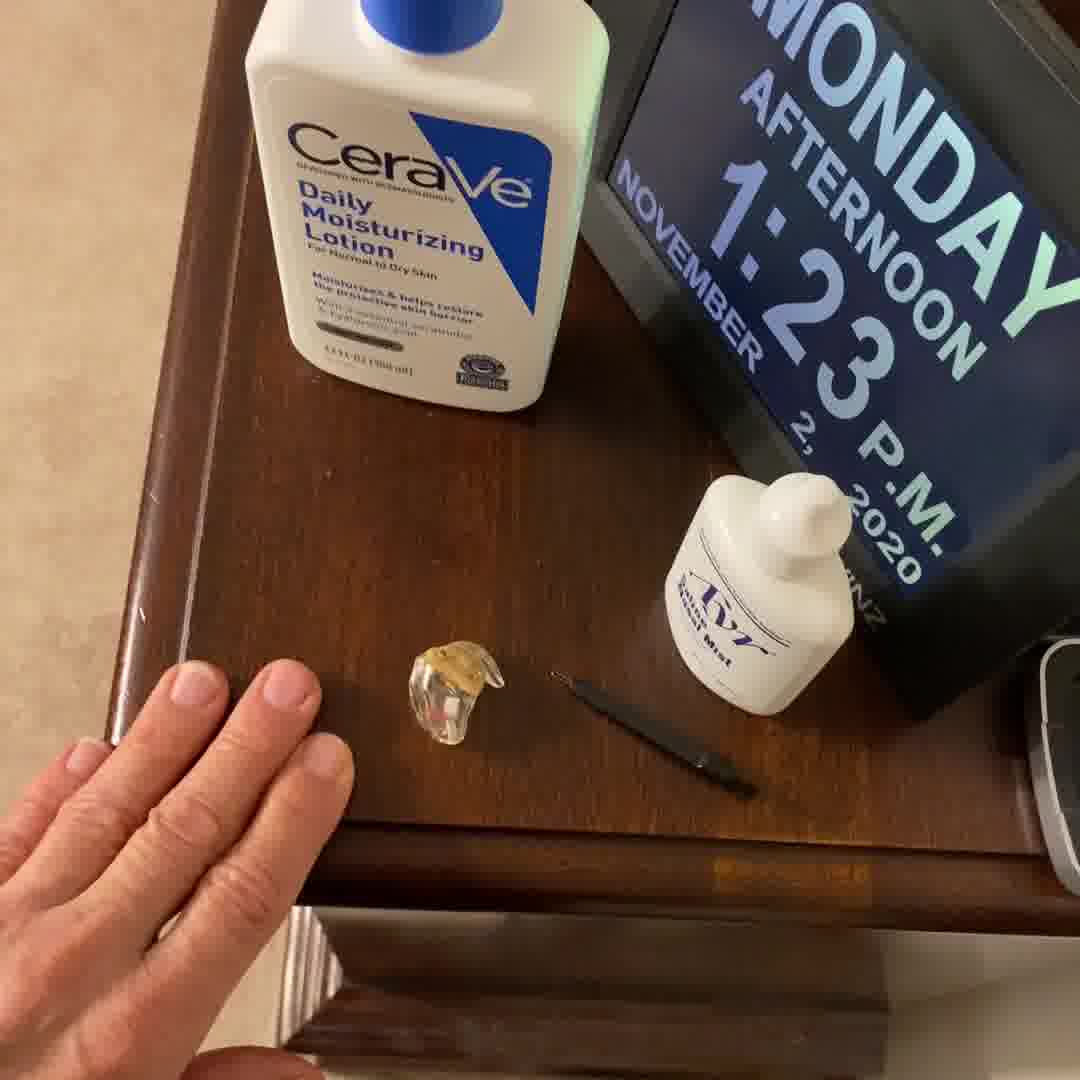}}
    \mbox{\includegraphics[width=0.095\textwidth]{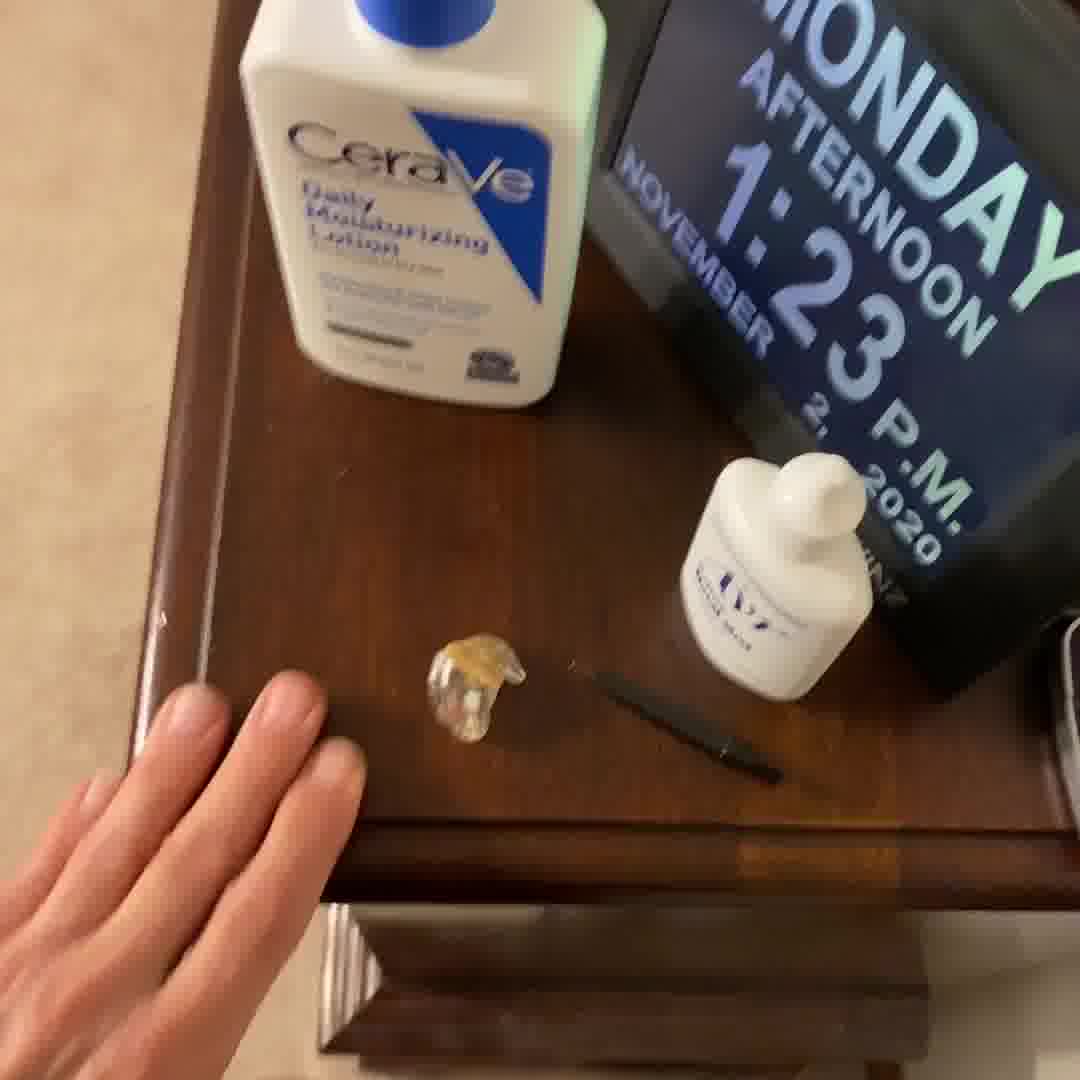}}
    \mbox{\includegraphics[width=0.095\textwidth]{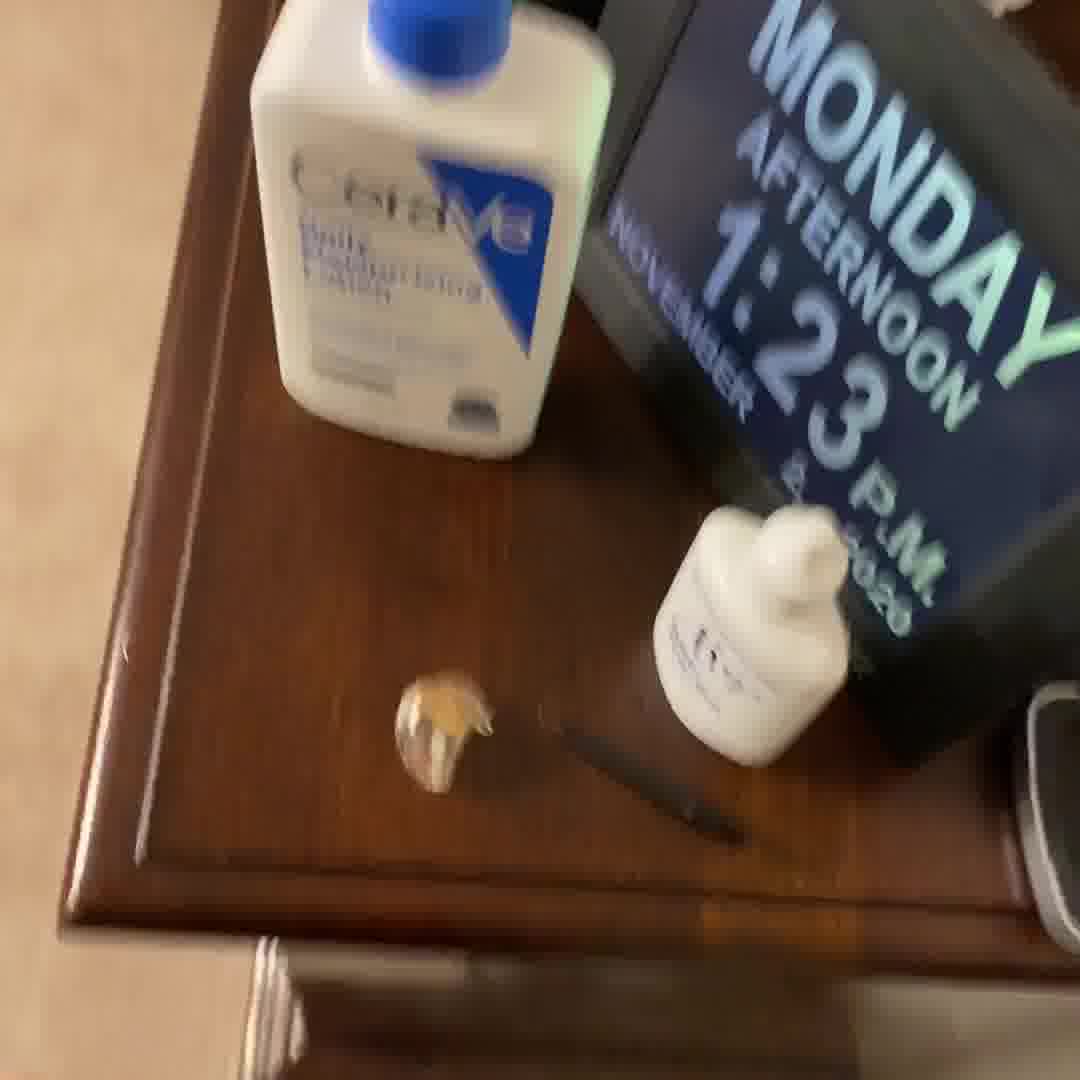}}
    \mbox{\includegraphics[width=0.095\textwidth]{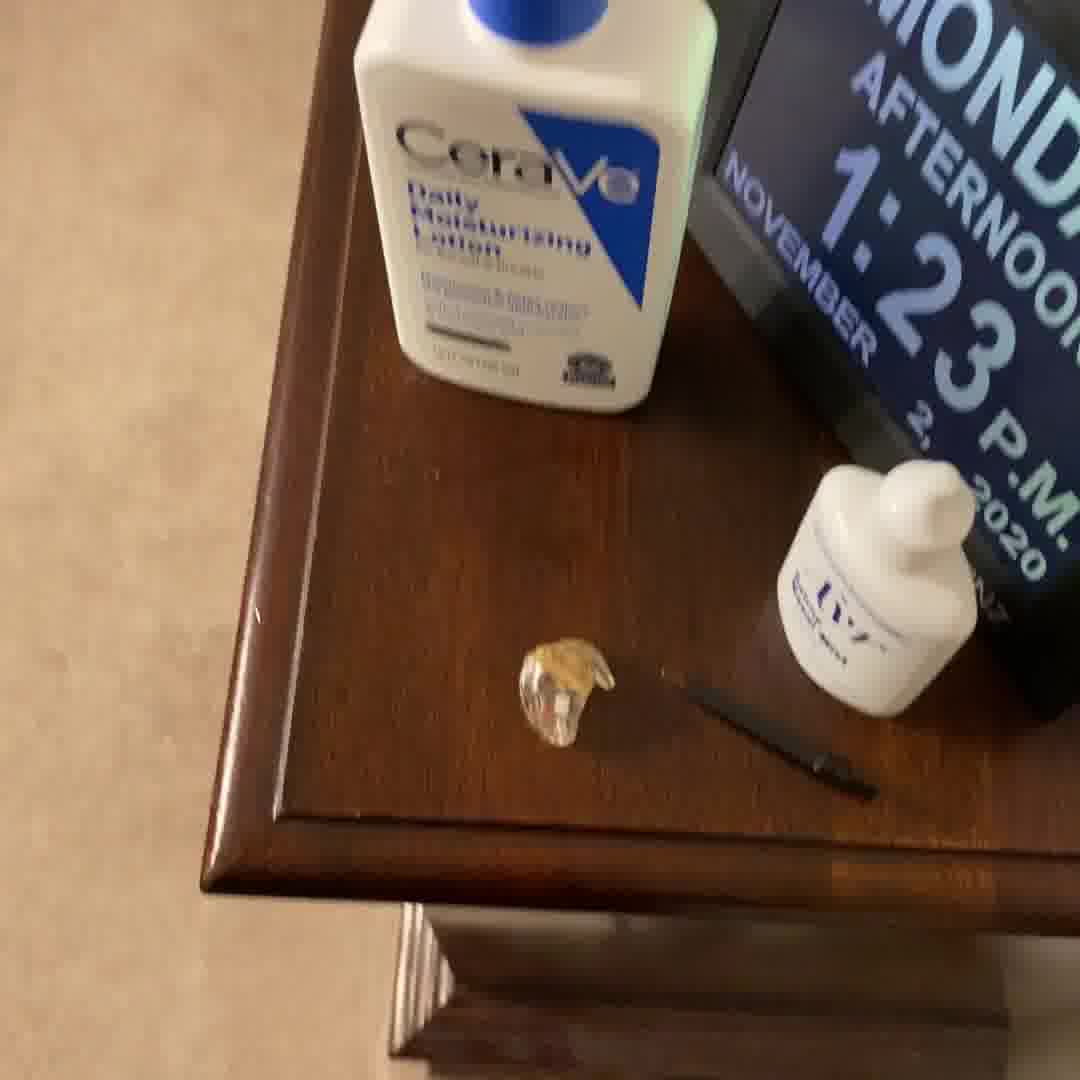}}
    \mbox{\includegraphics[width=0.095\textwidth]{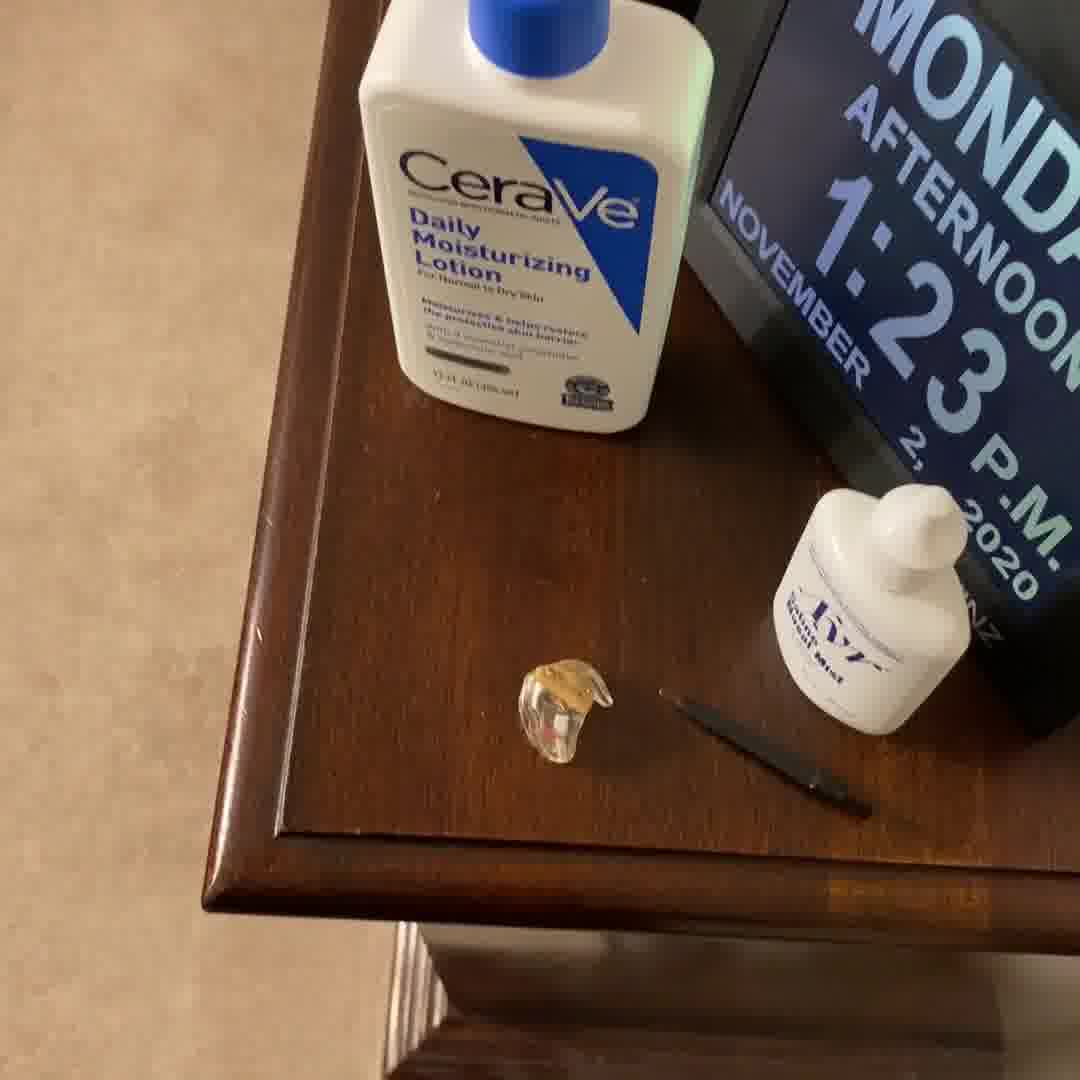}}
    \mbox{\includegraphics[width=0.095\textwidth]{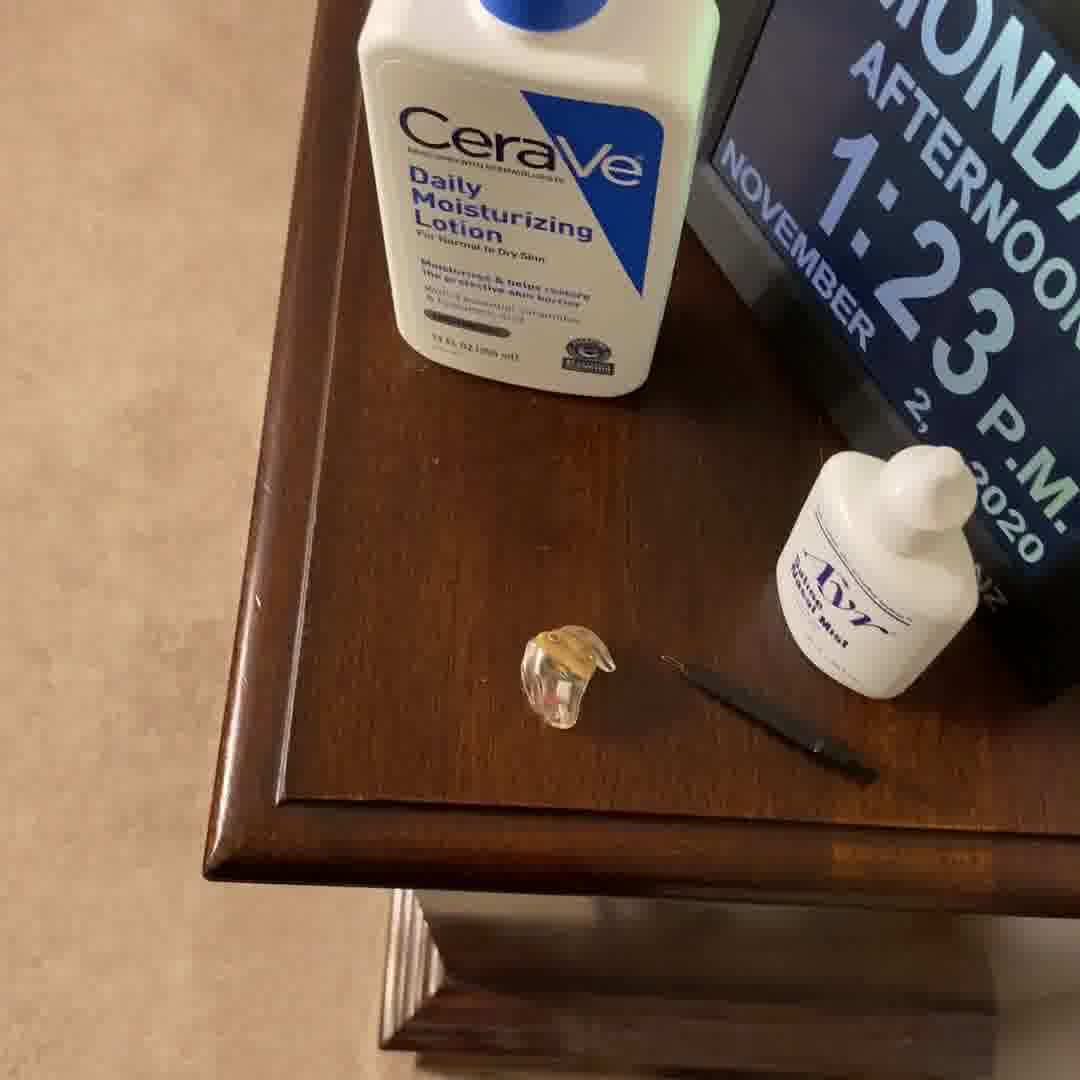}}
    \mbox{\includegraphics[width=0.095\textwidth]{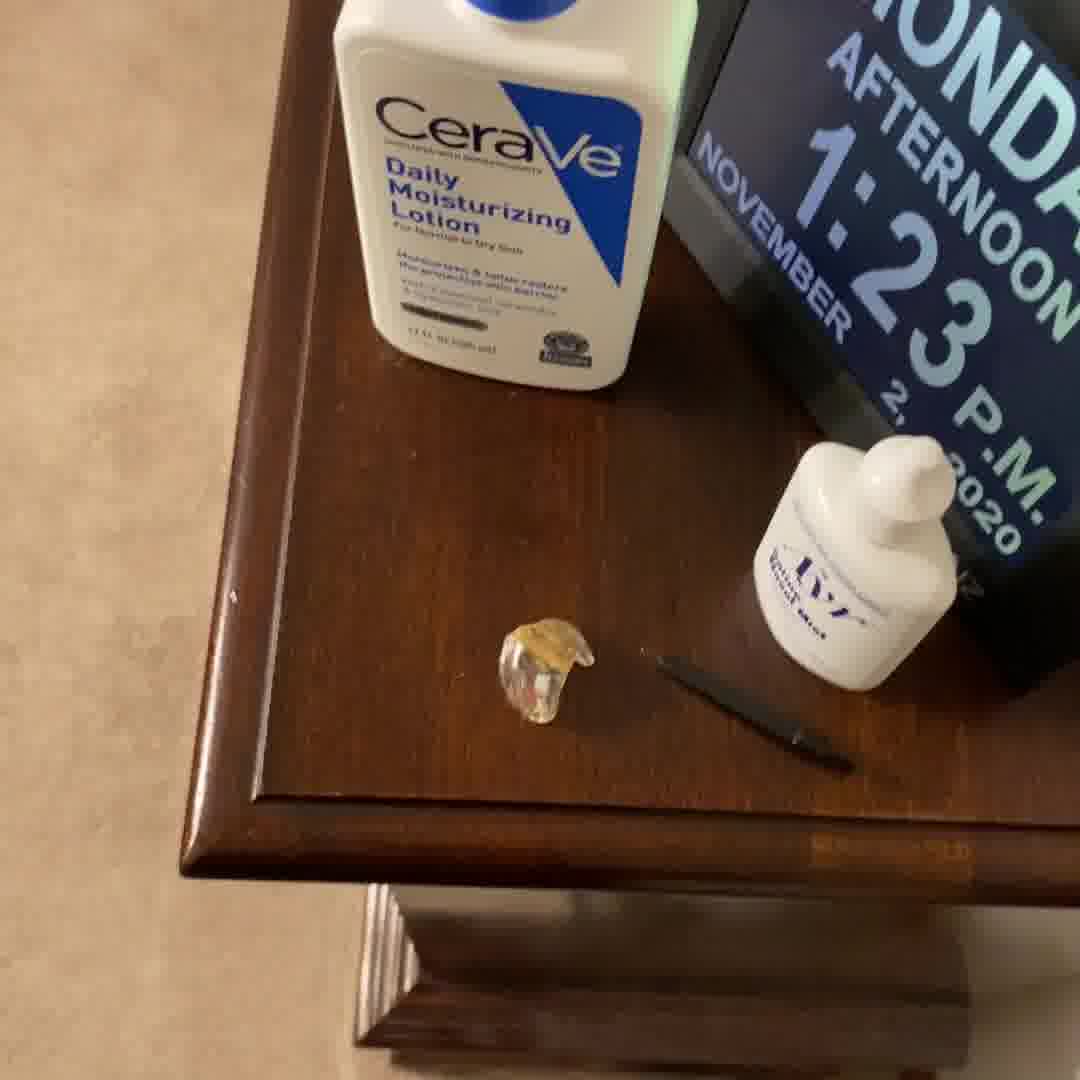}}
    \mbox{\includegraphics[width=0.095\textwidth]{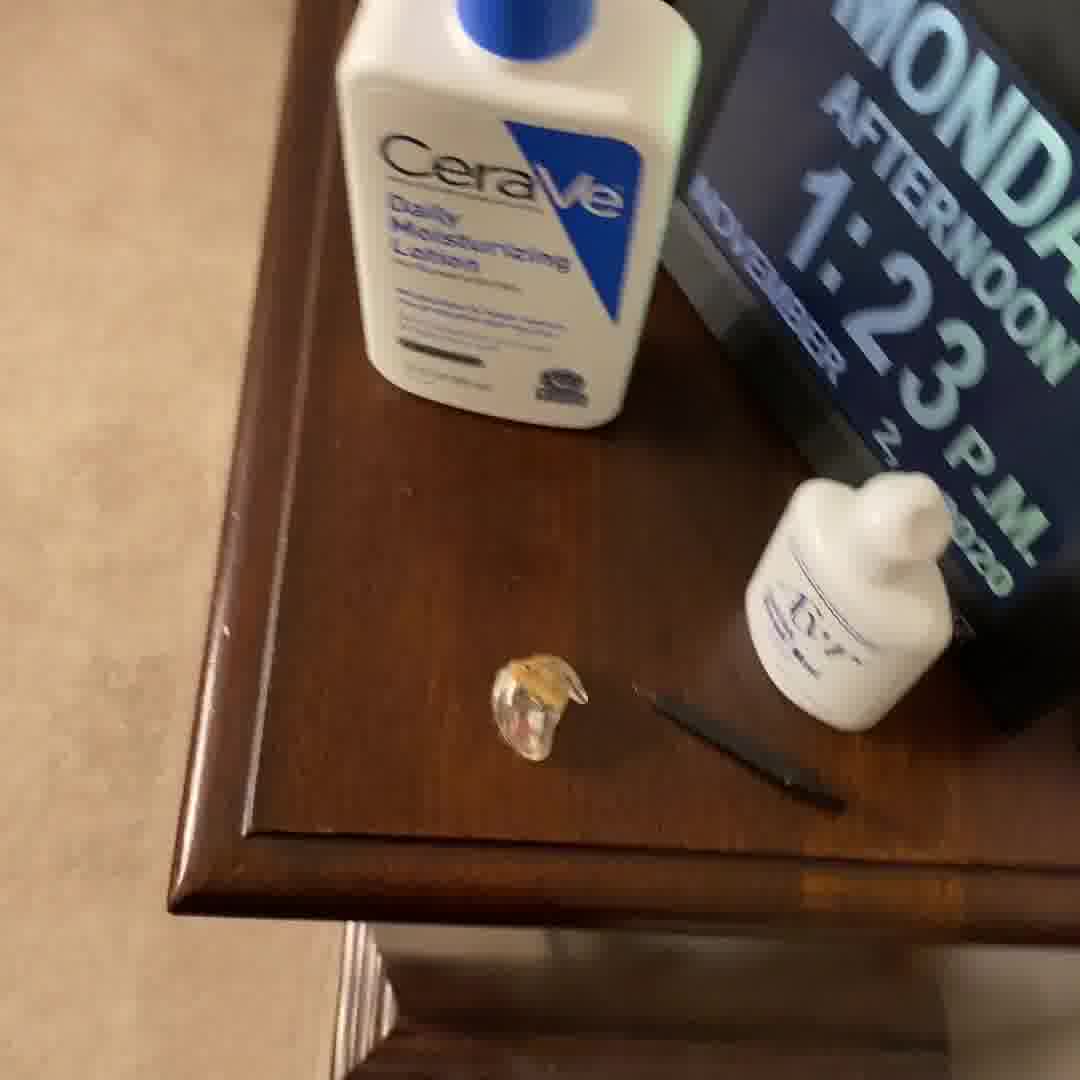}}
    \mbox{\includegraphics[width=0.095\textwidth]{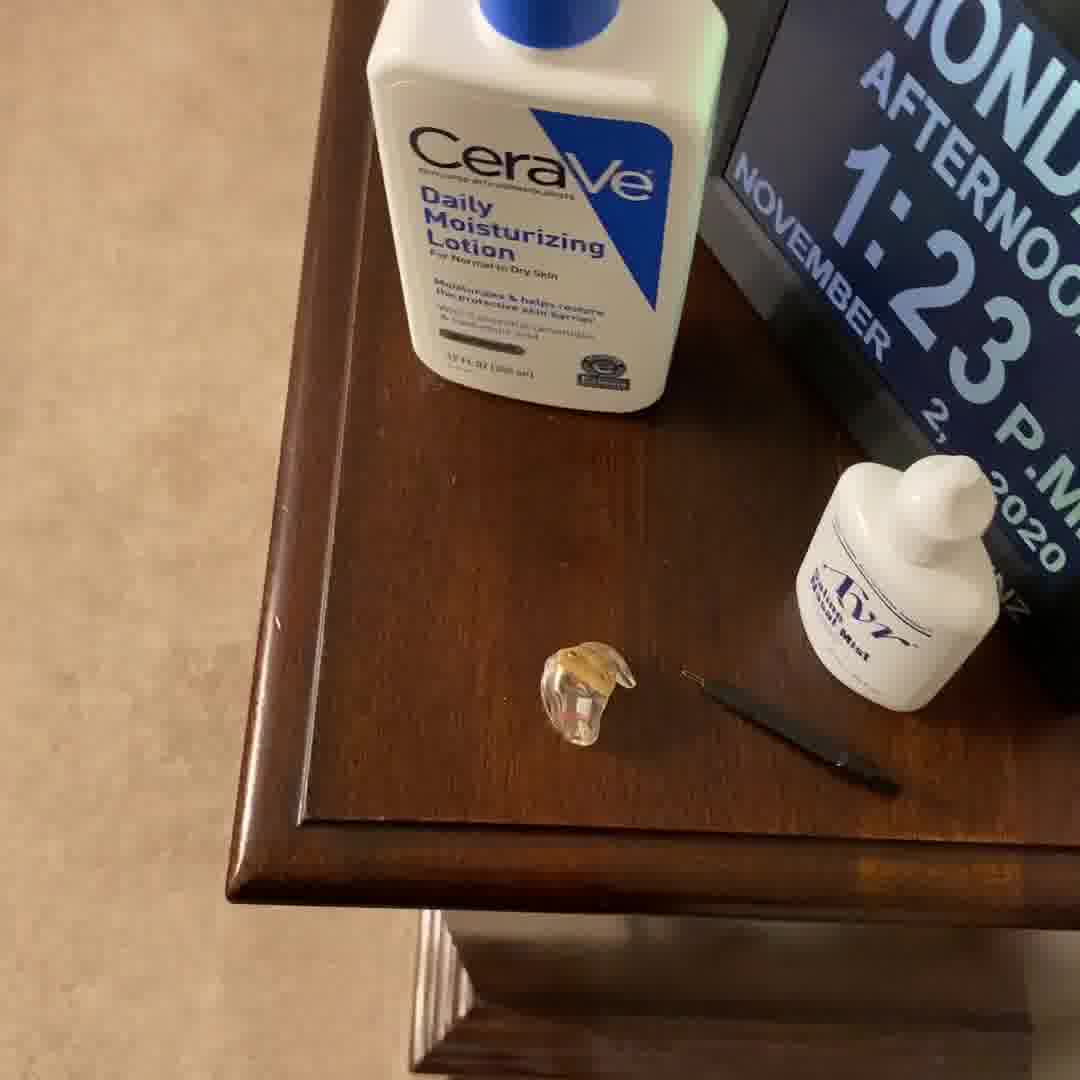}}}\\
    \vspace{-0.5em}
    \caption{Video clips from the ORBIT benchmark dataset. Each row comes from a different video (alternating clean and clutter). Rows 1,2: ``dog toy''. Rows 3,4: ``dog lead''. Rows 5,6: ``white cane''. Rows 7,8: ``victor stream book reader''. Rows 9,10: ``toddler cup''. Rows 11,12: ``hearing aid''.}
    \label{fig:more-qual-frame-examples}
\end{figure*}

\begin{figure*}[ht!]
    \centering
    \begin{subfigure}[t]{\textwidth}
        \centering
        \hspace{-1.2em}
        \includegraphics[width=\textwidth]{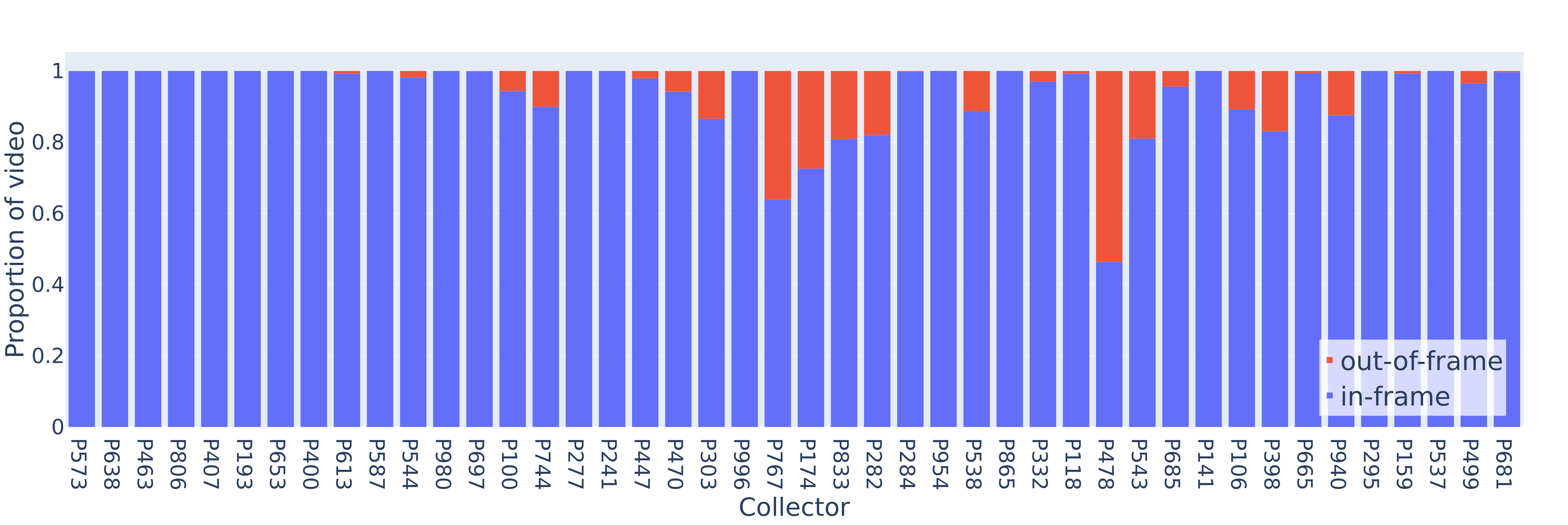}
        \vspace{-0.2em}
        \caption{Train users.}
        \vspace{-0.3em}
        \label{app:fig:bounding-box-summary-train}
    \end{subfigure}
    \begin{subfigure}[t]{0.3\textwidth}
        \centering
        \hspace{-1.1em}
        \includegraphics[width=0.65\textwidth]{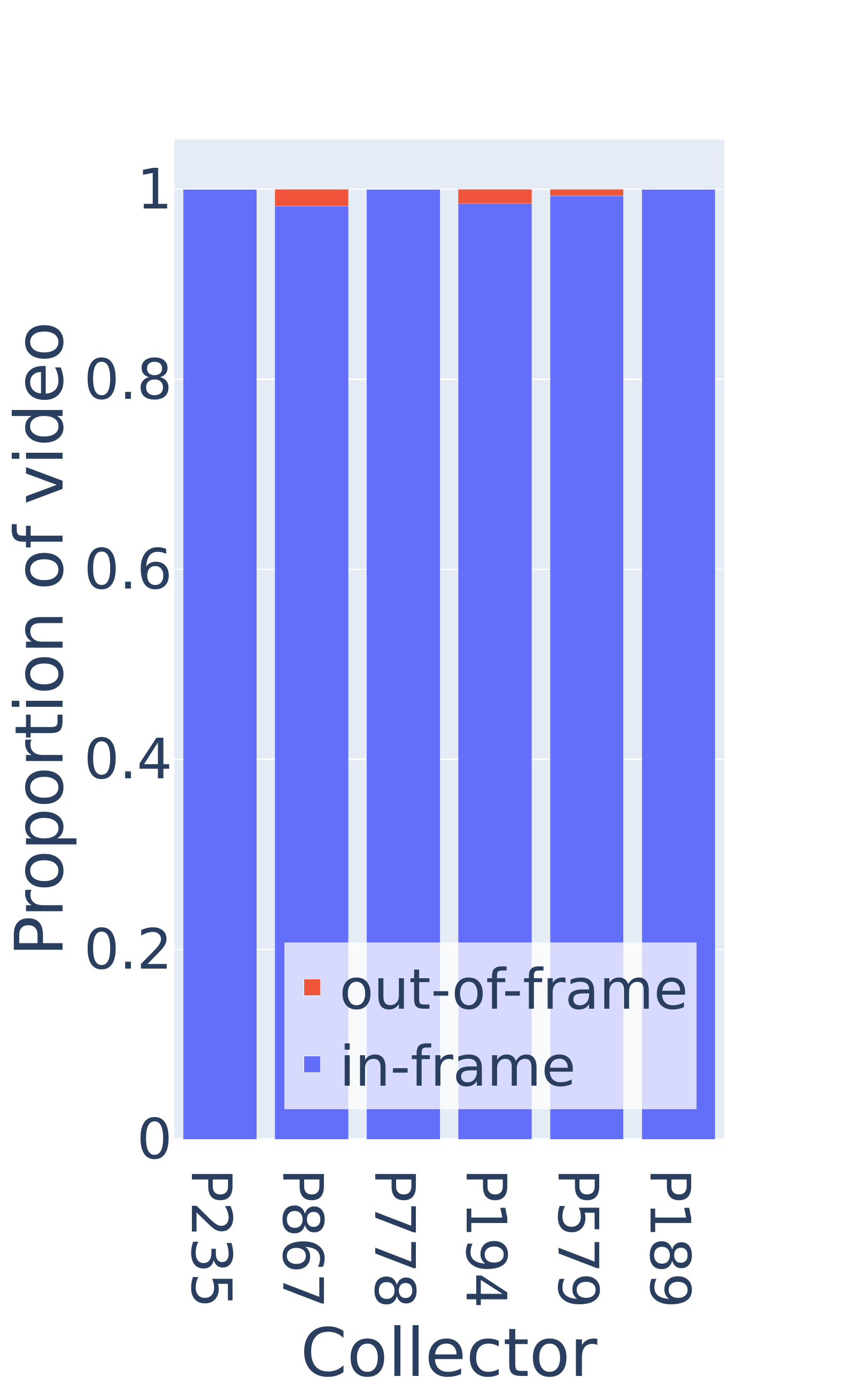}
        \vspace{-0.2em}
        \caption{Validation users.}
        \vspace{-0.4em}
        \label{app:fig:bounding-box-summary-validation}
    \end{subfigure}
    \begin{subfigure}[t]{0.69\textwidth}
        \centering
        \hspace{-1.2em}
        \includegraphics[width=0.63\textwidth]{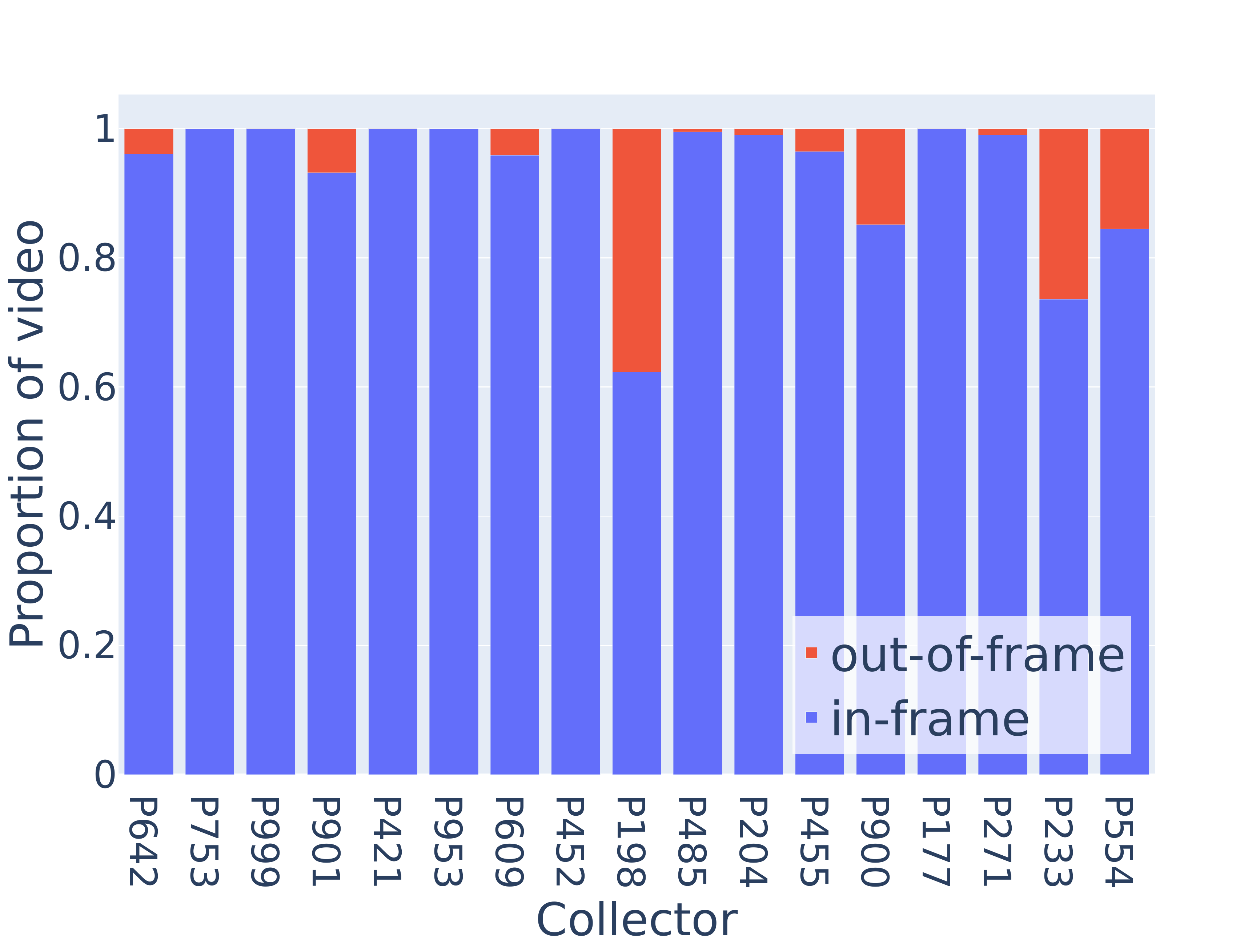}
        \caption{Test users.}
        \vspace{-0.4em}
        \label{app:fig:bounding-box-summary-test}
    \end{subfigure}
    \vspace{0em}
    \caption{The proportion of the video the target object spends in- versus out-of-frame, averaged over all clutter videos per collector and grouped by train, validation, and test users.}
    \label{app:fig:bounding-box-summary-allsets}
\end{figure*}

In \cref{sec:orbit-benchmark-dataset}, we summarized the benchmark dataset over all 67 users. 
Here, we report the same summaries but now broken down by train/validation/test users.
\vspace{-1em}
\paragraph{Number of videos and objects.}
Mirroring~\cref{tab:benchmark-dataset-summary}, we report the statistics for train/validation/test users in~\cref{tab:benchmark-dataset-summary-allsets}.
Extending~\cref{fig:num-vids-by-object-all-main,fig:num-objects-all-main}, we report the number of objects and videos per user in~\cref{fig:benchmark-num-objects-allsets,fig:benchmark-num-vids-by-object-allsets}, respectively.
\begin{table*}[ht!]
    \centering
    \begin{subtable}{\textwidth}
        \centering
        \scalebox{0.8}{
        \begin{tabular}{l|ccccccccc}
        & \textbf{Collectors} &  \textbf{Objects} & \textbf{Videos} & \multicolumn{3}{c}{\textbf{Videos per object}} & \multicolumn{3}{c}{\textbf{Frames per video}}\\
        & & &  & mean/std & 25/75\(^\text{th}\) perc. & min/max & mean/std & 25/75\(^\text{th}\) perc. & min/max \\
        \cmidrule{1-10}
        \textbf{Total} & 44 & 278 & 2277 & 8.2/6.0 & 7.0/7.0 & 3.0/46.0 & 716.8/448.6 & 378.0/898.0 & 34.0/3600.0 \\ 
        \cmidrule{1-10}
        \textbf{Clean} & & &  1814 & 6.5/6.0 & 5.0/6.0 & 2.0/44.0 & 774.0/471.6 & 394.5/899.0 & 34.0/3600.0\\ 
        \textbf{Clutter} & & & 463 & 1.7/0.8 & 1.0/2.0 & 1.0/7.0 & 492.7/235.5 & 324.5/599.0 & 47.0/2412.0\\ 
        \cmidrule{1-10}
        \textbf{Per-collector} & 1 & 6.3/2.6 & 51.8/55.2 & 7.6/4.8 & 6.5/7.4 & 3.4/38.4 & 735.5/232.0 & 629.7/805.7 & 213.1/1614.3 \\
        \cmidrule{1-10}
        \textbf{Clean} & & & 41.2/52.7 & 5.9/4.8 & 4.6/6.0 & 2.4/36.5 & 823.0/279.0 & 710.2/898.4 & 219.3/1872.6\\ 
        \textbf{Clutter} & & & 10.5/5.2 & 1.7/0.5 & 1.2/2.0 & 1.0/3.0 & 467.3/163.2 & 362.4/597.0 & 177.7/853.0\\
        \end{tabular}}
        \caption{Train users.}
        \label{tab:benchmark-dataset-summary-train}
    \end{subtable}\\
    \vspace{0.5em}
    \begin{subtable}{\textwidth}
        \centering
        \scalebox{0.8}{
        \begin{tabular}{l|ccccccccc}
        & \textbf{Collectors} &  \textbf{Objects} & \textbf{Videos} & \multicolumn{3}{c}{\textbf{Videos per object}} & \multicolumn{3}{c}{\textbf{Frames per video}}\\
        & & &  & mean/std & 25/75\(^\text{th}\) perc. & min/max & mean/std & 25/75\(^\text{th}\) perc. & min/max \\
        \cmidrule{1-10}
        \textbf{Total} & 6 & 50 & 347 & 6.9/1.0 & 7.0/7.0 & 4.0/10.0 & 638.4/227.6 & 561.0/775.5 & 33.0/2053.0\\
        \cmidrule{1-10}
        \textbf{Clean} & & & 284 & 5.7/1.0 & 5.0/6.0 & 3.0/9.0 & 689.6/210.3 & 626.0/835.0 & 33.0/2053.0\\
        \textbf{Clutter} & & & 63 & 1.3/0.6 & 1.0/1.0 & 1.0/4.0 & 407.7/144.1 & 289.0/593.5 & 176.0/618.0\\
        \cmidrule{1-10}
        \textbf{Per-collector} & 1 & 8.3/2.4 & 57.8/17.7 & 6.9/0.3 & 6.7/7.0 & 6.6/7.6 & 666.7/95.8 & 621.0/741.7 & 529.8/808.4\\ 
        \cmidrule{1-10}
        \textbf{Clean} & & & 47.3/17.5 & 5.5/0.7 & 5.0/5.7 & 4.6/6.6 & 730.5/114.1 & 673.9/819.0 & 565.0/899.9\\ 
        \textbf{Clutter} & & & 10.5/0.8 & 1.4/0.4 & 1.0/1.8 & 1.0/2.0 & 409.8/133.2 & 313.3/528.4 & 250.5/597.9\\
        \end{tabular}}
        \caption{Validation users.}
        \label{tab:benchmark-dataset-summary-val}
    \end{subtable}\\
    \vspace{0.5em}
    \begin{subtable}{\textwidth}
        \centering
        \scalebox{0.8}{
        \begin{tabular}{l|ccccccccc}
        & \textbf{Collectors} &  \textbf{Objects} & \textbf{Videos} & \multicolumn{3}{c}{\textbf{Videos per object}} & \multicolumn{3}{c}{\textbf{Frames per video}}\\
        & & &  & mean/std & 25/75\(^\text{th}\) perc. & min/max & mean/std & 25/75\(^\text{th}\) perc. & min/max \\
        \cmidrule{1-10}
        \textbf{Total} & 17 & 158 & 1198 & 7.6/2.4 & 7.0/7.0 & 3.0/19.0 & 696.4/384.8 & 428.0/900.0 & 40.0/3596.0 \\
        \cmidrule{1-10}
        \textbf{Clean} & & & 898 & 5.7/1.0 & 5.0/6.0 & 2.0/10.0 & 791.6/352.7 & 621.0/926.0 & 51.0/3596.0 \\
        \textbf{Clutter} & & & 300 & 1.9/2.3 & 1.0/2.0 & 1.0/13.0 & 411.5/332.5 & 175.0/598.0 & 40.0/3596.0 \\
        \cmidrule{1-10}
        \textbf{Per-collector} & 1 & 9.3/2.1 & 70.5/24.2 & 7.6/2.1 & 6.8/7.4 & 6.2/15.6 & 733.2/166.6 & 599.4/811.2 & 335.7/977.0 \\
        \cmidrule{1-10}
        \textbf{Clean} & & & 52.8/14.5 & 5.6/0.6 & 5.0/6.0 & 4.4/6.7 & 804.1/164.7 & 658.8/899.0 & 518.1/1095.6\\ 
        \textbf{Clutter} & & & 17.6/18.7 & 1.9/2.0 & 1.0/1.9 & 1.0/9.9 & 478.3/285.4 & 282.3/598.2 & 158.6/1258.8 \\ 
        \end{tabular}}
        \caption{Test users.}
        \label{tab:benchmark-dataset-summary-test}
    \end{subtable}\\
    \caption{The ORBIT benchmark dataset, grouped by train, validation, and test users.}
    \label{tab:benchmark-dataset-summary-allsets}
    \vspace{-1em}
\end{table*}
\vspace{-1em}
\paragraph{Types of objects.}
Following~\cref{app:sec:object-clustering}, we include the object cluster histogram for all benchmark users in~\cref{fig:benchmark-cluster-histo-all}, and for  train (\ref{fig:benchmark-cluster-histo-train}), validation (\ref{fig:benchmark-cluster-histo-val}) and test (\ref{fig:benchmark-cluster-histo-test}) users.
Extending~\cref{fig:qual-frame-examples}, we include more examples of clean and clutter frames from the ORBIT benchmark dataset in~\cref{fig:more-qual-frame-examples}.
\vspace{-1em}
\paragraph{Bounding box annotations.}
We provide bounding box annotations around the ground-truth object (as per the video label) in all clutter videos.
We compute the proportion of each clutter video for which the target object is in-frame (identified by if the frame contained a bounding box), and report these for train/validation/test users in~\cref{app:fig:bounding-box-summary-allsets}.
We use the summarization of test users (\cref{app:fig:bounding-box-summary-test}) as an upper bound on model performance in~\cref{tab:orbit-baselines}.

\section{Object clustering}
\label{app:sec:object-clustering}

\begin{figure*}[!ht]
    \centering
    \vspace{4em}
    \hspace{1em}
    \begin{subfigure}[t]{0.48\textwidth}
        \includegraphics[width=\textwidth]{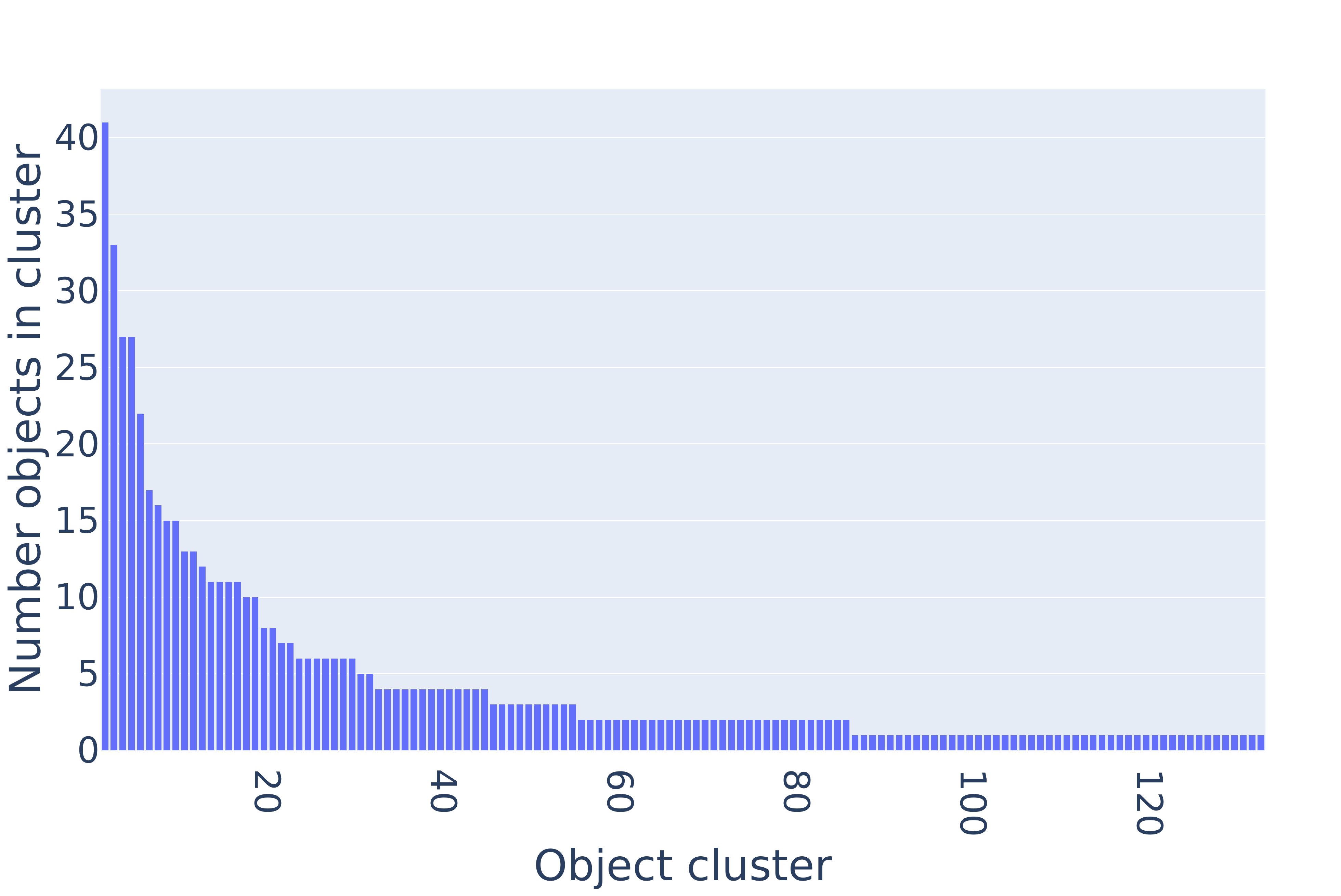}
        \caption{All users (unfiltered dataset).}
        \label{fig:full-cluster-histo-all}
    \end{subfigure}\\
    \begin{subfigure}[t]{0.48\textwidth}
        \includegraphics[width=\textwidth]{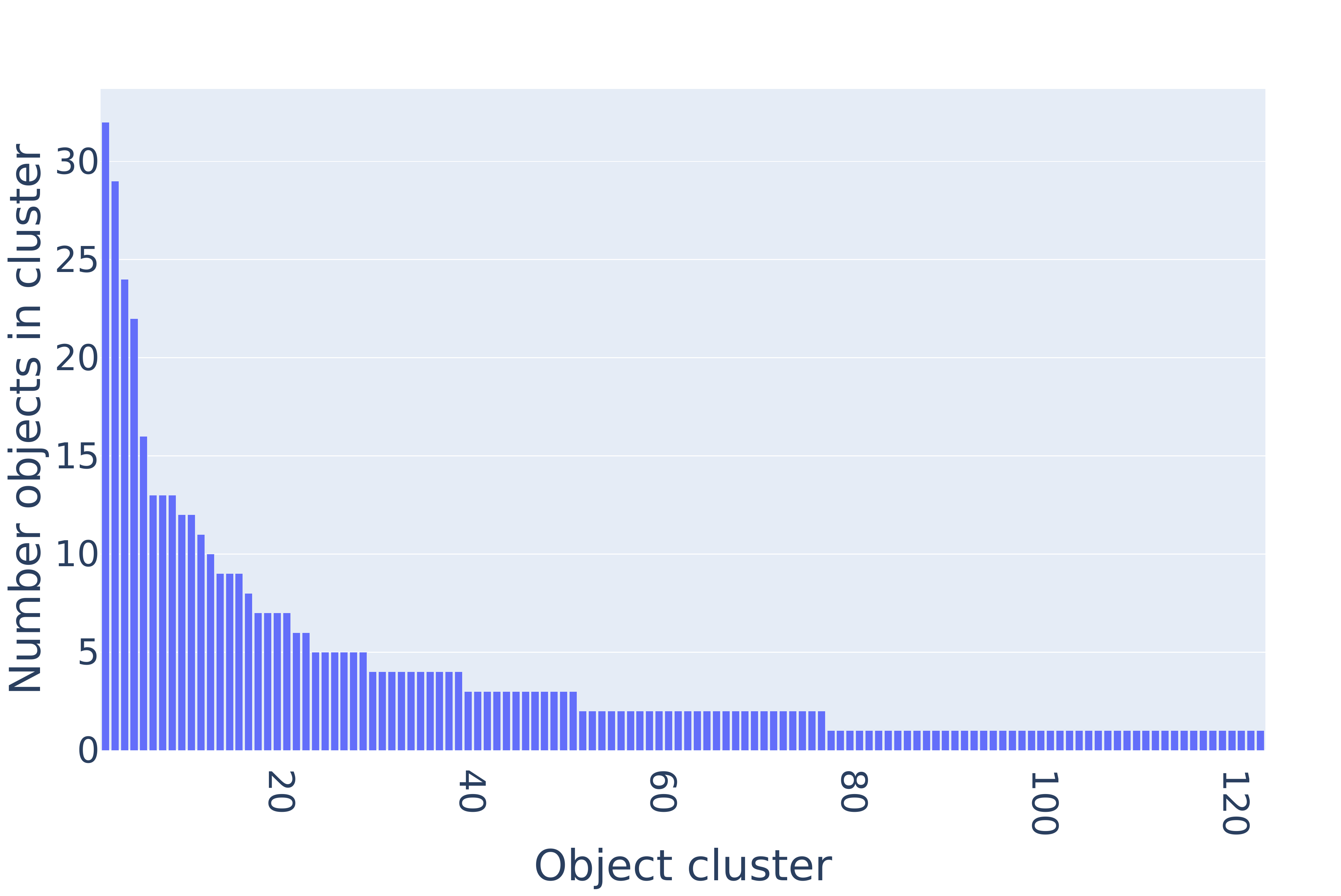}
        \caption{All users (benchmark dataset).}
        \label{fig:benchmark-cluster-histo-all}
    \end{subfigure}
    \begin{subfigure}[t]{0.48\textwidth}
        \centering
        \includegraphics[width=\textwidth]{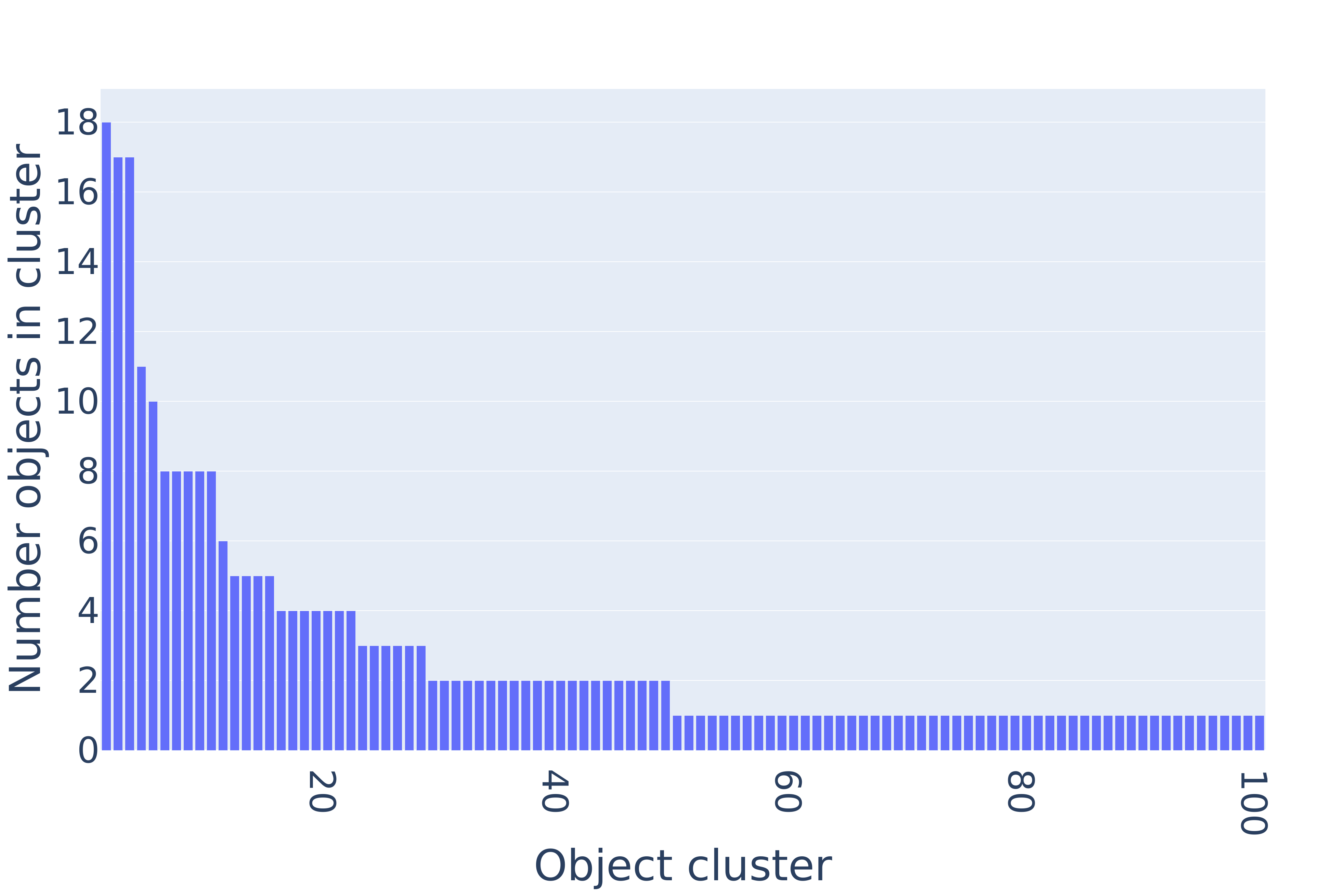}
        \caption{Train users (benchmark dataset).}
        \label{fig:benchmark-cluster-histo-train}
    \end{subfigure}\\
    \begin{subfigure}[t]{0.48\textwidth}
        \centering
        \includegraphics[width=\textwidth]{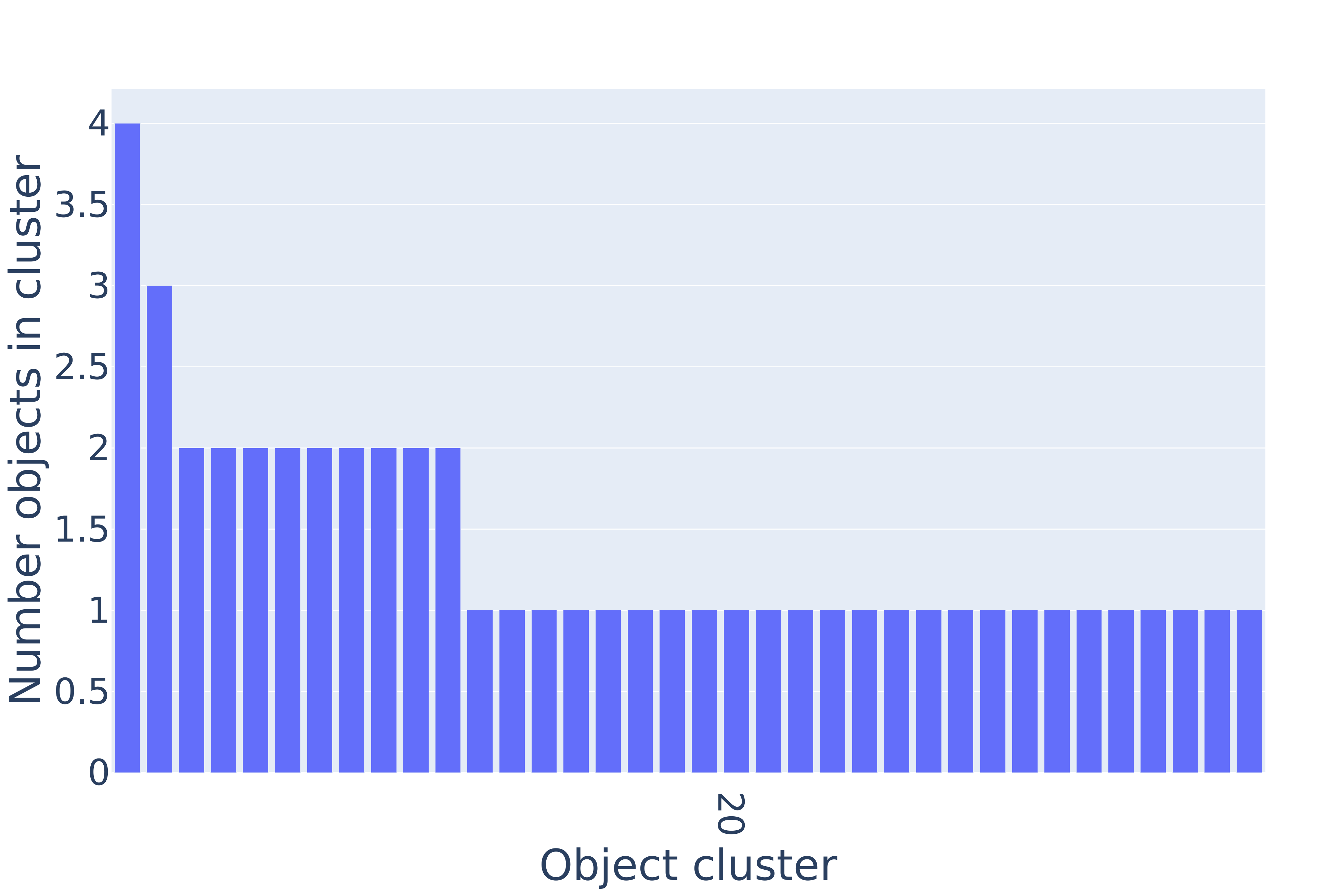}
        \caption{Validation users (benchmark dataset).}
        \label{fig:benchmark-cluster-histo-val}
    \end{subfigure}
    \begin{subfigure}[t]{0.48\textwidth}
        \centering
        \includegraphics[width=\textwidth]{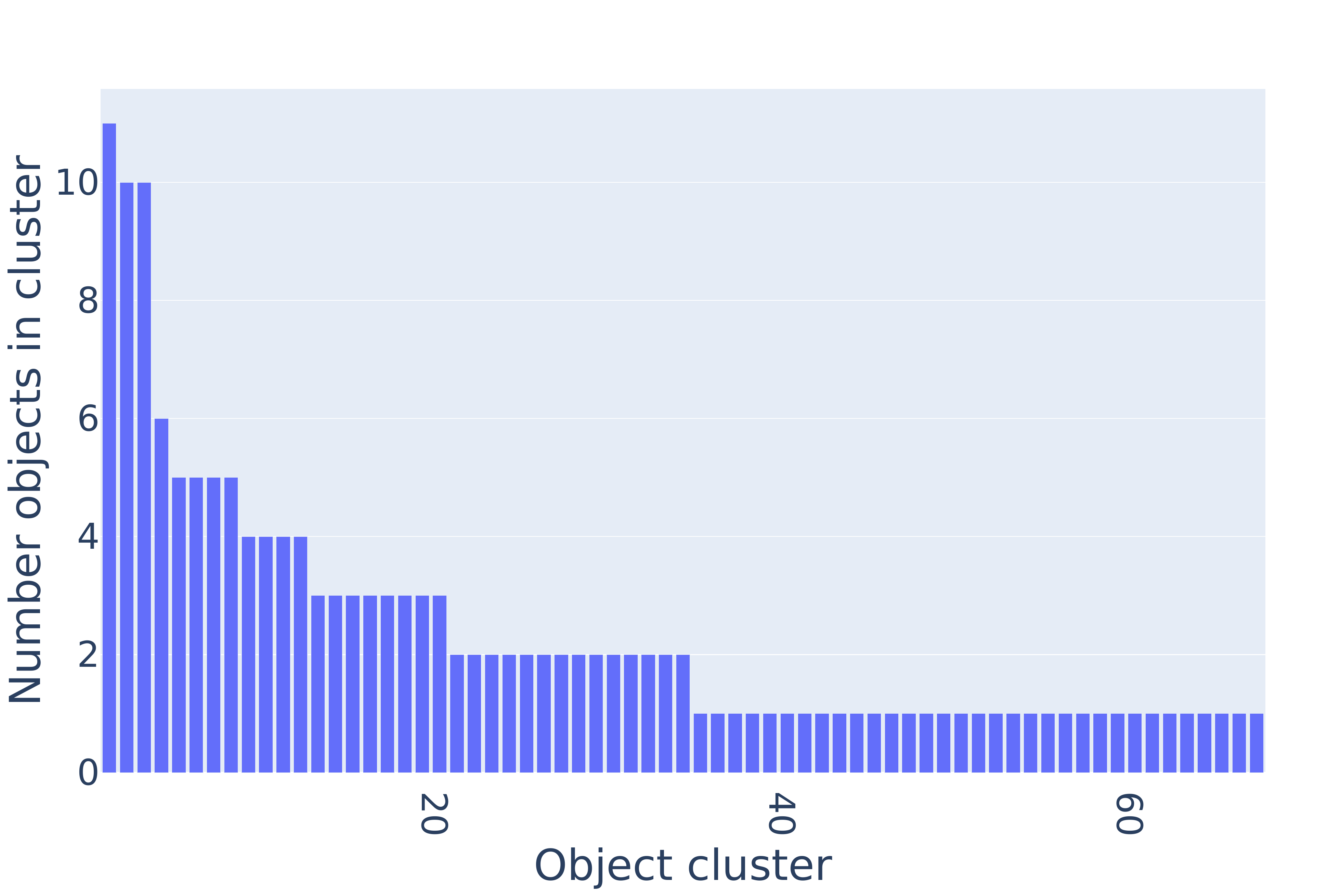}
        \caption{Test users (benchmark dataset).}
        \label{fig:benchmark-cluster-histo-test}
    \end{subfigure}
    \caption{Histogram of object clusters (grouped by object similarity) in the ORBIT dataset (benchmark and unfiltered). Objects in each cluster are listed in~\cref{app:sec:object-clustering}.}
    \label{fig:benchmark-cluster-histo-allsets}
\end{figure*}

The objects submitted by collectors varied widely in high-level object categories.
For summarization purposes, we grouped objects into clusters based on object similarity. 
\vspace{-1em}
\paragraph{Clustering algorithm}
Each object label (as entered by the user in the app) was converted to lowercase, tokenized and parts-of-speech tagged (using \textsc{nltk}).
A weighted average of all nouns (\textsc{nn, nns}) in the label was computed (each represented as a FastText~\cite{bojanowski2017enriching} embedding)\footnote{If the label had no nouns or only 1 word, the whole label was kept.} where each weight was the (normalized) frequency of the noun across all nouns in the dataset.
Intuitively, this encouraged more common nouns to dominate the object label's embedding (and down-weighted typos and collectors' idiosyncrasies).
We then clustered these embeddings using an agglomerative clustering algorithm based on pair-wise cosine distances (With a distance threshold of 0.4).
This produced cluster proposals which we then manually tweaked to obtain the final clusterings shown in~\cref{fig:benchmark-cluster-histo-allsets}.

\vspace{-1em}
\paragraph{Clusters for~\cref{fig:full-cluster-histo-all}}
[97 users, 588 objects, 132 clusters] 1: keys, house keys, radar key, my keys, front door keys, keychain, door keys, key, set of keys, 2: remote, tv remote, apple tv remote, amazon remote control, television remote control, remote control, virgin remote control, tv remote control, samsung tv remote control, presentation remote, sky q remote, clickr, fire stick remote, control, television control, remote tv, remote comtrol, apple tv ii remote control, t. v. remote control, 3: black small wallet, my purse, my wallet, ladies purse, money pouch, coin purse, wallet for bus pass cards and money, i d wallet, ipod in wallet, walletv, wallet, purse, 4: symbol cane, black mobility cane, my cane, white mobility came, folded white cane, long cane, cane, folded cane, folded long guide cane, p939411 white cane, white came, visibility stick, mobility, mobility cane, white cane, 5: earphones, apple headphones, apple earpods, airpods, my airpod pros case, iphone air pods, airpods pro charging case, airpod case, airpods, my airpods, airpods case, earpods, apple airpods case, 6: mug, personal mug, cup again, coffee mug, toddler cup, styrofoam cup, my mug, my cup, 7: front door, back door, house door, front door to house, door, my front door, shed door, front door from outside, 8: blue headphones, my headphones, headphone, bose wireless headphones, headset, my sennheiser pxc 350-2, headphones, 9: phone, mobile phone, iphone 6, apple mobile phone, i phone 11 pro, iphone in case, iphone, cell phone, work phone, phone case, 10: watch, wrist watch, apple watch, apple wath, risk watch, my apple watch, 11: backpack, my work backpack, work bag, bag, shoulder bag, bum bag, back pack, work backpack, sports bag, sport bago, numb bag, 12: sunglasses, my wraparound sunglasses, dark glasses, 13: slippers, nike trainers, my shoes, boot, trainers, trainer shoe, slipper, my trainers, shoes, running shoes, 14: medication, blister pack of painkillers, tablets, money, aspirin vs tylenol, aspirin, pill bottle, my pill dosette, buckleys, box of tablets, 15: victor stream book reader, orbit braille reader and notetaker, victor reader stream, orbit reader 20 braille display, braillepen slim braille keyboard, braille orbit reader, solo audiobook player, braille note, my braille displat, rnib talking book envelope, 16: guide dog play cola, dogs lead, guide dog harness, retractable dog lead, leash, guide dog harness, dog lead, dogs, running tether, 17: blue tooth keyboard, bluetooth keyboard, my keyboard, apple wireless keyboard, mini bluetooth keyboard, portable keyboard, keyboard, 18: my water bottle, bottle, pop bottle, green water bottle, water bottle, bottle of water, clean canteen stainless steel water bottle, 19: reading glasses, glasses, eye glasses, spectacles, 20: adaptive washing machine, washing machine, tumble dryer, adaptive dryer, tumble dry setting washing machine, washing machine setting, 21: baseball cap, cap, orange skullcap, my tilly hat, hat, my tilly hat upside down, 22: microwave, toaster, kettle, one cup kettle, 23: digital dab radio, dab radio, speaker, radio, my bose bluetooth speaker, 24: wheely bin, wheelie bin, black bin, recycling bin, bin, brown wheelie bin, 25: hairbrush, comb, hair brush, 26: necklace, brown leather bracelet, ladies silver bracelet, favourite earings, silver male wedding band, clear quartz, 27: screwdriver, small space screwdriver, small screwdriver, spanner, wood phillips head, pex plumbers pliers, wood screw phillips head, 28: 12 measuring cup, 13 measuring cup, 14 measuring cup, 1 cup measuring cup, measuring spoon, measuring spoons, 29: lime green marker, yellow marker, pink marker, migenta marker, reptile green marker, pen, 30: digital recorder, dictaphone, pen friend, handheld police scanner, 31: face mask, covid mask, my purple mask, blue facemask, 32: airpod pro, single airpod, apple airpods, pro, 33: wireless earphones, bone conducting headset, bose earpods, my muse s headband, 34: phone stand, nobile phone stand, ipod stand, iphone stand, 35: memory stick, usb c dongle, usb stick, usb, 36: apple phone charger, phone charger, 37: scissors, secateurs, 38: car, my sisters car, 39: socks, sock, 40: fridge, fridge freezer indicator, 41: tv unit, flat screen television, tv, smarttv, television, 42: prosecco, jd whisky bottle, bottle of alcoholic drink, vagabond ale bottle, 43: lighter, vape pen, cannabis vape battery, 44: pepper shaker, pink himalayan salt, mediterranean sea salt, salt grinder, 45: sunglasses case, glasses case, eyewear case, 46: lipstick, chap stick, lip balm, 47: tweezers, finger nail clipper, hook , 48: inhaler, insulin pen, spacer, 49: dining table setup, desk, garden table, 50: liquid level indicator, water level sensor, 51: bottle opener, corkscrew, 52: pint glass, wine glass, glass, 53: guide dog, my flee, 54: dog streetball, dog toy, adaptive tennis ball, 55: stylus, apple pencil, 56: my laptop, laptop, 57: ipad, ipad pro, 58: mouse, 59: local post box, post box, 60: black strappy vest, t-shirts, 61: gloves, winter gloves, 62: hand gel, lotion bottle, 63: perfume, deodorant, 64: hair clip, headband, 65: toothbrush, electric toothbrush, 66: alcohol wipe, skip prep, 67: my hearing aid, hearing aid, 68: magnifier, pocket magnifying glass, 69: blue artificial eye, green artificial eye, 70: hand saw, 71: tape measure, ruler, 72: electric sanding disc, miter saw, 73: valve oil, slide grease, 74: cooker, 75: shelf unit with things, wardrobe, 76: knitting basket, washing basket, basket, 77: home phone, cordless phone, 78: skipping rope, spinner, 79: stairgate, back patio gate, 80: large sewing needle, knitting needle, 81: mayonnaise jar, mustard, 82: rice, chicken instant noodles, 83: tea, cranberry cream tea, 84: ottawa bus stop, bus stop sign, 85: journal, book, 86: security fob, 87: money clip, 88: headphone case, 89: ps4 controller, 90: compact disc, 91: garden wall, 92: garden shed, 93: litter and dog waste bin, 94: my clock, 95: sleep mask, 96: glasses cleaning wipe, 97: tissue box, 98: condom box, 99: make up, 100: clear nail varnish, 101: eye drops, 102: battery drill, 103: bed, 104: sofa, 105: cushion, 106: garden bench, 107: mirror, 108: table fan, 109: tred mill, 110: exercise bench, 111: watering can, watering, 112: vase with flowers, 113: veg peeler, 114: sharp knife, 115: wall plug, 116: embroidery thread cone, 117: hole punch, 118: pencil case, 119: paperclips, 120: uno cards, 121: baked bean tin, 122: banana, 123: cappuccino pods, 124: mountain dew can, 125: pinesol cleaner, 126: fish food, 127: cat , 128: dog poo, 129: my slate, 130: av tambourine, 131: grinder, 132: fluffy blanket

\vspace{-1em}
\paragraph{Clusters for~\cref{fig:benchmark-cluster-histo-all}}
[67 users, 486 objects, 122 clusters] 1: keys, house keys, radar key, my keys, front door keys, keychain, door keys, key, set of keys, 2: remote, tv remote, apple tv remote, amazon remote control, television remote control, remote control, virgin remote control, tv remote control, samsung tv remote control, presentation remote, sky q remote, clickr, fire stick remote, control, television control, remote tv, 3: black small wallet, my purse, my wallet, ladies purse, money pouch, coin purse, wallet for bus pass cards and money, i d wallet, ipod in wallet, walletv, wallet, purse, 4: symbol cane, black mobility cane, my cane, white mobility came, folded white cane, long cane, cane, folded cane, folded long guide cane, p939411 white cane, white came, visibility stick, mobility, mobility cane, white cane, 5: earphones, apple headphones, apple earpods, airpods, my airpod pros case, iphone air pods, airpods pro charging case, airpod case, airpods, my airpods, airpods case, earpods, 6: blue headphones, my headphones, headphone, bose wireless headphones, headset, my sennheiser pxc 350-2, headphones, 7: phone, mobile phone, iphone 6, apple mobile phone, i phone 11 pro, iphone in case, iphone, cell phone, work phone, phone case, 8: mug, personal mug, cup again, coffee mug, toddler cup, styrofoam cup, my mug, 9: front door, back door, house door, front door to house, door, my front door, shed door, 10: sunglasses, my wraparound sunglasses, dark glasses, 11: watch, wrist watch, apple watch, apple wath, risk watch, my apple watch, 12: victor stream book reader, orbit braille reader and notetaker, victor reader stream, orbit reader 20 braille display, braillepen slim braille keyboard, braille orbit reader, solo audiobook player, braille note, my braille displat, 13: blue tooth keyboard, bluetooth keyboard, my keyboard, apple wireless keyboard, mini bluetooth keyboard, portable keyboard, keyboard, 14: medication, blister pack of painkillers, tablets, money, aspirin vs tylenol, aspirin, pill bottle, my pill dosette, buckleys, 15: backpack, my work backpack, work bag, bag, shoulder bag, bum bag, back pack, work backpack, 16: slippers, nike trainers, my shoes, boot, trainers, trainer shoe, slipper, 17: reading glasses, glasses, eye glasses, 18: baseball cap, cap, orange skullcap, my tilly hat, hat, my tilly hat upside down, 19: guide dog play cola, dogs lead, guide dog harness, retractable dog lead, leash, guide dog harness, dog lead, dogs, 20: my water bottle, bottle, pop bottle, green water bottle, water bottle, 21: hairbrush, comb, hair brush, 22: adaptive washing machine, washing machine, tumble dryer, adaptive dryer, 23: digital recorder, dictaphone, pen friend, handheld police scanner, 24: wheely bin, wheelie bin, black bin, recycling bin, bin, 25: face mask, covid mask, my purple mask, blue facemask, 26: screwdriver, small space screwdriver, small screwdriver, spanner, wood phillips head, pex plumbers pliers, 27: 12 measuring cup, 13 measuring cup, 14 measuring cup, 1 cup measuring cup,  measuring spoon, 28: lime green marker, yellow marker, pink marker, migenta marker, reptile green marker, 29: airpod pro, single airpod, apple airpods, pro, 30: wireless earphones, bone conducting headset, bose earpods, my muse s headband, 31: phone stand, nobile phone stand, ipod stand, iphone stand, 32: memory stick, usb c dongle, usb stick, usb, 33: scissors, secateurs, 34: socks, sock, 35: necklace, brown leather bracelet, ladies silver bracelet, favourite earings, 36: microwave, toaster, kettle, one cup kettle, 37: fridge, fridge freezer indicator, 38: prosecco, jd whisky bottle, bottle of alcoholic drink, vagabond ale bottle, 39: digital dab radio, dab radio, speaker, 40: apple phone charger, phone charger, 41: sunglasses case, glasses case, eyewear case, 42: lipstick, chap stick, lip balm, 43: tv unit, flat screen television, tv, smarttv, 44: dining table setup, desk, garden table, 45: liquid level indicator, water level sensor, 46: bottle opener, corkscrew, 47: pint glass, wine glass, glass, 48: lighter, vape pen, cannabis vape battery, 49: pepper shaker, pink himalayan salt, mediterranean sea salt, 50: dog streetball, dog toy, adaptive tennis ball, 51: stylus, apple pencil, 52: mouse, 53: car, 54: local post box, post box, 55: black strappy vest, t-shirts, 56: gloves, winter gloves, 57: hand gel, lotion bottle, 58: perfume, deodorant, 59: tweezers, finger nail clipper , 60: hair clip, headband, 61: inhaler, insulin pen, 62: alcohol wipe, skip prep, 63: my hearing aid, hearing aid, 64: magnifier, pocket magnifying glass, 65: hand saw, 66: tape measure, ruler, 67: electric sanding disc, miter saw, 68: cooker, 69: shelf unit with things, wardrobe, 70: knitting basket, washing basket, basket, 71: stairgate, back patio gate, 72: large sewing needle, knitting needle, 73: mayonnaise jar, mustard, 74: rice, chicken instant noodles, 75: tea, cranberry cream tea, 76: ottawa bus stop, bus stop sign, 77: money clip, 78: headphone case, 79: ps4 controller, 80: my laptop, 81: ipad, 82: compact disc, 83: garden wall, 84: garden shed, 85: litter and dog waste bin, 86: my clock, 87: sleep mask, 88: glasses cleaning wipe, 89: tissue box, 90: condom box, 91: make up, 92: clear nail varnish, 93: toothbrush, 94: eye drops, 95: battery drill, 96: bed, 97: sofa, 98: cushion, 99: mirror, 100: table fan, 101: tred mill, 102: exercise bench, 103: skipping rope, 104: watering can, watering, 105: vase with flowers, 106: veg peeler, 107: sharp knife, 108: wall plug, 109: embroidery thread cone, 110: hole punch, 111: pencil case, 112: paperclips, 113: baked bean tin, 114: banana, 115: mountain dew can, 116: pinesol cleaner, 117: fish food, 118: dog poo, 119: my slate, 120: av tambourine, 121: grinder, 122: journal

\vspace{-1em}
\paragraph{Clusters for~\cref{fig:benchmark-cluster-histo-train}}
[44 train users, 278 objects, 100 clusters] 1: keys, house keys, radar key, my keys, front door keys, keychain, door keys, key, set of keys, 2: remote, tv remote, apple tv remote, amazon remote control, television remote control, remote control, virgin remote control, tv remote control, samsung tv remote control, presentation remote, sky q remote, clickr, fire stick remote, control, television control, remote tv, 3: symbol cane, black mobility cane, my cane, white mobility came, folded white cane, long cane, cane, folded cane, folded long guide cane, p939411 white cane, white came, visibility stick, mobility, mobility cane, white cane, 4: earphones, apple headphones, apple earpods, airpods, my airpod pros case, iphone air pods, airpods pro charging case, airpod case, airpods, my airpods, airpods case, earpods, 5: black small wallet, my purse, my wallet, ladies purse, money pouch, coin purse, wallet for bus pass cards and money, i d wallet, ipod in wallet, walletv, wallet, purse, 6: blue headphones, my headphones, headphone, bose wireless headphones, headset, my sennheiser pxc 350-2, headphones, 7: phone, mobile phone, iphone 6, apple mobile phone, i phone 11 pro, iphone in case, iphone, cell phone, work phone, phone case, 8: sunglasses, my wraparound sunglasses, dark glasses, 9: medication, blister pack of painkillers, tablets, money, aspirin vs tylenol, aspirin, pill bottle, my pill dosette, buckleys, 10: mug, personal mug, cup again, coffee mug, toddler cup, styrofoam cup, my mug, 11: reading glasses, glasses, eye glasses, 12: slippers, nike trainers, my shoes, boot, trainers, trainer shoe, slipper, 13: watch, wrist watch, apple watch, apple wath, risk watch, my apple watch, 14: backpack, my work backpack, work bag, bag, shoulder bag, bum bag, back pack, work backpack, 15: guide dog play cola, dogs lead, guide dog harness, retractable dog lead, leash, guide dog harness, dog lead, dogs, 16: front door, back door, house door, front door to house, door, my front door, shed door, 17: blue tooth keyboard, bluetooth keyboard, my keyboard, apple wireless keyboard, mini bluetooth keyboard, portable keyboard, keyboard, 18: face mask, covid mask, my purple mask, blue facemask, 19: baseball cap, cap, orange skullcap, my tilly hat, hat, my tilly hat upside down, 20: hairbrush, comb, hair brush, 21: victor stream book reader, orbit braille reader and notetaker, victor reader stream, orbit reader 20 braille display, braillepen slim braille keyboard, braille orbit reader, solo audiobook player, braille note, my braille displat, 22: my water bottle, bottle, pop bottle, green water bottle, water bottle, 23: digital recorder, dictaphone, pen friend, handheld police scanner, 24: memory stick, usb c dongle, usb stick, usb, 25: sunglasses case, glasses case, eyewear case, 26: fridge, fridge freezer indicator, 27: adaptive washing machine, washing machine, tumble dryer, adaptive dryer, 28: pint glass, wine glass, glass, 29: airpod pro, single airpod, apple airpods, pro, 30: wireless earphones, bone conducting headset, bose earpods, my muse s headband, 31: mouse, 32: apple phone charger, phone charger, 33: scissors, secateurs, 34: local post box, post box, 35: wheely bin, wheelie bin, black bin, recycling bin, bin, 36: black strappy vest, t-shirts, 37: lipstick, chap stick, lip balm, 38: alcohol wipe, skip prep, 39: my hearing aid, hearing aid, 40: cooker, 41: microwave, toaster, kettle, one cup kettle, 42: bottle opener, corkscrew, 43: large sewing needle, knitting needle, 44: prosecco, jd whisky bottle, bottle of alcoholic drink, vagabond ale bottle, 45: lighter, vape pen, cannabis vape battery, 46: mayonnaise jar, mustard, 47: rice, chicken instant noodles, 48: tea, cranberry cream tea, 49: ottawa bus stop, bus stop sign, 50: money clip, 51: headphone case, 52: stylus, apple pencil, 53: my laptop, 54: compact disc, 55: digital dab radio, dab radio, speaker, 56: phone stand, nobile phone stand, ipod stand, iphone stand, 57: litter and dog waste bin, 58: socks, sock, 59: gloves, winter gloves, 60: sleep mask, 61: glasses cleaning wipe, 62: tissue box, 63: hand gel, lotion bottle, 64: make up, 65: clear nail varnish, 66: hair clip, headband, 67: toothbrush, 68: inhaler, insulin pen, 69: eye drops, 70: necklace, brown leather bracelet, ladies silver bracelet, favourite earings, 71: magnifier, pocket magnifying glass, 72: screwdriver, small space screwdriver, small screwdriver, spanner, wood phillips head, pex plumbers pliers, 73: hand saw, 74: battery drill, 75: tape measure, ruler, 76: electric sanding disc, miter saw, 77: tv unit, flat screen television, tv, smarttv, 78: bed, 79: sofa, 80: dining table setup, desk, garden table, 81: mirror, 82: tred mill, 83: exercise bench, 84: skipping rope, 85: stairgate, back patio gate, 86: liquid level indicator, water level sensor, 87: sharp knife, 88: embroidery thread cone, 89: hole punch, 90: pencil case, 91: baked bean tin, 92: banana, 93: pepper shaker, pink himalayan salt, mediterranean sea salt, 94: mountain dew can, 95: fish food, 96: dog poo, 97: dog streetball, dog toy, adaptive tennis ball, 98: av tambourine, 99: grinder, 100: journal

\vspace{-1em}
\paragraph{Clusters for~\cref{fig:benchmark-cluster-histo-val}}
[6 validation users, 50 objects, 36 clusters] 1: keys, house keys, radar key, my keys, front door keys, keychain, door keys, key, set of keys, 2: black small wallet, my purse, my wallet, ladies purse, money pouch, coin purse, wallet for bus pass cards and money, i d wallet, ipod in wallet, walletv, wallet, purse, 3: blue headphones, my headphones, headphone, bose wireless headphones, headset, my sennheiser pxc 350-2, headphones, 4: front door, back door, house door, front door to house, door, my front door, shed door, 5: remote, tv remote, apple tv remote, amazon remote control, television remote control, remote control, virgin remote control, tv remote control, samsung tv remote control, presentation remote, sky q remote, clickr, fire stick remote, control, television control, remote tv, 6: blue tooth keyboard, bluetooth keyboard, my keyboard, apple wireless keyboard, mini bluetooth keyboard, portable keyboard, keyboard, 7: phone, mobile phone, iphone 6, apple mobile phone, i phone 11 pro, iphone in case, iphone, cell phone, work phone, phone case, 8: watch, wrist watch, apple watch, apple wath, risk watch, my apple watch, 9: sunglasses, my wraparound sunglasses, dark glasses, 10: necklace, brown leather bracelet, ladies silver bracelet, favourite earings, 11: tv unit, flat screen television, tv, smarttv, 12: apple phone charger, phone charger, 13: car, 14: slippers, nike trainers, my shoes, boot, trainers, trainer shoe, slipper, 15: socks, sock, 16: my clock, 17: condom box, 18: baseball cap, cap, orange skullcap, my tilly hat, hat, my tilly hat upside down, 19: hairbrush, comb, hair brush, 20: perfume, deodorant, 21: tweezers, finger nail clipper , 22: hair clip, headband, 23: backpack, my work backpack, work bag, bag, shoulder bag, bum bag, back pack, work backpack, 24: symbol cane, black mobility cane, my cane, white mobility came, folded white cane, long cane, cane, folded cane, folded long guide cane, p939411 white cane, white came, visibility stick, mobility, mobility cane white cane, 25: victor stream book reader, orbit braille reader and notetaker, victor reader stream, orbit reader 20 braille display, braillepen slim braille keyboard, braille orbit reader, solo audiobook player, braille note, my braille displat, 26: magnifier, pocket magnifying glass, 27: guide dog play cola, dogs lead, guide dog harness, retractable dog lead, leash, guide dog harness, dog lead, dogs, 28: shelf unit with things, wardrobe, 29: dining table setup, desk, garden table, 30: vase with flowers, 31: mug, personal mug, cup again, coffee mug, toddler cup, styrofoam cup, my mug, 32: wall plug, 33: paperclips, 34: prosecco, jd whisky bottle, bottle of alcoholic drink, vagabond ale bottle, 35: pinesol cleaner, 36: my slate

\vspace{-1em}
\paragraph{Clusters for~\cref{fig:benchmark-cluster-histo-test}}
[17 test users, 158 objects, 67 clusters] 1: black small wallet, my purse, my wallet, ladies purse, money pouch, coin purse, wallet for bus pass cards and money, i d wallet, ipod in wallet, walletv, wallet, purse, 2: keys, house keys, radar key, my keys, front door keys, keychain, door keys, key, set of keys, 3: remote, tv remote, apple tv remote, amazon remote control, television remote control, remote control, virgin remote control, tv remote control, samsung tv remote control, presentation remote, sky q remote, clickr, fire stick remote, control, television control, remote tv, 4: front door, back door, house door, front door to house, door, my front door, shed door, 5: earphones, apple headphones, apple earpods, airpods, my airpod pros case, iphone air pods, airpods pro charging case, airpod case, airpods, my airpods, airpods case, earpods, 6: victor stream book reader, orbit braille reader and notetaker, victor reader stream, orbit reader 20 braille display, braillepen slim braille keyboard, braille orbit reader, solo audiobook player, braille note, my braille displat, 7: 12 measuring cup, 13 measuring cup, 14 measuring cup, 1 cup measuring cup,  measuring spoon, 8: lime green marker, yellow marker, pink marker, migenta marker, reptile green marker, 9: watch, wrist watch, apple watch, apple wath, risk watch, my apple watch, 10: symbol cane, black mobility cane, my cane, white mobility came, folded white cane, long cane, cane, folded cane, folded long guide cane, p939411 white cane, white came, visibility stick, mobility, mobility cane, white cane, 11: screwdriver, small space screwdriver, small screwdriver, spanner, wood phillips head, pex plumbers pliers, 12: mug, personal mug, cup again, coffee mug, toddler cup, styrofoam cup, my mug, 13: blue headphones, my headphones, headphone, bose wireless headphones, headset, my sennheiser pxc 350-2, headphones, 14: blue tooth keyboard, bluetooth keyboard, my keyboard, apple wireless keyboard, mini bluetooth keyboard, portable keyboard, keyboard, 15: phone, mobile phone, iphone 6, apple mobile phone, i phone 11 pro, iphone in case, iphone, cell phone, work phone, phone case, 16: phone stand, nobile phone stand, ipod stand, iphone stand, 17: wheely bin, wheelie bin, black bin, recycling bin, bin, 18: backpack, my work backpack, work bag, bag, shoulder bag, bum bag, back pack, work backpack, 19: adaptive washing machine, washing machine, tumble dryer, adaptive dryer, 20: my water bottle, bottle, pop bottle, green water bottle, water bottle, 21: airpod pro, single airpod, apple airpods, pro, 22: wireless earphones, bone conducting headset, bose earpods, my muse s headband, 23: digital recorder, dictaphone, pen friend, handheld police scanner, 24: digital dab radio, dab radio, speaker, 25: scissors, secateurs, 26: slippers, nike trainers, my shoes, boot, trainers, trainer shoe, slipper, 27: socks, sock, 28: sunglasses, my wraparound sunglasses, dark glasses, 29: baseball cap, cap, orange skullcap, my tilly hat, hat, my tilly hat upside down, 30: microwave, toaster, kettle, one cup kettle, 31: knitting basket, washing basket, basket, 32: liquid level indicator, water level sensor, 33: pepper shaker, pink himalayan salt, mediterranean sea salt, 34: dog streetball, dog toy, adaptive tennis ball, 35: stylus, apple pencil, 36: ps4 controller, 37: ipad, 38: memory stick, usb c dongle, usb stick, usb, 39: car, 40: garden wall, 41: garden shed, 42: gloves, winter gloves, 43: face mask, covid mask, my purple mask, blue facemask, 44: reading glasses, glasses, eye glasses, 45: hand gel, lotion bottle, 46: hairbrush, comb, hair brush, 47: perfume, deodorant, 48: lipstick, chap stick, lip balm, 49: tweezers, finger nail clipper , 50: medication, blister pack of painkillers, tablets, money, aspirin vs tylenol, aspirin, pill bottle, my pill dosette, buckleys, 51: inhaler, insulin pen, 52: necklace, brown leather bracelet, ladies silver bracelet, favourite earings, 53: guide dog play cola, dogs lead, guide dog harness, retractable dog lead, leash, guide dog harness, dog lead, dogs, 54: hand saw, 55: tape measure, ruler, 56: electric sanding disc, miter saw, 57: fridge, fridge freezer indicator, 58: cushion, 59: shelf unit with things, wardrobe, 60: dining table setup, desk, garden table, 61: table fan, 62: stairgate, back patio gate, 63: watering can, watering, 64: veg peeler, 65: bottle opener, corkscrew, 66: prosecco, jd whisky bottle, bottle of alcoholic drink, vagabond ale bottle, 67: lighter, vape pen, cannabis vape battery

\section{Datasheet for the ORBIT Dataset}
\label{app:sec:datasheet}
\glsresetall
\noindent Here we include a datasheet~\cite{gebru2018datasheets} for the ORBIT dataset.

\subsection{Motivation for dataset creation}

\paragraph{Why was the dataset created? (e.g., was there a specific task in mind?; was there a specific gap that needed to be filled?)}

The ORBIT dataset was created with two aims in mind: 1) to drive research in few-shot object recognition from high-variation videos --- compared to existing few-shot learning datasets which contain largely curated images and pose structured benchmark tasks, the ORBIT dataset contains realistic, high-variation videos captured by people who are blind/low-vision on their mobile phones in real-world settings; and 2) to enable the real-world application of \glspl{TOR} for people who are blind/low-vision.

\vspace{-1em}
\paragraph{Has the dataset been used already? If so, where are the results so others can compare (e.g., links to published papers)?}

No, this is a new dataset. This paper presents the first results on a few-shot learning benchmark task (see~\cref{sec:orbit-benchmark}) that was developed on this dataset.

\vspace{-1em}
\paragraph{What (other) tasks could the dataset be used for?}

The dataset could be used to explore a wide range of frame- and video-level recognition tasks including classification (this paper), object detection, object tracking, semantic segmentation, and instance segmentation.
Within each of these tasks, the dataset lends itself to exploring the robustness of models (e.g. performance, quantification of uncertainty) in the face of few high-variation examples. 
Beyond a few-shot setting, it could also be used to explore other learning paradigms, for example, continual learning where new objects are incrementally added over time.

In addition, this dataset supports research from a human and accessibility perspective.
It provides evidence of how people who are blind/low-vision take videos, what objects are of interest to them, and how they could be further supported in taking videos.
This work could be directly applied in assistive technology to support object recognition, and be extended to other application areas, such as navigation, question-answering, etc.

\vspace{-1em}
\paragraph{Who funded the creation of the dataset?}

The dataset was funded by the \href{https://www.microsoft.com/en-us/ai/ai-for-accessibility}{Microsoft AI for Accessibility} program.

\vspace{-1em}
\paragraph{Any other comment?}
None.

\subsection{Dataset composition}

\paragraph{What are the instances? (that is, examples; e.g. documents, images, people, countries) Are there multiple types of instances? (e.g. movies, users, ratings; people, interactions between them; nodes, edges)}

The ORBIT dataset is a collection of videos of objects recorded by people who are blind/low-vision on their mobile phones.
Each collector recorded 2 types of videos for each object: clean videos which show the object in isolation on a clear surface, and clutter videos which show the object in a realistic, multi-object scene.
The clutter videos were recorded with two different techniques: a zoom-out and a panning motion (see~\cref{app:sec:data-collection-protocol}).
The dataset is arranged into two parts: the unfiltered ORBIT dataset which contains \emph{all} videos contributed by collectors, and the ORBIT benchmark dataset which is a subset of the unfiltered dataset and is used to run the benchmark evaluation (see~\cref{sec:eval-protocol}).

\vspace{-1em}
\paragraph{How many instances are there in total (of each type, if appropriate)?}

The unfiltered dataset has 4,733 videos (3,161,718 frames, 97GB) of 588 objects. 3,356 of  these videos are clean videos and 1,377 are clutter videos. Of the clutter videos, 873 were recorded with a zoom-out technique, and 504 with a panning technique.
The benchmark dataset has 3,822 videos (2,687,934 frames, 83GB) of 486 objects. 2996 of these videos are clean videos and 826 videos are clutter videos. Here, the clutter videos were all recorded with the zoom-out technique.
For detailed statistics of the unfiltered and benchmark datasets, see~\cref{tab:full-dataset-summary} and~\cref{tab:benchmark-dataset-summary}, respectively.

\vspace{-1em}
\paragraph{What data does each instance consist of? “Raw” data (e.g. unprocessed text or images)? Features/attributes? If the instances related to people, are sub-populations identified (e.g. by age, gender, etc.) and what is their distribution?}

Each instance is an unprocessed video frame (i.e. a JPEG). Frames are grouped into folders by video. 
The video folders are then grouped by video type (clean/clutter), then by object, and finally by the (anonymized) collector who contributed them (see file hierarchy in~\cref{app:sec:file-structure}).
Sub-populations of collectors were not identified. 

\vspace{-1em}
\paragraph{Is there a label or target associated with each instance? If so, please provide a description.}
All frames from a video are labeled with the (video-level) object name entered by the collector prior to recording the video.
All clutter videos were further annotated (offline) with per-frame bounding boxes around the target object.

\vspace{-1em}
\paragraph{Is any information missing from individual instances? If so, please
provide a description, explaining why this information is missing (e.g., because it was unavailable). This does not include intentionally removed information, but might include, e.g., redacted text.}

No information is missing from individual instances (i.e.\ frames), however, in the collection process, whole videos were removed by a human annotator if:
\begin{inparaenum} [i)]
\item the object was not present in the video, 
\item the video contained \gls{PII}, or
\item the object or object name were inappropriate.
\end{inparaenum}
Note, if \gls{PII} was present in a small number of frames in a video, only these frames were removed.

\vspace{-1em}
\paragraph{Are relationships between individual instances made explicit (e.g. users’ movie ratings, social network links)? If so, please describe how these relationships are made explicit.}

Yes, frames are organized by video, video type, object and finally the (anonymized) collector who contributed them. These relationships are made explicit in the dataset's file structure (see~\cref{app:sec:file-structure}).

\vspace{-1em}
\paragraph{Does the dataset contain all possible instances or is it a sample (not necessarily random) of instances from a larger set? If the dataset is a sample, then what is the larger set? Is the sample representative of the larger set (e.g. geographic coverage)? If so, please describe how this representativeness was validated/verified. If it is not representative of the larger set, please describe why not (e.g. to cover a more diverse range of instances, because instances were withheld or unavailable).}

The dataset was collected by people who are blind/low-vision.
Although participation was open to the public, most collectors were recruited via blind/low-vision charities in Level 4\footnote{https://www.gapminder.org/fw/income-levels} countries (primarily UK, Canada, USA, Australia).
In addition, the data collection task required an iPhone with iOS  \(\geq\) 13.2, with sufficient space to store the videos before upload. To upload the videos to the server, collectors required a stable internet connection, and power supply.

\vspace{-1em}
\paragraph{Are there recommended data splits (e.g., training, development/validation, testing)? If so, please provide a description of these splits, explaining the rationale behind them.}

Yes, the recommended splits for the benchmark dataset are described in~\cref{app:sec:dataset-prep-set-creation}.
The unfiltered dataset has no recommended splits as it is not explicitly used for the benchmark evaluation. 

\vspace{-1em}
\paragraph{Are there any errors, sources of noise, or redundancies in the dataset? If so, please provide a description.}

By design, the dataset contains high-variation videos of objects.
Variation comes from objects being fully/partially out-of-frame, blurred (due to camera or object motion), too close or far from the camera, obstructed by hands or other objects, or under/overexposed.
All videos are unique, but there is always at least one video per unique object.
Some objects occur multiple times across different collectors, but each collector never has duplicate objects. 

\vspace{-1em}
\paragraph{Is the dataset self-contained, or does it link to or otherwise rely on external resources (e.g., websites, tweets, other datasets)? If it links to or relies on external resources, a) are there guarantees that they will exist, and remain constant, over time; b) are there official archival versions of the complete dataset (i.e., including the external resources as they existed at the time the dataset was created); c) are there any restrictions (e.g., licenses, fees) associated with any of the external resources that might apply to a future user? Please provide descriptions of all external resources and any restrictions associated with them, as well as links or other access points, as appropriate.}

The dataset and its additional annotations are self-contained and do not rely on any external resources.

\vspace{-1em}
\paragraph{Any other comments?}
None.

\subsection{Collection process}

\paragraph{What mechanisms or procedures were used to collect the data (e.g. hardware apparatus or sensor, manual human curation, software program, software API)? How were these mechanisms or procedures validated?}

The procedures used to collect the ORBIT data through manual curation are described in~\cref{app:sec:data-collection-protocol}. We validated these procedures through user testing and user feedback. 
The app used to collect the data is described in~\cref{app:sec:orbit-camera-app}.
It followed Apple's accessibility guidelines and was tested with blind/low-vision users. 
 
\vspace{-1em}
\paragraph{How was the data associated with each instance acquired? Was the data directly observable (e.g. raw text, movie ratings), reported by subjects (e.g. survey responses), or indirectly inferred/derived from other data (e.g. part-of-speech tags, model-based guesses for age or language)? If data was reported by subjects or indirectly inferred/derived from other data, was the data validated/verified? If so, please describe how.}

Videos were recorded and contributed manually by blind/low-vision collectors via a custom-built iOS app (see~\cref{app:sec:orbit-camera-app}).
 
\vspace{-1em}
\paragraph{If the dataset is a sample from a larger set, what was the sampling strategy (e.g. deterministic, probabilistic with specific sampling probabilities)?}

The dataset was not sampled from a larger set.

\vspace{-1em}
\paragraph{Over what time-frame was the data collected? Does this time-frame match the creation time-frame of the data associated with the instances (e.g. recent crawl of old news articles)? If not, please describe the time-frame in which the data associated with the instances was created.}

The dataset was collected in two phases. The first phase ran from June to August 2020 and involved 49 collectors who were blind/low-vision.
The second phase ran from November 2020 to January 2021 and involved 49 collectors who were blind/low-vision.
The time-frames for the dataset creation and the data itself overlap.

\vspace{-1em}
\paragraph{Who was involved in the data collection process (e.g. students, crowdworkers, contractors) and how were they compensated (e.g. how much were crowdworkers paid)?}

The dataset was collected by people who are blind/low-vision. In the first phase, collectors were based in the UK and were incentivized with a £50 voucher for their participation.
In the second phase, collectors were based globally and a donation of £25 (or the local currency equivalent) was made to one of five charities chosen by the collector.

\subsection{Data pre-processing}

\paragraph{Was any pre-processing/cleaning/labeling of the data done (e.g., discretization or bucketing, tokenization, part-of-speech tagging, SIFT feature extraction, removal of instances, processing of missing values)? If so, please provide a description. If not, you may skip the remainder of the questions in this section.}

During the data collection process, videos were expunged from the server if
\begin{inparaenum} [i)]
\item the object was not present in the video, 
\item the video contained \gls{PII}, or
\item the object or object name were inappropriate.
\end{inparaenum}
All remaining videos formed the unfiltered ORBIT dataset, and were not pre-processed in any other way.

The ORBIT benchmark dataset was constructed from the unfiltered dataset by removing 
\begin{inparaenum}[i)]
\item clutter videos recorded with the panning technique,
\item videos shorter than 30 frames, and
\item objects without at least 2 clean videos and 1 clutter video (see further details in~\cref{app:sec:dataset-prep}).
\end{inparaenum}
Finally, to run the baseline models on the benchmark task, video frames were resized from \(1080\times 1080\) to \(84\times 84\) pixels.

Regarding labeling, in addition to the object labels provided by collectors for each video, the clutter videos were further annotated with bounding boxes around the target object.
These were provided by a private company with whom we worked closely to ensure consistent and high-quality annotations across the dataset.

\vspace{-1em}
\paragraph{Was the ``raw'' data saved in addition to the pre-processed/cleaned/labeled data (e.g., to support unanticipated future uses)? If so, please provide a link or other access point to the
“raw” data.}

Both the unfiltered and benchmark datasets are available at \href{https://doi.org/10.25383/city.14294597}{https://doi.org/10.25383/city.14294597}.
For download details see the code repository at \href{https://github.com/microsoft/ORBIT-Dataset}{https://github.com/microsoft/ORBIT-Dataset}.
Note, videos expunged from the server during the data collection process were not saved.

\vspace{-1em}
\paragraph{Is the software used to pre-process/clean/label the instances available? If so, please provide a link or other access point.}

Pre-processing/cleaning involved removing videos that did not meet specific criteria from the unfiltered dataset to obtain the benchmark dataset (see~\cref{app:sec:dataset-prep-pre-proc}).
While scripts to do this are not explicitly provided, scripts to download (and resize frames) in the unfiltered and benchmark datasets are provided in the code repository.
Annotations beyond the video labels (e.g. bounding boxes) were collected by a private annotation company using proprietary software.

\vspace{-1em}
\paragraph{Does this dataset collection/processing procedure achieve the motivation for creating the dataset stated in the first section of this datasheet? If not, what are the limitations?}

Yes, by collecting videos from blind/low-vision collectors ``in situ'' and doing minimal pre-processing/cleaning of these videos has yielded a dataset that is highly representative of the high-variation real-world scenarios they encounter.
As a result, the unfiltered and benchmark datasets can be used to
\begin{inparaenum}[1)]
\item drive research in few-shot object recognition from high-variation videos, and
\item realize \glspl{TOR} for people who are blind/low-vision.
\end{inparaenum}

\vspace{-1em}
\paragraph{Any other comments}
None.
 
\subsection{Dataset distribution}

\paragraph{How will the dataset be distributed? (e.g., tarball on website, API, GitHub; does the data have a DOI and is it archived redundantly?)}

The ORBIT dataset is available as 4 zip files (train, validation, test, other) at DOI: \href{https://doi.org/10.25383/city.14294597}{https://doi.org/10.25383/city.14294597}.
We provide versions of each zip file with frames resized to \(224\times224\) pixels.

\vspace{-1em}
\paragraph{When will the dataset be released/first distributed? What license (if any) is it distributed under?}

The dataset was released on 7 April 2021 under an CC4.0 license. The code repository is released under an MIT license.

\vspace{-1em}
\paragraph{Are there any copyrights on the data?}
 
Under the CC4.0 license, the data can be shared and adapted, even commercially. The data must be correctly attributed.

\vspace{-1em}
\paragraph{Are there any fees or access/export restrictions?}

No, the dataset and code are open-source.
 
\vspace{-1em}
\paragraph{Any other comments?}
None.

\subsection{Dataset maintenance}

\paragraph{Who is supporting/hosting/maintaining the
dataset?}

The ORBIT dataset (unfiltered and benchmark) are hosted, and maintained by City, University of London. 
 
\vspace{-1em}
\paragraph{Will the dataset be updated? If so, how often and by whom?}

The dataset may be updated in the future with more videos collected by people who are blind/low-vision.
 
\vspace{-1em}
\paragraph{How will updates be communicated? (e.g. mailing list, GitHub)}

All updates will be communicated via the dataset's DOI (\href{https://doi.org/10.25383/city.14294597}{https://doi.org/10.25383/city.14294597}), code repository (\href{https://github.com/microsoft/ORBIT-Dataset}{https://github.com/microsoft/ORBIT-Dataset}), and website (\href{https://orbit.city.ac.uk}{https://orbit.city.ac.uk}), including if the dataset becomes obsolete.
 

\vspace{-1em}
\paragraph{Is there a repository to link to any/all papers/systems that use this dataset?}

No explicit repository has been built for this purpose, however the usage of this dataset can be tracked via the `cited by' feature available in most referencing tools.

\vspace{-1em}
\paragraph{If others want to extend/augment/build on this dataset, is there a mechanism for them to do so? If so, is there a process for tracking/assessing the quality of those contributions. What is the process for communicating/distributing these contributions
to users?}

Individuals are free to extend/augment/build on the dataset. This may include collecting more data/annotations, developing new models, or contributing to code in the repository.
For any questions, individuals can reach out to \href{mailto:info@orbit.city.ac.uk}{info@orbit.city.ac.uk} or use the issue tracker in the code repository. 
Any future updates will be communicated/distributed via the DOI, code repository, and website.

\subsection{Legal and ethical considerations}

\paragraph{Were any ethical review processes conducted (e.g., by an institutional review board)? If so, please provide a description of these review processes, including the outcomes, as well as a link or other access point to any supporting documentation.}

The pilot study for the ORBIT data collection was approved by the City, University of London, UK Computer Science and Library \& Information Science Research Ethics Committee (ETH1920-0331). The first and second phases of the data collection were approved by the City, University of London, UK Computer Science and Library \& Information Science Research Ethics Committee (ETH1920-1126 and ETH2021-0032, respectively).
 
\vspace{-1em}
\paragraph{Does the dataset contain data that might be considered confidential (e.g. data that is protected by legal privilege or by doctor patient confidentiality, data that includes the content of individuals non-public communications)? If so, please provide a description.}

No, all videos were checked by a human annotator during the data collection process and were expunged from the server if they contained any \gls{PII} or confidential content.

\vspace{-1em}
\paragraph{Does the dataset contain data that, if viewed directly, might be offensive, insulting, threatening, or might otherwise cause anxiety? If so, please describe why}

No, all videos were checked by a human annotator during the data collection process and were expunged from the server if they contained any offensive or inappropriate content.

\vspace{-1em}
\paragraph{Does the dataset relate to people? If not, you may skip the remaining
questions in this section.}

Yes, the dataset was collected by people who are blind/low-vision with the goal of realizing a real-world application, namely \glspl{TOR}.

\vspace{-1em}
\paragraph{Does the dataset identify any sub-populations (e.g., by age, gender)?
If so, please describe how these sub-populations are identified and provide a description of their respective distributions within the dataset.}

No, no sub-populations are identified in the dataset.
 
\vspace{-1em}
\paragraph{Is it possible to identify individuals (i.e., one or more natural persons), either directly or indirectly (i.e., in combination with other
data) from the dataset? If so, please describe how.}

No, collection was completely anonymized. No meta-data or \gls{PII} was collected that would allow individuals to be identified indirectly.
 
\vspace{-1em}
\paragraph{Does the dataset contain data that might be considered sensitive in any way (e.g. data that reveals racial or ethnic origins, sexual orientations, religious beliefs, political opinions or union memberships, or
locations; financial or health data; biometric or genetic data; forms of government identification, such as social security numbers; criminal
history)? If so, please provide a description.}

No, the dataset contains no sensitive data.
 
\vspace{-1em}
\paragraph{Did you collect the data from the individuals in question directly, or
obtain it via third parties or other sources (e.g. websites)?}

All videos were collected directly from the individuals in question.
 
\vspace{-1em}
\paragraph{Were the individuals in question notified about the data collection? If so, please describe (or show with screenshots or other information) how
notice was provided, and provide a link or other access point to, or otherwise reproduce, the exact language of the notification itself.}

Yes, collectors were invited to take part in the data collection through advertisements on social media, mailing lists, blogs, and podcasts.

\vspace{-1em}
\paragraph{Did the individuals in question consent to the collection and use of their data? If so, please describe (or show with screenshots or other
information) how consent was requested and provided, and provide a link or other access point to, or otherwise reproduce, the exact language to
which the individuals consented.}

Collectors were required to provide informed consent as part of the sign-up procedure and app download. Informed consent covered the following points:
\begin{compactitem}
    \item I confirm that I have read and understood the participant information dated [INSERT DATE AND VERSION NUMBER] for the above study. I have had the opportunity to consider the information and ask questions which have been answered satisfactorily.
    \item  I understand that, if approved, the videos I take with the iPhone app will become part of an open dataset.
    \item I understand that the researcher might choose not to publish one or more of my videos in the open dataset. 
    \item I understand that my participation is voluntary and that I am free to withdraw without giving a reason, and without being penalized or disadvantaged. 
    \item  I agree that, unless I decide to withdraw, the data collected up to this point will be used in the study.
    \item I agree to City, University of London recording and processing this information about me. I understand that this information will be used only for the purposes explained in the participant information, and my consent is conditional on City, University of London complying with its duties and obligations under the General Data Protection Regulation (GDPR).
\end{compactitem}

\vspace{-1em}
\paragraph{If consent was obtained, were the consenting individuals provided with a mechanism to revoke their consent in the future or for certain
uses? If so, please provide a description, as well as a link or other access point to the mechanism (if appropriate).}

Collectors were able to delete data through the iOS app before the data collection period ended.
Collectors were free to withdraw their consent at any time, and email the researchers to delete their data from the server. 

\vspace{-1em}
\paragraph{Has an analysis of the potential impact of the dataset and its use on data subjects (e.g. a data protection impact analysis) been conducted? If so, please provide a description of this analysis, including the outcomes, as well as a link or other access point to any supporting documentation.}

A Data Protection Impact Analysis was carried out as part of the Ethics approval process, and the public release of the dataset. It was determined low risk.   
 
\vspace{-1em}
\paragraph{Any other comments?}
None.

\section{Episodic \& non-episodic learning for \glspl{TOR}}
\label{app:sec:episodic-vs-non}

In the \textbf{(2) personalize} step, a few-shot recognition model, or a \gls{TOR}, aims to learn to recognize completely new objects from only a few examples (in our case videos) captured by the user themselves.
The recognition model can be trained to do this in a episodic or non-episodic manner (i.e. the \textbf{(1) train} step).
Here, we describe both classes of approach.

For a user \(\user\), we denote that each of their personal objects \(p \in \objectset^\user\) has \(N^\user_p\) context videos and \(M^\user_p\) target videos:
\begin{align}
    \allcleanvideos^\user_p = \{(\cleanvideo, \cleanvideolabel)_i \}_{i=1}^{N^\user_p} \quad\quad\quad
    \allcluttervideos^\user_p = \{(\cluttervideo, \cluttervideolabel)_i \}_{i=1}^{M^\user_p}
\end{align}
where \(\cleanvideo, \cluttervideo\) are context and target videos, respectively, and \(\cluttervideolabel\) is the object label.
We group the user's context and target videos over all their \(|\objectset^\user|\) objects as:
\begin{align}
\allcleanvideos^\user = \allcleanvideos^\user_1 \cup \cdots \cup \allcleanvideos^\user_{|\objectset^\user|} \quad\quad \allcluttervideos^\user = \allcluttervideos^\user_1 \cup \dots \cup \allcluttervideos^\user_{|\objectset^\user|} .
\end{align}

\subsection{Episodic few-shot learning}
\label{app:sec:episodic-few-shot}

An episodic approach~\cite{snell2017prototypical,vinyals2016matching,finn2017model,zintgraf2018cavia,requeima2019fast} is characterized by training the model in ``episodes'' or tasks.
For a train user \(\user \sim \trainusers\), a task is a random sample of their context videos \(\contextset^\user\) and target videos \(\targetset^\user\):
\begin{align}\label{eq:contexttargetset}
    \contextset^\user = \{(\cleanvideo, \cleanvideolabel)_i \}_{i=1}^N \quad\quad\quad \targetset^\user = \{(\cluttervideo, \cluttervideolabel)_i\}_{i=1}^M
\end{align}
where \(\cleanvideo, \cluttervideo\) are context and target videos, respectively, and \(\cluttervideolabel \in \objectset^\user\) is the object label.
Note, \(\contextset^\user \sim \allcleanvideos^\user\) and \(\targetset^\user \sim \allcluttervideos^\user\), where \(\allcleanvideos^\user\) and \(\allcluttervideos^\user\) are the total set of the user's context and target videos, respectively (see~\cref{app:sec:task-sampling} for further task sampling details).

A few-shot recognition model \(f\) is trained over many randomly sampled tasks per train user \(\user \sim \trainusers\), typically by maximizing the likelihood of predicting the user's correct personal object in each frame in a target video \(\cluttervideoframe_f \in \cluttervideo\) after seeing only that user's context videos \(\contextset^\user\):
\begin{align}\label{eq:episodic-theta-star}
\theta^*\!\!=\!\argmax_{\theta} 
\mathbb{E}_{\user \sim \trainusers} \notag
\bigg[ 
&\mathbb{E}_{\substack{\contextset^\user \sim \allcleanvideos^\user\\ \targetset^\user \sim \allcluttervideos^\user}}\!
\Big[ 
\mathbb{E}_{(\cluttervideo, \cluttervideolabel) \sim \targetset^\user}
\big[\\
&\!\!\sum\limits_{\cluttervideoframe_f \in \cluttervideo}\!\! \log p_{\theta^\user} (\clutterframelabel_f\! \given\! \cluttervideoframe_{f},\, \contextset^\user) 
\big] 
\!\Big]
\!\bigg]
\end{align}
where \(p_{\theta^\user} (\clutterframelabel_f \given v_{f},\, \contextset^\user)\) is the probability the user's personalized recognition model \(f_{\theta^\user}\) assigns to the ground-truth object label for frame \(v_f\).
The parameters are typically updated with \gls{SGD} and a cross entropy loss function.

The resulting model \(f_{\theta^*}\) can be thought of as knowing \textit{how} to personalize to a user.
That is, for a completely new test user \(\user \sim \testusers\) in the real-world, \(\theta^*\) can rapidly be updated to the user's personalized parameters \(\theta^\user\) with only the context videos  \(\contextset^\user\) of that user's personal objects.
The user can then scan \(f_{\theta^\user}\) around any new target scenario to identify their objects using~\cref{eq:frame-predict}.

\subsection{Non-episodic few-shot learning}
\label{app:sec:nonepisodic-few-shot}

A non-episodic approach~\cite{yosinski2014transferable,tian2020rethinking,chen2020new} is characterized by training the model in a standard batch-wise manner.
For a train user \(\user \sim \trainusers\), a batch \(\mathcal{B}^\user\) is the union of the user's context and target videos from~\cref{eq:contexttargetset}:
\begin{align}\label{eq:batch}
\mathcal{B}^\user = \contextset^\user \cup \targetset^\user = \{(\cluttervideo, p)_i\}_{i=1}^{N+M}   
\end{align}
where \(\cluttervideo\) is a context or target video, \(\cluttervideolabel \in \objectset^\text{all}\) is the object label, and \(\objectset^\text{all}\) is the set of personal objects pooled across all users in \(\trainusers\).

A recognition model \(f\) is trained over many randomly sampled batches per train user \(\user \sim \trainusers\), typically by maximizing the likelihood of predicting the user's correct personal object in each frame in a target video \(\cluttervideoframe_f \in \cluttervideo\):
\begin{align}\label{eq:non-episodic-theta-star}
\theta^*\!\! =\! \argmax_{\theta} \notag
\mathbb{E}_{\user \sim \trainusers}\! 
\bigg[
\mathbb{E}_{\substack{\contextset^\user \sim \allcleanvideos^\user\\ \targetset^\user \sim \allcluttervideos^\user}}\!
\Big[&
\mathbb{E}_{(\cluttervideo, p) \sim \mathcal{B}^\user}\!
\big[\\
&\!\!\sum\limits_{\cluttervideoframe_f \in \cluttervideo}\!\! \log p_{\theta} (\clutterframelabel_f\!\given\!\cluttervideoframe_{f}) 
\big] 
\!\Big]
\!\bigg]
\end{align}
where \(p_{\theta} (\clutterframelabel_f \given \cluttervideoframe_{f})\) is the probability the (generic) model \(f_{\theta}\) assigns to the ground-truth label \(y_f \in \objectset^\text{all}\) for frame \(\cluttervideoframe_f\).
As above, typically \gls{SGD} and a cross entropy loss are used.

Unlike an episodic approach, the resulting \(f_{\theta^*}\) is a generic recognizer for all the objects in \(\objectset^\text{all}\) and does not explicitly know how to personalize.
Instead, the generic model is personalized to a new test user \(\user \sim \testusers\) by fine-tuning \(f_{\theta^*}\) at test-time with the user's context videos \(\contextset^\user\) of their personal objects.
This is equivalent to transfer learning where the (generic) feature extractor is typically frozen and only a new linear classification layer is learned per user for their objects -- together giving the user's personalized parameters \(\theta^\user\).
As above, the user can then scan \(f_{\theta^\user}\) around to identify their objects in target scenarios using~\cref{eq:frame-predict}.

\section{Implementation Details}
\label{app:sec:implementation-details}

\subsection{Baselines}
\label{app:sec:baselines}

\noindent\textbf{ProtoNets~\cite{snell2017prototypical}.}
ProtoNets is a metric-based few-shot learner which classifies a target frame in a task based on the shortest distance (typically squared Euclidean) between its embedding and the mean embedding of each object class computed using the task's context set.
We use a ResNet18~\cite{he2016deep} feature extractor pre-trained on ILSVRC~\cite{ILSVRC15}.
We then train the model episodically on \(T^\text{train}\) tasks per train user (drawn from \(\trainusers\)) for 20 epochs with a learning rate of \(10^{\text{-}4}\) (reduced by \(0.1\) for the feature extractor).
Note, each epoch samples \(T^\text{train}\) \emph{new} tasks per train user.
We evaluate the trained model on all target videos per test user in \(\testusers\) (repeated for \(T^\text{test}\) tasks per user), and report the metrics over \(\targetset^\text{all}\), the flattened set of all target videos across all tasks for all test users.

\noindent\textbf{CNAPs~\cite{requeima2019fast}.}
CNAPs is an amortization-based few-shot learner which classifies a target frame in a task with a model that has been \textit{generated} by forward passing the task's context set through a hypernet.
In practice, the hypernet generates only a subset of the model's parameters following~\cite{requeima2019fast} -- the task's classifier, and \textit{FiLM} adapters~\cite{perez2018film} which are placed after convolution blocks in the feature extractor.
We use a ResNet18~\cite{he2016deep} feature extractor that is pre-trained on ILSVRC~\cite{ILSVRC15} and kept frozen.
We train the hypernet episodically on \(T^\text{train}\) tasks per train user (drawn from \(\trainusers\)) for 10 epochs with a learning rate of \(10^{\text{-}4}\).
Note, each epoch samples \(T^\text{train}\) \emph{new} tasks per train user.
We evaluate the trained model on all target videos per test user in \(\testusers\) (repeated for \(T^\text{test}\) tasks per user), and report the metrics over \(\targetset^\text{all}\).

\noindent\textbf{MAML~\cite{finn2017model}.}
MAML is an optimization-based few-shot learner which classifies a target frame in a task with a model after it has taken \(G\) gradient steps on the task's context set.
We use a ResNet18~\cite{he2016deep} feature extractor that is pre-trained on ILSVRC~\cite{ILSVRC15}.
We then train the model episodically on \(T^\text{train}\) tasks per train user (drawn from \(\trainusers\)) for 10 epochs with the Adam optimizer~\cite{kingma2014adam} and a learning rate of \(10^{\text{-}5}\) for the outer loop, and the \gls{SGD} optimizer and a learning rate of \(10^{\text{-}3}\) for the inner loop (rates reduced by \(0.1\) for the feature extractor in both loops).
Note, each epoch samples \(T^\text{train}\) \emph{new} tasks per train user.
We evaluate the trained model on all target videos per test user in \(\testusers\) (repeated for \(T^\text{test}\) tasks per user), and report the metrics over \(\targetset^\text{all}\).
Note that since each (train and test) task can have a different number of objects, we append a new classifier per task, initialized with zeroes following~\cite{triantafillou2019meta}.
The classifier is therefore not learned but instead the result after \(G\) gradient steps.
At both train and test time, \(G=15\).

\noindent\textbf{FineTuner~\cite{tian2020rethinking}}.
FineTuner is a non-episodic few-shot learner which classifies a target frame in a task with a model that has been finetuned using the task's context set, typically with its feature extractor frozen.
This is equivalent to a transfer learning approach.
We use a ResNet18~\cite{he2016deep} feature extractor pre-trained on ILSVRC~\cite{ILSVRC15}.
Following~\cite{tian2020rethinking}, we first train the extractor using \(\trainusers\): we pool the personal objects across \textit{all} train users in \(\trainusers\) and train a standard classification model\footnote{Because there are similar objects across the train users, we use the objects' cluster labels (i.e. 100-way classification task), as in~\cref{fig:benchmark-cluster-histo-train}} for 10 epochs using \(T^\text{train}\) tasks per train user.
Note, each epoch samples \(T^\text{train}\) \emph{new} tasks per train user.
Because the model is non-episodic, the context and target videos of each task are pooled and treated as a standard batch, following~\cref{eq:batch}. 
We use the Adam optimizer~\cite{kingma2014adam} and a learning rate of \(10^{\text{-}4}\) (reduced by \(0.1\) for the feature extractor).
At test time, this feature extractor is frozen, and for each test user drawn from \(\testusers\), a new classifier is appended, initialized with zeroes following~\cite{triantafillou2019meta} and tuned for \(G=50\) gradient steps on the user's context videos, using \gls{SGD} and a learning rate of \(0.1\).
We evaluate the trained model on all target videos per test user in \(\testusers\) (repeated for \(T^\text{test}\) tasks per user), and report the metrics over \(\targetset^\text{all}\).

\subsection{Sampling tasks} 
\label{app:sec:task-sampling}

The only strict requirement of the ORBIT evaluation protocol is that a model outputs a prediction for every frame in every target video for every test user.
How tasks are sampled from context/target videos during training, and context videos during testing is otherwise flexible.
In this section, we detail how we sample tasks, but acknowledge that other strategies could be chosen.
\vspace{-1em}
\paragraph{Train user tasks}
We sample \(T^\text{train}\) tasks (per epoch) for each train user \(\user \sim \trainusers\) as follows:
\begin{compactenum}
    \setstretch{1.2}
    \item \textbf{Way}: sample a set of the user's objects \(\ddot{\objectset}^\user \sim \objectset^\user\)
    \item \textbf{Shot}: for each object \(\cluttervideolabel \in \ddot{\objectset}^\user\), sample \(n_p\!\sim\![1, \dots, N^\user_\cluttervideolabel] \) and \(m_p \sim [1, \dots, M^\user_\cluttervideolabel] \)\footnote{\label{fn:setsizes}Note, \(N\) and \(M\) in~\cref{eq:contexttargetset} and~\cref{eq:batch} are \(N = \sum_{p \in \ddot{\objectset^\user}} n_p\) and \(M = \sum_{p \in \ddot{\objectset^\user}} m_p\), respectively.}
    \item Construct context set \(\contextset^\user = \{ (\cleanvideo, \cleanvideolabel) \overset{\scriptscriptstyle n_p}{\sim} \allcleanvideos^\user_p \given \forall p \in \ddot{\objectset}^\user\}\)
    \item Construct target set \(\targetset^\user = \{ (\cluttervideo, \cluttervideolabel) \overset{\scriptscriptstyle m_p}{\sim} \allcluttervideos^\user_p \given \forall p \in \ddot{\objectset}^\user\}\)
\end{compactenum}
For memory reasons, during training we impose a video cap per object of \(n_p = 5\) and \(m_p = 4\) if \(|\ddot{\objectset^\user}| \geq 6\) (otherwise the caps are doubled) and an object cap of \(|\ddot{\objectset}^\user| =10\). 
\vspace{-1em}
\paragraph{Test user tasks}
We sample \(T^\text{test}\) tasks (per epoch) for each test user \(\user \sim \testusers\) as follows:
\begin{compactenum}
    \setstretch{1.2}
    \item \textbf{Way}: select \textit{all} the user's objects, \(\ddot{\objectset}^\user = \objectset^\user\) (i.e. test personalization for all objects)
    \item \textbf{Shot}: for each object \(\cluttervideolabel \in \ddot{\objectset}^\user\), \(n_p\!=\!N^\user_\cluttervideolabel \) and \(m_p\!=\!M^\user_\cluttervideolabel \) (i.e. use all context and target videos)\footnoteref{fn:setsizes}
    \item Construct context set \(\contextset^\user = \{ (\cleanvideo, \cleanvideolabel) \overset{\scriptscriptstyle n_p}{\sim} \allcleanvideos^\user_p \given \forall p \in \ddot{\objectset}^\user\}\) 
    \item Construct target set \(\targetset^\user = \{ (\cluttervideo, \cluttervideolabel) \overset{\scriptscriptstyle m_p}{\sim} \allcluttervideos^\user_p \given \forall p \in \ddot{\objectset}^\user\}\) 
\end{compactenum}
For testing, no caps are imposed on the number of videos per object, or number of objects.

\subsection{Task sampling hyper-parameters}
\label{app:sec:task-hyperparameters}
The number of tasks sampled per train user (per epoch) was selected over a range of 5 to 500 (see~\cref{app:tab:num-train-tasks-analysis}).
The number of tasks sampled per test (and validation) user was selected to reduce variance in the reported metrics.
The number of clips per video and the clip length were selected based on available \textsc{gpu} memory during training/testing (\(2\times\) Dell NVIDIA Tesla V100 32GB \textsc{gpu}s).
\begin{compactitem}
\item Tasks per train user, \(T^\text{train}=50\)
\item Tasks per test (and validation) user, \(T^\text{test}=5\)
\item Clip length, \(L=8\) frames
\item Clips per context video in a train task, \(S_\contextset^\text{train}=4\)
\item Clips per target video in a train task, \(S_\targetset^\text{train}=4\)
\item Clips per context video in a test task, \(S_\contextset^\text{test} = 8\)
\end{compactitem}

\subsection{Optimization hyper-parameters} The learning rates were selected with a grid search over the range \(10^{\text{-}5}\) to \(10^{\text{-}1}\).
Unless otherwise specified, for all baselines we use the Adam optimizer~\cite{kingma2014adam} with default parameters \(\beta\)=(0.9, 0.999), \(\epsilon=10^{\text{-}8}\), and no weight decay.

\section{Extended analyses}
\label{app:sec:extended-analysis}

\noindent Here we extend analyses from~\cref{sec:evaluation:analysis} in the main paper.

\vspace{-1em}
\paragraph{Meta-training on other few-shot learning datasets.}
\Cref{app:tab:metadataset-test} shows the results when meta-training on Meta-Dataset and meta-testing on the ORBIT test users, for each of the baseline models (extends~\cref{tab:metadataset-test} in the main paper).

\vspace{-1em}
\paragraph{User-centric versus user-agnostic training.}
The baselines are trained in a user-centric manner whereby each task is sampled from only one train user's objects.
We compare this to a user-agnostic training regime in~\cref{tab:mixed-user} where we pool objects across \emph{all} train users in \(\trainusers\) and construct train tasks by randomly sampling objects from this pool (keeping all other sampling procedures the same).
This is akin to typical few-shot learning pipelines.
The performance of both regimes is overall equivalent, suggesting that imposing a user-centric hierarchy on task sampling does not negatively impact performance, relative to imposing no hierarchy.

\begin{figure*}[t]
    \centering
    \vspace{-4em}
    \begin{subfigure}{\textwidth}
    \centering
    \includegraphics[width=0.32\textwidth]{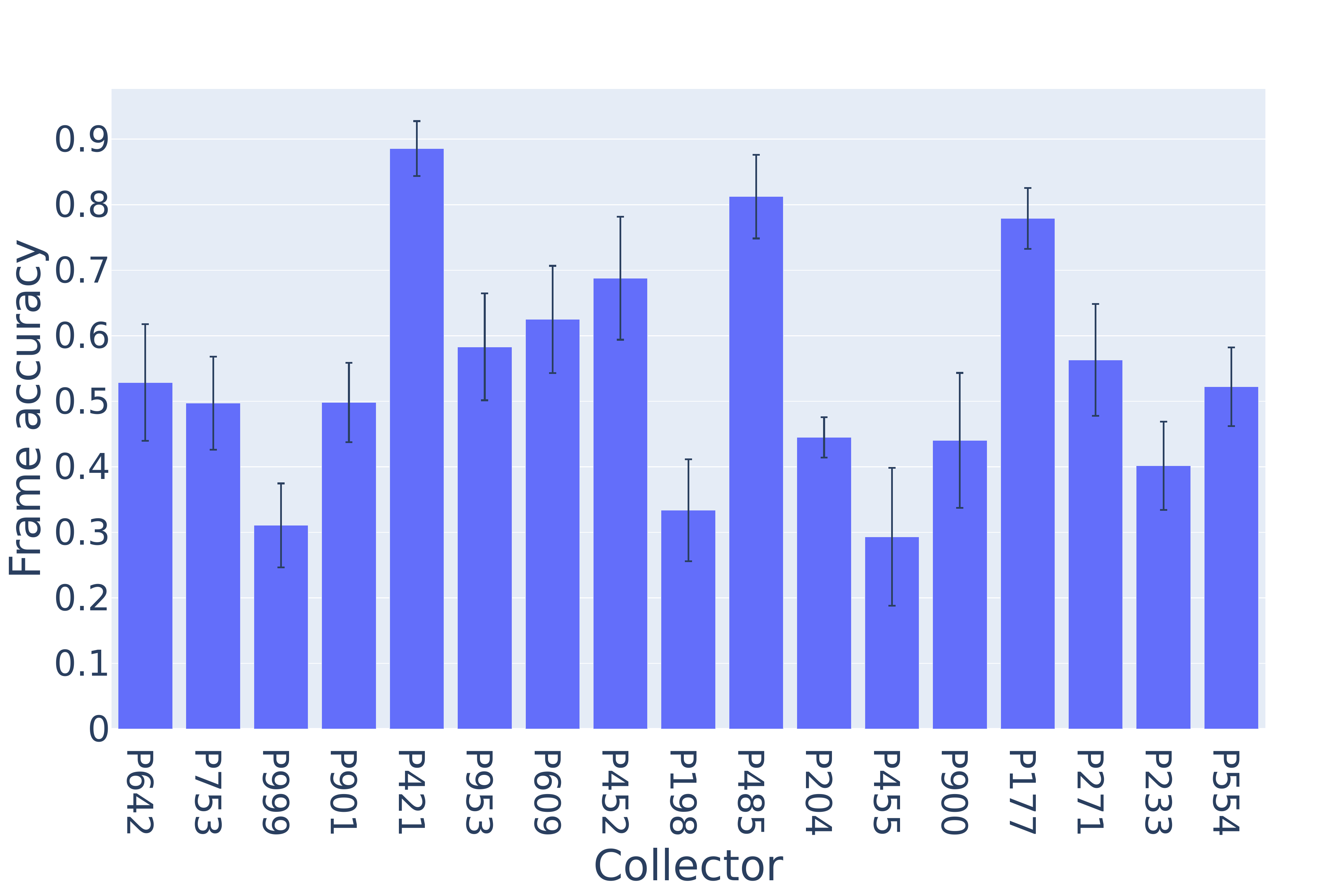}
    \includegraphics[width=0.32\textwidth]{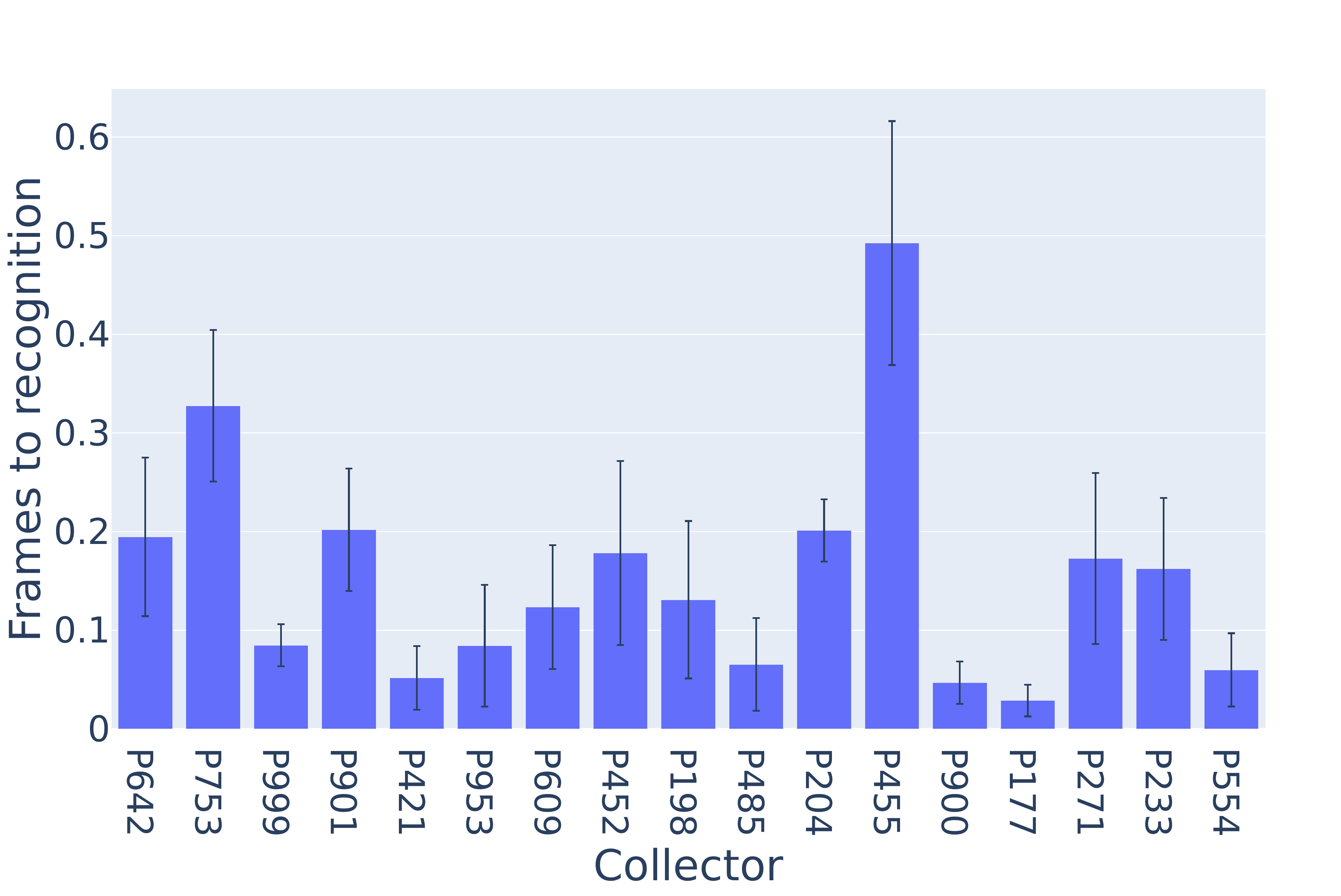}
    \includegraphics[width=0.32\textwidth]{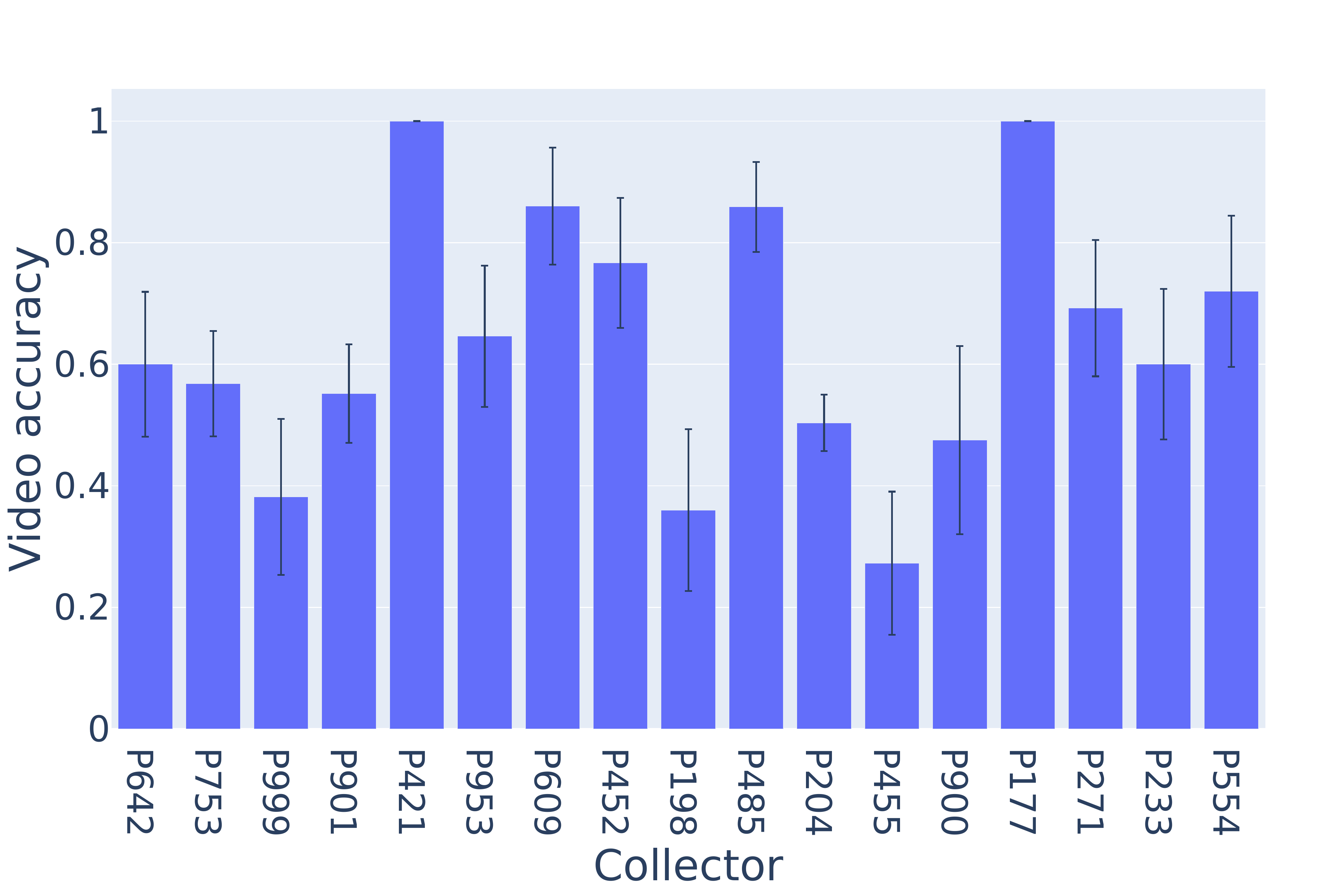}
    \caption{Results with CNAPs~\cite{requeima2019fast}.}
    \label{fig:orbitbaselines-per-user-cnaps}
    \end{subfigure}
    \begin{subfigure}{\textwidth}
    \centering
    \includegraphics[width=0.32\textwidth]{figures/frame_acc_per_user_orbit_cluve_protonets_resnet18_84.pdf}
    \includegraphics[width=0.32\textwidth]{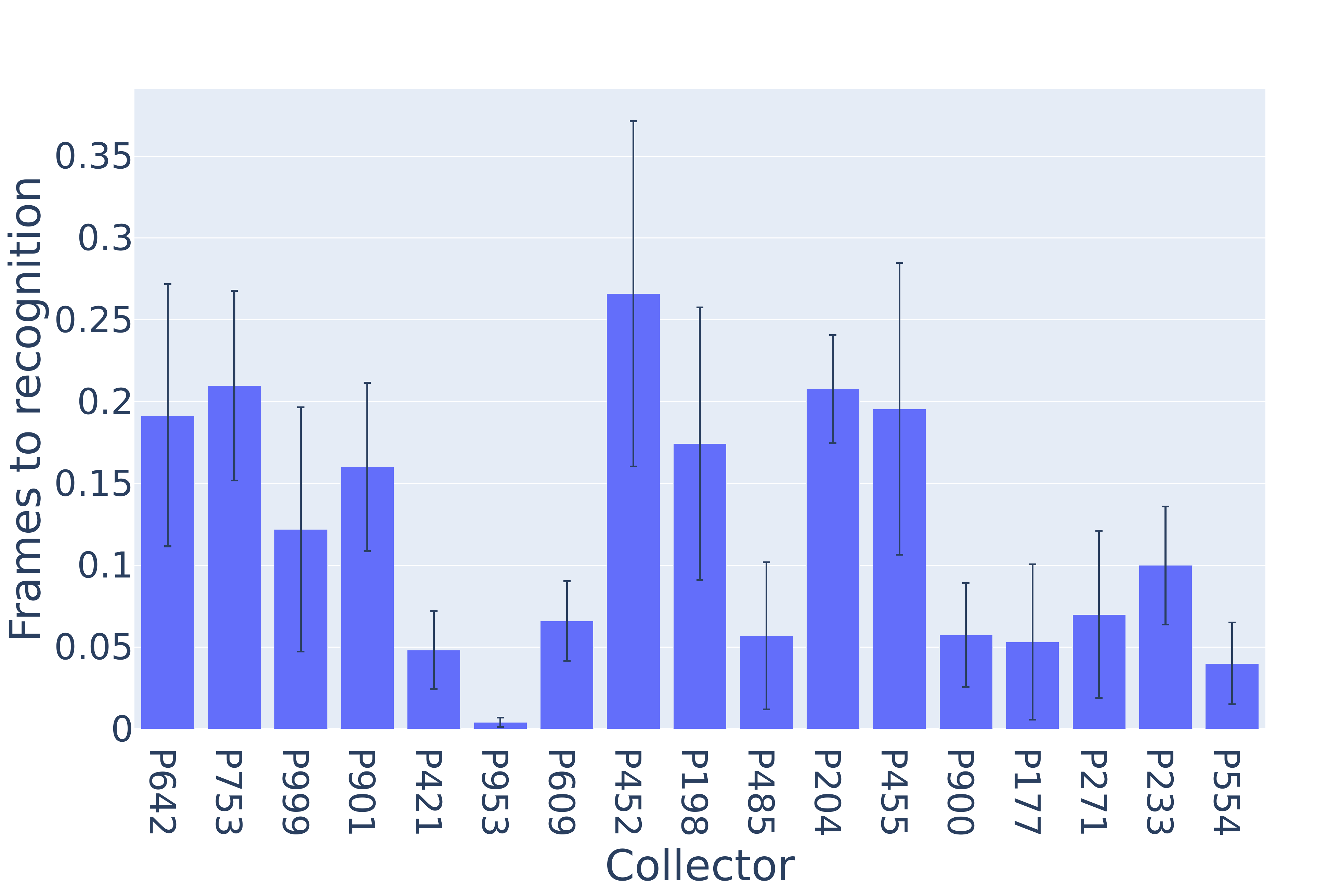}
    \includegraphics[width=0.32\textwidth]{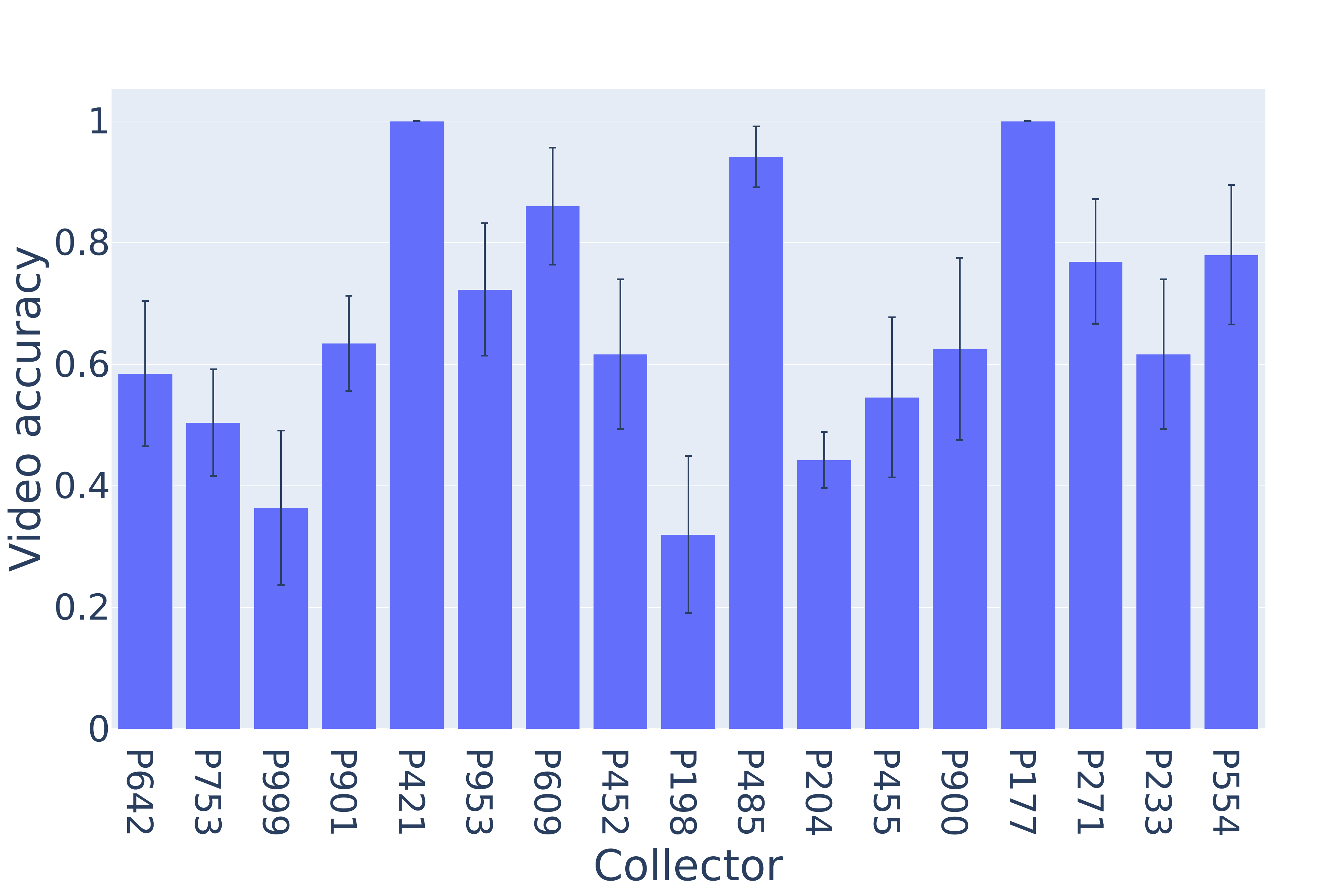}
    \caption{Results with ProtoNets~\cite{snell2017prototypical}.}
    \label{fig:orbitbaselines-per-user-protonets}
    \end{subfigure}
    \begin{subfigure}{\textwidth}
    \centering
    \includegraphics[width=0.32\textwidth]{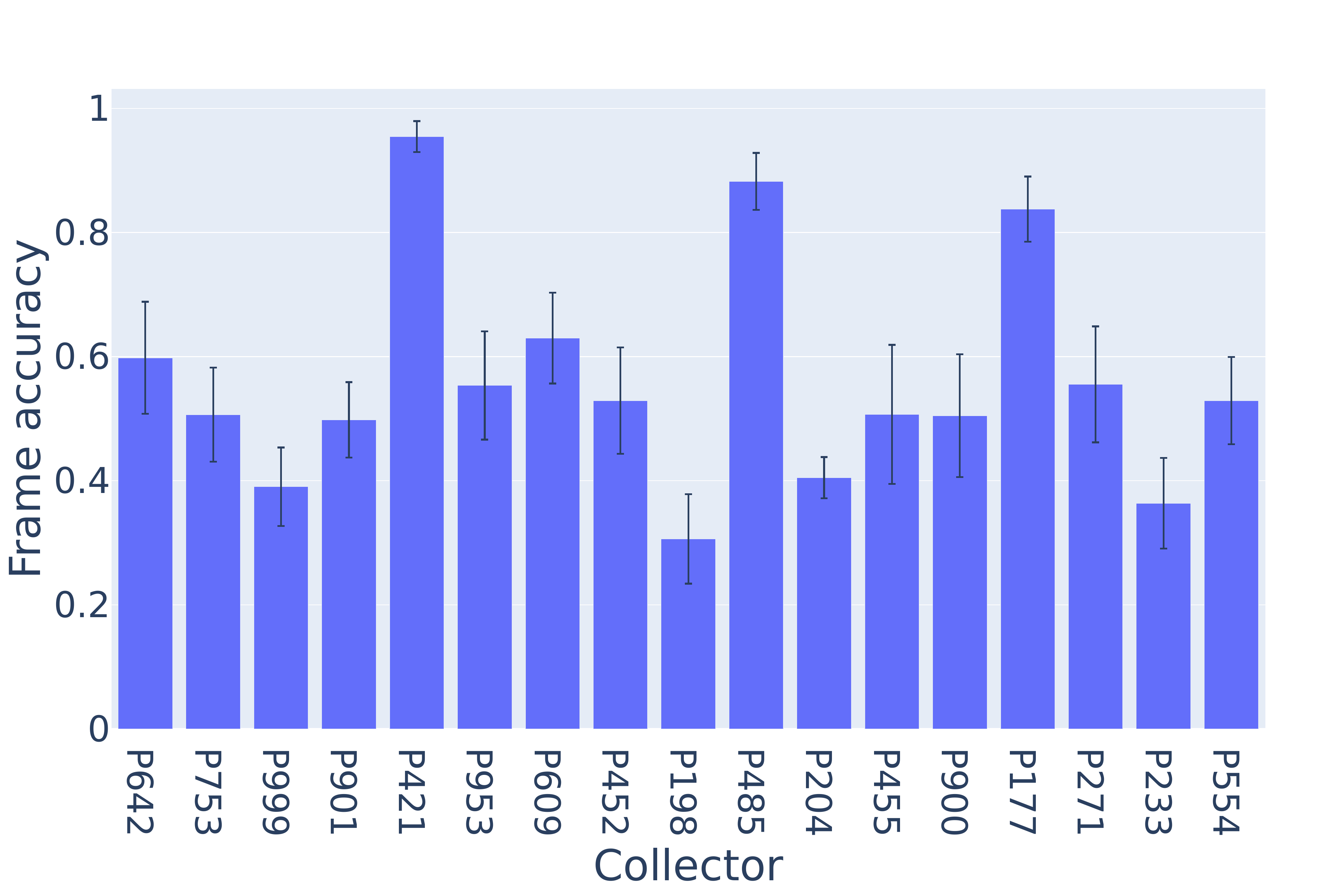}
    \includegraphics[width=0.32\textwidth]{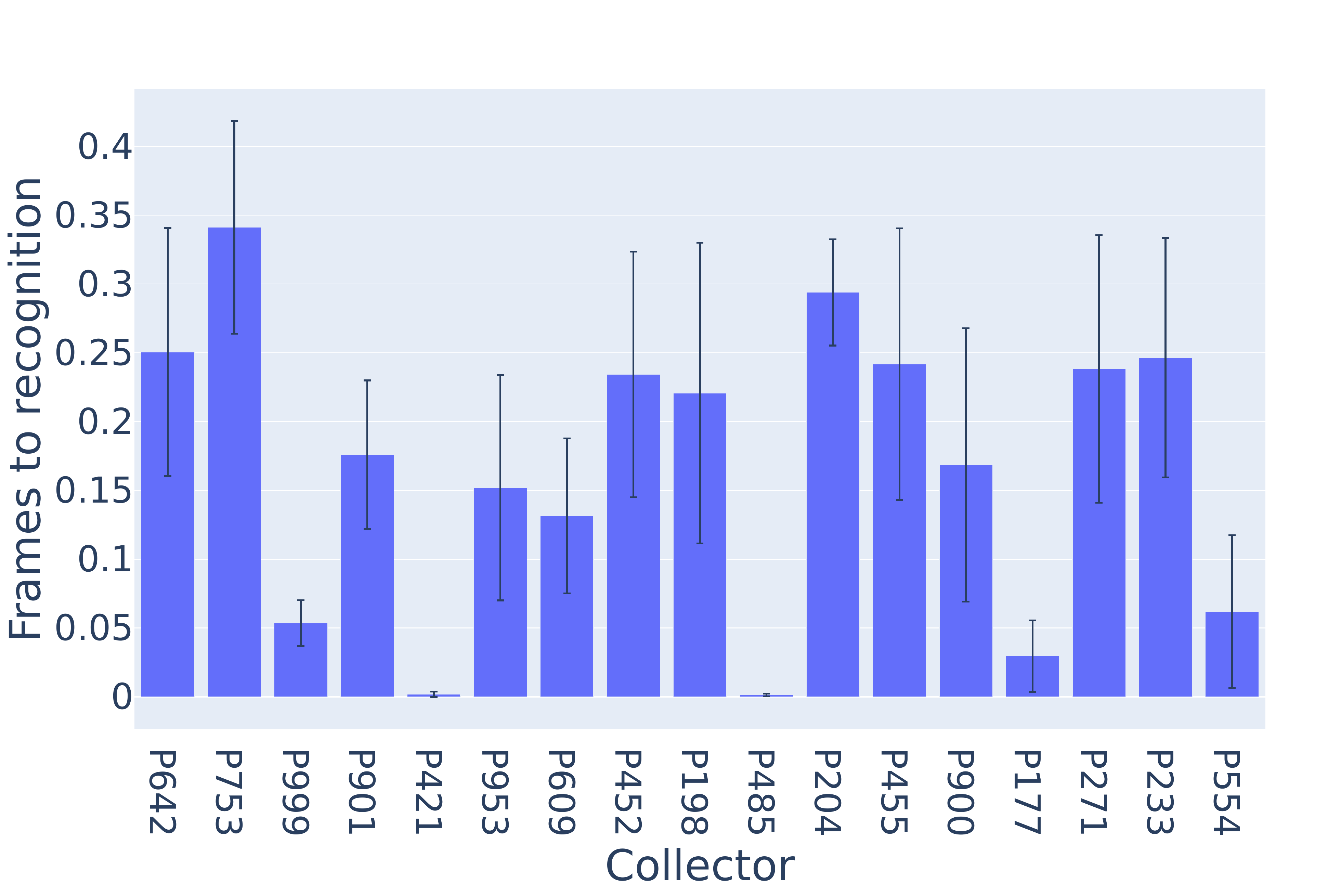}
    \includegraphics[width=0.32\textwidth]{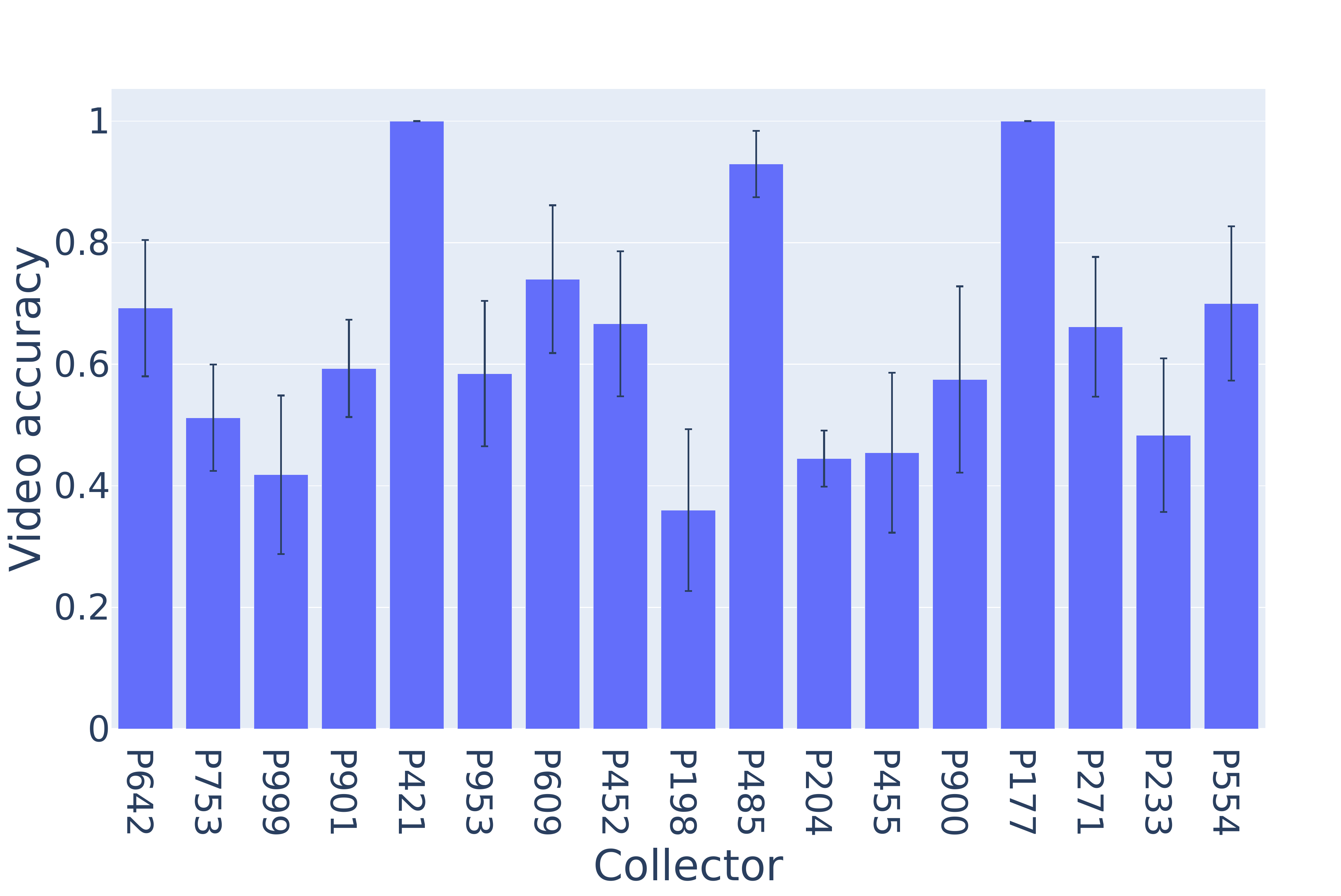}
    \caption{Results with MAML~\cite{finn2017model}.}
    \label{fig:orbitbaselines-per-user-maml}
    \end{subfigure}
    \begin{subfigure}{\textwidth}
    \centering
    \includegraphics[width=0.32\textwidth]{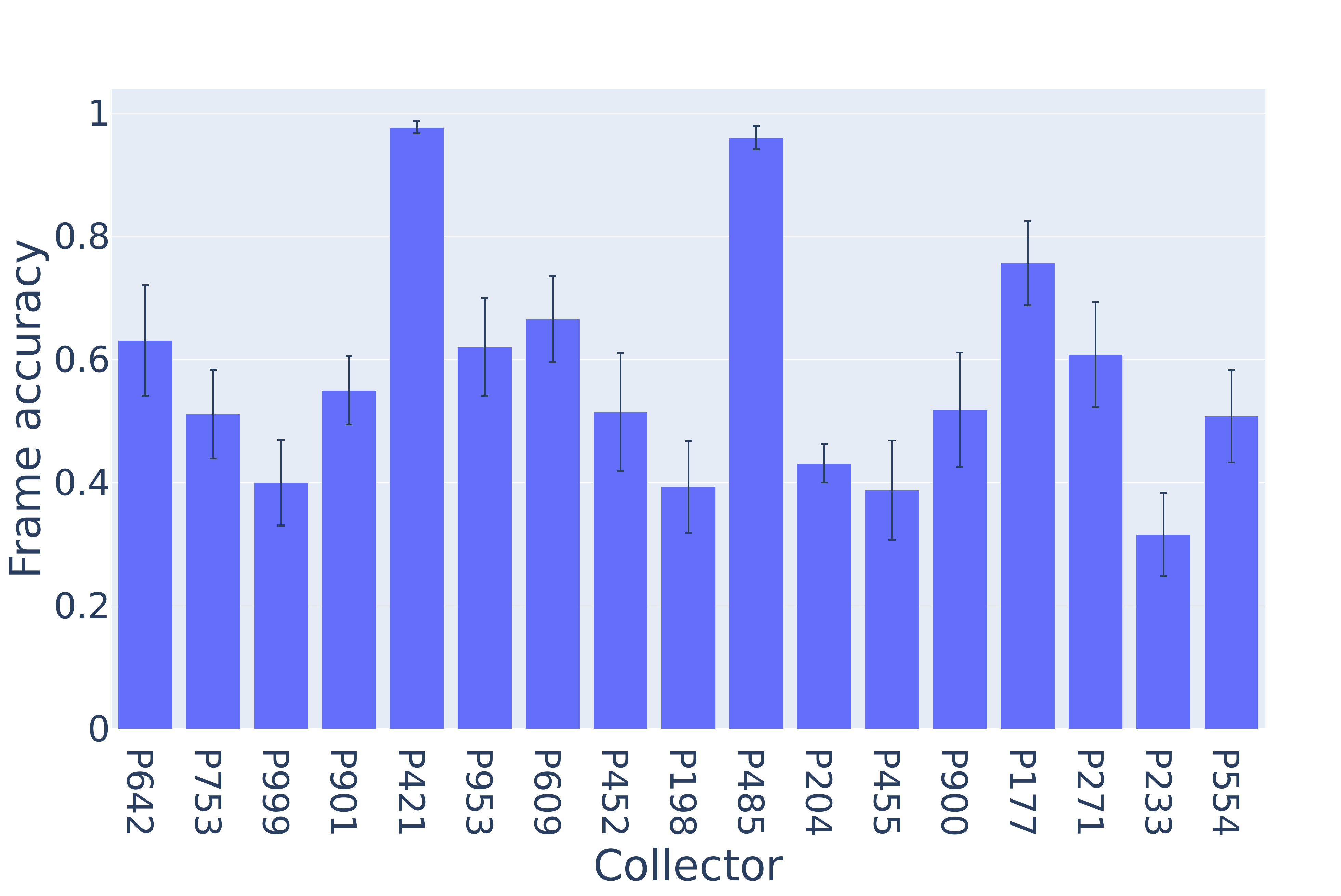}
    \includegraphics[width=0.32\textwidth]{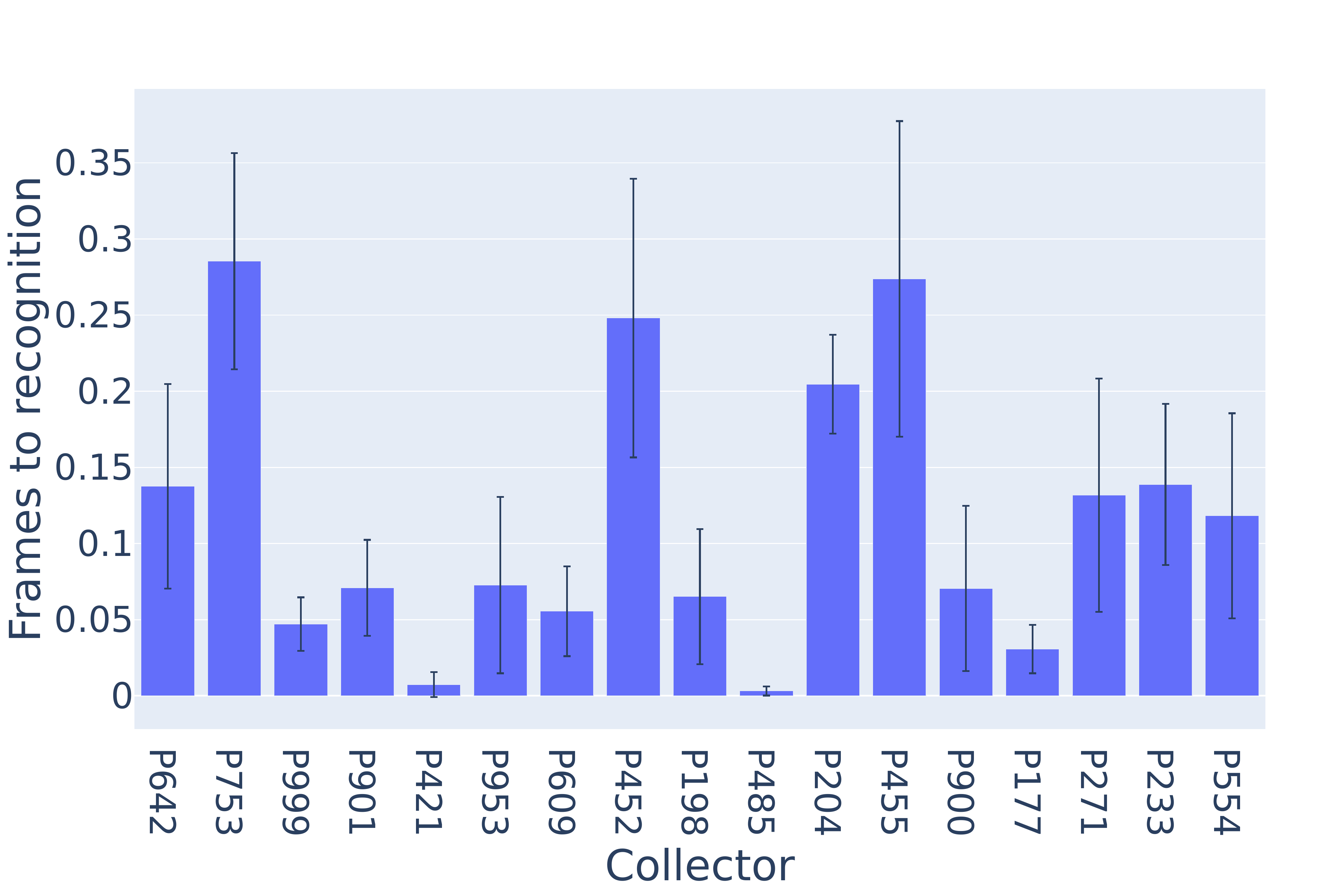}
    \includegraphics[width=0.32\textwidth]{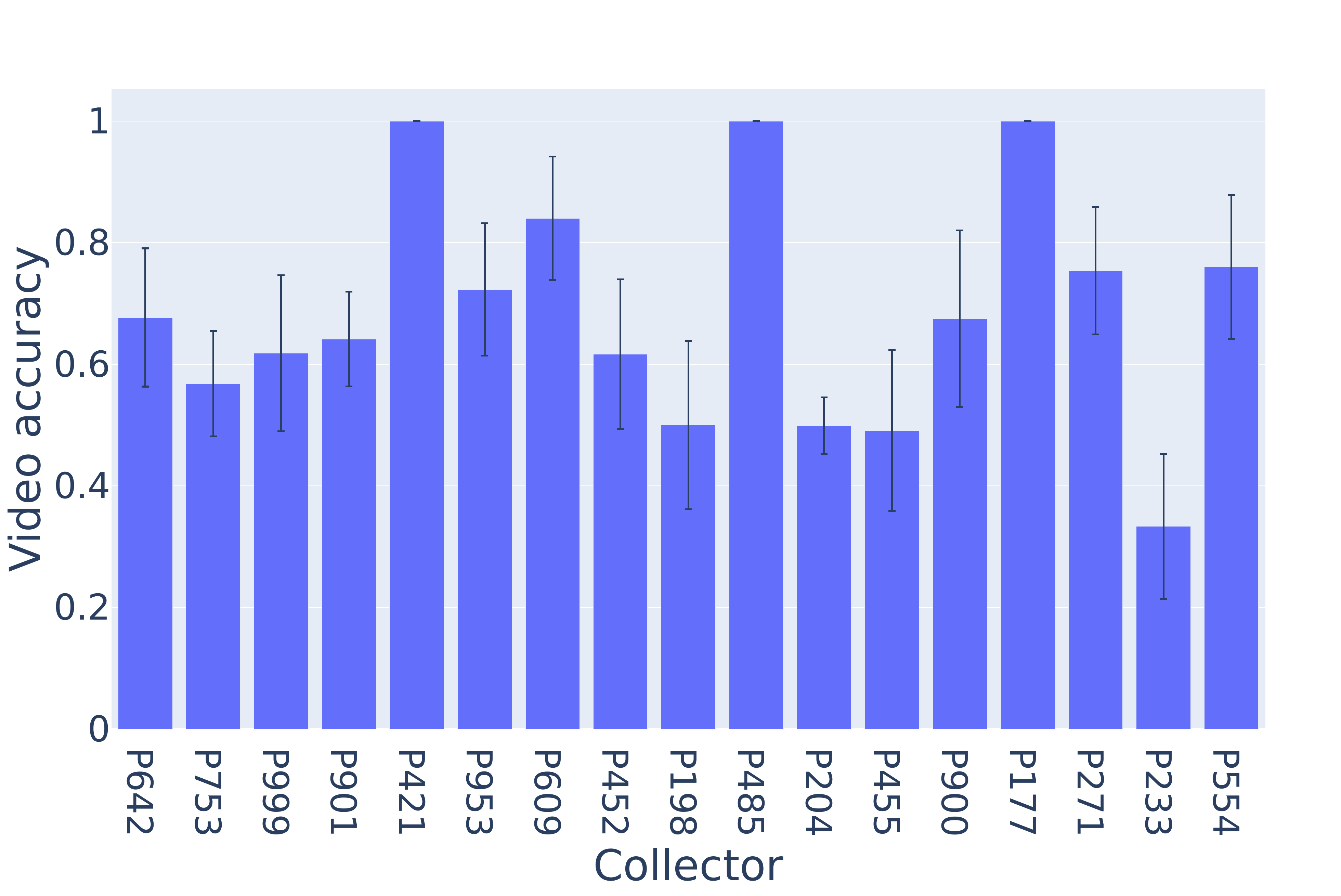}
    \caption{Results with FineTuner~\cite{tian2020rethinking}.}
    \label{fig:orbitbaselines-per-user-finetuner}
    \end{subfigure}
    \vspace{-0.5em}
    \caption{\textsc{clu-ve} performance varies widely across test users, for all metrics, across all baseline models. Error bars are 95\% confidence intervals.}
    \label{fig:orbit-baselines-per-user-all}
\end{figure*}

\vspace{-1em}
\paragraph{Per-user performance.}
Extending~\cref{fig:orbit-baselines-per-user} in the main paper, we plot \textsc{clu-ve} performance across the full suite of metrics (frame accuracy, frames-to-recognition and video accuracy) per user across all baseline models in~\cref{fig:orbit-baselines-per-user-all}.
Here, we see the same trend as the main paper --- personalization performance varies widely across users. 

\begin{table*}[!ht]
    \setlength\tabcolsep{3pt}
    \centering
    \scalebox{0.85}{
    \begin{tabular}{l|P{2cm}P{2cm}P{2cm}|P{2cm}P{2cm}P{2cm}}
    & \multicolumn{3}{c}{\textbf{Clean Video Evaluation (\textsc{cle-ve})}} & \multicolumn{3}{c}{\textbf{Clutter Video Evaluation (\textsc{clu-ve})}} \\
    \cmidrule{2-7}
    \textsc{\textbf{model}} & \textsc{\textbf{frame acc}} & \textsc{\textbf{ftr}} & \textsc{\textbf{video acc}} & \textsc{\textbf{frame acc}} & \textsc{\textbf{ftr}} & \textsc{\textbf{video acc}} \\
    \cmidrule{1-7}
    ProtoNets~\cite{snell2017prototypical} & \textbf{58.98 (2.23)} & \textbf{11.55 (1.79)} & \textbf{69.17 (3.01)} & \textbf{46.97 (1.84)} & \textbf{20.42 (1.71)} & \textbf{52.80 (2.53)} \\
    CNAPs~\cite{requeima2019fast} &  51.86 (2.49) & 20.81 (2.33) & 60.77 (3.18) & 41.59 (1.94) & 30.72 (2.13) & 43.00 (2.51) \\
    MAML~\cite{finn2017model} &  42.55 (2.67) & 37.28 (2.99) & 46.96 (3.25) & 24.35 (1.83) & 62.30 (2.34) & 25.73 (2.21)  \\ 
    FineTuner~\cite{tian2020rethinking} & \textbf{61.01 (2.24)} & \textbf{11.53 (1.82)} & \textbf{72.60 (2.91)} & \textbf{48.45 (1.86)} & \textbf{19.13 (1.69)} & \textbf{54.07 (2.52)}  \\
    \end{tabular}}
    \vspace{-0.5em}
    \caption{Models perform poorly when meta-trained on Meta-Dataset and meta-tested on ORBIT test users, suggesting that existing few-shot datasets may be insufficient for real-world adaptation. Results are reported as the average (95\% confidence interval) over all target videos pooled from 85 test tasks (5 tasks per test user, 17 test users).}
    \label{app:tab:metadataset-test}
\end{table*}

\begin{table*}[!ht]
    \setlength\tabcolsep{3pt}
    \centering
    \scalebox{0.85}{
    \begin{tabular}{l|P{2cm}P{2cm}P{2cm}|P{2cm}P{2cm}P{2cm}}
    & \multicolumn{3}{c}{\textbf{Clean Video Evaluation (\textsc{cle-ve})}} & \multicolumn{3}{c}{ \textbf{Clutter Video Evaluation (\textsc{clu-ve})}} \\
    \cmidrule{2-7}
    \textsc{\textbf{model}} & \textsc{\textbf{frame acc}} & \textsc{\textbf{ftr}} & \textsc{\textbf{video acc}} & \textsc{\textbf{frame acc}} & \textsc{\textbf{ftr}} & \textsc{\textbf{video acc}} \\
    \cmidrule{1-7}
    ProtoNets~\cite{snell2017prototypical} & \textbf{65.31 (2.10)} & \textbf{7.44 (1.42)} & \textbf{78.67 (2.67)} & \textbf{49.95 (1.76)} & \textbf{14.87 (1.50)} & \textbf{59.07 (2.49)} \\
    CNAPs~\cite{requeima2019fast} & \textbf{67.21 (2.14)} & \textbf{9.30 (1.59)} & \textbf{79.56 (2.63)} & \textbf{51.26 (1.76)} & \textbf{15.36 (1.54)} & \textbf{60.13 (2.48)} \\
    MAML~\cite{finn2017model} & \textbf{68.98 (2.16)} & \textbf{9.54 (1.68)} & \textbf{79.45 (2.63)} & \textbf{50.95 (1.92)} & 21.67 (1.89) & 55.00 (2.52)  \\
    FineTuner~\cite{tian2020rethinking} & \textbf{69.01 (2.19)} & \textbf{7.56 (1.48)} & \textbf{78.45 (2.68)} & \textbf{53.17 (1.85)} & \textbf{15.32 (1.57)} & \textbf{62.73 (2.45)}  \\
    \end{tabular}}
    \vspace{-0.5em}
    \caption{User-agnostic training on the ORBIT dataset (i.e.\ a train task is sampled from objects pooled across \emph{all} users in \(\trainusers\)) performs as well as user-centric training (i.e.\ a train task is sampled from only one user at a time). Results are reported as the average (95\% confidence interval) over all target videos pooled from 85 test tasks (5 tasks per test user, 17 test users).}
    \label{tab:mixed-user}
\end{table*}
\begin{figure*}
    \centering
        \includegraphics[width=0.32\textwidth]{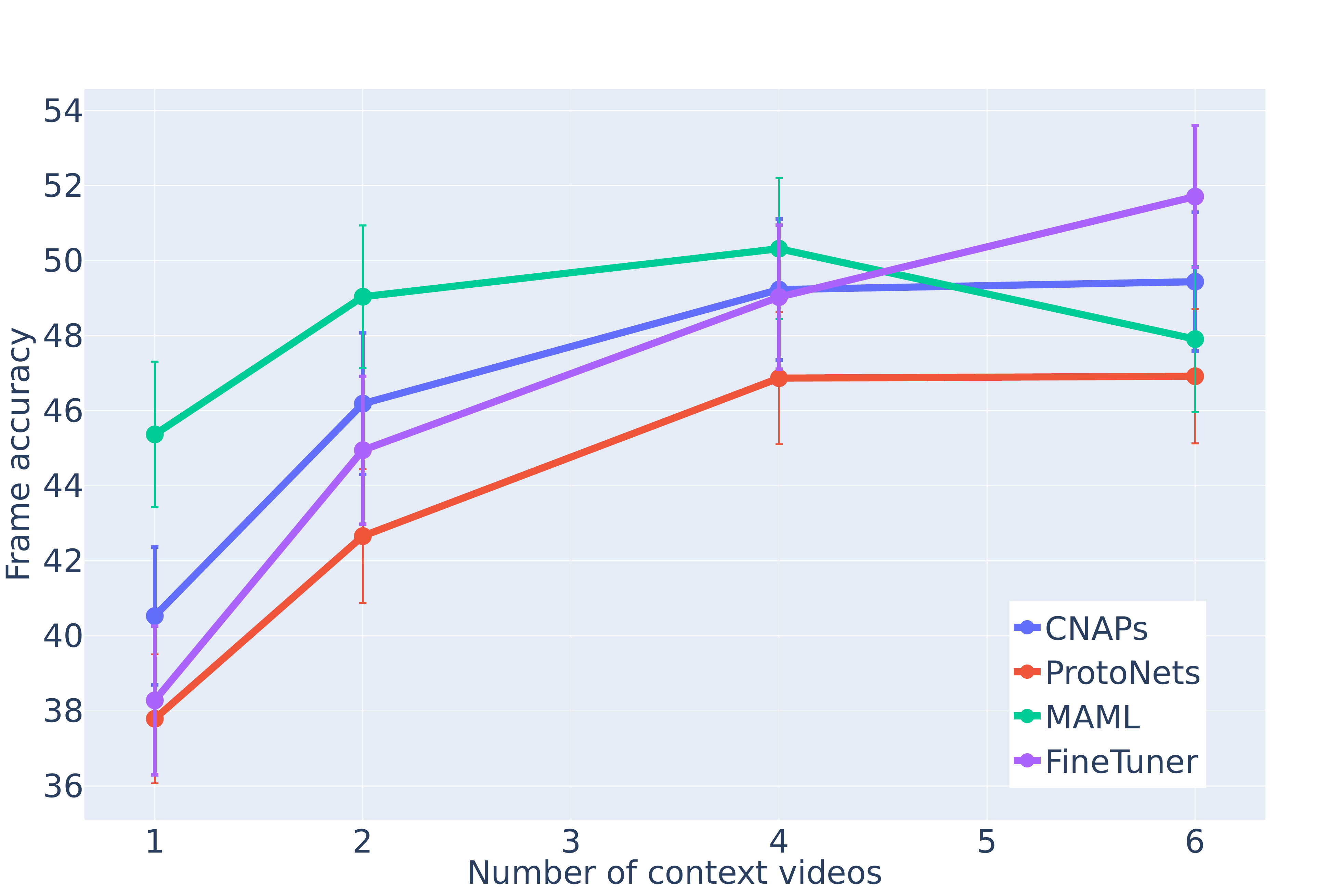}
        \includegraphics[width=0.32\textwidth]{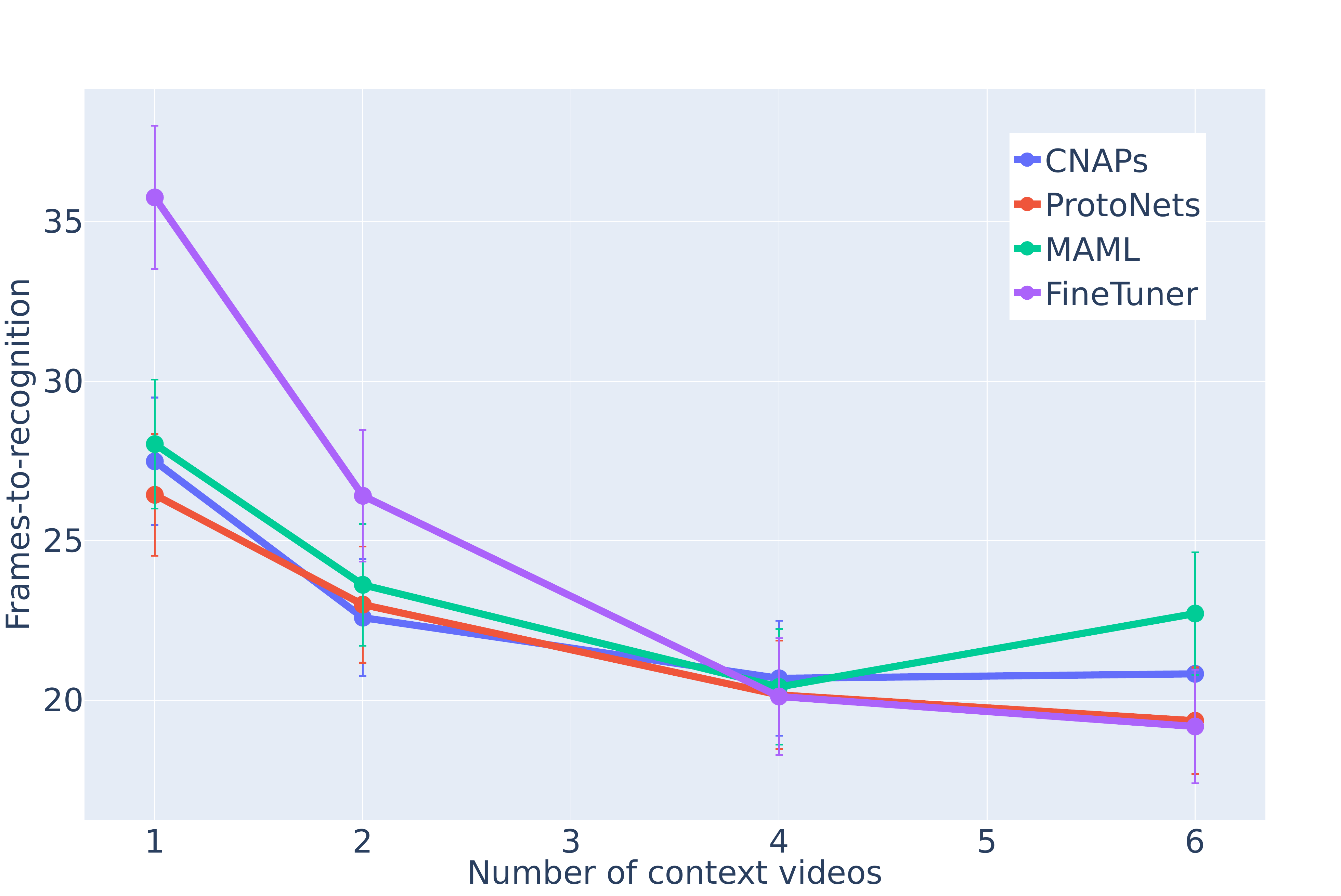}
        \includegraphics[width=0.32\textwidth]{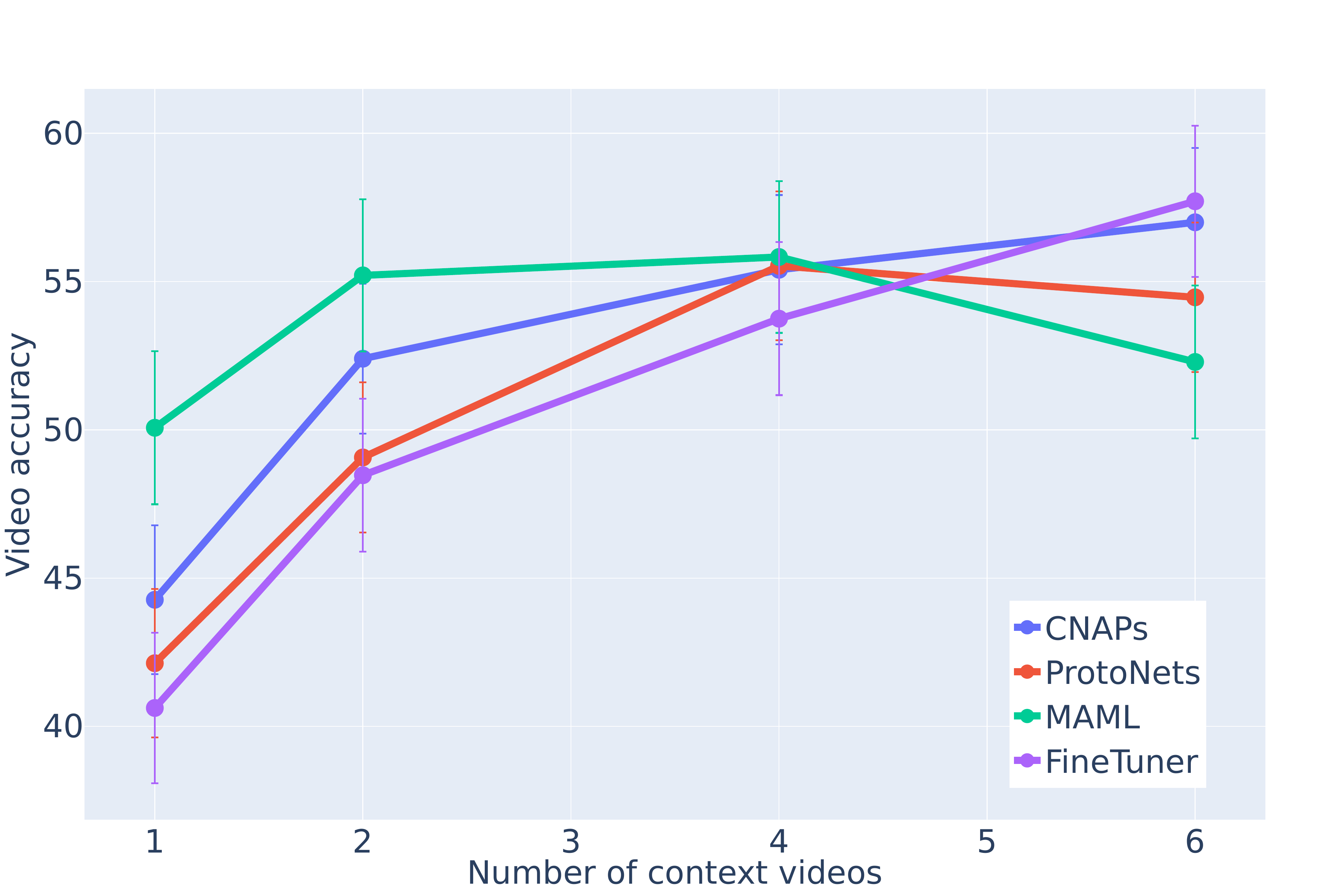}
    \caption{Meta-training with more context videos per object leads to better \textsc{clu-ve} performance (frame accuracy - left; frames-to-recognition - center; video accuracy - right). Frames are sampled from an increasing number of clean videos per object using the number of clips per video (\(S^\text{train}\)) to keep the total number of context frames fixed per train task. Extends~\cref{fig:num-context-vids-analysis} in the main paper.}
    \label{app:fig:num-context-vids-analysis}
\end{figure*}
\begin{figure*}
    \centering
    \includegraphics[width=0.32\textwidth]{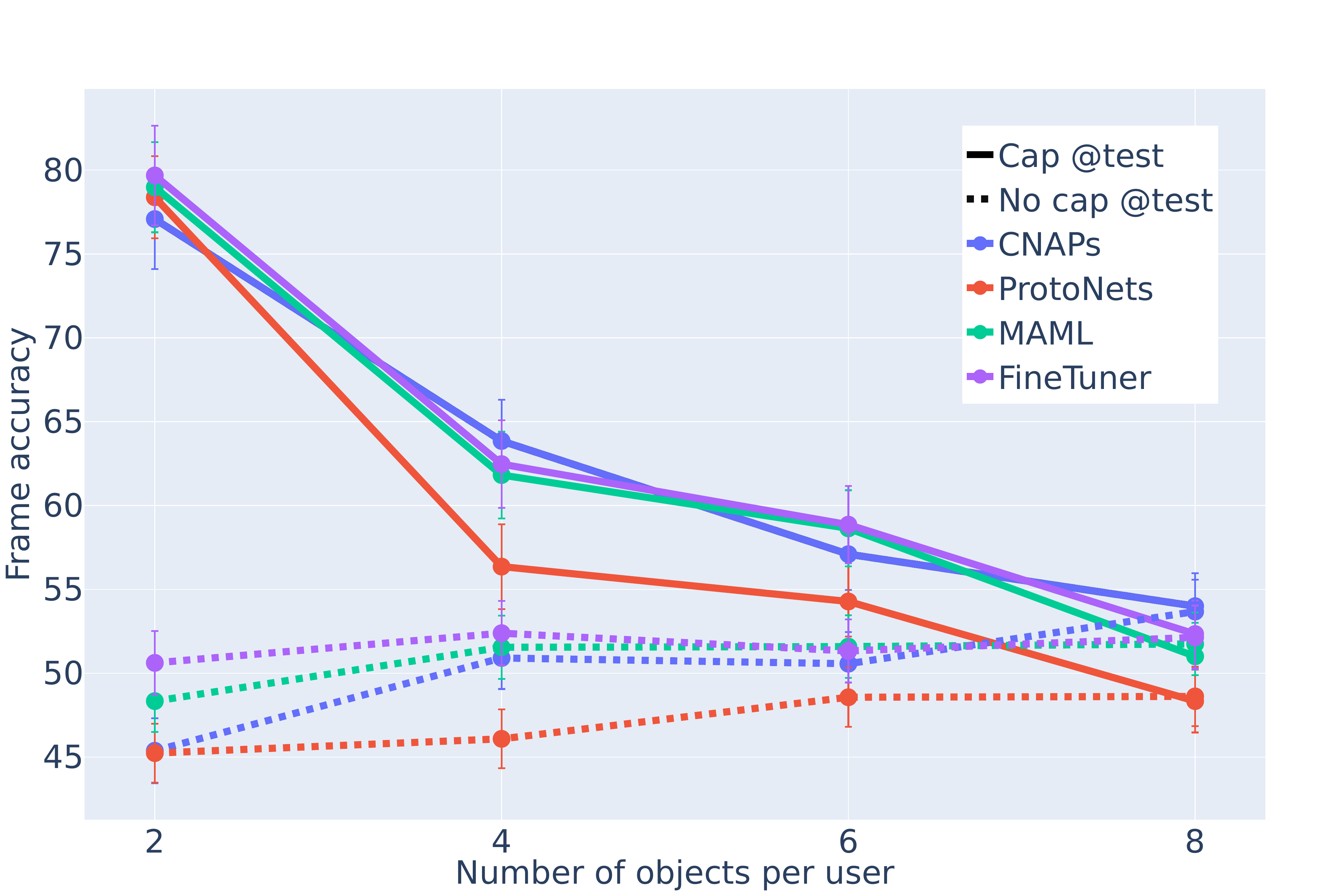}
    \includegraphics[width=0.32\textwidth]{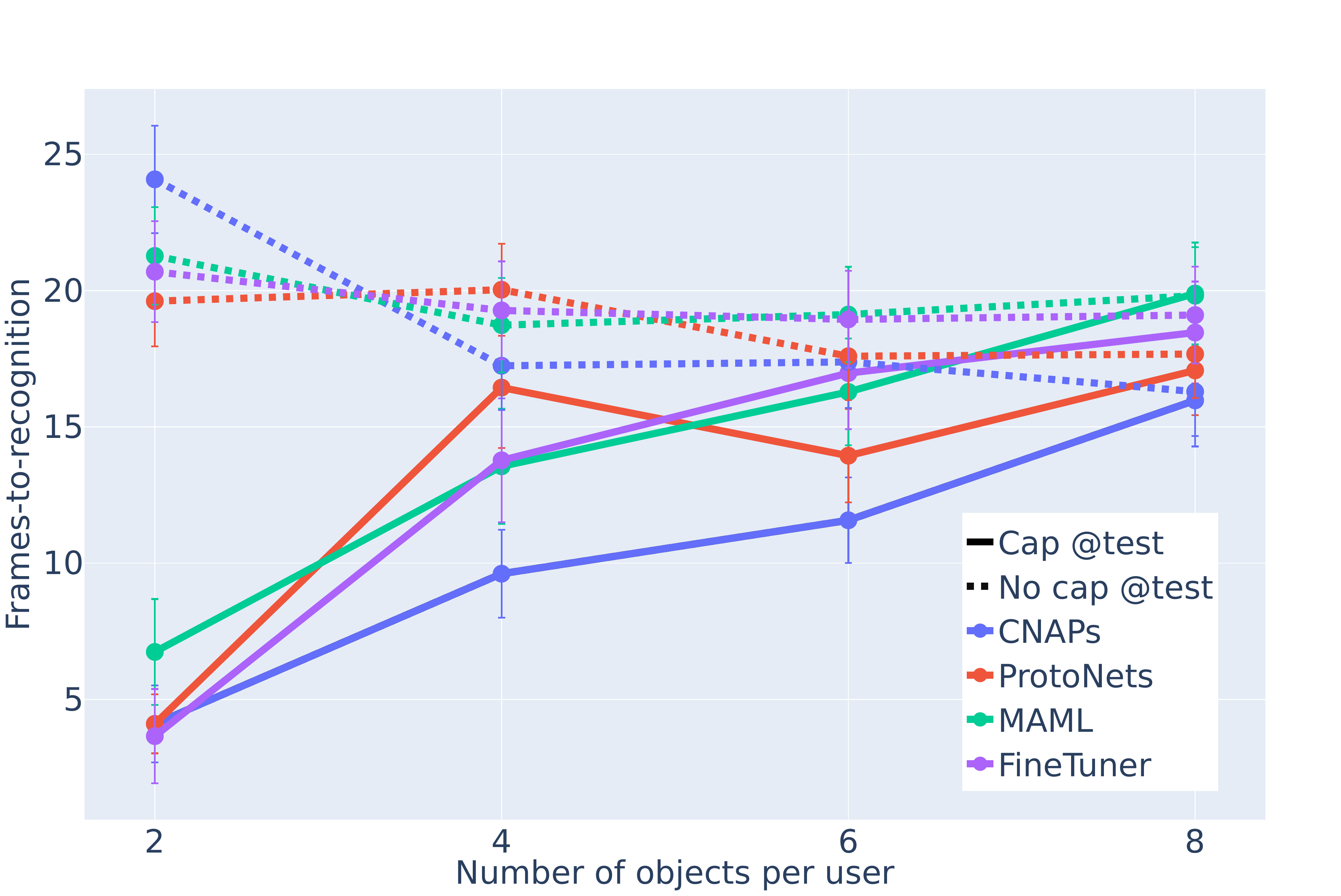}
    \includegraphics[width=0.32\textwidth]{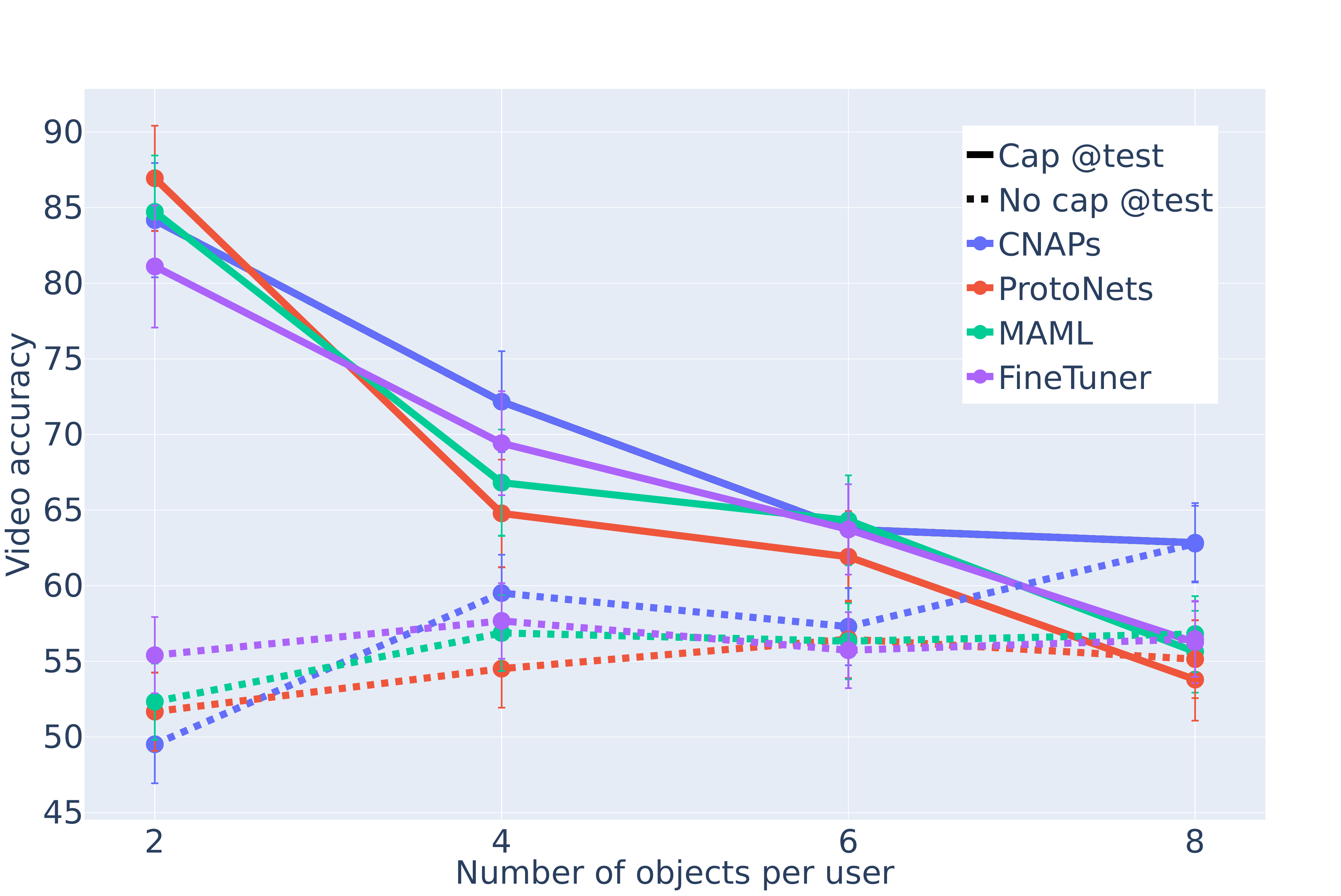}
    \caption{Meta-training and -testing with more objects per user poses a harder recognition problem (solid line), however, meta-training with fewer objects than encountered at meta-testing (dashed line) shows only a small \textsc{clu-ve} performance drop compared to~\cref{tab:orbit-baselines}, suggesting that models may be able to adapt to more objects in the real-world (frame accuracy - left; frames-to-recognition - center; video accuracy - right). Extends~\cref{fig:num-objects-analysis} in the main paper. }
    \label{app:fig:num-objects-analysis}
\end{figure*}

\vspace{-1em}
\paragraph{Train task composition.}
In~\cref{fig:num-context-vids-analysis,fig:num-objects-analysis} in the main paper, we investigate the impact of the number of context videos per object, and the number of objects per user, sampled in train tasks on \textsc{clu-ve} frame accuracy.
We extend these plots to the frames-to-recognition and video accuracy metrics in~\cref{app:fig:num-context-vids-analysis} and~\cref{app:fig:num-objects-analysis} and complement them with their corresponding tables in~\cref{tab:num-context-vids-analysis,tab:num-objects-analysis}.
\begin{table}[ht!]
    \centering
    \begin{subtable}[t]{0.48\textwidth}
    \centering
    \scalebox{0.8}{
    \begin{tabular}{ccc|ccc}
         \(N^\user_p\) & \(S_\contextset^\text{train}\) & \(L\) & \textbf{\textsc{frame acc}} & \textbf{\textsc{ftr}} & \textbf{\textsc{video acc}} \\
         \cmidrule{1-6}
         1 & 12 & 8 & 40.53 (1.84) & 27.49 (2.00) & 44.27 (2.51)  \\
         2 & 6 & 8 & 46.19 (1.89) & 22.59 (1.83) & 52.40 (2.53)  \\
         4 & 3 & 8 & 49.23 (1.88) & 20.69 (1.80) & 55.40 (2.52)\\
         6 & 2 & 8 & 49.44 (1.85) & 20.83 (1.78) & 57.00 (2.51)  \\
    \end{tabular}}
    \caption{Results with CNAPs~\cite{requeima2019fast}.}
    \label{tab:num-context-vids-cnaps}
    \vspace{1em}
    \end{subtable}
    \begin{subtable}[t]{0.48\textwidth}
    \centering
    \scalebox{0.8}{
    \begin{tabular}{ccc|ccc}
         \(N^\user_p\) & \(S_\contextset^\text{train}\) & \(L\) & \textbf{\textsc{frame acc}} & \textbf{\textsc{ftr}} & \textbf{\textsc{video acc}} \\
         \cmidrule{1-6}
         1 & 12 & 8 &  37.79 (1.72) & 26.44 (1.91) & 42.13 (2.50) \\
         2 & 6 & 8 & 42.66 (1.78) & 23.00 (1.82) & 49.07 (2.53) \\
         4 & 3 & 8 & 46.87 (1.76) & 20.17 (1.70) & 55.53 (2.51)  \\
         6 & 2 & 8 & 46.92 (1.79) & 19.36 (1.67) & 54.47 (2.52) \\
    \end{tabular}}
    \caption{Results with ProtoNets~\cite{snell2017prototypical}.}
    \label{tab:num-context-vids-proto}
    \vspace{1em}
    \end{subtable}
    \begin{subtable}[t]{0.48\textwidth}
    \centering
    \scalebox{0.8}{
    \begin{tabular}{ccc|ccc}
         \(N^\user_p\) & \(S_\contextset^\text{train}\) & \(L\) & \textbf{\textsc{frame acc}} & \textbf{\textsc{ftr}} & \textbf{\textsc{video acc}} \\
         \cmidrule{1-6}
         1 & 12 & 8 &  45.37 (1.94) & 28.03 (2.02) & 50.07 (2.58) \\
         2 & 6 & 8 & 49.04 (1.90) & 23.62 (1.91) & 55.21 (2.57) \\
         4 & 3 & 8 & 50.32 (1.88) & 20.42 (1.81) & 55.83 (2.56) \\
         6 & 2 & 8 &  47.91 (1.95) & 22.72 (1.92) & 52.29 (2.58) \\
    \end{tabular}}
    \caption{Results with MAML~\cite{finn2017model}.}
    \label{tab:num-context-vids-maml}
    \vspace{1em}
    \end{subtable}
    \begin{subtable}[t]{0.48\textwidth}
    \centering
    \scalebox{0.8}{
    \begin{tabular}{ccc|ccc}
         \(N^\user_p\) & \(S_\contextset^\text{train}\) & \(L\) & \textbf{\textsc{frame acc}} & \textbf{\textsc{ftr}} & \textbf{\textsc{video acc}} \\
         \cmidrule{1-6}
         1 & 12 & 8 & 38.28 (1.98) & 35.76 (2.25) & 40.62 (2.54) \\
         2 & 6 & 8 &  44.95 (1.97) & 26.41 (2.06) & 48.47 (2.58)\\
         4 & 3 & 8 & 49.03 (1.92) & 20.12 (1.83) & 53.75 (2.58)  \\
         6 & 2 & 8 & 51.71 (1.89) & 19.18 (1.78) & 57.71 (2.55) \\
    \end{tabular}}
    \caption{Results with FineTuner~\cite{tian2020rethinking}.}
    \label{tab:num-context-vids-finetuner}
    \vspace{1em}
    \end{subtable}
    \vspace{-1em}
    \caption{Meta-training with more context videos per object leads to better \textsc{clu-ve} performance. Frames are sampled from an increasing number of clean videos per object (\(N_p^\user\)) using the number of clips per video (\(S_\contextset^\text{train}\)) to keep the total number of context frames fixed per train task. Corresponds to~\cref{fig:num-context-vids-analysis} in the main paper.}
    \label{tab:num-context-vids-analysis}
\end{table}
\begin{table}[ht!]
    \centering
    \begin{subtable}[t]{0.48\textwidth}
        \centering
        \scalebox{0.8}{
        \begin{tabular}{P{1.3cm}P{1.2cm}|ccc}
         \#objs/ train user & \#objs/ test user  & \textbf{\textsc{frame acc}} & \textbf{\textsc{ftr}} & \textbf{\textsc{video acc}} \\
         \cmidrule{1-5}
         2 & 2 & 77.08 (2.98) & 4.10 (1.41) & 84.17 (3.77)  \\
         4 & 4 &  63.85 (2.45) & 9.61 (1.61) & 72.17 (3.34) \\
         6 & 6 & 57.10 (2.14) & 11.57 (1.57) & 63.72 (2.99)  \\
         8 & 8 & 54.00 (1.96) & 15.97 (1.69) & 62.84 (2.62) \\
         2 & no-cap & 45.38 (1.94) & 24.08 (1.97) & 49.51 (2.58) \\
         4 & no-cap & 50.91 (1.85) & 17.24 (1.62) & 59.51 (2.54) \\
         6 & no-cap & 50.57 (1.89) & 17.38 (1.69) & 57.29 (2.55)\\
         8 & no-cap &  53.68 (1.90) & 16.29 (1.63) & 62.78 (2.50)  \\
    \end{tabular}}
    \caption{Results with CNAPs~\cite{requeima2019fast}.}
    \label{tab:num-objects-cnaps}
    \vspace{1em}
    \end{subtable}
    \begin{subtable}[t]{0.48\textwidth}
        \centering
        \scalebox{0.8}{
        \begin{tabular}{P{1.3cm}P{1.2cm}|ccc}
         \#objs/ train user & \#objs/ test user  & \textbf{\textsc{frame acc}} & \textbf{\textsc{ftr}} & \textbf{\textsc{video acc}} \\
         \cmidrule{1-5}
         2 & 2 &  78.38 (2.45) & 4.10 (1.08) & 86.94 (3.48)  \\
         4 & 4 &  56.36 (2.53) & 16.44 (2.22) & 64.78 (3.56) \\
         6 & 6 & 54.28 (2.09) & 13.94 (1.71) & 61.91 (3.02) \\
         8 & 8 & 48.34 (1.87) & 17.07 (1.64) & 53.79 (2.71)  \\
         2 & no-cap & 45.24 (1.75) & 19.61 (1.66) & 51.67 (2.58) \\
         4 & no-cap & 46.09 (1.75) & 20.03 (1.69) & 54.51 (2.57) \\
         6 & no-cap & 48.57 (1.76) & 17.59 (1.61) & 56.46 (2.56) \\
         8 & no-cap &  48.62 (1.77) & 17.67 (1.61) & 55.14 (2.57) \\
    \end{tabular}}
    \caption{Results with ProtoNets~\cite{snell2017prototypical}.}
    \label{tab:num-objects-proto}
    \vspace{1em}
    \end{subtable}
    \begin{subtable}[t]{0.48\textwidth}
        \centering
        \scalebox{0.8}{
        \begin{tabular}{P{1.3cm}P{1.2cm}|ccc}
         \#objs/ train user & \#objs/ test user  & \textbf{\textsc{frame acc}} & \textbf{\textsc{ftr}} & \textbf{\textsc{video acc}} \\
         \cmidrule{1-5}
         2 & 2 &  78.98 (2.69) & 6.74 (1.94) & 84.72 (3.72) \\
         4 & 4 & 61.82 (2.59) & 13.55 (2.11) & 66.81 (3.51) \\
         6 & 6 &  58.64 (2.27) & 16.28 (1.96) & 64.32 (2.98) \\
         8 & 8 & 51.01 (1.99) & 19.89 (1.87) & 55.63 (2.70) \\
         2 & no-cap & 48.34 (1.84) & 21.27 (1.79) & 52.33 (2.53) \\
         4 & no-cap & 51.55 (1.89) & 18.73 (1.73) & 56.87 (2.51) \\
         6 & no-cap & 51.59 (1.87) & 19.12 (1.75) & 56.33 (2.51) \\
         8 & no-cap &  51.75 (1.87) & 19.81 (1.78) & 56.80 (2.51)   \\
    \end{tabular}}
    \caption{Results with MAML~\cite{finn2017model}.}
    \label{tab:num-objects-maml}
    \vspace{1em}
    \end{subtable}
    \begin{subtable}[t]{0.48\textwidth}
        \centering
        \scalebox{0.8}{
        \begin{tabular}{P{1.3cm}P{1.2cm}|ccc}
         \#objs/ train user & \#objs/ test user  & \textbf{\textsc{frame acc}} & \textbf{\textsc{ftr}} & \textbf{\textsc{video acc}} \\
         \cmidrule{1-5}
         2 & 2 & 79.68 (2.97) & 3.65 (1.73) & 81.11 (4.04) \\
         4 & 4 &  62.47 (2.61) & 13.77 (2.27) & 69.42 (3.44) \\
         6 & 6 &  58.86 (2.31) & 16.97 (2.06) & 63.72 (2.99) \\
         8 & 8 &  52.34 (2.00) & 18.46 (1.87) & 56.25 (2.69)\\
         2 & no-cap &  50.62 (1.90) & 20.69 (1.85) & 55.40 (2.52) \\
         4 & no-cap & 52.39 (1.92) & 19.27 (1.80) & 57.67 (2.50) \\
         6 & no-cap & 51.33 (1.88) & 18.94 (1.78) & 55.73 (2.51)  \\
         8 & no-cap &  52.14 (1.90) & 19.10 (1.78) & 56.47 (2.51) \\
    \end{tabular}}
    \caption{Results with FineTuner~\cite{tian2020rethinking}.}
    \label{tab:num-objects-finetuner}
    \vspace{1em}
    \end{subtable}
    \vspace{-1em}
    \caption{Meta-training and -testing with more objects per user poses a harder recognition problem, however, meta-training with fewer objects than encountered at meta-testing shows only a small \textsc{clu-ve} performance drop compared to~\cref{tab:orbit-baselines}, suggesting that models may be able to adapt to more objects in the real-world. Corresponds to~\cref{fig:num-objects-analysis} in the  main paper.}
    \label{tab:num-objects-analysis}
\end{table}

\vspace{-1em}
\paragraph{Number of tasks per train user.}
In~\cref{app:tab:num-train-tasks-analysis}), we investigate the impact of the number of tasks sampled per train user (per epoch), \(T^\text{train}\) during meta-training, and observe that \textsc{clu-ve} performance increases with more tasks, but levels off at around 50 tasks per user.
We expect that bigger gains could be achieved by accounting for the informativeness of frames, instead of uniformly sampling frames from videos.
This is likely because clean and clutter videos are noisy and contain many redundant frames.

\begin{table}[ht!]
    \begin{subtable}[t]{0.48\textwidth}
        \centering
        \scalebox{0.8}{
        \begin{tabular}{l|ccc}
        \(T^\text{train}\) & \textbf{\textsc{frame acc}} & \textbf{\textsc{ftr}} & \textbf{\textsc{video acc}} \\
         \cmidrule{1-4}
         5 & 48.08 (1.85) & 22.21 (1.84) & 55.07 (2.52)  \\
         25 & 51.85 (1.82) & 16.58 (1.59) & 60.47 (2.47) \\ 
         50 & 51.47 (1.81) & 17.87 (1.69) & 59.53 (2.48) \\ 
         100 & 51.24 (1.86) & 18.96 (1.71)& 59.53 (2.48) \\
         200 & 50.83 (1.82) & 20.25 (1.72) & 60.20 (2.48) \\
         500 & 49.72 (1.81) & 18.83 (1.69) & 56.07 (2.51) \\
         \end{tabular}}
        \caption{Results with CNAPs~\cite{requeima2019fast}.}
        \label{tab:num-train-tasks-cnaps}
        \vspace{1em}
    \end{subtable}
    \begin{subtable}[t]{0.48\textwidth}
        \centering
        \scalebox{0.8}{
        \begin{tabular}{l|ccc}
        \(T^\text{train}\) & \textbf{\textsc{frame acc}} & \textbf{\textsc{ftr}} & \textbf{\textsc{video acc}} \\
         \cmidrule{1-4}
         5 & 43.90 (1.71) & 19.42 (1.63) & 49.07 (2.53) \\
         25 & 46.89 (1.76) & 20.40 (1.72) & 54.00 (2.52) \\
         50 & 50.34 (1.74) & 14.93 (1.52) & 59.93 (2.48) \\ 
         100 & 49.59 (1.79) & 19.38 (1.64) & 59.33 (2.49) \\
         200 & 49.78 (1.74) & 15.51 (1.46) & 58.20 (2.50)  \\
         500 & 48.70 (1.79) & 18.28 (1.61) & 54.67 (2.52)  \\
        \end{tabular}}
        \caption{Results with ProtoNets~\cite{snell2017prototypical}.}
        \label{tab:num-train-tasks-protonets}
        \vspace{1em}
    \end{subtable}
    \begin{subtable}[t]{0.48\textwidth}
        \centering
        \scalebox{0.8}{
        \begin{tabular}{l|ccc}
        \(T^\text{train}\) & \textbf{\textsc{frame acc}} & \textbf{\textsc{ftr}} & \textbf{\textsc{video acc}} \\
         \cmidrule{1-4}
         5 & 44.33 (2.05) & 32.60 (2.20) & 47.27 (2.53)\\
         25 & 51.52 (1.89) & 19.16 (1.75) & 57.33 (2.50)\\
         50 & 51.67 (1.86) & 19.74 (1.76) & 57.47 (2.50) \\
         100 & 51.64 (1.86) & 20.09 (1.80) & 56.87 (2.51) \\
         200 & 50.61 (1.92) & 21.64 (1.85) & 56.13 (2.51)\\
         500 & 49.78 (1.89) & 21.86 (1.84) & 56.27 (2.51) \\
        \end{tabular}}
        \caption{Results with MAML~\cite{finn2017model}.}
        \label{tab:num-train-tasks-maml}
        \vspace{1em}
    \end{subtable}
    \begin{subtable}[t]{0.48\textwidth}
        \centering
        \scalebox{0.8}{
        \begin{tabular}{l|ccc}
        \(T^\text{train}\) & \textbf{\textsc{frame acc}} & \textbf{\textsc{ftr}} & \textbf{\textsc{video acc}} \\
         \cmidrule{1-4}
         5 & 48.99 (1.82) & 19.32 (1.75) & 56.47 (2.51)\\
         25 & 51.90 (1.87) & 19.71 (1.81) & 58.13 (2.50) \\
         50 & 52.52 (1.88) & 18.60 (1.76) & 58.00 (2.50) \\
         100 & 52.20 (1.89) & 17.99 (1.72) & 57.13 (2.50) \\
         200 & 54.79 (1.87) & 16.86 (1.67) & 61.13 (2.47) \\
         500 & 52.90 (1.88) & 18.18 (1.74) & 57.87 (2.50) \\
        \end{tabular}}
        \caption{Results with FineTuner~\cite{tian2020rethinking}.}
        \label{tab:num-train-tasks-finetuner}
        \vspace{1em}
        \end{subtable}
        \vspace{-1em}
        \caption{Sampling more tasks per train user has limited benefits on \textsc{clu-ve} performance beyond \(T^\text{train}=50\) suggesting that sampling techniques that account for frame informativeness may be required.}
        \label{app:tab:num-train-tasks-analysis}
\end{table}

\end{document}